\documentclass{article}

\usepackage[table]{xcolor}         

\usepackage{wrapfig}
\usepackage{tikz}
\usepackage{comment}
\usepackage{amsmath,amssymb} 
\usepackage{color}
\usepackage{graphicx}
\usepackage{amsmath}
\usepackage{amssymb}
\usepackage{booktabs}
\usepackage{tabularx}
\usepackage{pifont}
\usepackage{multirow}
\usepackage{caption}
\usepackage{subcaption}
\usepackage{booktabs}
\usepackage{lipsum}
\usepackage[export]{adjustbox}
\usepackage{longtable}
\usepackage{titletoc}
\usepackage{float}
\usepackage{xspace}
\usepackage{enumitem}
\usepackage{xr}
\usepackage[percent]{overpic}

\newcommand\crule[3][black]{\textcolor{#1}{\rule{#2}{#3}}}

\usepackage[accsupp]{axessibility}  

\newcommand{\thor}{\mbox{\sc{AI2-THOR}}\xspace}
\newcommand{\env}{\mbox{\sc{ProcTHOR}}\xspace}
\newcommand{\envb}{\mbox{\sc{\textbf{ProcTHOR}}}\xspace}
\newcommand{\elienv}{\mbox{\sc{ArchitecTHOR}}\xspace}
\newcommand{\tenk}{\mbox{\sc{ProcTHOR-10k}}\xspace}

\newcommand{\eai}{\mbox{\sc{E-AI}}\xspace}

\newcommand{\eg}{\emph{e.g.}\xspace}
\newcommand{\ie}{\emph{i.e.}\xspace}
\newcommand{\etc}{\emph{etc.}\xspace}
\newcommand{\etal}{\emph{et al.}\xspace}

\newcommand\blfootnote[1]{%
  \begingroup
  \renewcommand\thefootnote{}\footnote{#1}%
  \addtocounter{footnote}{-1}%
  \endgroup
}

\definecolor{Color4}{HTML}{FFEEEE}

    \usepackage[preprint,nonatbib]{neurips_2022}

\usepackage[utf8]{inputenc} 
\usepackage[T1]{fontenc}    
\usepackage{url}            
\usepackage{booktabs}       
\usepackage{amsfonts}       
\usepackage{nicefrac}       
\usepackage{microtype}      
\definecolor{citecolor}{HTML}{2980b9}
\definecolor{linkcolor}{HTML}{c0392b}
\usepackage[pagebackref=true,breaklinks=true,colorlinks,bookmarks=false,citecolor=citecolor,linkcolor=linkcolor]{hyperref}

\definecolor{bathroomColor}{HTML}{f0b975}
\definecolor{bedroomColor}{HTML}{5ebdeb}
\definecolor{livingRoomColor}{HTML}{52ebbd}
\definecolor{kitchenColor}{HTML}{eb6a9d}

\title{
    \raisebox{-0.15\height}{\includegraphics[width=0.25in]{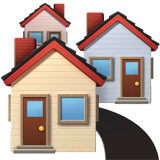}}
    ProcTHOR: Large-Scale Embodied AI Using Procedural Generation
}

\author{%
  Matt Deitke$^{\dagger\psi}$, Eli VanderBilt$^\dagger$, Alvaro Herrasti$^\dagger$, Luca Weihs$^\dagger$\\ \textbf{Jordi Salvador$^\dagger$, Kiana Ehsani$^\dagger$, Winson Han$^\dagger$, Eric Kolve$^\dagger$}\\ \textbf{Ali Farhadi$^{\psi}$, Aniruddha Kembhavi$^{\dagger\psi}$, Roozbeh Mottaghi$^{\dagger\psi}$}\\
  $^\dagger$PRIOR @ Allen Institute for AI, $^\psi$University of Washington, Seattle\\
  \href{https://procthor.allenai.org}{\texttt{procthor.allenai.org}}
}

\begin{document}

\maketitle

\begin{abstract}

Massive datasets and high-capacity models have driven many recent advancements in computer vision and natural language understanding. This work presents a platform to enable similar success stories in Embodied AI. We propose \env, a framework for procedural generation of Embodied AI environments. \env enables us to sample arbitrarily large datasets of diverse, interactive, customizable, and performant virtual environments to train and evaluate embodied agents across navigation, interaction, and manipulation tasks. We demonstrate the power and potential of \env via a sample of 10,000 generated houses and a simple neural model. Models trained using only RGB images on \env, with no explicit mapping and no human task supervision produce state-of-the-art results across 6 embodied AI benchmarks for navigation, rearrangement, and arm manipulation, including the presently running Habitat 2022, AI2-THOR Rearrangement 2022, and RoboTHOR challenges. We also demonstrate strong 0-shot results on these benchmarks, via pre-training on \env with no fine-tuning on the downstream benchmark, often beating previous state-of-the-art systems that access the downstream training data.
\end{abstract}


\blfootnote{Correspondence to <mattd@allenai.org>.}

\section{Introduction}
\label{sec:introduction}

Computer vision and natural language processing models have become increasingly powerful through the use of large-scale training data. Recent models such as CLIP~\cite{Radford2021LearningTV}, DALL-E~\cite{Ramesh2021ZeroShotTG}, GPT-3~\cite{Brown2020LanguageMA}, and Flamingo~\cite{flamingo} use massive amounts of task agnostic data to pre-train large neural architectures that perform remarkably well at downstream tasks, including in zero and few-shot settings. In comparison, the Embodied AI (\eai) research community predominantly trains agents in simulators with far fewer scenes~\cite{Ramakrishnan2021HabitatMatterport3D,kolve2017ai2,Deitke2020RoboTHORAO}. Due to the complexity of tasks and the need for long planning horizons, the best performing \eai models continue to overfit on the limited training scenes and thus generalize poorly to unseen environments.

In recent years, \eai simulators have become increasingly more powerful with support for physics, manipulators, object states, deformable objects, fluids, and real-sim counterparts \cite{kolve2017ai2,savva2019habitat,Shen2020iGibsonAS,Gan2020ThreeDWorldAP,xiang2020sapien}, but scaling them up to tens of thousands of scenes has remained challenging. Existing \eai  environments are either designed manually \cite{kolve2017ai2,Gan2020ThreeDWorldAP} or obtained via 3D scans of real structures \cite{savva2019habitat,Ramakrishnan2021HabitatMatterport3D}. The former approach requires 3D artists to spend a significant amount of time designing 3D assets, arranging them in sensible configurations within large spaces, and carefully configuring the right textures and lighting in these environments. The latter involves moving specialized cameras through many real-world environments and then stitching the resulting images together to form 3D reconstructions of the scenes. These approaches are not scalable, and expanding existing scene repositories multiple orders of magnitude is not practical.

\begin{figure}[tp]
    \centering
    \vspace{-0.35in}
    \makebox[\textwidth][c]{
        \includegraphics[width=1\textwidth]{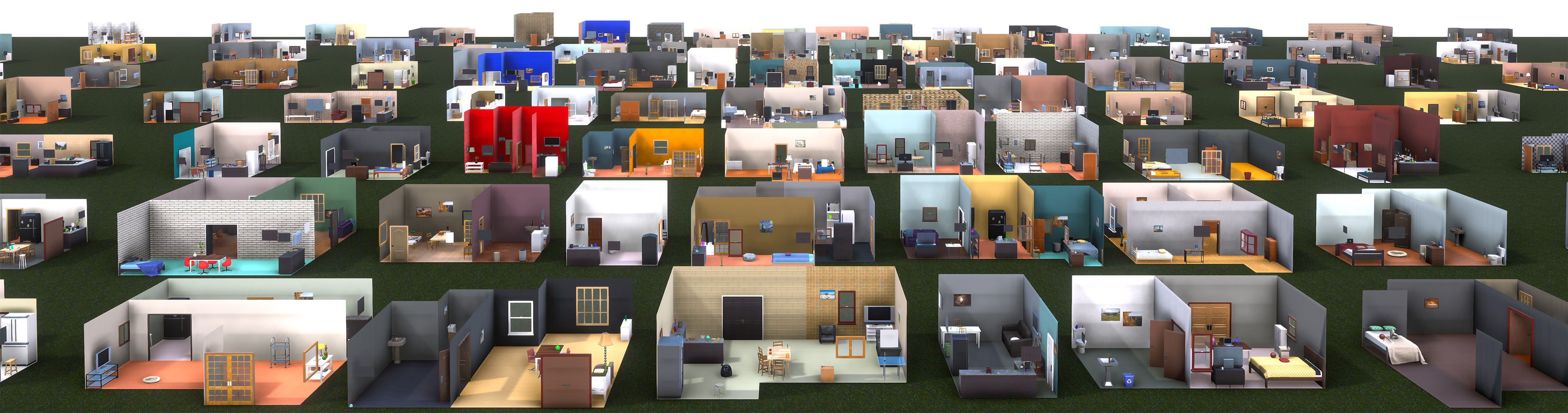}
    }
    \caption{We propose \env{}, a framework to procedurally generate a large variety of diverse, interactable, and customizable houses.}
    \vspace{-0.2in}
    \label{fig:examples}
\end{figure}

We present \env, a framework built off of AI2-THOR~\cite{kolve2017ai2}, to procedurally generate fully-interactive, physics-enabled environments for \eai research. Given a room specification (e.g., a house with 3 bedrooms, 3 baths, and 1 kitchen), \env can produce a large and diverse set of floorplans that meet these requirements (Fig.~\ref{fig:examples}). A large asset library of 108 object types and 1633 fully interactable instances is used to automatically populate each floorplan, ensuring that object placements are physically plausible, natural, and realistic. One can also vary the intensity and color of lighting elements (both artificial lighting and simulated skyboxes) in each scene, to simulate variations in indoor lighting and the time of the day. Assets (such as furniture and fruit) and larger structures such as walls and doors can be assigned a variety of colors and textures, sampled from sets of plausible colors and materials for each asset category. Together, the diversity of layouts, assets, placements, and lighting leads to an arbitrarily large set of environments -- allowing \env to scale orders of magnitude beyond the number of scenes currently supported by present-day simulators. In addition, \env supports dynamic material randomizations, whereby colors and materials of individual assets can be randomized each time an environment is loaded into memory for training. Importantly, in contrast to environments produced using 3D scans, scenes produced by \env contain objects that both support a variety of different object states (\eg open, closed, broken, \etc) and are fully interactive so that they can be physically manipulated by agents with robotic arms. We also present \elienv, a 3D artist-designed set of 10 high quality fully interactable houses, meant to be used as a test-only environment for research within household environments. In contrast to AI2-iTHOR (single rooms) and RoboTHOR (lesser visual diversity) environments, \elienv contains larger, diverse, and realistic houses.

We demonstrate the ease and effectiveness of \env by sampling an environment of 10,000 houses (named \tenk), composed of diverse layouts ranging from small 1-room houses to larger 10-room houses. We train agents with very simple neural architectures (CNN+RNN) -- \emph{without} a depth sensor, and instead only employing RGB channels, with no explicit mapping and no human task supervision -- on \tenk and produce state-of-the-art (SoTA) models on several navigation and interaction benchmarks. As of 10am PT on June 14th, 2022 we obtain (1) \textbf{RoboTHOR ObjectNav Challenge}~\cite{robothor-challenge} --  0-shot performance superior to the previous SoTA which uses RoboTHOR training scenes -- with fine-tuning we obtain an 8.8 point improvement in SPL over the previous SoTA; (2) \textbf{Habitat ObjectNav Challenge 2022}~\cite{habitat-challenge} -- top of the leaderboard results with a ${>}3$ point gain in SPL over the next best submission; (3) \textbf{1-phase Rearrangement Challenge 2022}~\cite{roomr-challenge} -- top of the leaderboard results with Prop Fixed Strict improving from 0.19 to 0.245; (4) \textbf{AI2-iTHOR ObjectNav} -- 0-shot numbers which already outperform a previous model that trains on AI2-iTHOR, with fine-tuning we achieve a success rate of 77.5\%; (5) \textbf{ArmPointNav}~\cite{manipulathor} -- 0-shot number that beats previous SoTA results when using RGB; and (6) \textbf{ArchitecTHOR ObjectNav} -- a large success rate improvement from 18.5\% to 31.4\%. Finally, an ablation analysis clearly shows the advantages of scaling up from 10 to 100 to 1K and finally to 10K scenes and indicates that further improvements can be obtained by invoking \env to produce even larger environments.

In summary, our contributions are (1) \env, a framework that allows for the performant procedural generation of an unbounded number of diverse, fully-interactive, simulated environments, (2) \elienv, a new, 3D artist-designed set of houses for \eai evaluation, and (3) SoTA results across six \eai benchmarks covering manipulation and navigation tasks, including strong 0-shot results.
\env will be open-sourced and the code used in this work will be released.

\section{Related Work}
\label{sec:relatedwork}

\noindent \textbf{Embodied AI platforms.} Various Embodied AI platforms have been developed over the past several years~\cite{kolve2017ai2,savva2019habitat,Shen2020iGibsonAS,xiang2020sapien,Gan2020ThreeDWorldAP,house3d}. These platforms target different design goals. AI2-THOR~\cite{kolve2017ai2} and its variants (ManipulaTHOR~\cite{manipulathor} and RoboTHOR~\cite{Deitke2020RoboTHORAO}) are built in the Unity game engine and focus on agent-object interactions, object state changes, and accurate physics simulation. Unlike AI2-THOR, Habitat~\cite{savva2019habitat} provides scenes constructed from 3D scans of houses, however, objects and scenes are not interactable. A more recent version, Habitat 2.0~\cite{szot2021habitat}, introduces object interactions at the expense of being limited to one floorplan and synthetic scenes. iGibson~\cite{Shen2020iGibsonAS} includes photo-realistic scenes, but with limited interactions such as pushing. iGibson 2.0~\cite{Li2021iGibson2O} extends iGibson by focusing on household tasks and object state changes in synthetic scenes and includes a virtual reality interface. ThreeDWorld~\cite{Gan2020ThreeDWorldAP} targets high-fidelity physics simulation such as liquid and deformable object simulation. VirtualHome~\cite{Puig2018VirtualHomeSH} is designed for simulating human activities via programs. RLBench~\cite{james2020rlbench}, RoboSuite~\cite{zhu2020robosuite} and Sapien~\cite{xiang2020sapien} target fine-grained manipulation. The main advantage of \env is that we can generate a diverse set of \emph{interactive} scenes procedurally, enabling studies of data augmentation and large-scale training in the context of Embodied AI. 
 
\noindent \textbf{Large-scale datasets.} Large-scale datasets have resulted in major breakthroughs in different domains such as image classification~\cite{imagenet,Kuznetsova2020TheOI}, vision and language~\cite{Changpinyo2021Conceptual1P,Thomee2016YFCC100MTN}, 3D understanding~\cite{shapenet2015,Xiang2016ObjectNet3DAL}, autonomous driving~\cite{nuscenes,sun2020scalability}, and robotic object manipulation~\cite{pinto2016supersizing,mu2021maniskill}. However, there are not many interactive large-scale datasets for Embodied AI research. \env includes interactive houses generated procedurally. Hence, there are an arbitrarily large number of scenes in the framework. The closest works to ours are~\cite{Ramakrishnan2021HabitatMatterport3D,Petrenko2021MegaverseSE,li2021openrooms}. HM3D~\cite{Ramakrishnan2021HabitatMatterport3D} is a recent framework that includes 1,000 scenes generated using 3D scans of real environments. \env has a number of key distinctions: (1) unlike HM3D which includes static scenes, the scenes in \env are interactive i.e., objects can move and change state, the lighting and texture of objects can change, and a physics engine determines the future states of the scenes; (2) it is challenging to scale up HM3D as it requires scanning a house and cleaning up the data, while we can procedurally generate more houses; (3) HM3D can be used only for navigation tasks (as there is no physics simulation and object interaction), while \env can be used for tasks other than navigation. OpenRooms~\cite{li2021openrooms} is similar to HM3D in terms of the source of the data (3D scans) and dataset size. However, OpenRooms is interactive. OpenRooms is also confined to the set of scanned houses, and it takes a significant amount of time to annotate a new scene (e.g., labeling materials for one object takes 1 minute), while \env does not suffer from these issues. Megaverse~\cite{Petrenko2021MegaverseSE} is another large-scale Embodied AI platform that includes procedurally generated environments. Although it is impressive in terms of simulation speed, it includes only game-like environments with a simplified appearance. In contrast, \env mimics real-world houses in terms of the complexity of appearance, physics, and object interactions. 

\noindent \textbf{Scene generation.} Indoor scene synthesis has been studied extensively in computer vision and graphics communities. \cite{Chang2015TextT3,Chang2014LearningSK,chang2014interactive} address generating 3D scenes from text descriptions. \cite{wu2019data,Hu2020Graph2PlanLF,Nauata2021HouseGANGA} learn to generate house floorplans. \cite{Ritchie2019FastAF,Zhang2020DeepGM,Chaudhuri2019LearningGM,Wang2021SceneFormerIS,keshavarzi2020scenegen,li2019grains} use generative models for indoor scene generation. \cite{zhou2019scenegraphnet,Savva2017SceneSuggestC3} propose potential objects for a query location in an indoor scene. Others have used procedural generation \cite{khalifa2020pcgrl, earle2021learning} and unsupervised learning \cite{dennis2020emergent} to synthesize grid-world environments for AI. \env is specifically designed for Embodied AI research in the sense that (1) all scenes are interactive and physics-enabled, and the placement of objects respects the physics of the world (e.g., there are no two objects that clip through each other), (2) there are various forms of scene augmentation such as randomization of object placements while following certain commonsense rules, variation in the appearance of objects and structures, and variation in lighting. 

\vspace{-0.1in}
\section{\env}
\label{sec:generation}
\vspace{-0.1in}

\env is a framework to procedurally generate \eai environments. It extends \thor and, thereby, inherits \thor's large asset library, robotic agents, and accurate physics simulation. Just as in scenes painstakingly created by designers in \thor, environments in \env are fully interactive and support navigation, object manipulation, and multi-agent interaction.

\begin{minipage}{\linewidth}
    \centering
    \vspace{0.05in}
    \begin{overpic}[width=\textwidth]{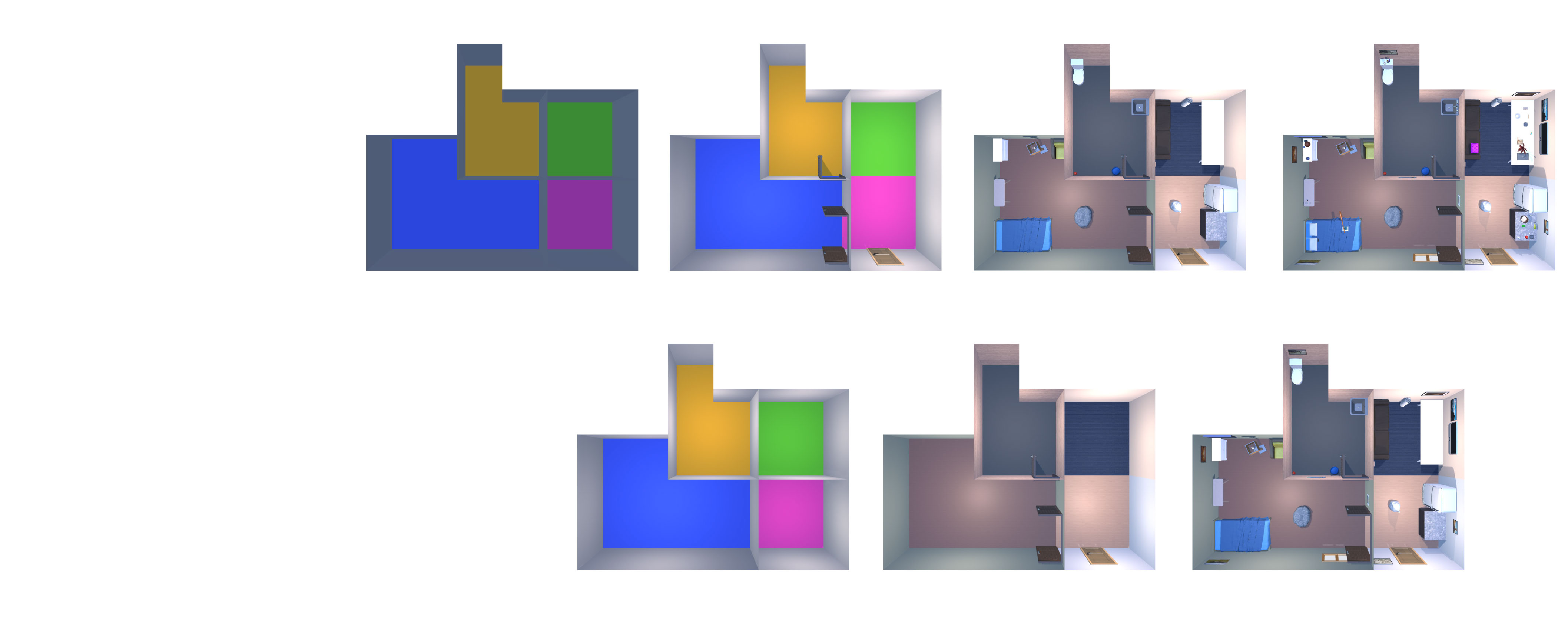}
        \put(0,0){\includegraphics[width=1\textwidth]{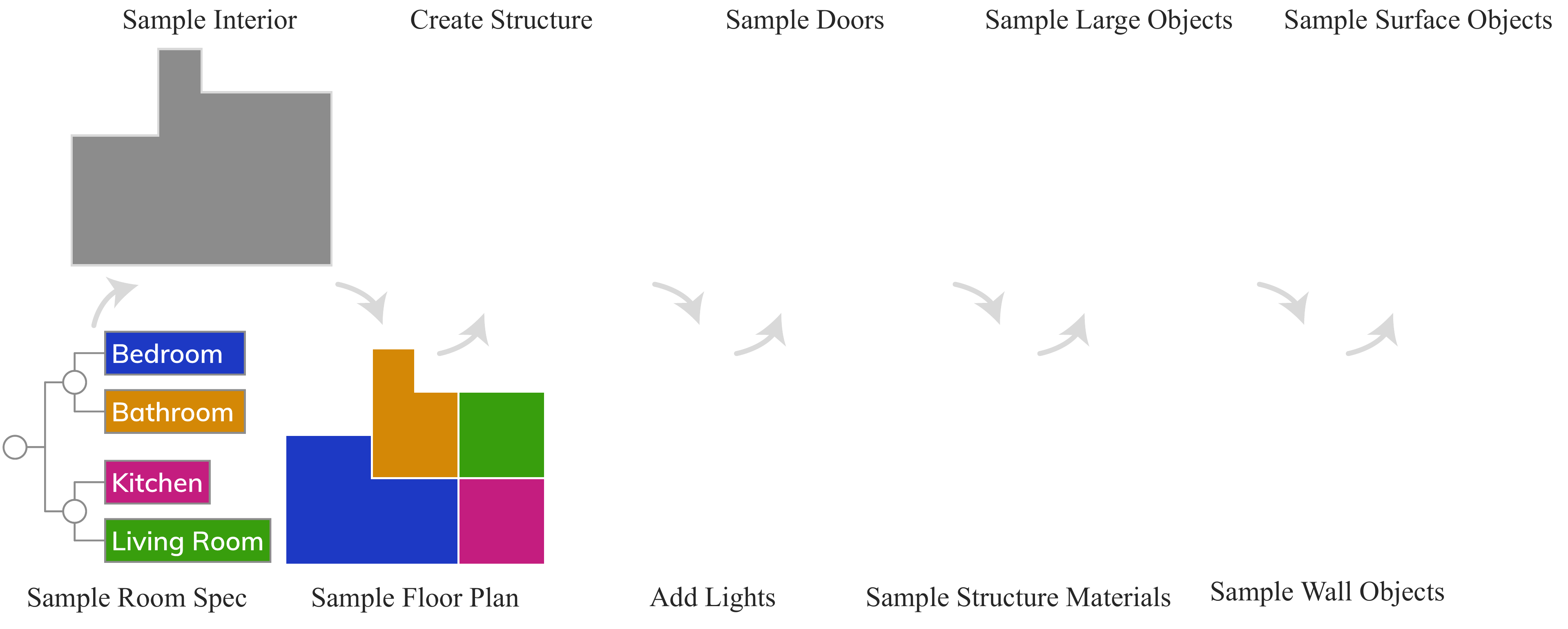}}
    \end{overpic}
    \captionof{figure}{Procedurally generating a house using \env.}
    \label{fig:genOverview}
\end{minipage}

Fig.~\ref{fig:genOverview} shows a high-level schematic of the procedure used by \env to generate a scene. Given a room specification (\eg house with 1 bedroom + 1 bathroom), \env uses multi-stage conditional sampling to, iteratively, generate a floor plan, create an external wall structure, sample lighting, and doors, then sample assets including large, small and wall objects, pick colors and textures, and determine appropriate placements for assets within the scene. We refer the reader to the appendix for details regarding our procedural generation and sampling mechanism, but highlight five key characteristics of \env: \textbf{Diversity}, \textbf{Interactivity}, \textbf{Customizability}, \textbf{Scale}, and \textbf{Efficiency}.

\noindent \textbf{Diversity.}
\env enables the creation of rich and diverse environments. Mirroring the success of pre-training models with diverse data in the vision and NLP domains, we demonstrate the utility of this diversity on several \eai tasks. Scenes in \env exhibit diversity across several facets:

\noindent \emph{Diversity of floor plans.} 
Given a room specification, we first employ iterative boundary cutting to obtain an external scene layout (that can range from a simple rectangle to a complex polygon). The recursive layout generation algorithm by Lopes \etal~\cite{lopes2010constrained} is then used to divide the scene into the desired rooms. Finally, we determine connectivity between rooms using a set of user-defined constraints. These procedures result in natural room layouts (e.g., bedrooms are often connected to adjoining bathrooms via a door, bathrooms more often have a single entrance, etc). As exemplified in Fig.~\ref{fig:floorplanExamples}, \env generates hugely diverse floor plans using this procedure.

\begin{minipage}{\linewidth}
    \centering
    \vspace{0.05in}
    \includegraphics[width=0.9\textwidth]{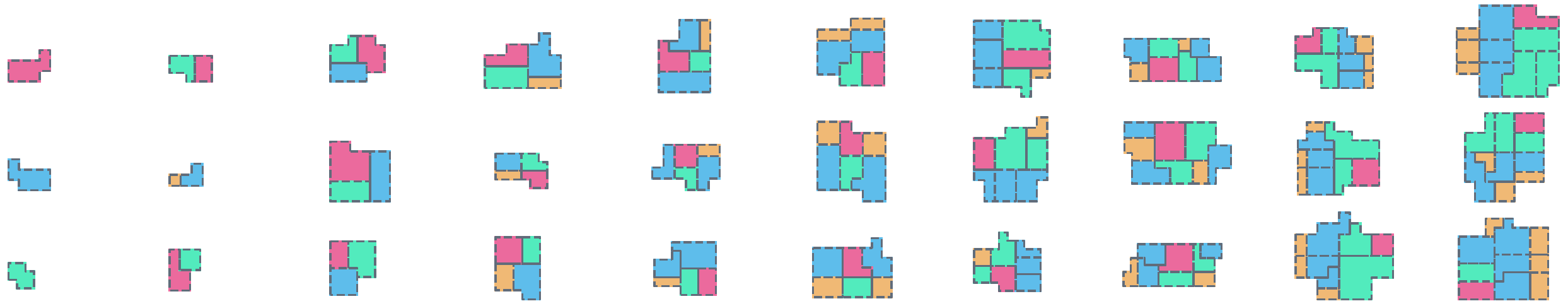}
    \captionof{figure}{\textbf{Floorplan diversity.} Examples showing the diversity of the generated floorplans. Rooms in the house are colored by \crule[bedroomColor]{2.5mm}{2.5mm} Bedroom, \crule[bathroomColor]{2.5mm}{2.5mm} Bathroom, \crule[kitchenColor]{2.5mm}{2.5mm} Kitchen, and \crule[livingRoomColor]{2.5mm}{2.5mm} Living Room.}
    \label{fig:floorplanExamples}
\end{minipage}

\noindent \emph{Diversity of assets.} 
\env populates scenes with small and large assets from its database of 1633 household assets across 108 categories (examples in Fig.~\ref{fig:assetsDiversity}). While many assets are inherited from \thor, we also introduce new assets such as windows, doors, and countertops, hand-designed by 3D graphic designers. Asset instances are split into train/val/test subsets and are interactable, i.e. objects can be picked and placed within the scenes, some objects have multiple states (\eg a light can be on or off) and several objects consists of parts with rigid body motions (\eg door on a microwave).

\begin{minipage}{\linewidth}
    \centering
    \includegraphics[width=1\textwidth]{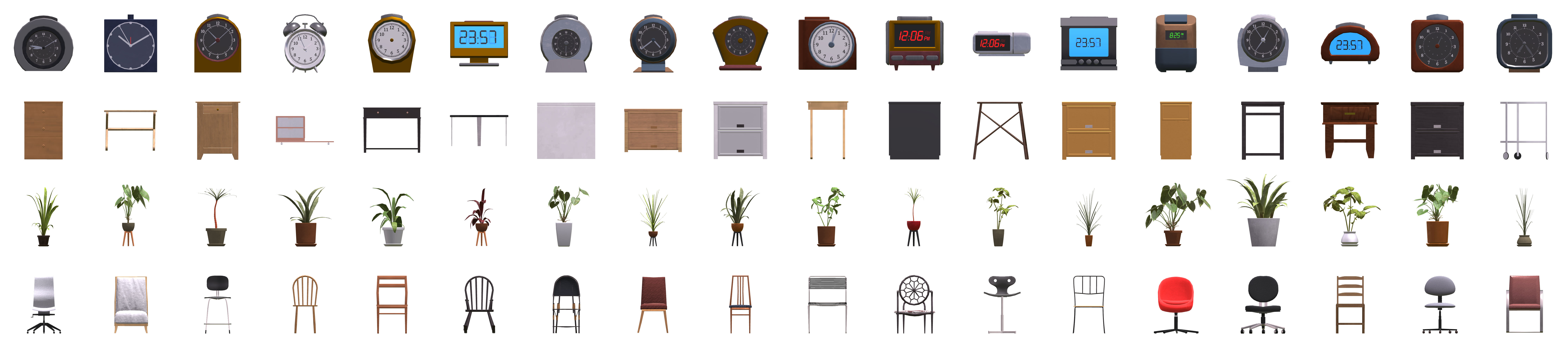}
    \begin{center}
        \vspace{-0.05in}
        {\large $\cdots$}
        \vspace{-0.05in}
    \end{center}
    
    \captionof{figure}{\textbf{Object diversity.} A subset of instances for four object categories.}
    \label{fig:assetsDiversity}
\end{minipage}

\noindent \emph{Diversity of materials.} 
Walls can have two kinds of materials -- one of 40 solid (and popular) colors or one of 122 wall textures such as brick and tile. 
We also provide 55 floor materials. The ceiling material for the entire house is sampled from the set of wall materials. \env also provides the ability to randomize materials of objects. Materials are only randomized within categories, which ensures objects still look and behave like the class they represent.

\begin{minipage}{\linewidth}
    \centering
    \includegraphics[width=1\textwidth]{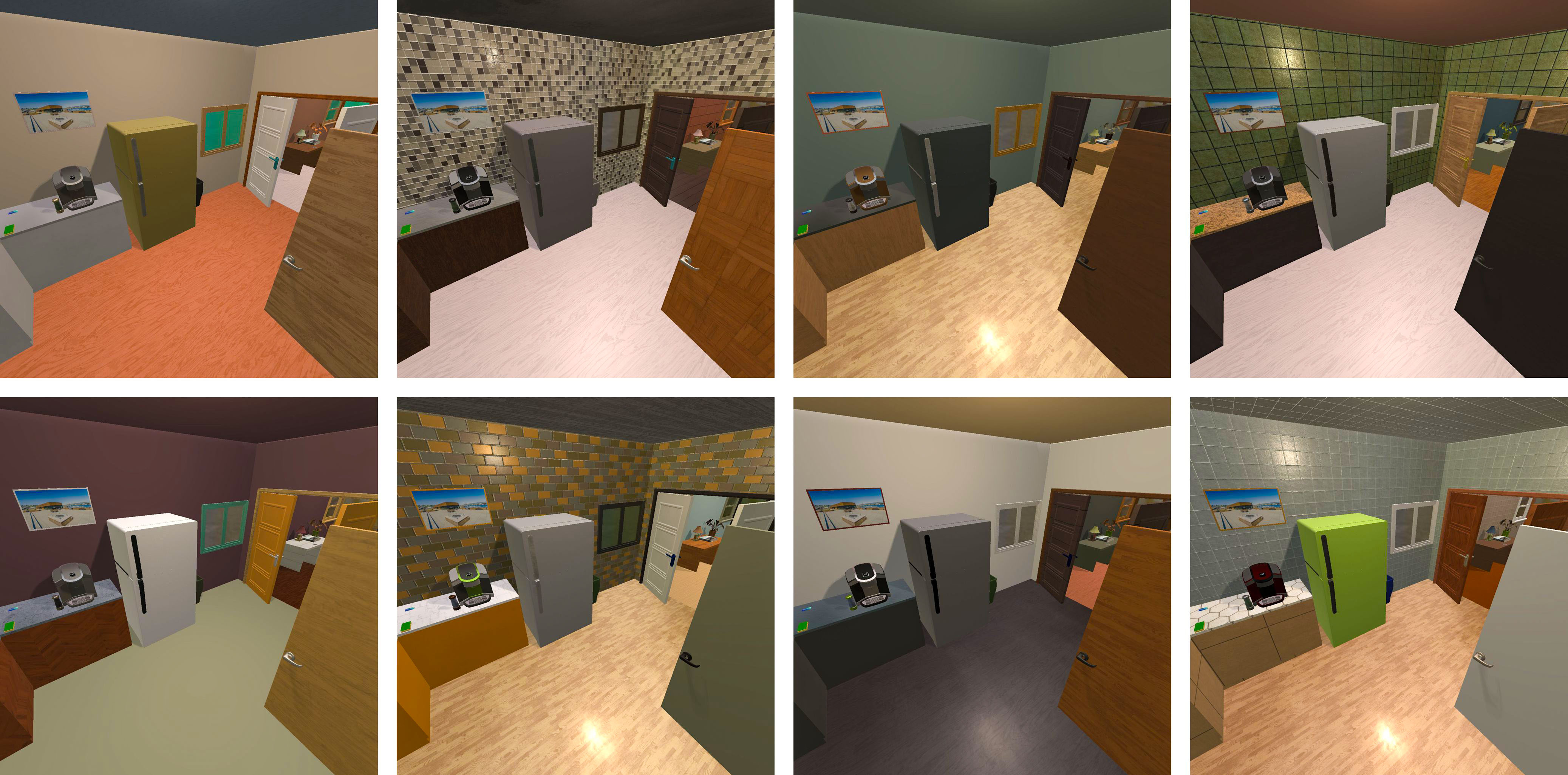}
    \captionof{figure}{\textbf{Material augmentation}. Different materials for objects and structural elements like walls and floors.}
    \label{fig:textureDiversity}
\end{minipage}

\noindent \emph{Diversity of object placements.} 
Asset categories have several soft annotations that help place them realistically within a house. These include room assignments (\eg couch in a living room but not a bathroom) and location assignments (\eg fridge along a wall, TV not on the floor). We also develop the notion of a Semantic Asset Group (SAG) -- groups of assets that typically co-occur (\eg dining table with four chairs) and thus must be sampled and placed using dependent sampling. Given a layout, individual assets and SAGs that lie on the floor are sampled and placed iteratively, ensuring that rooms continue to have adequate floor space for agents to navigate and manipulate objects. Then wall objects such as windows and paintings get placed, and finally, surface objects (ones found on top of other assets) are placed (\eg cups on the kitchen counter). This sampling allows for a large and diverse set of object choices and placements within any layout. Fig.~\ref{fig:placementDiversity} shows such variations.

\begin{minipage}{\linewidth}
    \centering
    \vspace{0.05in}
    \includegraphics[width=1\textwidth]{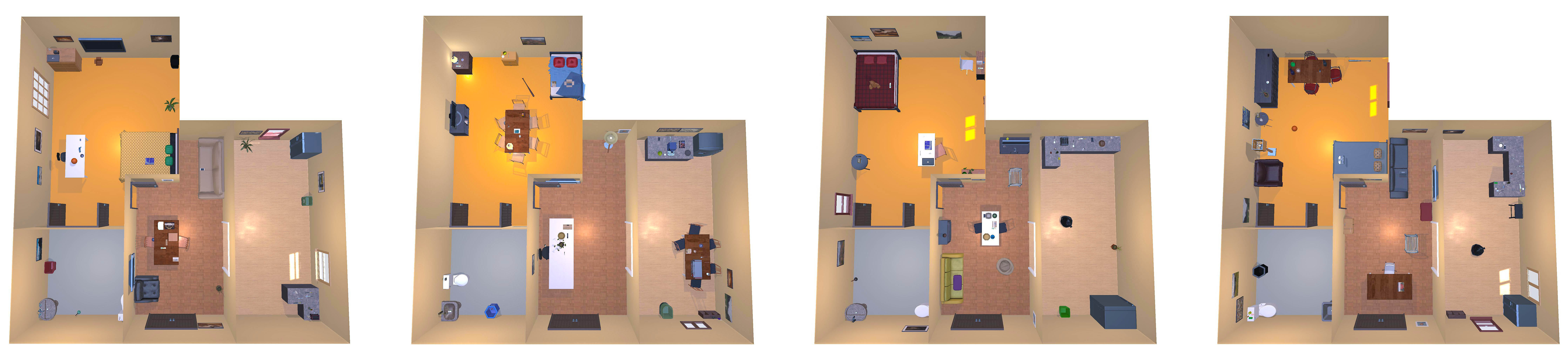}
    \captionof{figure}{\textbf{Object placement.} Four examples of object placement within the same room layout.}
    \label{fig:placementDiversity}
\end{minipage}

\noindent \emph{Diversity of lighting.}
\env supports a single directional light (analogous to the sun) and several point lights (analogous to lightbulbs). Varying the color, intensity, and placement of these sources allows us to simulate different artificial lighting, typically observed in houses, and also at different times of the day. Lighting has a significant effect on the rendered images as seen in Fig.~\ref{fig:lightingDiversity}.

\begin{minipage}{\linewidth}
    \centering
    \vspace{0.05in}
    \includegraphics[width=1\textwidth]{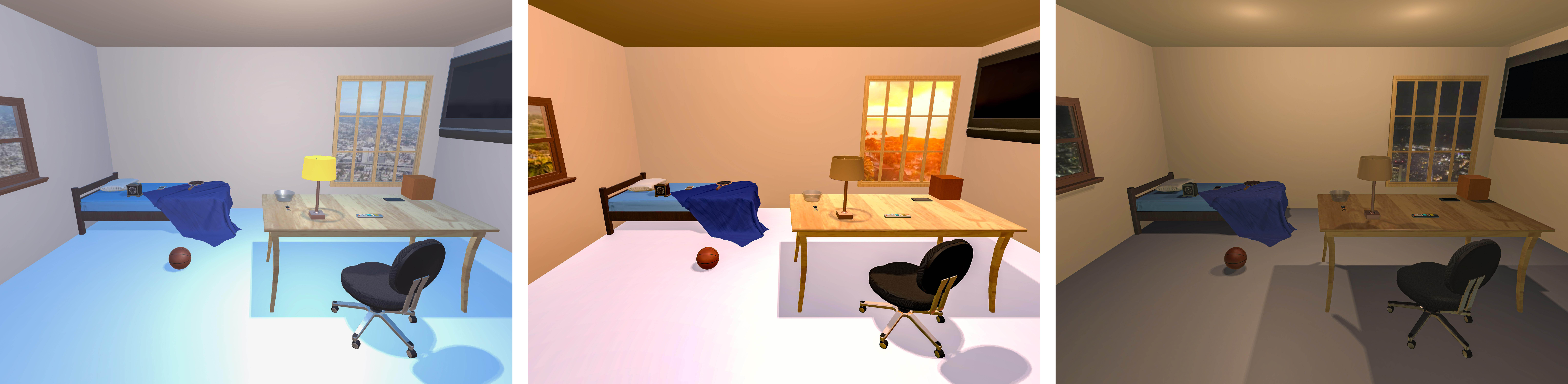}
    \captionof{figure}{\textbf{Lighting variation}. Morning, dusk, and night lighting for an example scene.}
    \label{fig:lightingDiversity}
\end{minipage}

\noindent \textbf{Interactivity.} A key property of \env is the ability to interact with objects to change their location or state (Fig.~\ref{fig:interactivity}). This capability is fundamental to many Embodied AI tasks. Datasets like HM3D~\cite{Ramakrishnan2021HabitatMatterport3D} that are created from static 3D scans do not possess this capability. \env supports agents with arms capable of manipulating objects and interacting with each other.

\begin{minipage}{\linewidth}
    \centering
    \vspace{0.05in}
    \includegraphics[width=1\textwidth]{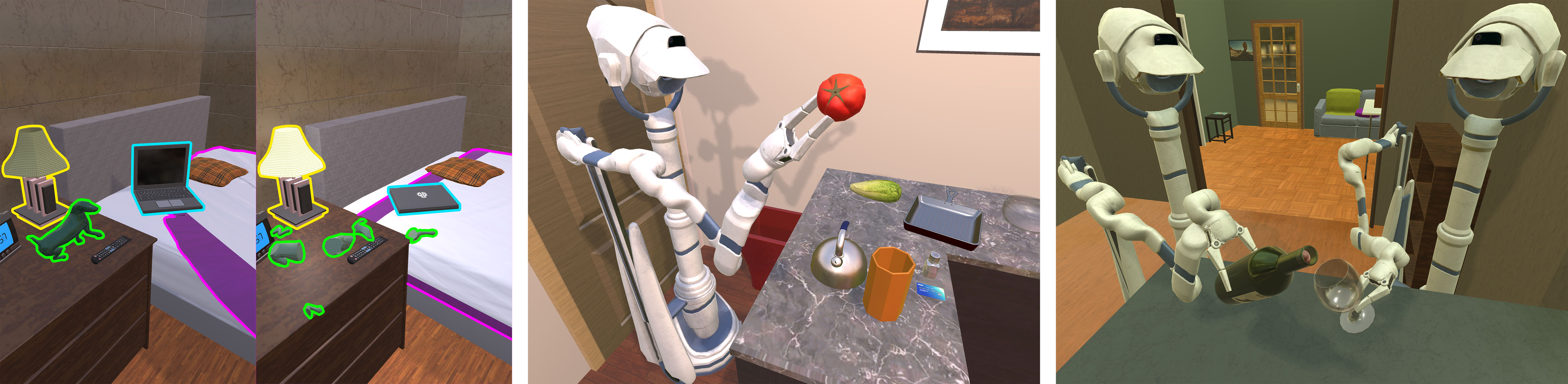}
    \captionof{figure}{\textbf{Interactivity.} Object states can change (e.g., the laptop or the lamp in the left panel), and the agents can interact with objects and other agents (middle and right panels).}
    \label{fig:interactivity}
\end{minipage}

\noindent \textbf{Customizability.} \env supports many room, asset, material, and lighting specifications. With a few simple lines of specification, one can easily generate customized environments of interest. Fig.~\ref{fig:customize} shows examples of such varied scenes (classroom, library, and office).  

\begin{minipage}{\linewidth}
    \centering
    \vspace{0.05in}
    \includegraphics[width=1\textwidth]{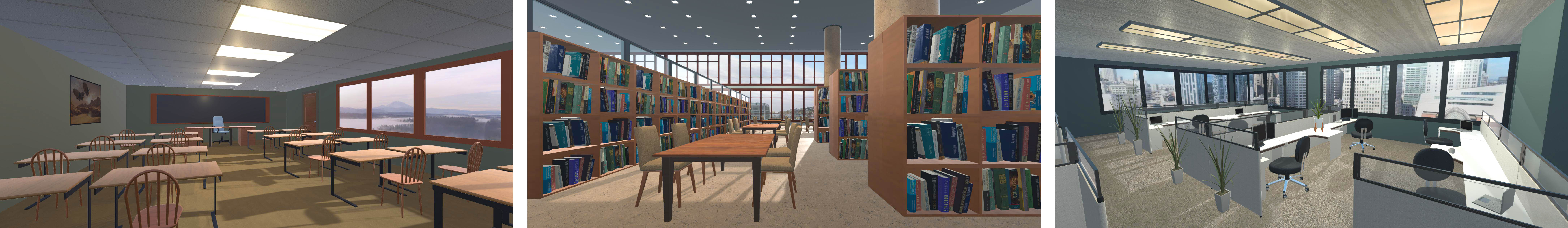}
    \captionof{figure}{\textbf{Customizability.} \env can be used to construct custom scene types such as classrooms, libraries, and offices.}
    \label{fig:customize}
\end{minipage}

\noindent \textbf{Scale and Efficiency.}
\env currently uses 16 different scene specifications to seed the scene generation process. These can result in over 100 billion layouts. \env uses 18 different Semantic Asset groups and 1633 assets. These can result in roughly 20 million unique asset groups. Each of these assets can be placed in numerous locations. In addition, each house gets scaled and uses a variety of lighting. This diversity of layouts, assets, materials, placements, and lighting enables the generation of \emph{arbitrarily large} sets of houses -- either statically generated and stored as a dataset or dynamically generated at each iteration of training. Scenes are efficiently represented in a JSON specification and are loaded into \thor at runtime, making the memory overhead of storing houses incredibly efficient. Moreover, the scene generation process is fully automatic and fast and \env provides high framerates for training \eai models (see Sec.~\ref{sec:analysis} for details).



\section{\envb{}-10K}
\label{sec:analysis}

\begin{figure}[t]
     \centering
     \begin{subfigure}[h]{0.325\textwidth}
         \centering
         \includegraphics[width=\textwidth]{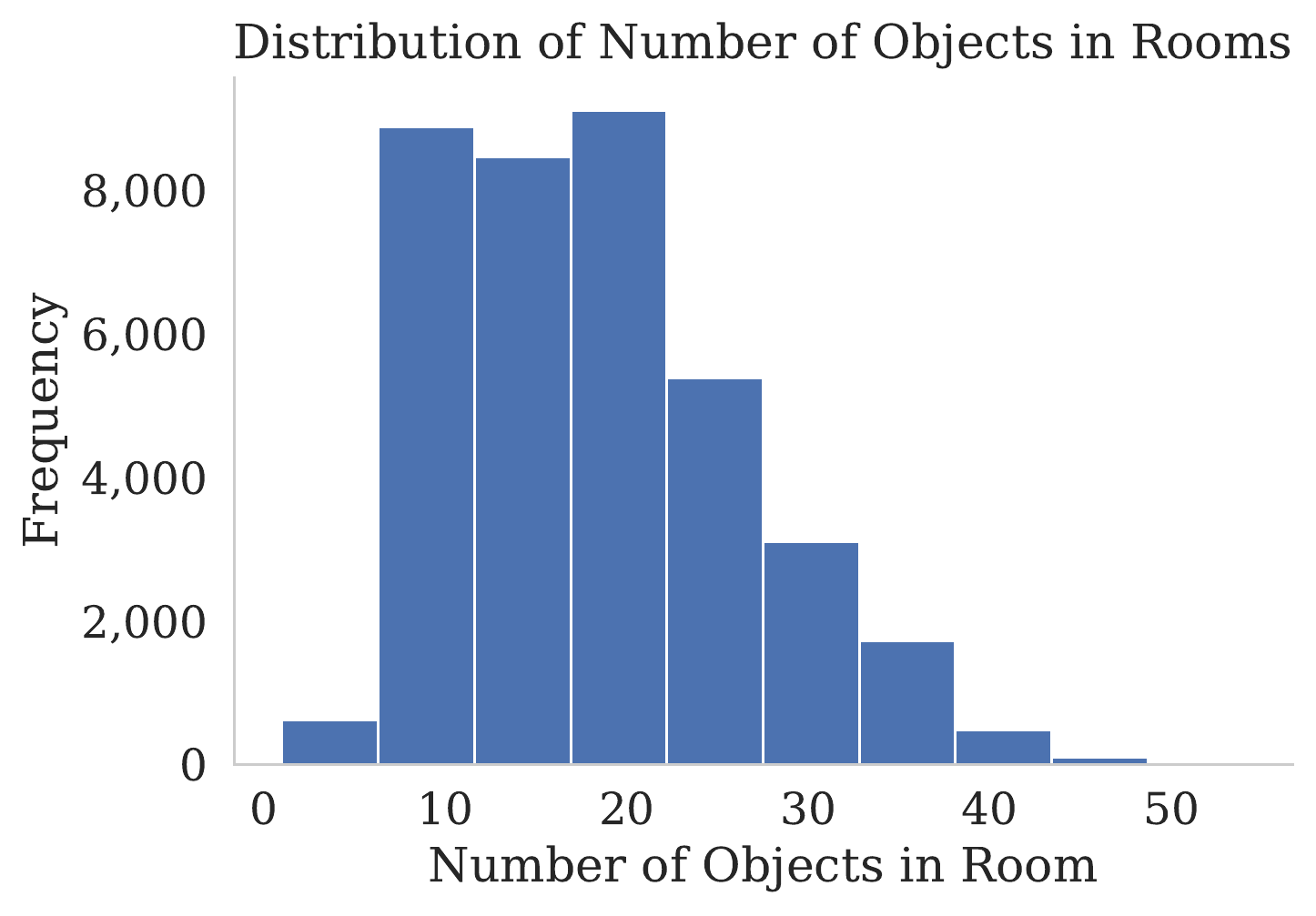}
     \end{subfigure}
     \hfill
     \begin{subfigure}[h]{0.325\textwidth}
         \centering
         \includegraphics[width=\textwidth]{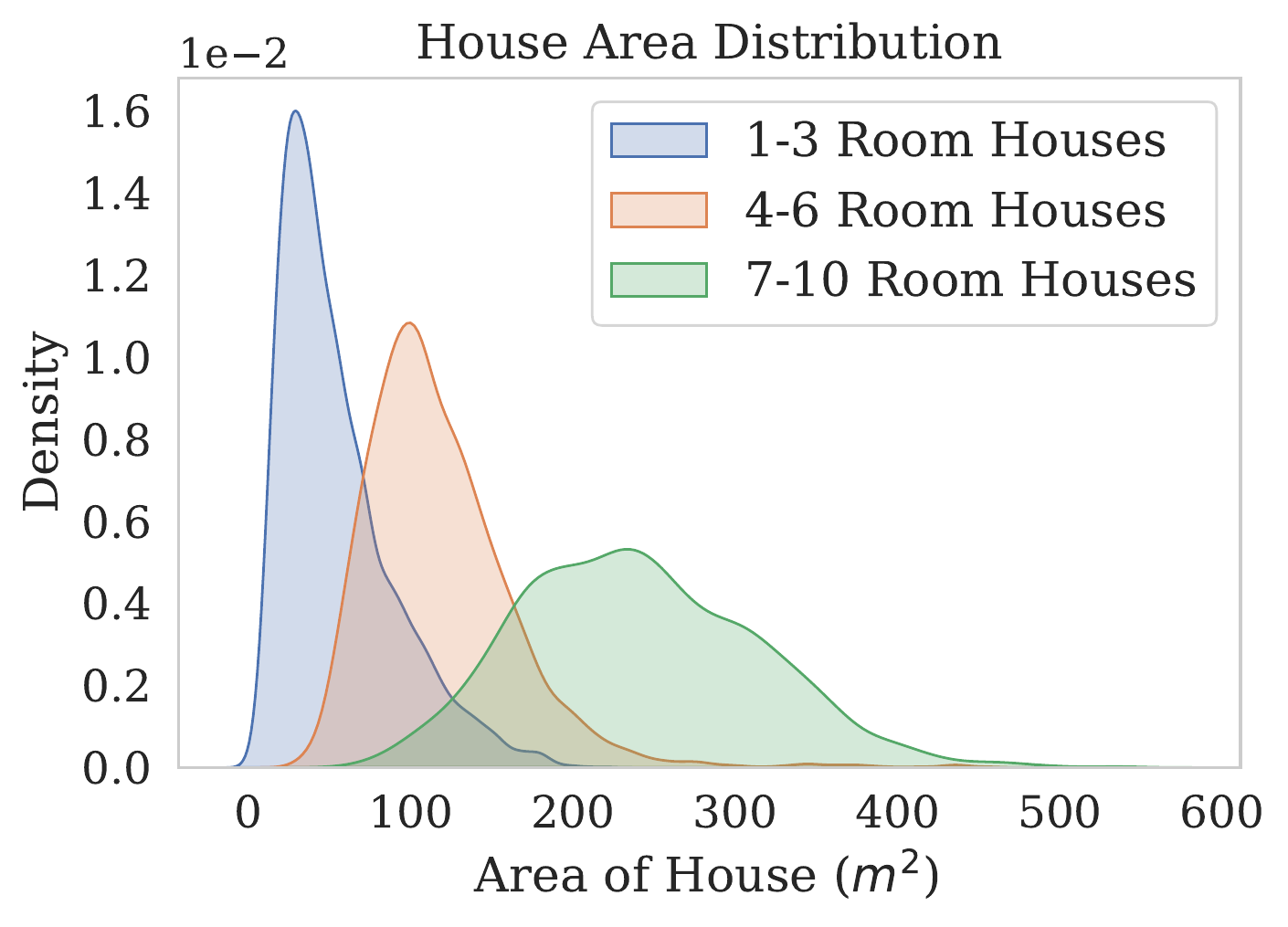}
     \end{subfigure}
     \hfill
     \begin{subfigure}[h]{0.325\textwidth}
         \centering
         \includegraphics[width=\textwidth]{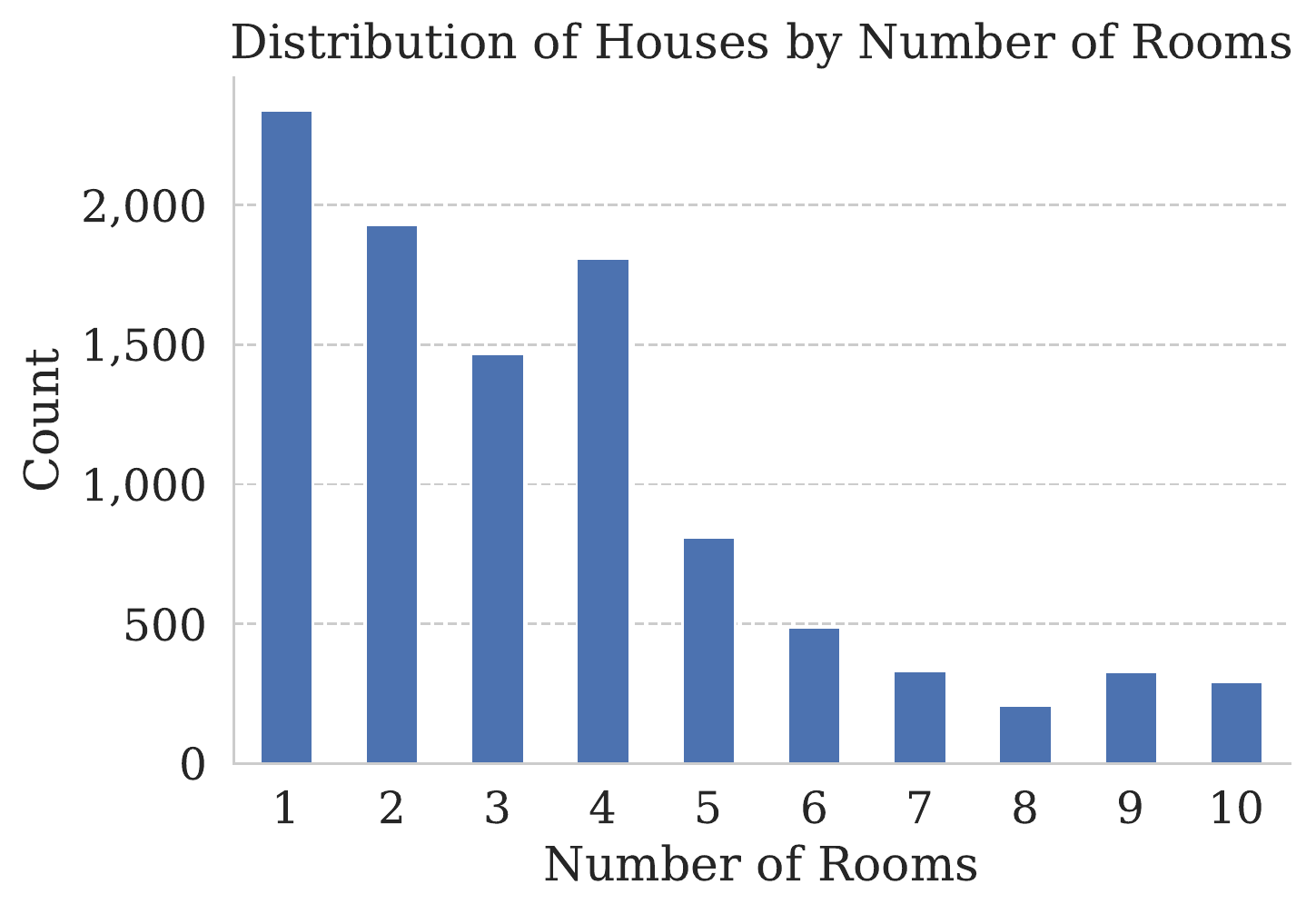}
     \end{subfigure}
    \caption{\textbf{\envb{}-10K statistics.} \emph{Left:} distribution of the number of objects in each room; \emph{Middle:} distribution of the area of each house, bucketed into small, medium, and large houses; \emph{Right:} bar plot showing the distribution over the number of rooms that make up each house.}
    \label{fig:dists}
    \vspace{-0.2in}
\end{figure}

\begin{figure}[b]
    \centering
    \vspace{-0.2in}
    \includegraphics[width=\textwidth]{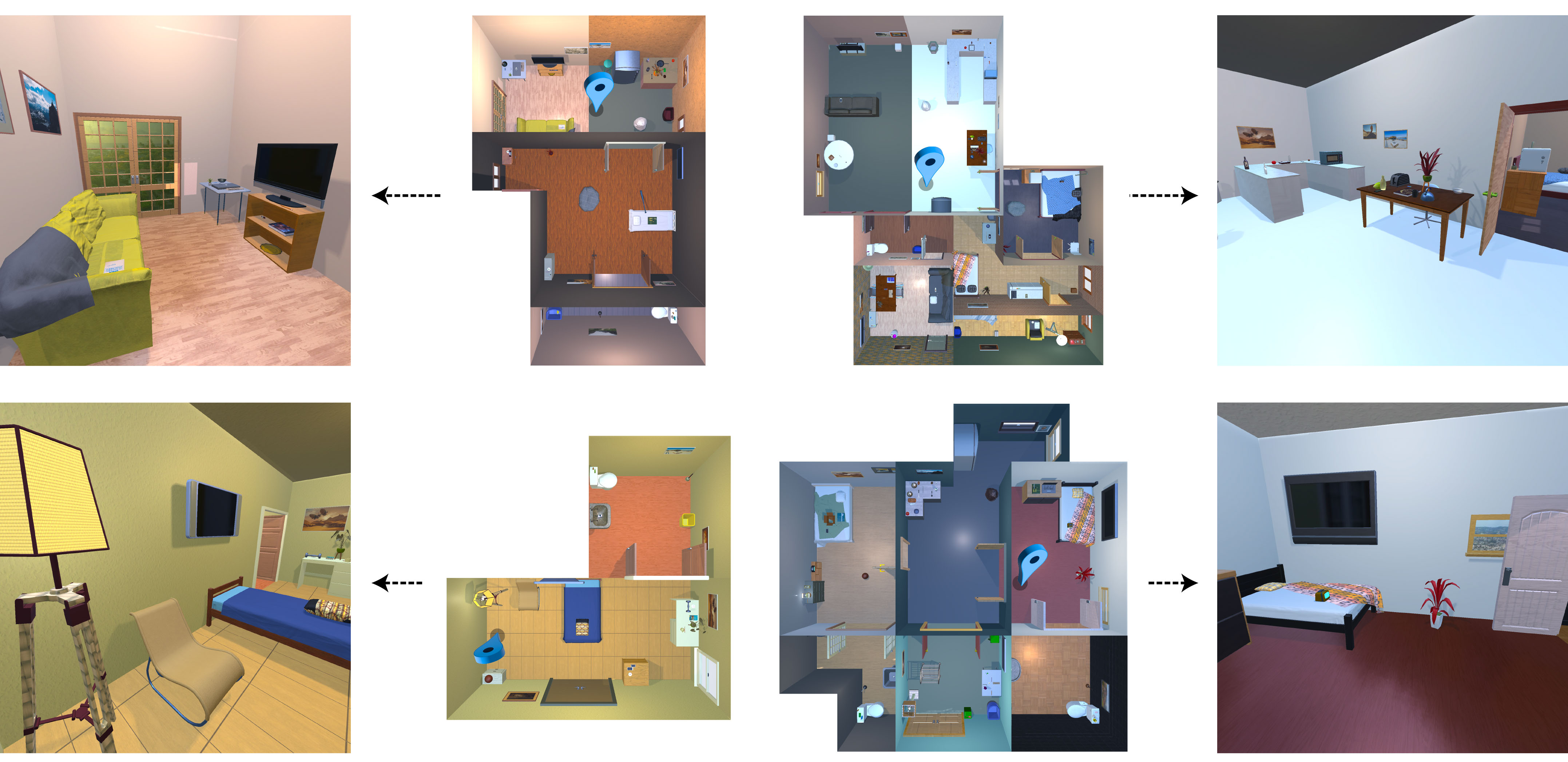}
    \caption{\textbf{Example scenes} in \env{}-10K with top-down and an egocentric view.}
    \label{fig:10k}
\end{figure}

We demonstrate the power and potential of \env using a sampled set of 10,000 fully interactive houses obtained by the procedural generation process described in Section~\ref{sec:generation} -- which we label \env-10K. An additional set of 1,000 validation and 1,000 testing houses are available for evaluation. Asset splits across train/val/test are detailed in the Appendix. All houses are fully navigable, allowing an agent to traverse through each room without any interaction. In terms of scale, \env-10K is one of the largest sets of interactive home environments for Embodied AI -- as a comparison, AI2-iTHOR~\cite{kolve2017ai2} includes 120 scenes, RoboTHOR~\cite{Deitke2020RoboTHORAO} has 89 scenes, iGibson~\cite{Shen2020iGibsonAS} has 15 scenes, Habitat Matterport 3D~\cite{Ramakrishnan2021HabitatMatterport3D} has 1,000 static (non-interactive) scenes, and Habitat 2.0~\cite{szot2021habitat} has 105 scene layouts. Scaling beyond 10K houses is straightforward and inexpensive. This set of 10K houses was generated in 1 hour on a local workstation with 4 NVIDIA RTX A5000 GPUs. Fig.~\ref{fig:10k} shows examples of ego-centric and top-down views of houses present in \env{}-10K.

\noindent \textbf{Scene statistics.} Houses in \env-10K are generated using 16 different room specifications. An example room spec is: \emph{A house with 1 bedroom connected to 1 bathroom, 1 kitchen, and 1 living room} and is visualized in Fig.~\ref{fig:genOverview}. Houses in this dataset have as few as 1 room and as many as 10. 
Fig.~\ref{fig:dists} shows the distribution of areas (middle) and the number of rooms (right) of these generated houses. Our use of room specifications enables us to change the distribution of the size and complexity of houses fairly easily. \env-10K encompasses a wider spectrum of scenes than AI2-iTHOR~\cite{kolve2017ai2} and \textsc{RoboTHOR}~\cite{Deitke2020RoboTHORAO} (biased towards room-sized scenes) and Gibson \cite{xia2018gibson} and HM3D \cite{Ramakrishnan2021HabitatMatterport3D} (biased towards large houses).

Rooms in each of these houses contain objects from 95 different categories including common household objects such as fridges, countertops, beds, toilets, and house plants, and structure objects such as doorways and windows. Fig.~\ref{fig:dists} (left) shows the distribution of the number of objects per room per house, which shows that houses in \env-10K are well populated. They also contain objects sampled via 18 different Semantic Asset groups. Examples of Semantic asset groups (SAG) are a \emph{Dining Table with 4 Chairs} or \emph{Bed with 2 Pillows}. Given our large asset library and SAGs, we can create 19.3 million combinations of group instantiations.

\begin{table}
    \centering
    \small
    \begin{tabular}{l cc c cc c cc}
        \toprule
        & \multicolumn{2}{c}{Navigation FPS} && \multicolumn{2}{c}{Isolated Interaction FPS} && \multicolumn{2}{c}{Environment Query FPS} \\
        \cmidrule{2-3}\cmidrule{5-6}\cmidrule{8-9}
        Compute & Small & Large && Small & Large && Small & Large \\
        \midrule
        8 GPUs & 8,599{\scriptsize$\pm$359} & 3,208{\scriptsize$\pm$127} && 6,488{\scriptsize$\pm$250} & 2,861{\scriptsize$\pm$107} && 480,205{\scriptsize$\pm$19,684} & 433,587{\scriptsize$\pm$18,729} \\
        1 GPU & 1,427{\scriptsize$\pm$74} & 6,280{\scriptsize$\pm$40} && 1,265{\scriptsize$\pm$71}  & 597{\scriptsize$\pm$37} && 160,622{\scriptsize$\pm$2,846} & 157,567{\scriptsize$\pm$2,689}\\
        1 Process & 240{\scriptsize$\pm$69} & 115{\scriptsize$\pm$19} && 180{\scriptsize$\pm$42} & 93{\scriptsize$\pm$15} &&
        14,825{\scriptsize$\pm$199} & 14,916{\scriptsize$\pm$186}\\
        \bottomrule
    \end{tabular}%
    \vspace{0.05in}
    \caption{\textbf{Rendering speed.} Benchmarking FPS for navigation (\eg moving/rotating), interaction (\eg pushing an object), and querying the environment for data (\eg checking the dimensions of the agent). We report FPS for Small and Large houses. See Appendix for details.}
    \label{tab:fps}
    \vspace{-0.2in}
\end{table}

\noindent \textbf{Rendering speed.} A crucial requirement for large-scale training is high rendering speed since the training algorithms require millions of iterations to converge. Table~\ref{tab:fps} shows these statistics. Experiments were run on a server with 8 NVIDIA Quadro RTX 8000 GPUs. For the 1 GPU experiments, we use 15 processes and for the 8 GPU experiments, we use 120 processes, evenly distributed across the GPUs. \env provides framerates comparable to iTHOR and RoboTHOR environments in spite of having larger houses (See Appendix for details), rendering it fast enough for training large models for hundreds of millions of steps in a reasonable amount of time.

\section{Experiments}
\label{sec:experiments}

\noindent \textbf{Tasks.}
We now present results for models pre-trained on \env{}-10K on several navigation and manipulation benchmarks to demonstrate the benefits of large-scale training. We consider ObjectNav (navigation towards a specific object category) in \env{}, \elienv{}, RoboTHOR~\cite{Deitke2020RoboTHORAO}, HM3D~\cite{Ramakrishnan2021HabitatMatterport3D}, and AI2-iTHOR~\cite{kolve2017ai2}. We also consider two manipulation-based tasks: ArmPointNav~\cite{manipulathor} and 1-phase Room Rearrangement~\cite{weihs2021visual}. In ArmPointNav, the agent moves an object using a robotic arm from a source location to a destination location specified in the 3D coordinate frame. In Room Rearrangement, the goal is to move objects or change their state to reach a target scene state.

\noindent \textbf{Models.}
Our models for all tasks consist of a CNN to encode visual information and a GRU to capture temporal information. We deliberately use a simple architecture across all tasks to show the benefits of large-scale training. Our ObjectNav and Rearrangement models use the CLIP-based architectures of \cite{khandelwal2021simple}. Our ArmPointNav model uses a simpler visual encoder with 3 convolutional layers; we found this more effective than the CLIP encoder. All models are trained with the AllenAct~\cite{weihs2020allenact} framework, see the Appendix for training details.

\noindent \textbf{Results.}
\label{sec:results}
We present results in two settings: zero-shot and after fine-tuning on the training scenes provided by the downstream benchmark. Zero-shot experiments show us how well models trained on \env generalize to new environments, whereas fine-tuning experiments tell us if representations learned from \env can serve as a good initialization for quick tuning. For all experiments, we use only RGB images (no depth and other modalities is used). 

Zero-shot is particularly challenging since other environments have different appearance statistics, layouts, and object distributions compared to \env{}. \elienv{} and AI2-iTHOR~\cite{kolve2017ai2} are high-fidelity artist-designed scenes with high-quality shadows and lighting. HM3D is constructed from 3D scans of houses which can differ quite a bit from synthetic environments. RoboTHOR~\cite{Deitke2020RoboTHORAO} houses use wall panels and floors with very specific textures.

\noindent \textbf{\emph{Zero-shot transfer results.}} Models trained only on \env and evaluated zero-shot outperform previous SoTA models on 3 benchmarks (refer to \emph{zero-shot} rows of Table~\ref{tab:allresults}). These are very strong results since the models generalize to not only unseen objects and scenes, but also different appearance and layout statistics.

\noindent \textbf{\emph{Fine-tuning results.}} Further fine-tuning of the model using each benchmark's training data, achieves state-of-the-art results on all benchmarks (refer to \emph{fine-tune} rows of Table~\ref{tab:allresults}). Notably, our model is ranked first on three public leaderboards as of 10am PT, June 14th 2022: Habitat 2022 ObjectNav challenge, AI2-THOR Rearrangement 2022 challenge, and RoboTHOR ObjectNav challenge. It should be noted that our model achieves these results using a very simple architecture and only RGB images. Other techniques typically use more complex architectures that include mapping or visual odometry modules and use additional perception sensors such as depth images.

\noindent \textbf{Scale ablation.} To evaluate the effect of scale we train the models on 10, 100, 1,000, and 10,000 houses. For this experiment, we do not use any material augmentations. As shown in Table~\ref{tab:ablation}, the performance improves as we use more houses for training, demonstrating the benefits of large-scale data for Embodied AI tasks.

\begin{table}[h!]
    \centering
    \resizebox{0.9\textwidth}{!}{%
    \begin{tabular}{ cclcc }
        \toprule
        Task & Benchmark & Method & \multicolumn{2}{c}{Metrics} \\
        
        \midrule
        & & & Success & SPL \\
        \cmidrule{4-5}
        ObjectNav & RoboTHOR Challenge & EmbCLIP~\cite{khandelwal2021simple}$^a$  & 47.0\% & 0.200 \\
        & & \cellcolor{Color4}ProcTHOR 0-shot  & \cellcolor{Color4}55.0\% & \cellcolor{Color4}0.237 \\
        & & ProcTHOR + fine-tune  & \textbf{65.2}\% & \textbf{0.288} \\
        
        \midrule
        & & & Success & SPL \\
        \cmidrule{4-5}
        
        & & MLNLC$^c$  & 52.0\% & 0.280 \\
        & Habitat Challenge & FusionNav (AIRI)$^c$  & 54.0\% & 0.270 \\
        ObjectNav & (2022) & \cellcolor{Color4}ProcTHOR 0-shot & \cellcolor{Color4} 9.00\% & \cellcolor{Color4} 0.055\\
        & \emph{HM3D-Semantics} & ProcTHOR + fine-tune  & 53.0\% & 0.270 \\
        & & \cellcolor{Color4}ProcTHOR + Large$^d$ + 0-shot & \cellcolor{Color4} 13.2\% & \cellcolor{Color4} 0.077\\
        & & ProcTHOR + Large$^d$ + fine-tune  & \textbf{54.4}\% & \textbf{0.318} \\
        
        
        \midrule
        & & & Success & SPL \\
        \cmidrule{4-5}
        ObjectNav & AI2-iTHOR & EmbCLIP~\cite{khandelwal2021simple}$^b$  & 68.4\% & 0.516 \\
        & & \cellcolor{Color4}ProcTHOR 0-shot  & \cellcolor{Color4}75.7\% & \cellcolor{Color4}\textbf{0.644} \\
        & & ProcTHOR + fine-tune  & \textbf{77.5}\% & 0.621 \\
        
        \midrule
        & & & Success & SPL \\
        \cmidrule{4-5}
        ObjectNav & \elienv{} & EmbCLIP~\cite{khandelwal2021simple}$^b$  & 18.5\% & 0.118 \\
        & & ProcTHOR  & \textbf{31.4}\% & \textbf{0.195} \\

        \midrule
        \multirow{4}{*}{Rearrangement} & & & Success & \% Fixed Strict \\
        \cmidrule{4-5}
         & AI2-THOR Challenge & EmbCLIP~\cite{khandelwal2021simple}  & 7.10\% & 0.190 \\
        & \emph{1-phase} (2022) & \cellcolor{Color4}ProcTHOR 0-shot  & \cellcolor{Color4}3.80\% & \cellcolor{Color4}0.156 \\
        & & ProcTHOR + fine-tune  & \textbf{7.40}\% & \textbf{0.245} \\ 
        
        \midrule
        & & & Success & \% PickUp SR \\
        \cmidrule{4-5}
        ArmPointNav & ManipulaTHOR & iTHOR-SimpleConv~\cite{manipulathor}$^e$ & 29.2\% & 73.4 \\
        &  & \cellcolor{Color4}ProcTHOR 0-shot  & \cellcolor{Color4}\textbf{37.9}\% & \cellcolor{Color4}\textbf{74.8} \\
        \bottomrule
    \end{tabular}
    }
    \vspace{0.5em}
    \caption{Results for models trained on ProcTHOR and evaluated 0-shot and with fine-tuning on several \eai benchmarks. For each benchmark we also compare to the relevant baselines (previous SoTA or leaderboard submissions where applicable). $^a$EmbCLIP~\cite{khandelwal2021simple} trained on \textsc{RoboTHOR}, $^b$EmbCLIP~\cite{khandelwal2021simple} trained on AI2-iTHOR, $^c$submission on the Habitat 2022 ObjectNav leaderboard~\cite{habitat-challenge}. $^d$ For HM3D we present results when pretraining using the standard EmbCLIP architecture (which uses a CLIP-pretrained ResNet50 backbone) as well as with a ``Large'' model which uses a larger CLIP backbone CNN as well as a wider RNN, see supplement for details.
    $^e$uses the model from \cite{manipulathor} but retrains on the complete iTHOR data with RGB inputs. \crule[Color4]{5mm}{2.5mm} 0-shot results, whereby models are pre-trained on \env{}-10K and do not use any training data from the benchmark that they are evaluated on.}
    \label{tab:allresults}
\end{table}

\begin{table}[h!]
    \vspace{-0.2in}
    \centering
    \footnotesize
    \resizebox{0.905\columnwidth}{!}{
        \begin{adjustbox}{center}
            \begin{tabular}{ r cc cc cc cc }
                \toprule\\[-0.125in]
                & 
                \multicolumn{2}{c}{\shortstack{\elienv{}\\Test}} & \multicolumn{2}{c}{\shortstack{\textsc{RoboTHOR}\\Test (0-Shot)}} & \multicolumn{2}{c}{\shortstack{\textsc{HM3D}\\Valid (0-Shot)}} & \multicolumn{2}{c}{\shortstack{AI2-iTHOR\\Test (0-Shot)}} \\
                \cmidrule{2-9}
                \textsc{\# Houses} & SPL & SR & SPL & SR & SPL & SR & SPL & SR\\
                \midrule
                10 Houses 
                & 0.077 & 11.3\% 
                & 0.040 & 8.53\% 
                & 0.007 & 1.60\%
                & 0.249 & 28.7\% \\
                100 Houses 
                & 0.102 & 18.6\% 
                & 0.076 & 20.9\% 
                & 0.050 & \textbf{10.4\%}
                & 0.352 & 42.0\% \\
                1,000 Houses 
                & 0.122 & 17.2\% 
                & 0.157 & 33.1\% 
                & 0.027 & 4.65\%
                & 0.456 & 53.0\% \\
                10,000 Houses 
                & \textbf{0.185} & \textbf{27.0\%} 
                & \textbf{0.210} & \textbf{44.5\%} 
                & \textbf{0.060} & 9.70\%
                & \textbf{0.554} & \textbf{64.9\%} \\
                \toprule\\[-0.05in]
            \end{tabular}
        \end{adjustbox}
    }
    \caption{Ablation study to evaluate the effect of the number of training houses. Each model is trained to 80\% success during training. Test performance increases with the number of training houses. }
    \label{tab:ablation}
    \vspace{-0.2in}
\end{table}

\section{Conclusion}
\label{sec:conclusion}

We propose \env{}, a framework to procedurally generate \emph{arbitrarily large} sets of interactive, physics-enabled houses for Embodied AI research. 
We pre-train simple models on 10,000 generated houses and show state-of-the-art results across 6 embodied navigation and manipulation benchmarks with strong 0-shot results, even outperforming prior state-of-the-art on 3 of these benchmarks.



\section*{Acknowledgements}

We would like to thank the teams behind the open-source packages used in this project, including AI2-THOR~\cite{kolve2017ai2}, AllenAct~\cite{weihs2020allenact}, Habitat~\cite{savva2019habitat}, \raisebox{-0.1\height}{\includegraphics[width=0.025\textwidth]{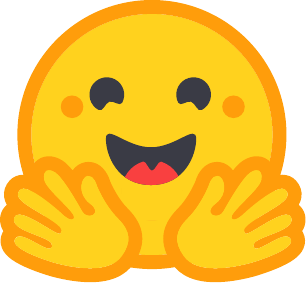}} Datasets~\cite{Lhoest2021DatasetsAC}, NumPy~\cite{harris2020array}, PyTorch~\cite{paszke2017automatic}, Pandas~\cite{mckinney2011pandas}, Wandb~\cite{wandb}, Shapely~\cite{shapely2007}, Hydra~\cite{Yadan2019Hydra}, SciPy~\cite{virtanen2020scipy}, UMAP~\cite{mcinnes2018umap-software}, NetworkX~\cite{hagberg2008exploring}, EvalAI~\cite{EvalAI}, TensorFlow~\cite{abadi2016tensorflow}, OpenAI Gym~\cite{Brockman2016OpenAIG}, Seaborn~\cite{waskom2021seaborn}, PySAT~\cite{imms-sat18}, and Matplotlib~\cite{hunter2007matplotlib}.

{
\small

\bibliographystyle{ieee}
\bibliography{egbib}

\begin{thebibliography}{100}\itemsep=-1pt

\bibitem{abadi2016tensorflow}
Mart{\'i}n Abadi, Paul Barham, Jianmin Chen, Z. Chen, Andy Davis, Jeffrey Dean,
  Matthieu Devin, Sanjay Ghemawat, Geoffrey Irving, Michael Isard, Manjunath
  Kudlur, Josh Levenberg, Rajat Monga, Sherry Moore, Derek~Gordon Murray,
  Benoit Steiner, Paul~A. Tucker, Vijay Vasudevan, Pete Warden, Martin Wicke,
  Yuan Yu, and Xiaoqiang Zhang.
\newblock Tensorflow: A system for large-scale machine learning.
\newblock In {\em OSDI}, 2016.

\bibitem{team2021open}
Adam, Anuj Mahajan, Catarina Barros, Charlie Deck, Jakob Bauer, Jakub
  Sygnowski, Maja Trebacz, Max Jaderberg, Michael Mathieu, Nat McAleese,
  Nathalie Bradley-Schmieg, Nathaniel Wong, Nicolas Porcel, Roberta Raileanu,
  Steph Hughes-Fitt, Valentin Dalibard, and Wojciech~Marian Czarnecki.
\newblock Open-ended learning leads to generally capable agents.
\newblock {\em arXiv}, 2021.

\bibitem{flamingo}
Jean-Baptiste Alayrac, Jeff Donahue, Pauline Luc, Antoine Miech, Iain Barr,
  Yana Hasson, Karel Lenc, Arthur Mensch, Katie Millican, Malcolm Reynolds,
  Roman Ring, Eliza Rutherford, Serkan Cabi, Tengda Han, Zhitao Gong, Sina
  Samangooei, Marianne Monteiro, Jacob Menick, Sebastian Borgeaud, Andrew
  Brock, Aida Nematzadeh, Sahand Sharifzadeh, Mikolaj Binkowski, Ricardo
  Barreira, Oriol Vinyals, Andrew Zisserman, and Karen Simonyan.
\newblock Flamingo: a visual language model for few-shot learning.
\newblock {\em arXiv}, 2022.

\bibitem{roomr-challenge}
{Allen Institute for AI}.
\newblock {Rearrangement Challenge 2022}.
\newblock
  \url{https://leaderboard.allenai.org/ithor_rearrangement_1phase_2022}.

\bibitem{robothor-challenge}
{Allen Institute for AI}.
\newblock {RoboTHOR ObjectNav Challenge}.
\newblock \url{https://github.com/allenai/robothor-challenge}.

\bibitem{Anderson2018OnEO}
Peter Anderson, Angel~X. Chang, Devendra~Singh Chaplot, Alexey Dosovitskiy,
  Saurabh Gupta, Vladlen Koltun, Jana Kosecka, Jitendra Malik, Roozbeh
  Mottaghi, Manolis Savva, and Amir~Roshan Zamir.
\newblock On evaluation of embodied navigation agents.
\newblock {\em arXiv}, 2018.

\bibitem{batra2020objectnav}
Dhruv Batra, Aaron Gokaslan, Aniruddha Kembhavi, Oleksandr Maksymets, Roozbeh
  Mottaghi, Manolis Savva, Alexander Toshev, and Erik Wijmans.
\newblock Objectnav revisited: On evaluation of embodied agents navigating to
  objects.
\newblock {\em arXiv}, 2020.

\bibitem{wandb}
Lukas Biewald.
\newblock Experiment tracking with weights and biases, 2020.
\newblock Software available from wandb.com.

\bibitem{Brockman2016OpenAIG}
Greg Brockman, Vicki Cheung, Ludwig Pettersson, Jonas Schneider, John Schulman,
  Jie Tang, and Wojciech Zaremba.
\newblock Openai gym.
\newblock {\em arXiv}, 2016.

\bibitem{Brown2020LanguageMA}
Tom~B. Brown, Benjamin Mann, Nick Ryder, Melanie Subbiah, Jared Kaplan,
  Prafulla Dhariwal, Arvind Neelakantan, Pranav Shyam, Girish Sastry, Amanda
  Askell, Sandhini Agarwal, Ariel Herbert-Voss, Gretchen Krueger, T.~J.
  Henighan, Rewon Child, Aditya Ramesh, Daniel~M. Ziegler, Jeff Wu, Clemens
  Winter, Christopher Hesse, Mark Chen, Eric Sigler, Mateusz Litwin, Scott
  Gray, Benjamin Chess, Jack Clark, Christopher Berner, Sam McCandlish, Alec
  Radford, Ilya Sutskever, and Dario Amodei.
\newblock Language models are few-shot learners.
\newblock In {\em NeurIPS}, 2020.

\bibitem{nuscenes}
Holger Caesar, Varun Bankiti, Alex~H. Lang, Sourabh Vora, Venice~Erin Liong,
  Qiang Xu, Anush Krishnan, Yu Pan, Giancarlo Baldan, and Oscar Beijbom.
\newblock nuscenes: A multimodal dataset for autonomous driving.
\newblock In {\em CVPR}, 2020.

\bibitem{locobot}
{Carnegie Mellon University}.
\newblock Locobot: an open source low cost robot.
\newblock \url{http://www.locobot.org/}.

\bibitem{chang2014interactive}
Angel Chang, Manolis Savva, and Christopher~D Manning.
\newblock Interactive learning of spatial knowledge for text to 3d scene
  generation.
\newblock In {\em ACL Workshop on Interactive Language Learning, Visualization,
  and Interfaces}, 2014.

\bibitem{shapenet2015}
Angel~X. Chang, Thomas Funkhouser, Leonidas Guibas, Pat Hanrahan, Qixing Huang,
  Zimo Li, Silvio Savarese, Manolis Savva, Shuran Song, Hao Su, Jianxiong Xiao,
  Li Yi, and Fisher Yu.
\newblock {ShapeNet: An Information-Rich 3D Model Repository}.
\newblock {\em arXiv}, 2015.

\bibitem{chang2015shapenet}
Angel~X. Chang, Thomas~A. Funkhouser, Leonidas~J. Guibas, Pat Hanrahan,
  Qi{-}Xing Huang, Zimo Li, Silvio Savarese, Manolis Savva, Shuran Song, Hao
  Su, Jianxiong Xiao, Li Yi, and Fisher Yu.
\newblock Shapenet: An information-rich 3d model repository.
\newblock {\em arXiv}, 2015.

\bibitem{Chang2015TextT3}
Angel~X. Chang, Will Monroe, Manolis Savva, Christopher Potts, and
  Christopher~D. Manning.
\newblock Text to 3d scene generation with rich lexical grounding.
\newblock In {\em ACL}, 2015.

\bibitem{Chang2014LearningSK}
Angel~X. Chang, Manolis Savva, and Christopher~D. Manning.
\newblock Learning spatial knowledge for text to 3d scene generation.
\newblock In {\em EMNLP}, 2014.

\bibitem{Changpinyo2021Conceptual1P}
Soravit Changpinyo, Piyush~Kumar Sharma, Nan Ding, and Radu Soricut.
\newblock Conceptual 12m: Pushing web-scale image-text pre-training to
  recognize long-tail visual concepts.
\newblock In {\em CVPR}, 2021.

\bibitem{Charco2021CameraPE}
Jorge~L. Charco, Angel~Domingo Sappa, Boris~Xavier Vintimilla, and Henry~O.
  Velesaca.
\newblock Camera pose estimation in multi-view environments: From virtual
  scenarios to the real world.
\newblock {\em Image Vis. Comput.}, 2021.

\bibitem{Chaudhuri2019LearningGM}
Siddhartha Chaudhuri, Daniel Ritchie, Kai Xu, and Hao Zhang.
\newblock Learning generative models of 3d structures.
\newblock In {\em Eurographics}, 2019.

\bibitem{chen2021savi}
Changan Chen, Ziad Al-Halah, and Kristen Grauman.
\newblock Semantic audio-visual navigation.
\newblock In {\em CVPR}, 2021.

\bibitem{chen2020soundspaces}
Changan Chen, Unnat Jain, Carl Schissler, Sebastia Vicenc~Amengual Gari, Ziad
  Al-Halah, Vamsi~Krishna Ithapu, Philip Robinson, and Kristen Grauman.
\newblock Soundspaces: Audio-visual navigation in 3d environments.
\newblock In {\em ECCV}, 2020.

\bibitem{Chitnis2021LearningNe}
Rohan Chitnis, Tom Silver, Joshua~B. Tenenbaum, Tomas Lozano-Perez, and
  Leslie~Pack Kaelbling.
\newblock {Learning Neuro-Symbolic Relational Transition Models for Bilevel
  Planning}.
\newblock {\em arXiv}, 2021.

\bibitem{cho2014learning}
Kyunghyun Cho, Bart Van~Merri{\"e}nboer, Caglar Gulcehre, Dzmitry Bahdanau,
  Fethi Bougares, Holger Schwenk, and Yoshua Bengio.
\newblock Learning phrase representations using rnn encoder-decoder for
  statistical machine translation.
\newblock In {\em EMNLP}, 2014.

\bibitem{chung2014empirical}
Junyoung Chung, Caglar Gulcehre, KyungHyun Cho, and Yoshua Bengio.
\newblock Empirical evaluation of gated recurrent neural networks on sequence
  modeling.
\newblock {\em arXiv}, 2014.

\bibitem{collins2021abo}
Jasmine Collins, Shubham Goel, Kenan Deng, Achleshwar Luthra, Leon Xu, Erhan
  Gundogdu, Xi Zhang, Tomas~F Yago~Vicente, Thomas Dideriksen, Himanshu Arora,
  Matthieu Guillaumin, and Jitendra Malik.
\newblock Abo: Dataset and benchmarks for real-world 3d object understanding.
\newblock In {\em CVPR}, 2022.

\bibitem{Deitke2020RoboTHORAO}
Matt Deitke, Winson Han, Alvaro Herrasti, Aniruddha Kembhavi, Eric Kolve,
  Roozbeh Mottaghi, Jordi Salvador, Dustin Schwenk, Eli VanderBilt, Matthew
  Wallingford, Luca Weihs, Mark Yatskar, and Ali Farhadi.
\newblock Robothor: An open simulation-to-real embodied ai platform.
\newblock In {\em CVPR}, 2020.

\bibitem{imagenet}
Jia Deng, Wei Dong, Richard Socher, Li-Jia Li, Kai Li, and Li Fei-Fei.
\newblock Imagenet: A large-scale hierarchical image database.
\newblock In {\em CVPR}, 2009.

\bibitem{dennis2020emergent}
Michael Dennis, Natasha Jaques, Eugene Vinitsky, Alexandre Bayen, Stuart
  Russell, Andrew Critch, and Sergey Levine.
\newblock Emergent complexity and zero-shot transfer via unsupervised
  environment design.
\newblock In {\em NeurIPS}, 2020.

\bibitem{downs2022google}
Laura Downs, Anthony Francis, Nate Koenig, Brandon Kinman, Ryan Hickman, Krista
  Reymann, Thomas~B McHugh, and Vincent Vanhoucke.
\newblock Google scanned objects: A high-quality dataset of 3d scanned
  household items.
\newblock In {\em ICRA}, 2022.

\bibitem{earle2021learning}
Sam Earle, Maria Edwards, Ahmed Khalifa, Philip Bontrager, and Julian Togelius.
\newblock Learning controllable content generators.
\newblock In {\em CoG}, 2021.

\bibitem{ehsani2022object}
Kiana Ehsani, Ali Farhadi, Aniruddha Kembhavi, and Roozbeh Mottaghi.
\newblock Object manipulation via visual target localization.
\newblock {\em arXiv}, 2022.

\bibitem{manipulathor}
Kiana Ehsani, Winson Han, Alvaro Herrasti, Eli VanderBilt, Luca Weihs, Eric
  Kolve, Aniruddha Kembhavi, and Roozbeh Mottaghi.
\newblock {ManipulaTHOR: A Framework for Visual Object Manipulation}.
\newblock In {\em CVPR}, 2021.

\bibitem{ehsani2018segan}
Kiana Ehsani, Roozbeh Mottaghi, and Ali Farhadi.
\newblock Segan: Segmenting and generating the invisible.
\newblock In {\em CVPR}, 2018.

\bibitem{Feng2021DeepMO}
Di Feng, Christian Haase-Schuetz, Lars Rosenbaum, Heinz Hertlein, Fabian
  Duffhauss, Claudius Gl{\"a}ser, Werner Wiesbeck, and Klaus C.~J. Dietmayer.
\newblock Deep multi-modal object detection and semantic segmentation for
  autonomous driving: Datasets, methods, and challenges.
\newblock {\em IEEE Trans. on Intelligent Transportation Systems}, 2021.

\bibitem{gadre2022continuous}
Samir~Yitzhak Gadre, Kiana Ehsani, Shuran Song, and Roozbeh Mottaghi.
\newblock Continuous scene representations for embodied ai.
\newblock In {\em CVPR}, 2022.

\bibitem{Gan2020ThreeDWorldAP}
Chuang Gan, Jeremy Schwartz, Seth Alter, Martin Schrimpf, James Traer,
  Julian~De Freitas, Jonas Kubilius, Abhishek Bhandwaldar, Nick Haber, Megumi
  Sano, Kuno Kim, Elias Wang, Damian Mrowca, Michael Lingelbach, Aidan Curtis,
  Kevin~T. Feigelis, Daniel Bear, Dan Gutfreund, David Cox, James~J. DiCarlo,
  Josh~H. McDermott, Joshua~B. Tenenbaum, and Daniel L.~K. Yamins.
\newblock Threedworld: A platform for interactive multi-modal physical
  simulation.
\newblock In {\em NeurIPS (dataset track)}, 2021.

\bibitem{gan2020look}
Chuang Gan, Yiwei Zhang, Jiajun Wu, Boqing Gong, and Joshua~B Tenenbaum.
\newblock Look, listen, and act: Towards audio-visual embodied navigation.
\newblock In {\em ICRA}, 2020.

\bibitem{gan2021threedworld}
Chuang Gan, Siyuan Zhou, Jeremy Schwartz, Seth Alter, Abhishek Bhandwaldar, Dan
  Gutfreund, Daniel~LK Yamins, James~J DiCarlo, Josh McDermott, Antonio
  Torralba, et~al.
\newblock The threedworld transport challenge: A visually guided
  task-and-motion planning benchmark for physically realistic embodied ai.
\newblock {\em arXiv}, 2021.

\bibitem{Gebru2021DatasheetsFD}
Timnit Gebru, Jamie~H. Morgenstern, Briana Vecchione, Jennifer~Wortman Vaughan,
  Hanna~M. Wallach, Hal Daum{\'e}, and Kate Crawford.
\newblock Datasheets for datasets.
\newblock {\em Comm. of the ACM}, 2021.

\bibitem{shapely2007}
Sean Gillies et~al.
\newblock Shapely: manipulation and analysis of geometric objects, 2007.

\bibitem{gordon2018iqa}
Daniel Gordon, Aniruddha Kembhavi, Mohammad Rastegari, Joseph Redmon, Dieter
  Fox, and Ali Farhadi.
\newblock Iqa: Visual question answering in interactive environments.
\newblock In {\em CVPR}, 2018.

\bibitem{greff2022kubric}
Klaus Greff, Francois Belletti, Lucas Beyer, Carl Doersch, Yilun Du, Daniel
  Duckworth, David~J Fleet, Dan Gnanapragasam, Florian Golemo, Charles
  Herrmann, Thomas Kipf, Abhijit Kundu, Dmitry Lagun, Issam Laradji,
  Hsueh-Ti~(Derek) Liu, Henning Meyer, Yishu Miao, Derek Nowrouzezahrai, Cengiz
  Oztireli, Etienne Pot, Noha Radwan, Daniel Rebain, Sara Sabour, Mehdi S.~M.
  Sajjadi, Matan Sela, Vincent Sitzmann, Austin Stone, Deqing Sun, Suhani Vora,
  Ziyu Wang, Tianhao Wu, Kwang~Moo Yi, Fangcheng Zhong, and Andrea
  Tagliasacchi.
\newblock Kubric: a scalable dataset generator.
\newblock In {\em CVPR}, 2022.

\bibitem{hagberg2008exploring}
Aric Hagberg, Pieter Swart, and Daniel S~Chult.
\newblock Exploring network structure, dynamics, and function using networkx.
\newblock Technical report, Los Alamos National Lab, 2008.

\bibitem{harris2020array}
Charles~R. Harris, K.~Jarrod Millman, St{\'{e}}fan~J. van~der Walt, Ralf
  Gommers, Pauli Virtanen, David Cournapeau, Eric Wieser, Julian Taylor,
  Sebastian Berg, Nathaniel~J. Smith, Robert Kern, Matti Picus, Stephan Hoyer,
  Marten~H. van Kerkwijk, Matthew Brett, Allan Haldane, Jaime~Fern{\'{a}}ndez
  del R{\'{\i}}o, Mark Wiebe, Pearu Peterson, Pierre G{\'{e}}rard-Marchant,
  Kevin Sheppard, Tyler Reddy, Warren Weckesser, Hameer Abbasi, Christoph
  Gohlke, and Travis~E. Oliphant.
\newblock Array programming with numpy.
\newblock {\em Nature}, 2020.

\bibitem{higgins2022head}
Padraig Higgins, Ryan Barron, and Cynthia Matuszek.
\newblock Head pose as a proxy for gaze in virtual reality.
\newblock In {\em Workshop on Virtual, Augmented, and Mixed Reality for HRI},
  2022.

\bibitem{Hu2020Graph2PlanLF}
Ruizhen Hu, Zeyu Huang, Yuhan Tang, Oliver~Matias van Kaick, Hao Zhang, and Hui
  Huang.
\newblock Graph2plan: Learning floorplan generation from layout graphs.
\newblock {\em ACM Trans. on Graphics}, 2020.

\bibitem{huang2022language}
Wenlong Huang, Pieter Abbeel, Deepak Pathak, and Igor Mordatch.
\newblock Language models as zero-shot planners: Extracting actionable
  knowledge for embodied agents.
\newblock {\em arXiv}, 2022.

\bibitem{hunter2007matplotlib}
John~D Hunter.
\newblock Matplotlib: A 2d graphics environment.
\newblock {\em Computing in science \& engineering}, 2007.

\bibitem{imms-sat18}
Alexey Ignatiev, Antonio Morgado, and Joao Marques{-}Silva.
\newblock {PySAT:} {A} {Python} toolkit for prototyping with {SAT} oracles.
\newblock In {\em SAT}, pages 428--437, 2018.

\bibitem{jain2020cordial}
Unnat Jain, Luca Weihs, Eric Kolve, Ali Farhadi, Svetlana Lazebnik, Aniruddha
  Kembhavi, and Alexander Schwing.
\newblock A cordial sync: Going beyond marginal policies for multi-agent
  embodied tasks.
\newblock In {\em ECCV}, 2020.

\bibitem{jain2019two}
Unnat Jain, Luca Weihs, Eric Kolve, Mohammad Rastegari, Svetlana Lazebnik, Ali
  Farhadi, Alexander~G Schwing, and Aniruddha Kembhavi.
\newblock Two body problem: Collaborative visual task completion.
\newblock In {\em CVPR}, 2019.

\bibitem{james2020rlbench}
Stephen James, Zicong Ma, David~Rovick Arrojo, and Andrew~J Davison.
\newblock Rlbench: The robot learning benchmark \& learning environment.
\newblock {\em IEEE Robotics and Automation Letters}, 2020.

\bibitem{kadian2020sim2real}
Abhishek Kadian, Joanne Truong, Aaron Gokaslan, Alexander Clegg, Erik Wijmans,
  Stefan Lee, Manolis Savva, Sonia Chernova, and Dhruv Batra.
\newblock Sim2real predictivity: Does evaluation in simulation predict
  real-world performance?
\newblock {\em IEEE Robotics and Automation Letters}, 2020.

\bibitem{karamcheti2020learning}
Siddharth Karamcheti, Dorsa Sadigh, and Percy Liang.
\newblock Learning adaptive language interfaces through decomposition.
\newblock {\em arXiv}, 2020.

\bibitem{keshavarzi2020scenegen}
Mohammad Keshavarzi, Aakash Parikh, Xiyu Zhai, Melody Mao, Luisa Caldas, and
  Allen~Y Yang.
\newblock Scenegen: Generative contextual scene augmentation using scene graph
  priors.
\newblock {\em arXiv}, 2020.

\bibitem{khalifa2020pcgrl}
Ahmed Khalifa, Philip Bontrager, Sam Earle, and Julian Togelius.
\newblock Pcgrl: Procedural content generation via reinforcement learning.
\newblock In {\em AIIDE}, 2020.

\bibitem{khandelwal2021simple}
Apoorv Khandelwal, Luca Weihs, Roozbeh Mottaghi, and Aniruddha Kembhavi.
\newblock Simple but effective: Clip embeddings for embodied ai.
\newblock In {\em CVPR}, 2021.

\bibitem{kim2020learning}
Seung~Wook Kim, Yuhao Zhou, Jonah Philion, Antonio Torralba, and Sanja Fidler.
\newblock Learning to simulate dynamic environments with gamegan.
\newblock In {\em CVPR}, 2020.

\bibitem{kingma2014adam}
Diederik~P Kingma and Jimmy Ba.
\newblock Adam: A method for stochastic optimization.
\newblock {\em arXiv}, 2014.

\bibitem{koh2022simple}
Jing~Yu Koh, Harsh Agrawal, Dhruv Batra, Richard Tucker, Austin Waters, Honglak
  Lee, Yinfei Yang, Jason Baldridge, and Peter Anderson.
\newblock Simple and effective synthesis of indoor 3d scenes.
\newblock {\em arXiv}, 2022.

\bibitem{koh2021pathdreamer}
Jing~Yu Koh, Honglak Lee, Yinfei Yang, Jason Baldridge, and Peter Anderson.
\newblock Pathdreamer: A world model for indoor navigation.
\newblock In {\em ICCV}, 2021.

\bibitem{kolve2017ai2}
Eric Kolve, Roozbeh Mottaghi, Winson Han, Eli VanderBilt, Luca Weihs, Alvaro
  Herrasti, Daniel Gordon, Yuke Zhu, Abhinav Gupta, and Ali Farhadi.
\newblock Ai2-thor: An interactive 3d environment for visual ai.
\newblock {\em arXiv}, 2017.

\bibitem{kotar2022interactron}
Klemen Kotar and Roozbeh Mottaghi.
\newblock Interactron: Embodied adaptive object detection.
\newblock In {\em CVPR}, 2022.

\bibitem{krantz2020beyond}
Jacob Krantz, Erik Wijmans, Arjun Majumdar, Dhruv Batra, and Stefan Lee.
\newblock Beyond the nav-graph: Vision-and-language navigation in continuous
  environments.
\newblock In {\em ECCV}, 2020.

\bibitem{kumar2021rma}
Ashish Kumar, Zipeng Fu, Deepak Pathak, and Jitendra Malik.
\newblock Rma: Rapid motor adaptation for legged robots.
\newblock In {\em RSS}, 2021.

\bibitem{Kuznetsova2020TheOI}
Alina Kuznetsova, Hassan Rom, Neil~Gordon Alldrin, Jasper R.~R. Uijlings, Ivan
  Krasin, Jordi Pont-Tuset, Shahab Kamali, Stefan Popov, Matteo Malloci,
  Alexander Kolesnikov, Tom Duerig, and Vittorio Ferrari.
\newblock The open images dataset v4.
\newblock {\em IJCV}, 2020.

\bibitem{Lhoest2021DatasetsAC}
Quentin Lhoest, Albert~Villanova del Moral, Yacine Jernite, Abhishek Thakur,
  Patrick von Platen, Suraj Patil, Julien Chaumond, Mariama Drame, Julien Plu,
  Lewis Tunstall, Joe Davison, Mario vSavsko, Gunjan Chhablani, Bhavitvya
  Malik, Simon Brandeis, Teven~Le Scao, Victor Sanh, Canwen Xu, Nicolas Patry,
  Angelina McMillan-Major, Philipp Schmid, Sylvain Gugger, Clement Delangue,
  Th'eo Matussiere, Lysandre Debut, Stas Bekman, Pierric Cistac, Thibault
  Goehringer, Victor Mustar, Franccois Lagunas, Alexander~M. Rush, and Thomas
  Wolf.
\newblock Datasets: A community library for natural language processing.
\newblock {\em arXiv}, 2021.

\bibitem{Li2021iGibson2O}
Chengshu Li, Fei Xia, Roberto Mart'in-Mart'in, Michael Lingelbach, Sanjana
  Srivastava, Bokui Shen, Kent Vainio, Cem Gokmen, Gokul Dharan, Tanish Jain,
  Andrey Kurenkov, Karen Liu, Hyowon Gweon, Jiajun Wu, Li Fei-Fei, and Silvio
  Savarese.
\newblock igibson 2.0: Object-centric simulation for robot learning of everyday
  household tasks.
\newblock In {\em CoRL}, 2021.

\bibitem{li2019grains}
Manyi Li, Akshay~Gadi Patil, Kai Xu, Siddhartha Chaudhuri, Owais Khan, Ariel
  Shamir, Changhe Tu, Baoquan Chen, Daniel Cohen-Or, and Hao Zhang.
\newblock Grains: Generative recursive autoencoders for indoor scenes.
\newblock {\em ACM Trans. on Graphics}, 2019.

\bibitem{li20223d}
Yunzhu Li, Shuang Li, Vincent Sitzmann, Pulkit Agrawal, and Antonio Torralba.
\newblock 3d neural scene representations for visuomotor control.
\newblock In {\em CoRL}, 2022.

\bibitem{li2021openrooms}
Zhengqin Li, Ting Yu, Shen Sang, Sarah Wang, Mengcheng Song, Yuhan Liu, Yu-Ying
  Yeh, Rui Zhu, Nitesh~B. Gundavarapu, Jia Shi, Sai Bi, Hong-Xing Yu, Zexiang
  Xu, Kalyan Sunkavalli, Milos Hasan, Ravi Ramamoorthi, and Manmohan
  Chandraker.
\newblock Openrooms: An open framework for photorealistic indoor scene
  datasets.
\newblock In {\em CVPR}, 2021.

\bibitem{lopes2010constrained}
Ricardo Lopes, Tim Tutenel, Ruben~M Smelik, Klaas~Jan De~Kraker, and Rafael
  Bidarra.
\newblock A constrained growth method for procedural floor plan generation.
\newblock In {\em Game-ON}, 2010.

\bibitem{luo2022stubborn}
Haokuan Luo, Albert Yue, Zhang-Wei Hong, and Pulkit Agrawal.
\newblock Stubborn: A strong baseline for indoor object navigation.
\newblock {\em arXiv}, 2022.

\bibitem{marson2010automatic}
Fernando Marson and Soraia~Raupp Musse.
\newblock Automatic real-time generation of floor plans based on squarified
  treemaps algorithm.
\newblock {\em International Journal of Computer Games Technology}, 2010.

\bibitem{mata2022standardsim}
Cristina Mata, Nick Locascio, Mohammed~Azeem Sheikh, Kenny Kihara, and Dan
  Fischetti.
\newblock Standardsim: A synthetic dataset for retail environments.
\newblock In {\em ICIAP}, 2022.

\bibitem{mcinnes2018umap-software}
Leland McInnes, John Healy, Nathaniel Saul, and Lukas Grossberger.
\newblock Umap: Uniform manifold approximation and projection.
\newblock {\em The Journal of Open Source Software}, 2018.

\bibitem{mckinney2011pandas}
Wes McKinney et~al.
\newblock pandas: a foundational python library for data analysis and
  statistics.
\newblock {\em Python for high performance and scientific computing}, 2011.

\bibitem{habitat-challenge}
{Meta AI}.
\newblock {Habitat ObjectNav Challenge 2022}.
\newblock \url{https://aihabitat.org/challenge/2022}.

\bibitem{mildenhall2020nerf}
Ben Mildenhall, Pratul~P Srinivasan, Matthew Tancik, Jonathan~T Barron, Ravi
  Ramamoorthi, and Ren Ng.
\newblock Nerf: Representing scenes as neural radiance fields for view
  synthesis.
\newblock In {\em ECCV}, 2020.

\bibitem{Mo_2019_CVPR}
Kaichun Mo, Shilin Zhu, Angel~X. Chang, Li Yi, Subarna Tripathi, Leonidas~J.
  Guibas, and Hao Su.
\newblock {PartNet}: A large-scale benchmark for fine-grained and hierarchical
  part-level {3D} object understanding.
\newblock In {\em CVPR}, 2019.

\bibitem{mu2021maniskill}
Tongzhou Mu, Zhan Ling, Fanbo Xiang, Derek Yang, Xuanlin Li, Stone Tao, Zhiao
  Huang, Zhiwei Jia, and Hao Su.
\newblock {M}ani{S}kill: {G}eneralizable {M}anipulation {S}kill {B}enchmark
  with {L}arge-{S}cale {D}emonstrations.
\newblock In {\em NeurIPS (dataset track)}, 2021.

\bibitem{murnane2021simulator}
Mark Murnane, Padraig Higgins, Monali Saraf, Francis Ferraro, Cynthia Matuszek,
  and Don Engel.
\newblock A simulator for human-robot interaction in virtual reality.
\newblock In {\em VRW}, 2021.

\bibitem{narang2022factory}
Yashraj Narang, Kier Storey, Iretiayo Akinola, Miles Macklin, Philipp Reist,
  Lukasz Wawrzyniak, Yunrong Guo, Adam Moravanszky, Gavriel State, Michelle Lu,
  Ankur Handa, and Dieter Fox.
\newblock Factory: Fast contact for robotic assembly.
\newblock In {\em RSS}, 2022.

\bibitem{Nauata2021HouseGANGA}
Nelson Nauata, Sepidehsadat Hosseini, Kai-Hung Chang, Hang Chu, Chin-Yi Cheng,
  and Yasutaka Furukawa.
\newblock House-gan++: Generative adversarial layout refinement network towards
  intelligent computational agent for professional architects.
\newblock In {\em CVPR}, 2021.

\bibitem{ni2021towards}
Tianwei Ni, Kiana Ehsani, Luca Weihs, and Jordi Salvador.
\newblock Towards disturbance-free visual mobile manipulation.
\newblock {\em arXiv}, 2021.

\bibitem{padmakumar2021teach}
Aishwarya Padmakumar, Jesse Thomason, Ayush Shrivastava, Patrick Lange, Anjali
  Narayan-Chen, Spandana Gella, Robinson Piramuthu, Gokhan Tur, and Dilek
  Hakkani-Tur.
\newblock Teach: Task-driven embodied agents that chat.
\newblock In {\em AAAI}, 2022.

\bibitem{paszke2017automatic}
Adam Paszke, Sam Gross, Francisco Massa, Adam Lerer, James Bradbury, Gregory
  Chanan, Trevor Killeen, Zeming Lin, Natalia Gimelshein, Luca Antiga, et~al.
\newblock Pytorch: An imperative style, high-performance deep learning library.
\newblock {\em Advances in neural information processing systems}, 32, 2019.

\bibitem{perez2021robot}
Claudia P{\'e}rez-D’Arpino, Can Liu, Patrick Goebel, Roberto
  Mart{\'\i}n-Mart{\'\i}n, and Silvio Savarese.
\newblock Robot navigation in constrained pedestrian environments using
  reinforcement learning.
\newblock In {\em ICRA}, 2021.

\bibitem{Petrenko2021MegaverseSE}
Aleksei Petrenko, Erik Wijmans, Brennan Shacklett, and Vladlen Koltun.
\newblock Megaverse: Simulating embodied agents at one million experiences per
  second.
\newblock In {\em ICML}, 2021.

\bibitem{pinto2016supersizing}
Lerrel Pinto and Abhinav Gupta.
\newblock Supersizing self-supervision: Learning to grasp from 50k tries and
  700 robot hours.
\newblock In {\em ICRA}, 2016.

\bibitem{Puig2018VirtualHomeSH}
Xavier Puig, Kevin~Kyunghwan Ra, Marko Boben, Jiaman Li, Tingwu Wang, Sanja
  Fidler, and Antonio Torralba.
\newblock Virtualhome: Simulating household activities via programs.
\newblock In {\em CVPR}, 2018.

\bibitem{Radford2021LearningTV}
Alec Radford, Jong~Wook Kim, Chris Hallacy, Aditya Ramesh, Gabriel Goh,
  Sandhini Agarwal, Girish Sastry, Amanda Askell, Pamela Mishkin, Jack Clark,
  Gretchen Krueger, and Ilya Sutskever.
\newblock Learning transferable visual models from natural language
  supervision.
\newblock In {\em ICML}, 2021.

\bibitem{Ramakrishnan2021HabitatMatterport3D}
Santhosh~K. Ramakrishnan, Aaron Gokaslan, Erik Wijmans, Oleksandr Maksymets,
  Alexander Clegg, John Turner, Eric Undersander, Wojciech Galuba, Andrew
  Westbury, Angel~Xuan Chang, Manolis Savva, Yili Zhao, and Dhruv Batra.
\newblock Habitat-matterport 3d dataset (hm3d): 1000 large-scale 3d
  environments for embodied ai.
\newblock {\em arXiv}, 2021.

\bibitem{Ramesh2021ZeroShotTG}
Aditya Ramesh, Mikhail Pavlov, Gabriel Goh, Scott Gray, Chelsea Voss, Alec
  Radford, Mark Chen, and Ilya Sutskever.
\newblock Zero-shot text-to-image generation.
\newblock In {\em ICML}, 2021.

\bibitem{Ramrakhya2022HabitatWebLE}
Ram Ramrakhya, Eric Undersander, Dhruv Batra, and Abhishek Das.
\newblock Habitat-web: Learning embodied object-search strategies from human
  demonstrations at scale.
\newblock In {\em CVPR}, 2022.

\bibitem{reizenstein2021common}
Jeremy Reizenstein, Roman Shapovalov, Philipp Henzler, Luca Sbordone, Patrick
  Labatut, and David Novotny.
\newblock Common objects in 3d: Large-scale learning and evaluation of
  real-life 3d category reconstruction.
\newblock In {\em ICCV}, 2021.

\bibitem{Ritchie2019FastAF}
Daniel Ritchie, Kai Wang, and Yu-An Lin.
\newblock Fast and flexible indoor scene synthesis via deep convolutional
  generative models.
\newblock In {\em CVPR}, 2019.

\bibitem{Ross2011ARO}
St{\'e}phane Ross, Geoffrey~J. Gordon, and J.~Andrew Bagnell.
\newblock A reduction of imitation learning and structured prediction to
  no-regret online learning.
\newblock In {\em AISTATS}, 2011.

\bibitem{Savva2017SceneSuggestC3}
Manolis Savva, Angel~X. Chang, and Maneesh Agrawala.
\newblock Scenesuggest: Context-driven 3d scene design.
\newblock {\em arXiv}, 2017.

\bibitem{savva2019habitat}
Manolis Savva, Abhishek Kadian, Oleksandr Maksymets, Yili Zhao, Erik Wijmans,
  Bhavana Jain, Julian Straub, Jia Liu, Vladlen Koltun, Jitendra Malik, et~al.
\newblock Habitat: A platform for embodied ai research.
\newblock In {\em ICCV}, 2019.

\bibitem{schulman2015high}
John Schulman, Philipp Moritz, Sergey Levine, Michael Jordan, and Pieter
  Abbeel.
\newblock High-dimensional continuous control using generalized advantage
  estimation.
\newblock In {\em ICLR}, 2016.

\bibitem{schulman2017proximal}
John Schulman, Filip Wolski, Prafulla Dhariwal, Alec Radford, and Oleg Klimov.
\newblock Proximal policy optimization algorithms.
\newblock {\em arXiv}, 2017.

\bibitem{Shen2020iGibsonAS}
Bokui Shen, Fei Xia, Chengshu Li, Roberto Mart'in-Mart'in, Linxi~(Jim) Fan,
  Guanzhi Wang, S. Buch, Claudia.~P{\'e}rez D'Arpino, Sanjana Srivastava,
  Lyne~P. Tchapmi, Micael~Edmond Tchapmi, Kent Vainio, Li Fei-Fei, and Silvio
  Savarese.
\newblock igibson, a simulation environment for interactive tasks in large
  realistic scenes.
\newblock In {\em IROS}, 2021.

\bibitem{shridhar2020alfred}
Mohit Shridhar, Jesse Thomason, Daniel Gordon, Yonatan Bisk, Winson Han,
  Roozbeh Mottaghi, Luke Zettlemoyer, and Dieter Fox.
\newblock Alfred: A benchmark for interpreting grounded instructions for
  everyday tasks.
\newblock In {\em CVPR}, 2020.

\bibitem{Srivastava2021BEHAVIORBF}
Sanjana Srivastava, Chengshu Li, Michael Lingelbach, Roberto Mart'in-Mart'in,
  Fei Xia, Kent Vainio, Zheng Lian, Cem Gokmen, S. Buch, C.~Karen Liu, Silvio
  Savarese, Hyowon Gweon, Jiajun Wu, and Li Fei-Fei.
\newblock Behavior: Benchmark for everyday household activities in virtual,
  interactive, and ecological environments.
\newblock In {\em CoRL}, 2021.

\bibitem{sun2020scalability}
Pei Sun, Henrik Kretzschmar, Xerxes Dotiwalla, Aurelien Chouard, Vijaysai
  Patnaik, Paul Tsui, James Guo, Yin Zhou, Yuning Chai, Benjamin Caine, et~al.
\newblock Scalability in perception for autonomous driving: Waymo open dataset.
\newblock In {\em CVPR}, 2020.

\bibitem{szegedy2015going}
Christian Szegedy, Wei Liu, Yangqing Jia, Pierre Sermanet, Scott Reed, Dragomir
  Anguelov, Dumitru Erhan, Vincent Vanhoucke, and Andrew Rabinovich.
\newblock Going deeper with convolutions.
\newblock In {\em CVPR}, 2015.

\bibitem{szot2021habitat}
Andrew Szot, Alexander Clegg, Eric Undersander, Erik Wijmans, Yili Zhao, John
  Turner, Noah Maestre, Mustafa Mukadam, Devendra~Singh Chaplot, Oleksandr
  Maksymets, Aaron Gokaslan, Vladimir Vondrus, Sameer Dharur, Franziska Meier,
  Wojciech Galuba, Angel~Xuan Chang, Zsolt Kira, Vladlen Koltun, Jitendra
  Malik, Manolis Savva, and Dhruv Batra.
\newblock Habitat 2.0: Training home assistants to rearrange their habitat.
\newblock In {\em NeurIPS}, 2021.

\bibitem{tancik2022block}
Matthew Tancik, Vincent Casser, Xinchen Yan, Sabeek Pradhan, Ben Mildenhall,
  Pratul~P Srinivasan, Jonathan~T Barron, and Henrik Kretzschmar.
\newblock Block-nerf: Scalable large scene neural view synthesis.
\newblock In {\em CVPR}, 2022.

\bibitem{Thomee2016YFCC100MTN}
Bart Thomee, David~A. Shamma, Gerald Friedland, Benjamin Elizalde, Karl~S. Ni,
  Douglas~N. Poland, Damian Borth, and Li-Jia Li.
\newblock Yfcc100m: the new data in multimedia research.
\newblock {\em Comm. of the ACM}, 2016.

\bibitem{virtanen2020scipy}
Pauli Virtanen, Ralf Gommers, Travis~E Oliphant, Matt Haberland, Tyler Reddy,
  David Cournapeau, Evgeni Burovski, Pearu Peterson, Warren Weckesser, Jonathan
  Bright, et~al.
\newblock Scipy 1.0: fundamental algorithms for scientific computing in python.
\newblock {\em Nature methods}, 2020.

\bibitem{Wang2021SceneFormerIS}
Xinpeng Wang, Chandan Yeshwanth, and Matthias Nie{\ss}ner.
\newblock Sceneformer: Indoor scene generation with transformers.
\newblock In {\em 3DV}, 2021.

\bibitem{waskom2021seaborn}
Michael~L Waskom.
\newblock Seaborn: statistical data visualization.
\newblock {\em Journal of Open Source Software}, 2021.

\bibitem{weihs2021visual}
Luca Weihs, Matt Deitke, Aniruddha Kembhavi, and Roozbeh Mottaghi.
\newblock Visual room rearrangement.
\newblock In {\em CVPR}, 2021.

\bibitem{weihs2020allenact}
Luca Weihs, Jordi Salvador, Klemen Kotar, Unnat Jain, Kuo-Hao Zeng, Roozbeh
  Mottaghi, and Aniruddha Kembhavi.
\newblock {AllenAct: A framework for embodied AI research}.
\newblock {\em arXiv}, 2020.

\bibitem{wijmans2019dd}
Erik Wijmans, Abhishek Kadian, Ari Morcos, Stefan Lee, Irfan Essa, Devi Parikh,
  Manolis Savva, and Dhruv Batra.
\newblock Dd-ppo: Learning near-perfect pointgoal navigators from 2.5 billion
  frames.
\newblock In {\em ICLR}, 2019.

\bibitem{wortsman2019learning}
Mitchell Wortsman, Kiana Ehsani, Mohammad Rastegari, Ali Farhadi, and Roozbeh
  Mottaghi.
\newblock Learning to learn how to learn: Self-adaptive visual navigation using
  meta-learning.
\newblock In {\em CVPR}, 2019.

\bibitem{wu2021communicative}
Qi Wu, Cheng-Ju Wu, Yixin Zhu, and Jungseock Joo.
\newblock Communicative learning with natural gestures for embodied navigation
  agents with human-in-the-scene.
\newblock In {\em IROS}, 2021.

\bibitem{wu2019data}
Wenming Wu, Xiao-Ming Fu, Rui Tang, Yuhan Wang, Yu-Hao Qi, and Ligang Liu.
\newblock Data-driven interior plan generation for residential buildings.
\newblock {\em ACM Trans. on Graphics}, 2019.

\bibitem{house3d}
Yi Wu, Yuxin Wu, Georgia Gkioxari, and Yuandong Tian.
\newblock Building generalizable agents with a realistic and rich 3d
  environment.
\newblock {\em arXiv}, 2018.

\bibitem{Xia2021ReLMoGenIM}
Fei Xia, Chengshu Li, Roberto Mart'in-Mart'in, Or Litany, Alexander Toshev, and
  Silvio Savarese.
\newblock Relmogen: Integrating motion generation in reinforcement learning for
  mobile manipulation.
\newblock In {\em ICRA}, 2021.

\bibitem{xia2018gibson}
Fei Xia, Amir~R Zamir, Zhiyang He, Alexander Sax, Jitendra Malik, and Silvio
  Savarese.
\newblock Gibson env: Real-world perception for embodied agents.
\newblock In {\em CVPR}, 2018.

\bibitem{xiang2020sapien}
Fanbo Xiang, Yuzhe Qin, Kaichun Mo, Yikuan Xia, Hao Zhu, Fangchen Liu, Minghua
  Liu, Hanxiao Jiang, Yifu Yuan, He Wang, Li Yi, Angel X.Chang, Leonidas
  Guibas, and Hao Su.
\newblock {S}{A}{P}{I}{E}{N}: {A} {S}imul{A}ted {P}art-based {I}nteractive
  {E}{N}vironment.
\newblock In {\em CVPR}, 2020.

\bibitem{Xiang2016ObjectNet3DAL}
Yu Xiang, Wonhui Kim, Wei Chen, Jingwei Ji, Christopher~Bongsoo Choy, Hao Su,
  Roozbeh Mottaghi, Leonidas~J. Guibas, and Silvio Savarese.
\newblock Objectnet3d: A large scale database for 3d object recognition.
\newblock In {\em ECCV}, 2016.

\bibitem{Yadan2019Hydra}
Omry Yadan.
\newblock Hydra - a framework for elegantly configuring complex applications.
\newblock Github, 2019.

\bibitem{EvalAI}
Deshraj Yadav, Rishabh Jain, Harsh Agrawal, Prithvijit Chattopadhyay, Taranjeet
  Singh, Akash Jain, Shiv~Baran Singh, Stefan Lee, and Dhruv Batra.
\newblock Evalai: Towards better evaluation systems for ai agents.
\newblock {\em arXiv}, 2019.

\bibitem{yitzhak2022clip}
Samir Yitzhak~Gadre, Mitchell Wortsman, Gabriel Ilharco, Ludwig Schmidt, and
  Shuran Song.
\newblock Clip on wheels: Zero-shot object navigation as object localization
  and exploration.
\newblock {\em arXiv}, 2022.

\bibitem{Zhang2020DeepGM}
Zaiwei Zhang, Zhenpei Yang, Chongyang Ma, Linjie Luo, Alexander~G. Huth,
  Etienne Vouga, and Qixing Huang.
\newblock Deep generative modeling for scene synthesis via hybrid
  representations.
\newblock {\em ACM Trans. on Graphics}, 2020.

\bibitem{Zheng2022TowardsOp}
Kaiyu Zheng, Rohan Chitnis, Yoonchang Sung, George Konidaris, and Stefanie
  Tellex.
\newblock {Towards Optimal Correlational Object Search}.
\newblock In {\em ICRA}, 2022.

\bibitem{zhou2019scenegraphnet}
Yang Zhou, Zachary While, and Evangelos Kalogerakis.
\newblock Scenegraphnet: Neural message passing for 3d indoor scene
  augmentation.
\newblock In {\em ICCV}, 2019.

\bibitem{zhu2017target}
Yuke Zhu, Roozbeh Mottaghi, Eric Kolve, Joseph~J Lim, Abhinav Gupta, Li
  Fei-Fei, and Ali Farhadi.
\newblock Target-driven visual navigation in indoor scenes using deep
  reinforcement learning.
\newblock In {\em ICRA}, 2017.

\bibitem{zhu2020robosuite}
Yuke Zhu, Josiah Wong, Ajay Mandlekar, and Roberto Mart{\'\i}n-Mart{\'\i}n.
\newblock robosuite: A modular simulation framework and benchmark for robot
  learning.
\newblock {\em arXiv}, 2020.

\end{thebibliography}

}

\newpage

\section*{Contributions}

\textbf{Matt Deitke} designed and implemented the procedure to generate houses, implemented ObjectNav pre-training experiments and fine-tuning experiments, built the website, advised and implemented parts of the Unity backend, built the platform to visualize assets and create semantic asset groups, contributed to visuals, and wrote the paper.

\textbf{Kiana Ehsani} implemented ArmPointNav experiments and wrote parts of the paper.

\textbf{Ali Farhadi} advised on the research direction.

\textbf{Alvaro Herrasti} led the Unity backend development that creates a house from a JSON specification.

\textbf{Aniruddha Kembhavi} advised on research direction, the \elienv{} development, and the house generation process and wrote the paper.

\textbf{Eric Kolve} advised on the Unity backend development.

\textbf{Roozbeh Mottaghi} advised on the research direction, the Unity backend, the \elienv{} development, and the house generation process and wrote the paper.

\textbf{Jordi Salvador} implemented rearrangement experiments, advised on multi-node training experiments, and wrote parts of the paper.

\textbf{Eli VanderBilt} standardized AI2-THOR's asset and material database to make it usable with \env{}, led the development of \elienv{}, implemented parts of the Unity backend, created new 3D assets and skyboxes, advised on lighting the houses, and contributed to visuals.

\textbf{Winson Han} implemented parts of \elienv{}, implemented parts of the Unity backend, and contributed to visuals.

\textbf{Luca Weihs} advised the work on experiments, assisted with rearragement experiments, implemented ObjectNav fine-tuning on HM3D-Semantics, and wrote parts of the paper.

\appendix


{
\section*{Appendix}

\startcontents[sections]
\printcontents[sections]{l}{1}{\setcounter{tocdepth}{3}}
}





\newpage



\newpage

\section{ProcTHOR Assets}
\label{sec:procthorObjects}

\vspace{-0.15in}

\begin{figure}[htbp]
     \centering
     \hfill
     \begin{subfigure}[b]{0.57\textwidth}
         \begin{subfigure}{0.2\textwidth}
             \centering
             \includegraphics[width=\textwidth, cfbox=black!30]{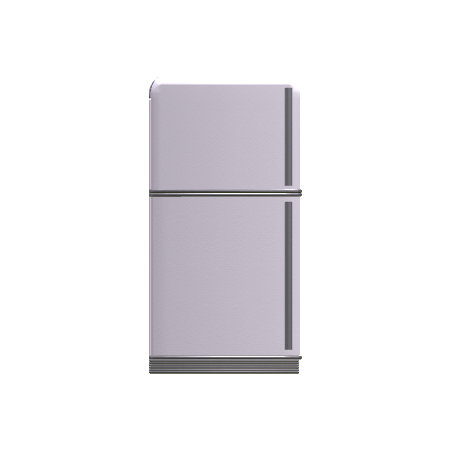}
             \caption*{\textsc{Fridge\_1}}
         \end{subfigure}
         \hfill
         \begin{subfigure}{0.2\textwidth}
             \centering
             \includegraphics[width=\textwidth, cfbox=black!30]{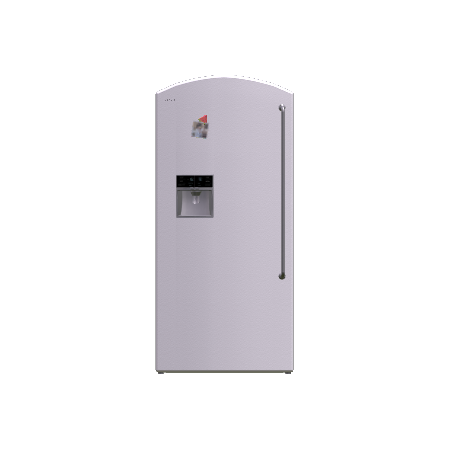}
             \caption*{\textsc{Fridge\_10}}
         \end{subfigure}
         \hfill
         \begin{subfigure}{0.2\textwidth}
             \centering
             \includegraphics[width=\textwidth, cfbox=black!30]{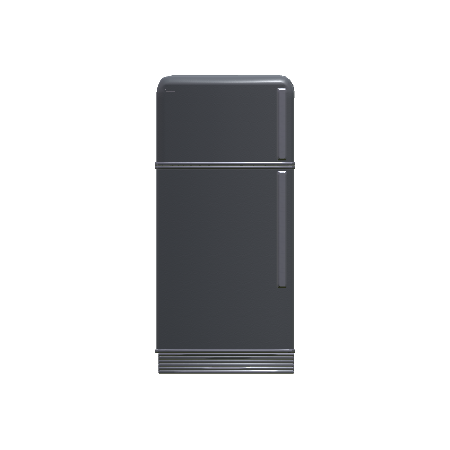}
             \caption*{\textsc{Fridge\_11}}
         \end{subfigure}
         \hfill
         \begin{subfigure}{0.2\textwidth}
             \centering
             \includegraphics[width=\textwidth, cfbox=black!30]{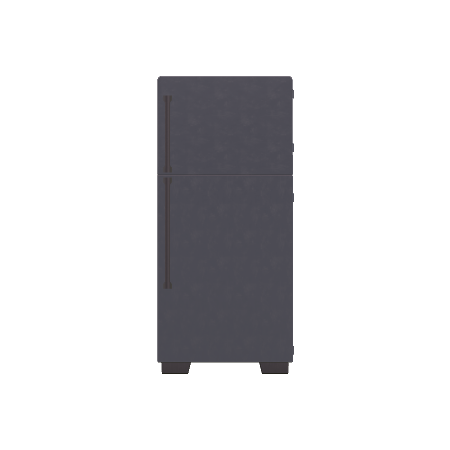}
             \caption*{\textsc{Fridge\_12}}
         \end{subfigure}
         \\[0.1in]

         \begin{subfigure}{0.2\textwidth}
             \centering
             \includegraphics[width=\textwidth, cfbox=black!30]{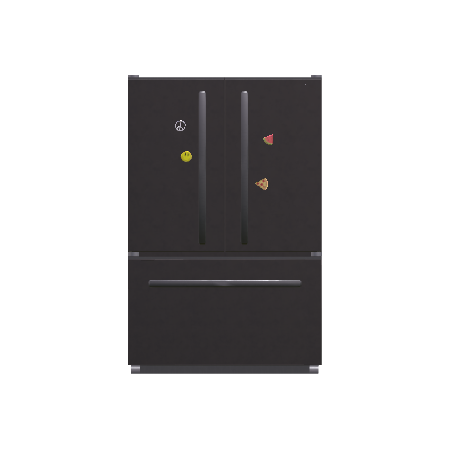}
             \caption*{\textsc{Fridge\_13}}
         \end{subfigure}
         \hfill
         \begin{subfigure}{0.2\textwidth}
             \centering
             \includegraphics[width=\textwidth, cfbox=black!30]{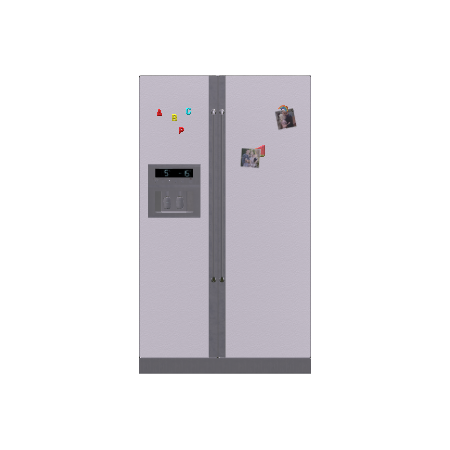}
             \caption*{\textsc{Fridge\_14}}
         \end{subfigure}
         \hfill
         \begin{subfigure}{0.2\textwidth}
             \centering
             \includegraphics[width=\textwidth, cfbox=black!30]{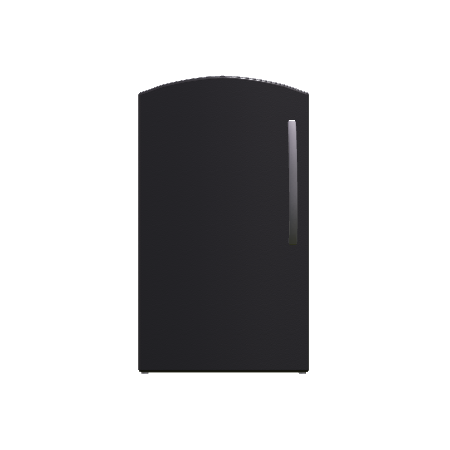}
             \caption*{\textsc{Fridge\_15}}
         \end{subfigure}
         \hfill
         \begin{subfigure}{0.2\textwidth}
             \centering
             \includegraphics[width=\textwidth, cfbox=black!30]{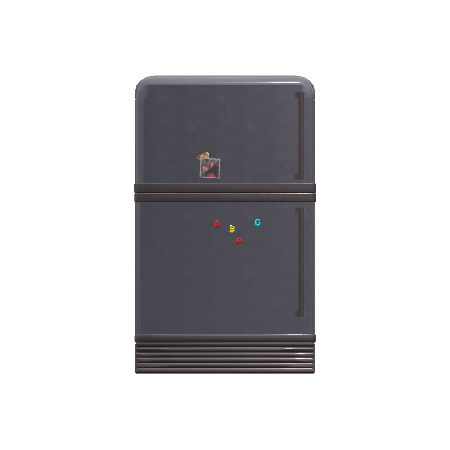}
             \caption*{\textsc{Fridge\_16}}
         \end{subfigure}
         \\[0.22in]

         \begin{subfigure}{0.2\textwidth}
             \centering
             \includegraphics[width=\textwidth, cfbox=black!30]{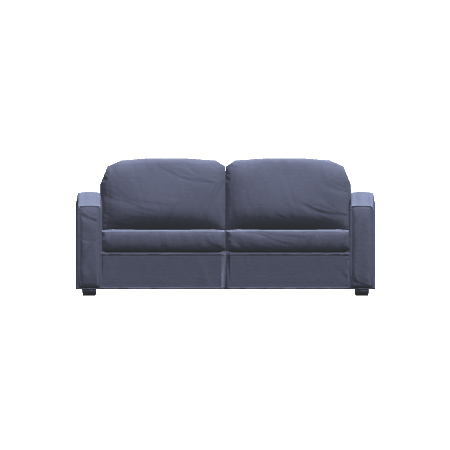}
             \caption*{\textsc{Sofa\_201\_2}}
         \end{subfigure}
         \hfill
         \begin{subfigure}{0.2\textwidth}
             \centering
             \includegraphics[width=\textwidth, cfbox=black!30]{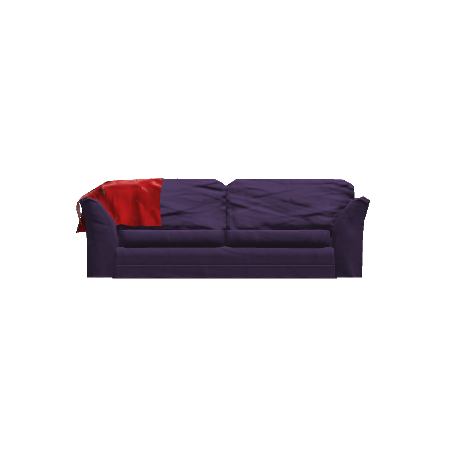}
             \caption*{\textsc{Sofa\_203\_1}}
         \end{subfigure}
         \hfill
         \begin{subfigure}{0.2\textwidth}
             \centering
             \includegraphics[width=\textwidth, cfbox=black!30]{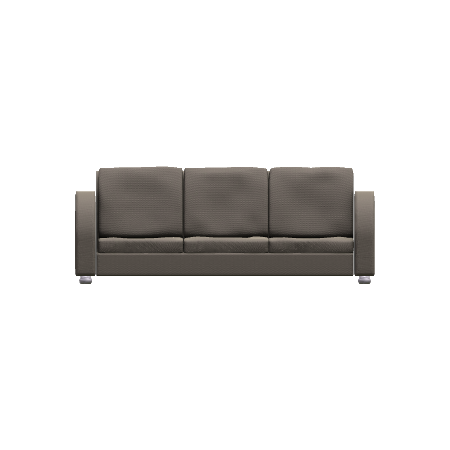}
             \caption*{\textsc{Sofa\_204\_1}}
         \end{subfigure}
         \hfill
         \begin{subfigure}{0.2\textwidth}
             \centering
             \includegraphics[width=\textwidth, cfbox=black!30]{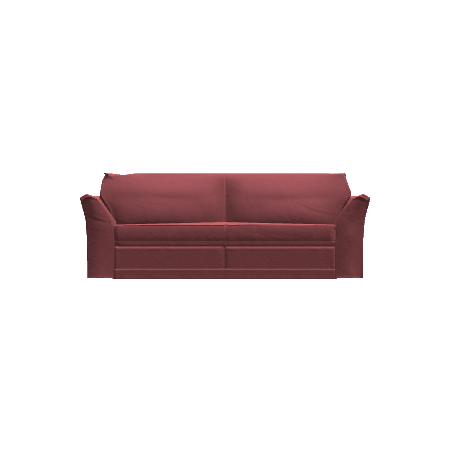}
             \caption*{\textsc{Sofa\_205\_1}}
         \end{subfigure}
         \\[0.1in]

         \begin{subfigure}{0.2\textwidth}
             \centering
             \includegraphics[width=\textwidth, cfbox=black!30]{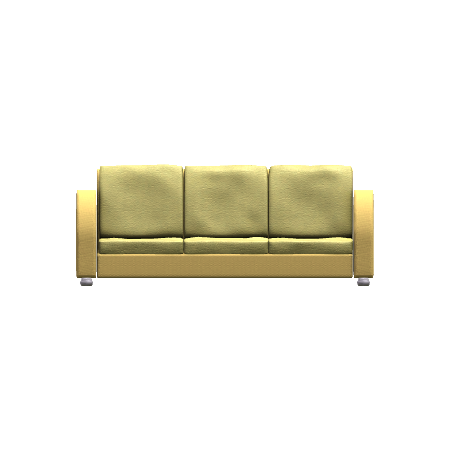}
             \caption*{\textsc{Sofa\_207\_3}}
         \end{subfigure}
         \hfill
         \begin{subfigure}{0.2\textwidth}
             \centering
             \includegraphics[width=\textwidth, cfbox=black!30]{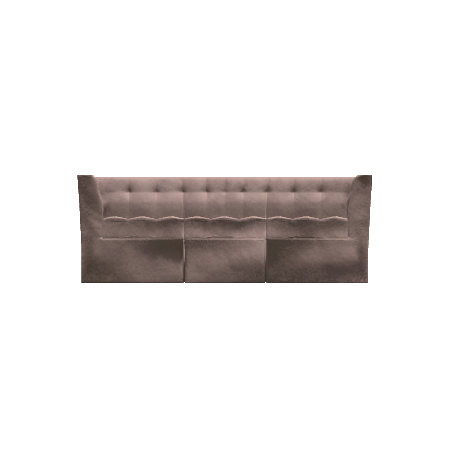}
             \caption*{\textsc{Sofa\_214\_2}}
         \end{subfigure}
         \hfill
         \begin{subfigure}{0.2\textwidth}
             \centering
             \includegraphics[width=\textwidth, cfbox=black!30]{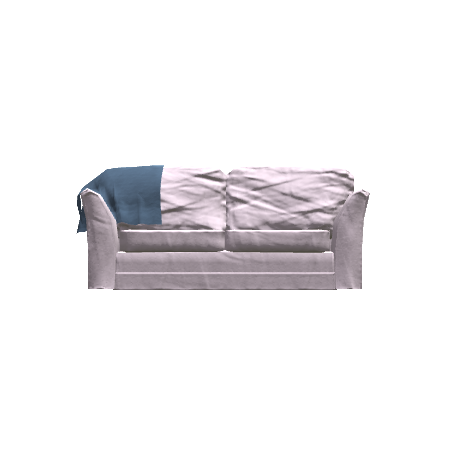}
             \caption*{\textsc{Sofa\_218\_1}}
         \end{subfigure}
         \hfill
         \begin{subfigure}{0.2\textwidth}
             \centering
             \includegraphics[width=\textwidth, cfbox=black!30]{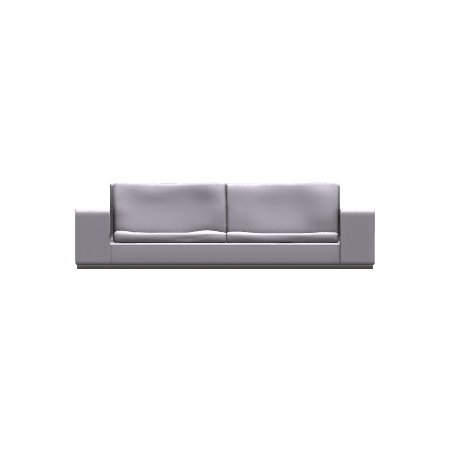}
             \caption*{\textsc{Sofa\_227\_1}}
         \end{subfigure}
         \\[0.22in]

         \begin{subfigure}{0.2\textwidth}
             \centering
             \includegraphics[width=\textwidth, cfbox=black!30]{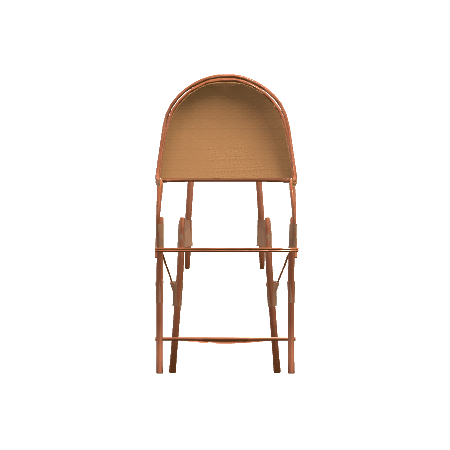}
             \caption*{\textsc{Chair\_002\_1}}
         \end{subfigure}
         \hfill
         \begin{subfigure}{0.2\textwidth}
             \centering
             \includegraphics[width=\textwidth, cfbox=black!30]{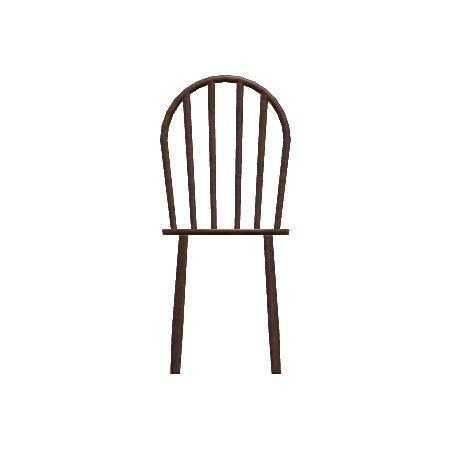}
             \caption*{\textsc{Chair\_007\_1}}
         \end{subfigure}
         \hfill
         \begin{subfigure}{0.2\textwidth}
             \centering
             \includegraphics[width=\textwidth, cfbox=black!30]{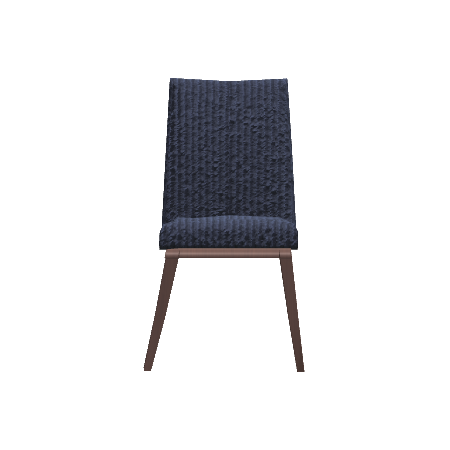}
             \caption*{\textsc{Chair\_201\_1}}
         \end{subfigure}
         \hfill
         \begin{subfigure}{0.2\textwidth}
             \centering
             \includegraphics[width=\textwidth, cfbox=black!30]{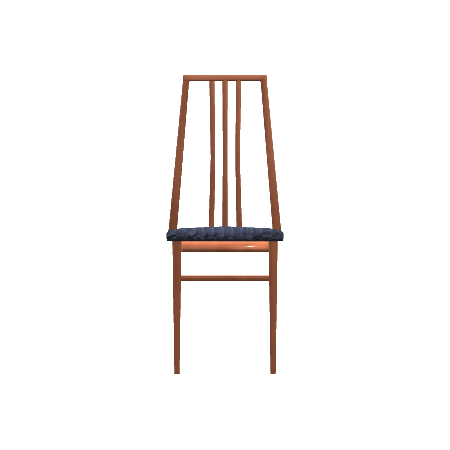}
             \caption*{\textsc{Chair\_203\_1}}
         \end{subfigure}
         \\[0.15in]

         \begin{subfigure}{0.2\textwidth}
             \centering
             \includegraphics[width=\textwidth, cfbox=black!30]{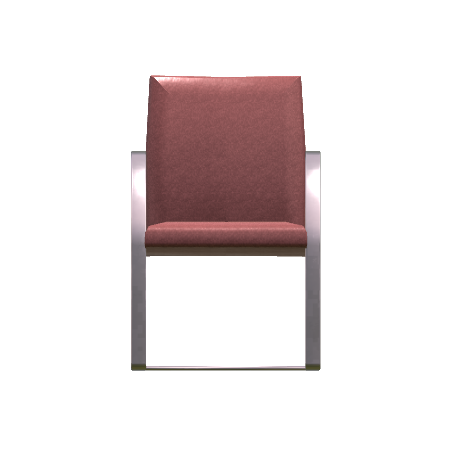}
             \caption*{\textsc{Chair\_204\_1}}
         \end{subfigure}
         \hfill
         \begin{subfigure}{0.2\textwidth}
             \centering
             \includegraphics[width=\textwidth, cfbox=black!30]{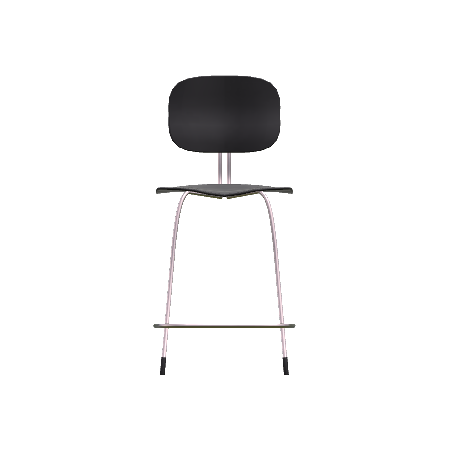}
             \caption*{\textsc{Chair\_205\_1}}
         \end{subfigure}
         \hfill
         \begin{subfigure}{0.2\textwidth}
             \centering
             \includegraphics[width=\textwidth, cfbox=black!30]{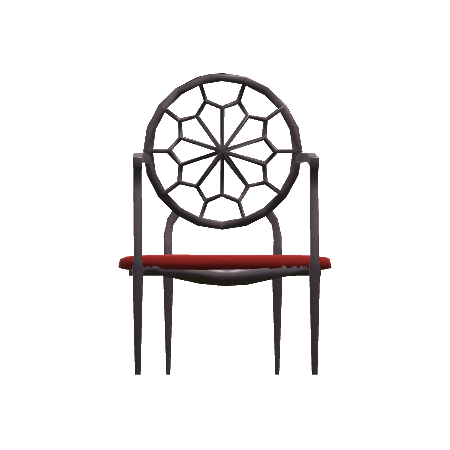}
             \caption*{\textsc{Chair\_210\_1}}
         \end{subfigure}
         \hfill
         \begin{subfigure}{0.2\textwidth}
             \centering
             \includegraphics[width=\textwidth, cfbox=black!30]{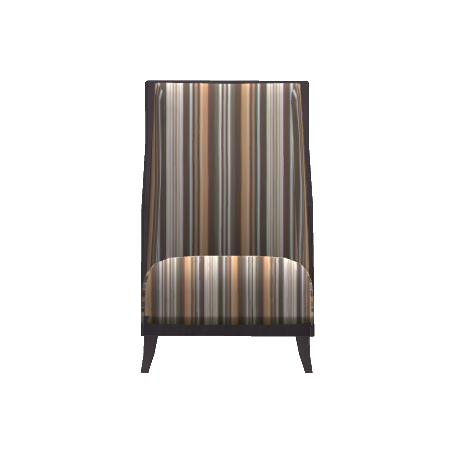}
             \caption*{\textsc{Chair\_215\_1}}
         \end{subfigure}

         \caption{Examples of assets in the asset database. The forward-facing direction for each asset is consistent across all assets within its type, which allows us to do things like not spawn fridges facing the wall.}

     \end{subfigure}
     \hfill\;
     \begin{subfigure}[b]{0.37\textwidth}
         \centering
         \includegraphics[width=\textwidth]{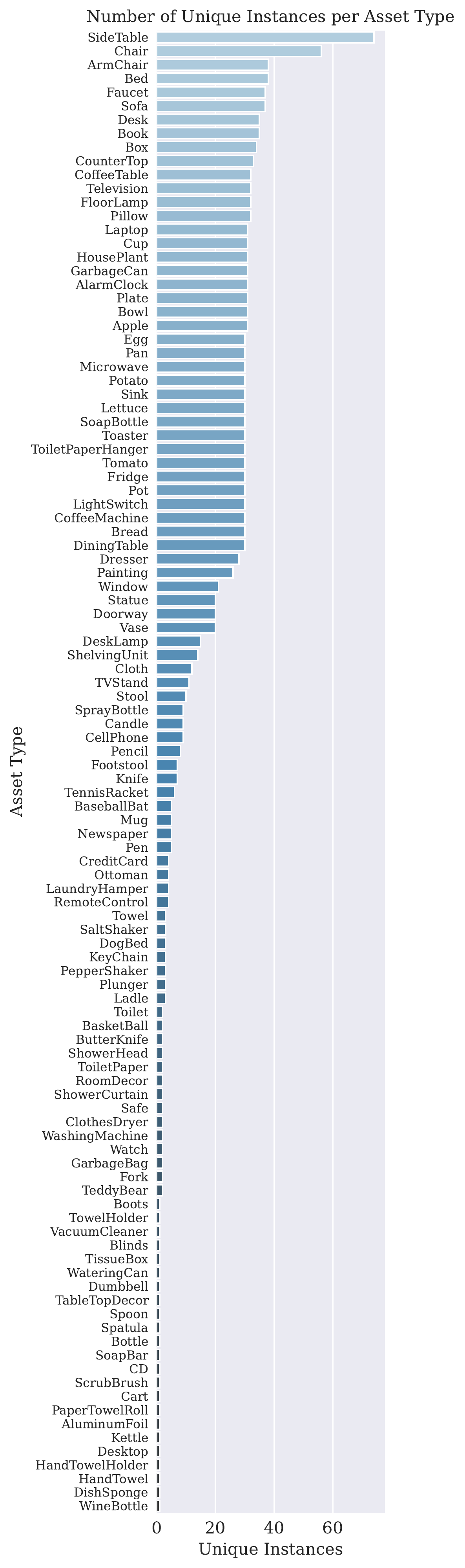}
         \caption{The number of unique 3D modeled assets for each of the 108 asset types. There are 1,633 unique assets in total.}
     \end{subfigure}
    \caption{Examples and statistics of assets in the asset database.}
\end{figure}


\begin{minipage}{\textwidth}

\section{House Generation}
\label{sec:houseGeneration}

This section gives more details about the process we developed to procedurally sample houses.

\subsection{Examples}

\end{minipage}

\newpage

\subsubsection{3-Room Houses}

\begin{figure}[H]
    \centering
    \vspace{0.015in}
    \makebox[\textwidth][c]{
        \includegraphics[width=1.07\textwidth]{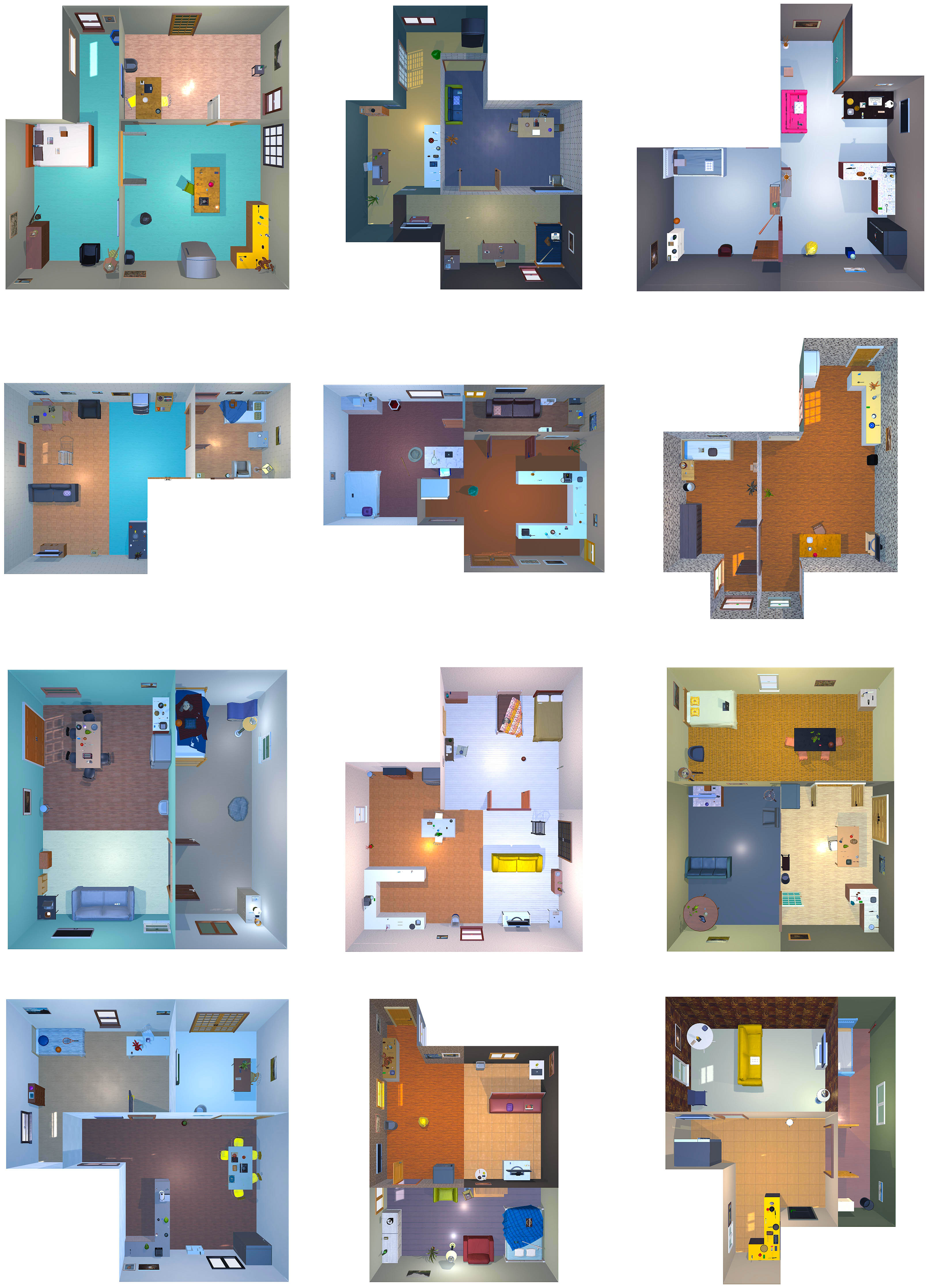}
    }
    \vspace{0.025in}
    \caption{Examples of 3-room houses generated in \env-10K.}
\end{figure}

\newpage

\subsubsection{4-Room Houses}

\begin{figure}[H]
    \centering
    \vspace{0.015in}
    \makebox[\textwidth][c]{
        \includegraphics[width=1.07\textwidth]{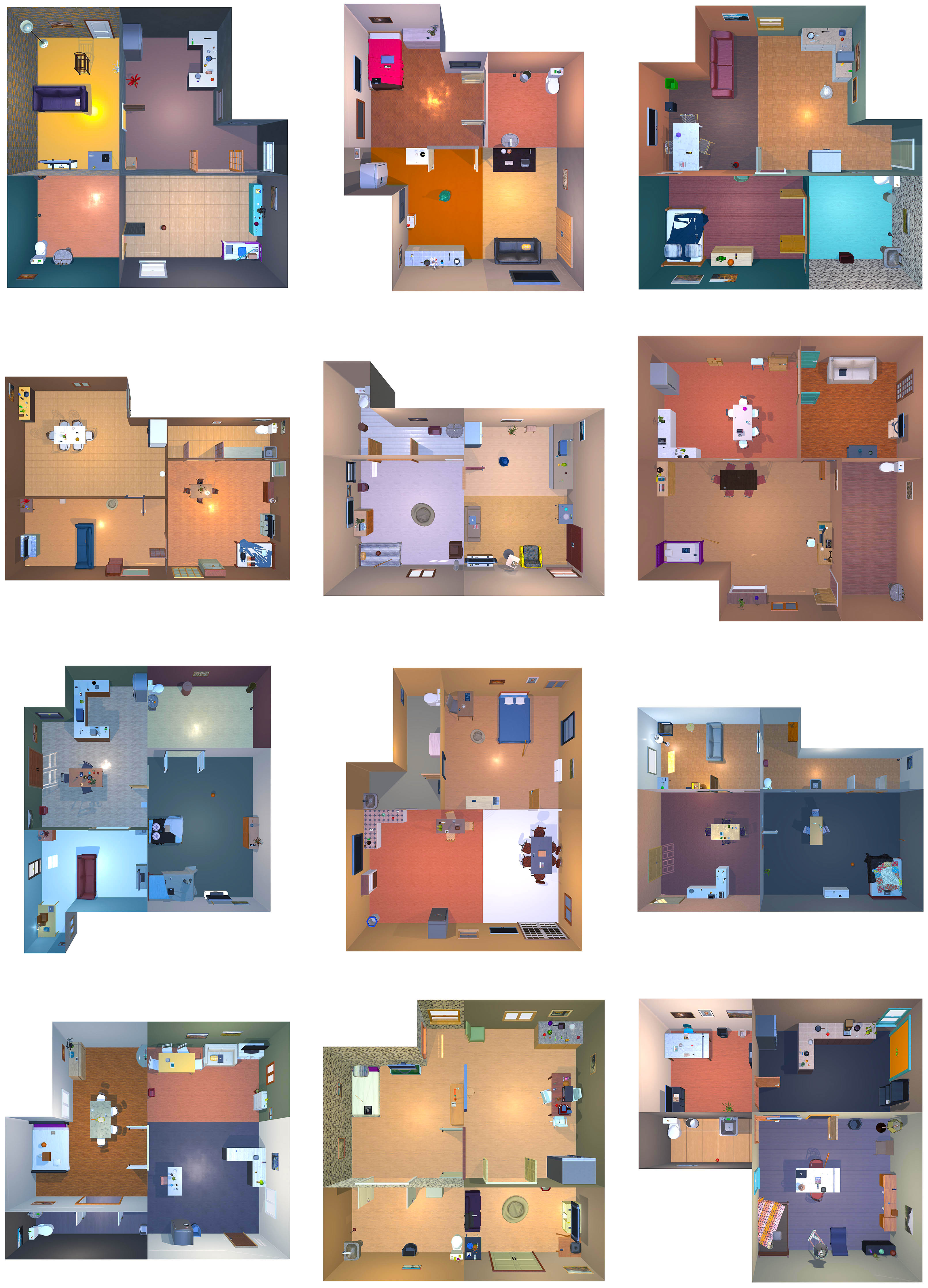}
    }
    \vspace{0.025in}
    \caption{Examples of 4-room houses generated in \env-10K.}
\end{figure}

\newpage

\subsubsection{5-Room Houses}

\begin{figure}[H]
    \centering
    \vspace{0.015in}
    \makebox[\textwidth][c]{
        \includegraphics[width=1.07\textwidth]{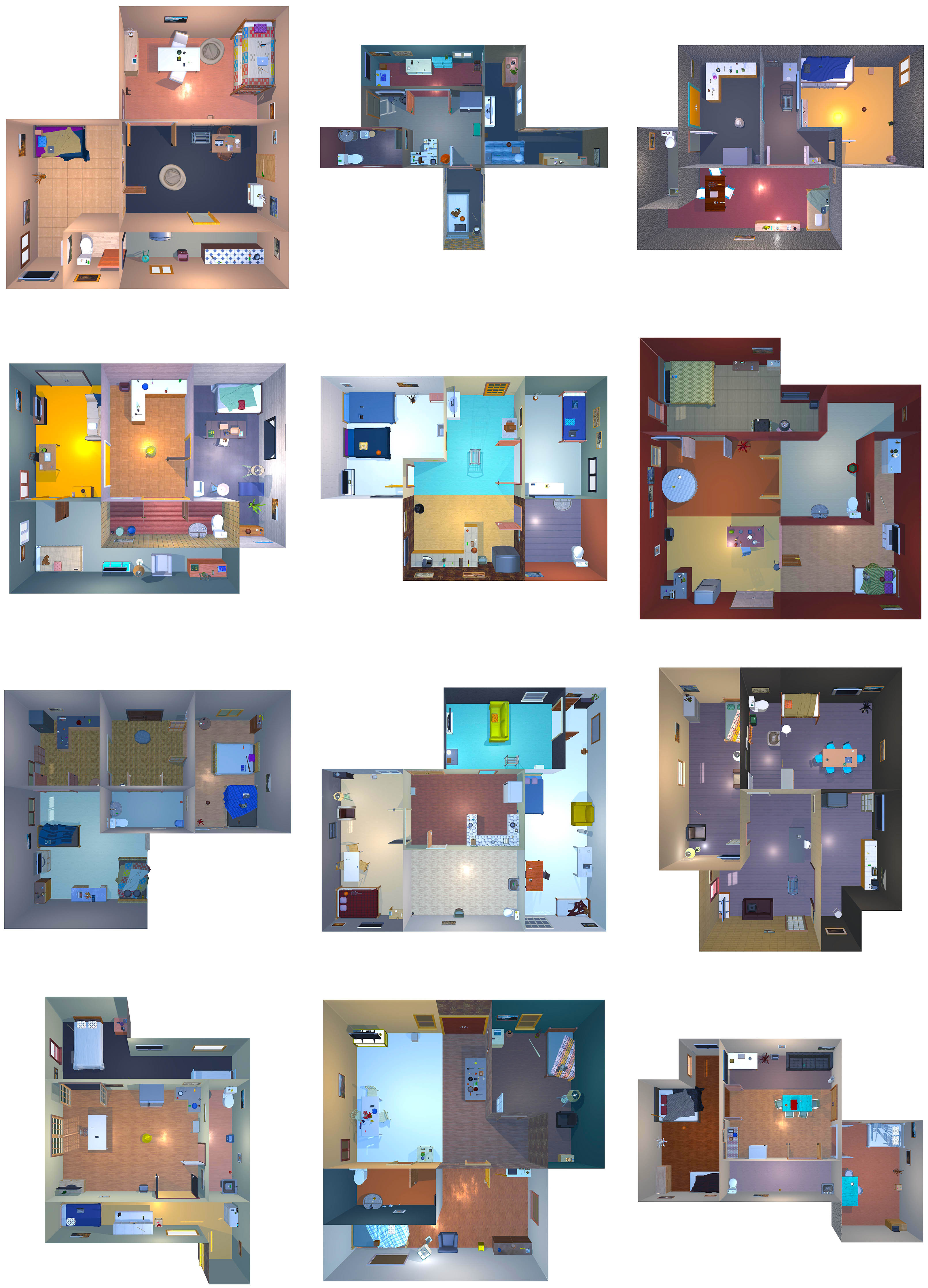}
    }
    \vspace{0.025in}
    \caption{Examples of 5-room houses generated in \env-10K.}
\end{figure}

\newpage

\subsubsection{6-Room Houses}

\begin{figure}[H]
    \centering
    \vspace{0.015in}
    \makebox[\textwidth][c]{
        \includegraphics[width=1.07\textwidth]{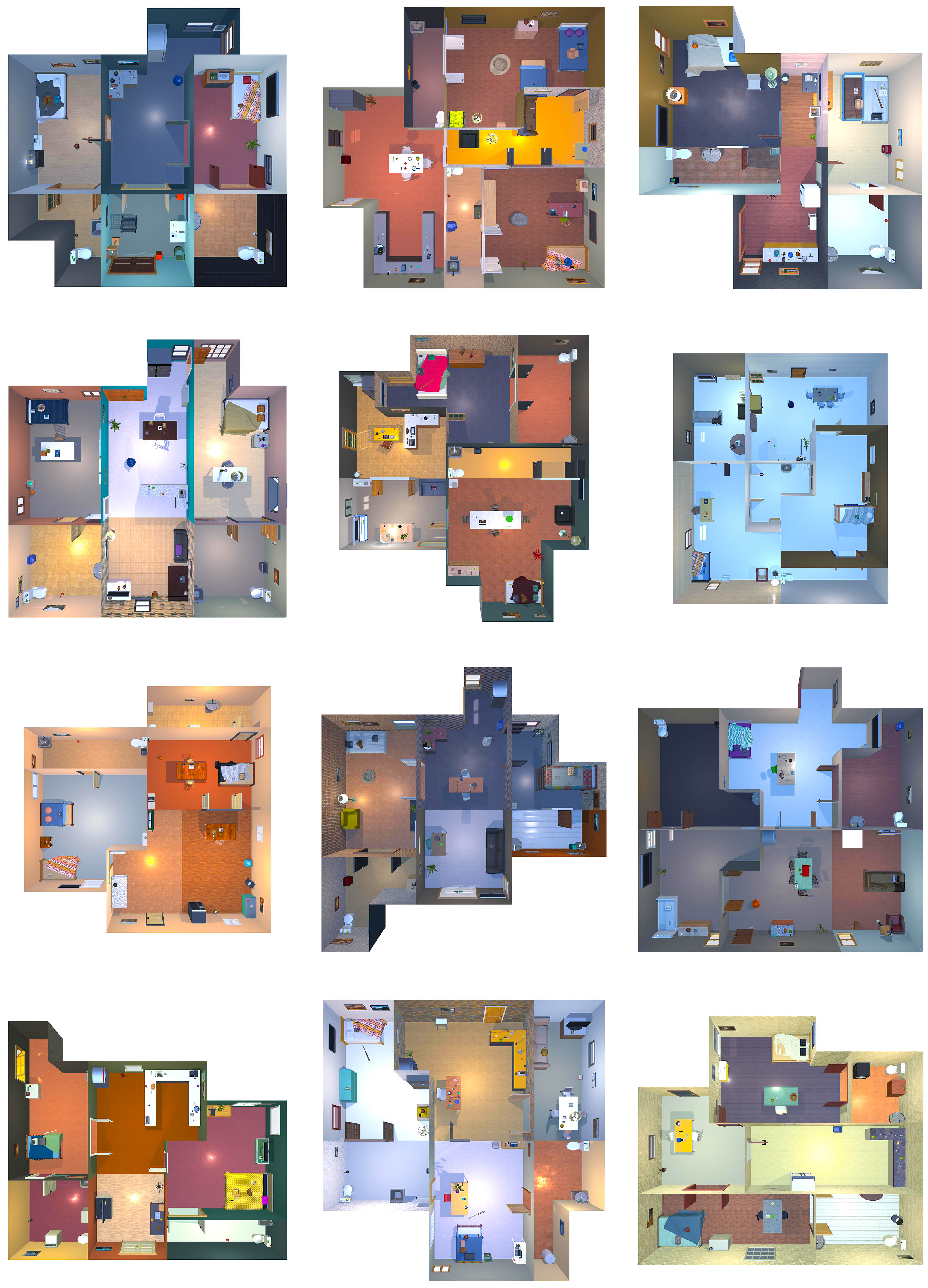}
    }
    \vspace{0.025in}
    \caption{Examples of 6-room houses generated in \env-10K.}
\end{figure}

\newpage

\subsubsection{7+ Room Houses}

\begin{figure}[H]
    \centering
    \vspace{0.015in}
    \makebox[\textwidth][c]{
        \includegraphics[width=1.07\textwidth]{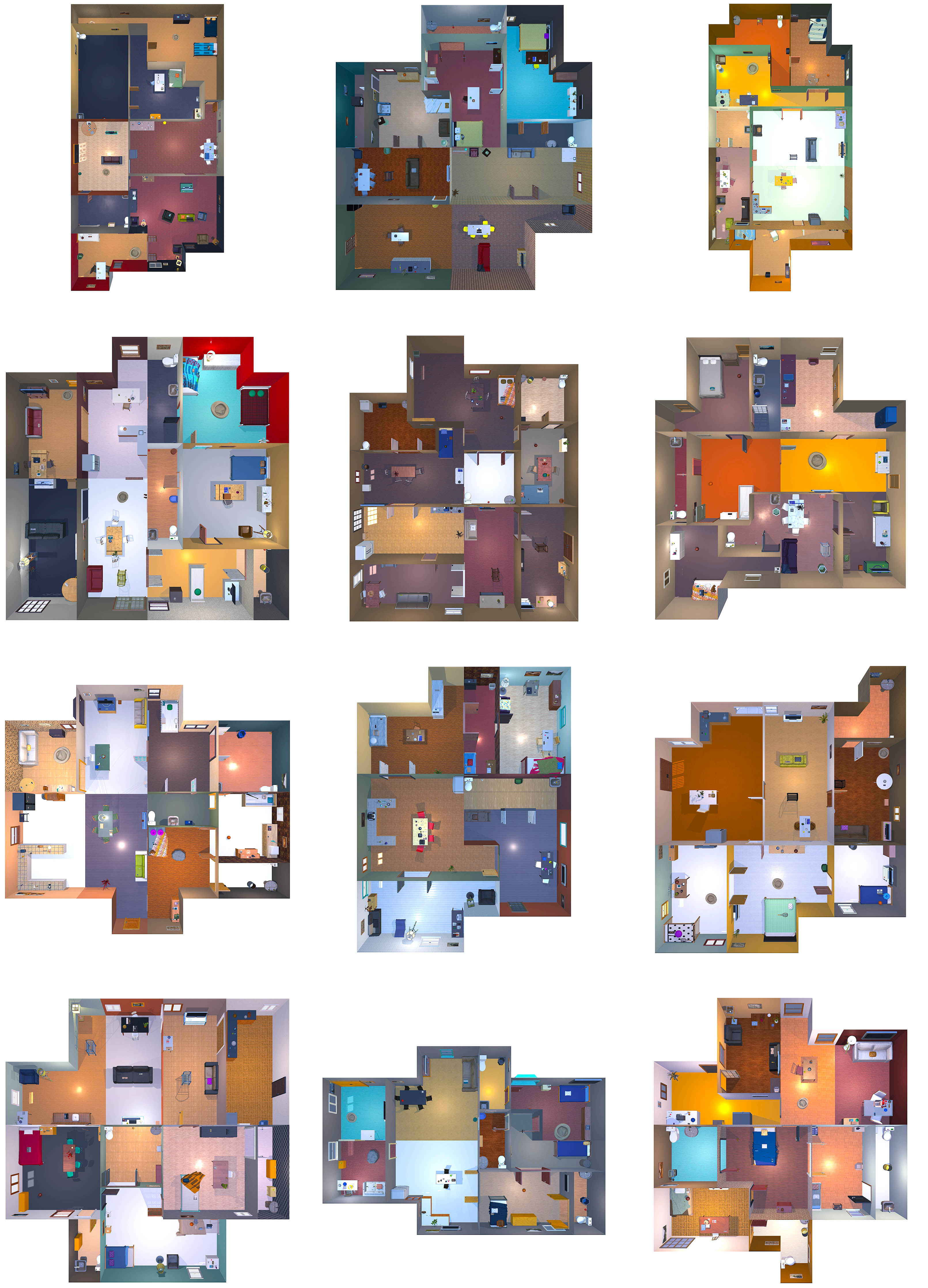}
    }
    \vspace{0.025in}
    \caption{Examples of 7+ room houses generated in \env-10K.}
\end{figure}

\newpage

\subsection{Room Specs}

\begin{figure}
    \centering
    \begin{subfigure}[b]{0.32\textwidth}
        \centering
        \includegraphics[width=\textwidth]{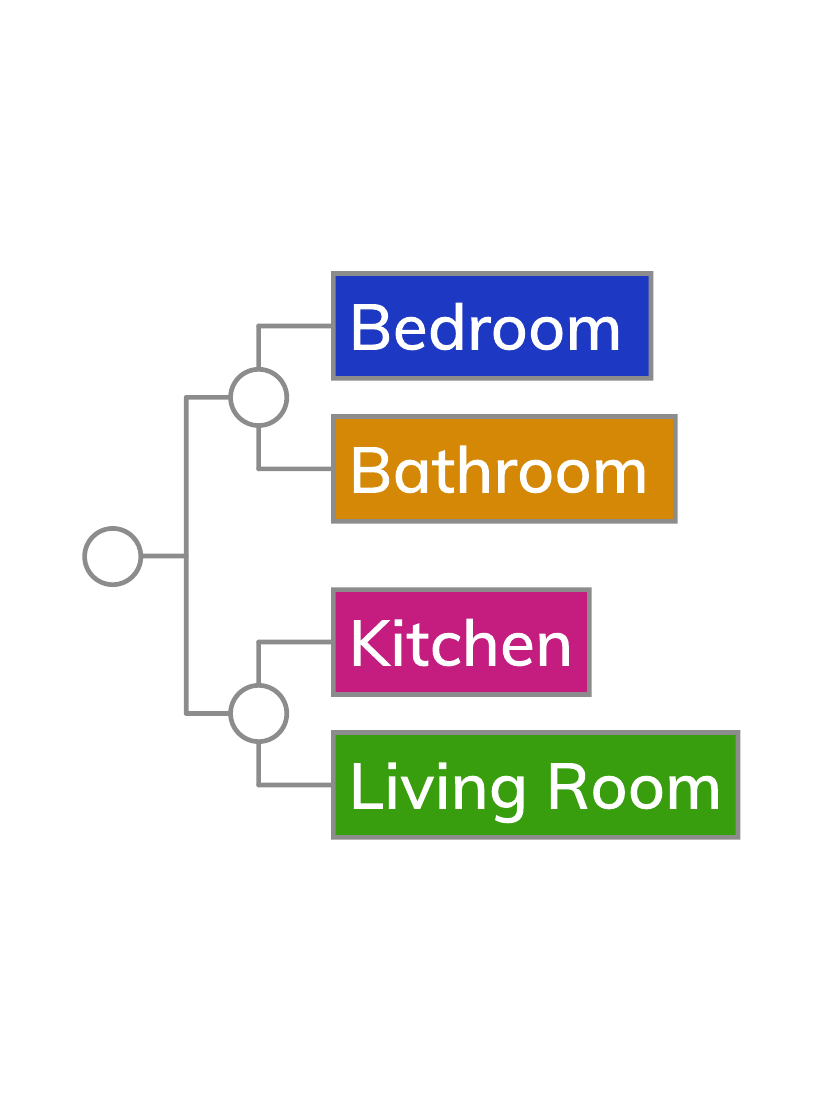}
        \caption{4-Room House}
        \label{fig:roomSpecExsA}
    \end{subfigure}
    \hfill
    \begin{subfigure}[b]{0.32\textwidth}
        \centering
        \includegraphics[width=\textwidth]{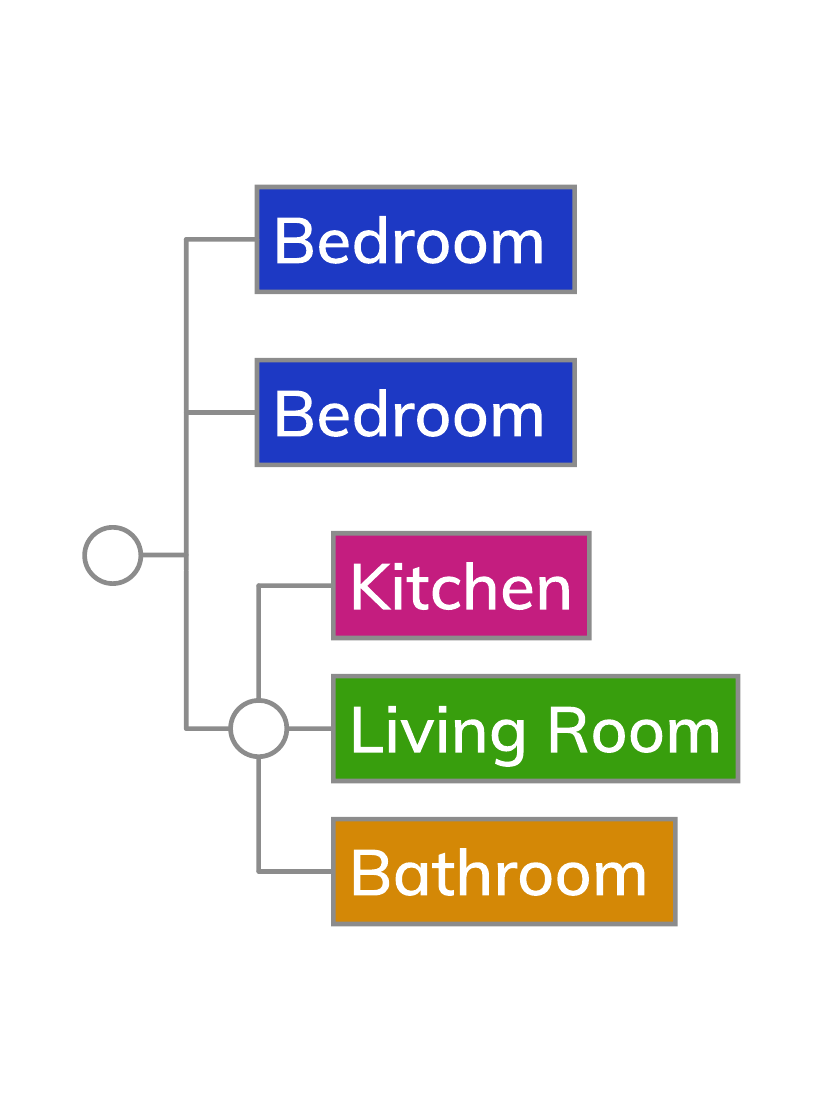}
        \caption{5-Room House}
    \end{subfigure}
    \hfill
    \begin{subfigure}[b]{0.32\textwidth}
        \centering
        \includegraphics[width=\textwidth]{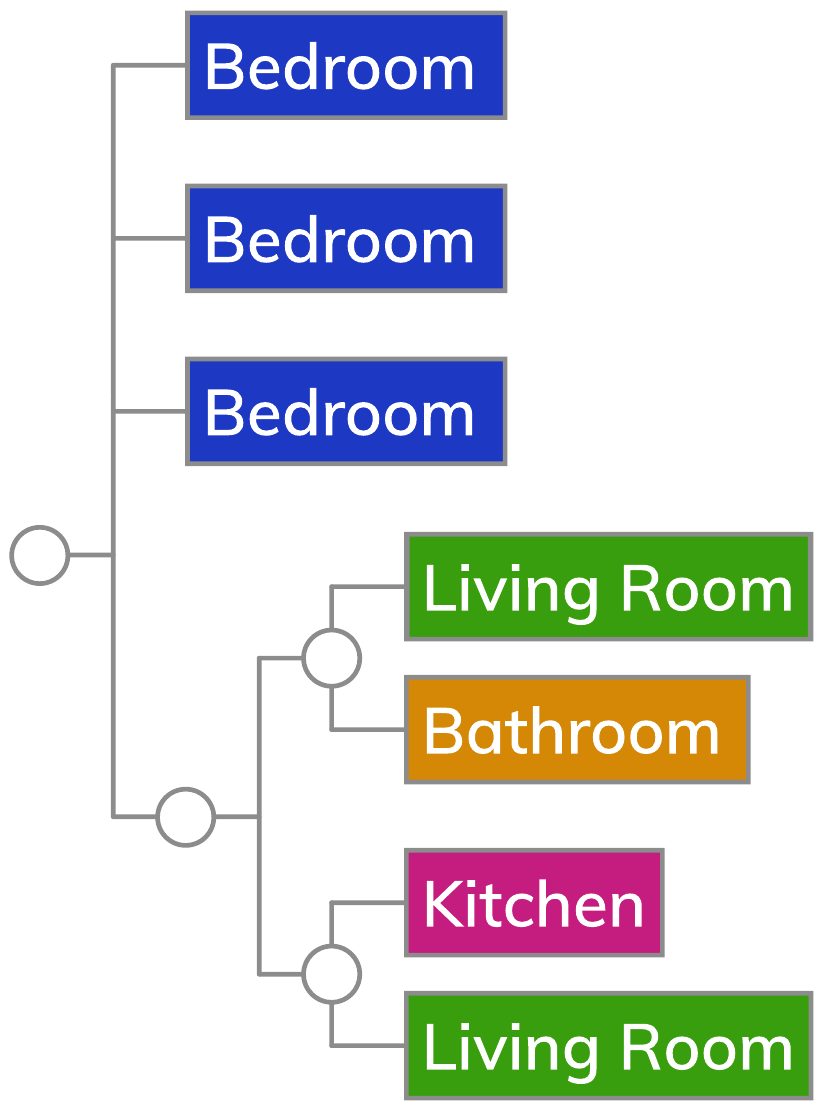}
        \caption{7-Room House}
    \end{subfigure}
    \caption{Examples of room spec hierarchies used to sample differently sized houses.}
    \label{fig:roomSpecExs}
\end{figure}

Room specs provide the ability to specify the rooms that appear in a house, the relative size of each room, and how the rooms are connected with doors. Their idea was first proposed in \cite{marson2010automatic}. A room spec is manually specified with a tree data structure.

Figure \ref{fig:roomSpecExsA} shows a simplified example of a room spec with four rooms: bedroom, bathroom, kitchen, and living room. In this room spec, there are two subtrees, comprising $\mathcal Z_{bb} = \{$\text{bedroom}, \text{bathroom}$\}$ and $\mathcal Z_{klv} = \{$\text{kitchen}, \text{living room}$\}$. At each level of the tree, there is a constraint that there must be a direct path connecting every child node of a parent. Thus, in our example, there will be a path between the bedroom and the bathroom, a path between the kitchen and the living room, and another path connecting $\mathcal Z_{bb}$ to $\mathcal Z_{klv}$. We can also specify which room types we would prefer not to have a path between it and the parent. For example, we typically do not want the bathroom to have 2 doors, such as between it and the bedroom and between it and a room in $\mathcal Z_{klv}$.

Each tree node, below the root of the tree, is also assigned a growth weight, which approximates the relative size of the node compared to all other nodes that share the same parent. For instance, we might assign both $\mathcal Z_{bb}$ and $\mathcal Z_{klv}$ a growth rate of $1$, to be roughly the same size. But, if we want the bedroom to take up roughly $60\%$ of the $\mathcal Z_{bb}$'s area, then we might assign the bedroom a growth rate of $3$ and the bathroom a growth rate of $2$.

Room specs allow us to flexibly choose the distribution of houses we sample, allowing us to specify massive mansions, studio apartments, and anything in-between. Moreover, just a few room specs can go a long way. To generate our houses, we use 16 room specs, which each uses between 1 to 10 rooms. To generate the houses dataset, we assign a sampling weight to each of our room specs, and then use weighted sampling to sample a room spec for each house.

\subsection{Sampling Floor Plans}

\begin{figure}[htbp]
    \centering
    \includegraphics[width=\textwidth]{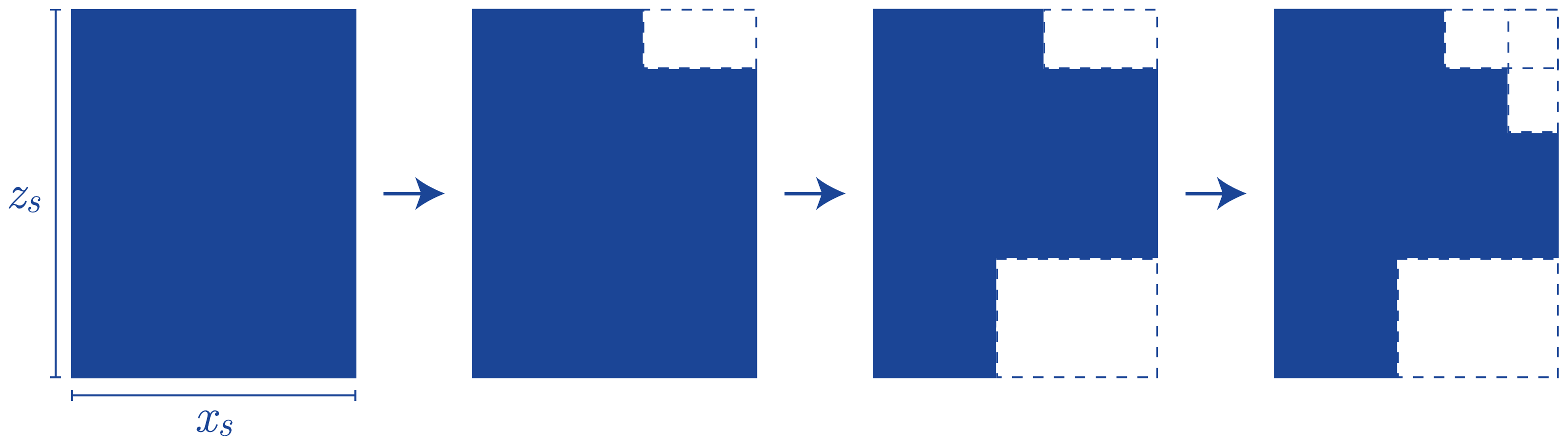}
    \caption{An example of the interior boundary cut algorithm. The images show a top-down view of the house's floor plan. First, we sample an interior boundary rectangle $(x_s, z_s)$, which is shown on the left. Then, we make $n_c$ rectangular cuts to the corners of the rectangle to make the interior boundary of the house a more complex polygon. In this case, we make $n_c=3$ cuts to form the final interior boundary, which is shown on the right.}
    \label{fig:interior}
\end{figure}

The size and shape of the house are sampled to form the interior boundaries. Room specs specify the distribution over the dimensions of the house. Figure \ref{fig:interior} visualizes the process of sampling an interior boundary, where we first sample the size of the boundary and then make cuts to the corners to add randomness. The sampling starts off by choosing the initial upper bound of the top-down $x$ and $z$ size of the house, in meters, respectively denoted as $x_s$ and $z_s$. Each dimension is an integer. In most room specs, each dimension is independently sampled from the discrete uniform distribution $x_s, z_s \sim U(\max(\ell_{\min}, \mu_a\sqrt{n_r} - \nicefrac{\mu_a}{2}), \mu_a \sqrt{n_r} + \nicefrac{\mu_a}{2})$, inclusive. However, individual room specs can override the $x_s$ and $z_s$ sampling distributions. Here, $n_r$ represents the number of rooms in the house, $\ell_{\min}$ is set to $2$ and represents the minimum size of $x_s$ and $z_s$, and $\mu_a$ is set to $3$ and represents the average size of $x_s$ and $z_s$ per room.

\begin{figure}[htbp]
    \centering
    \includegraphics[width=\textwidth]{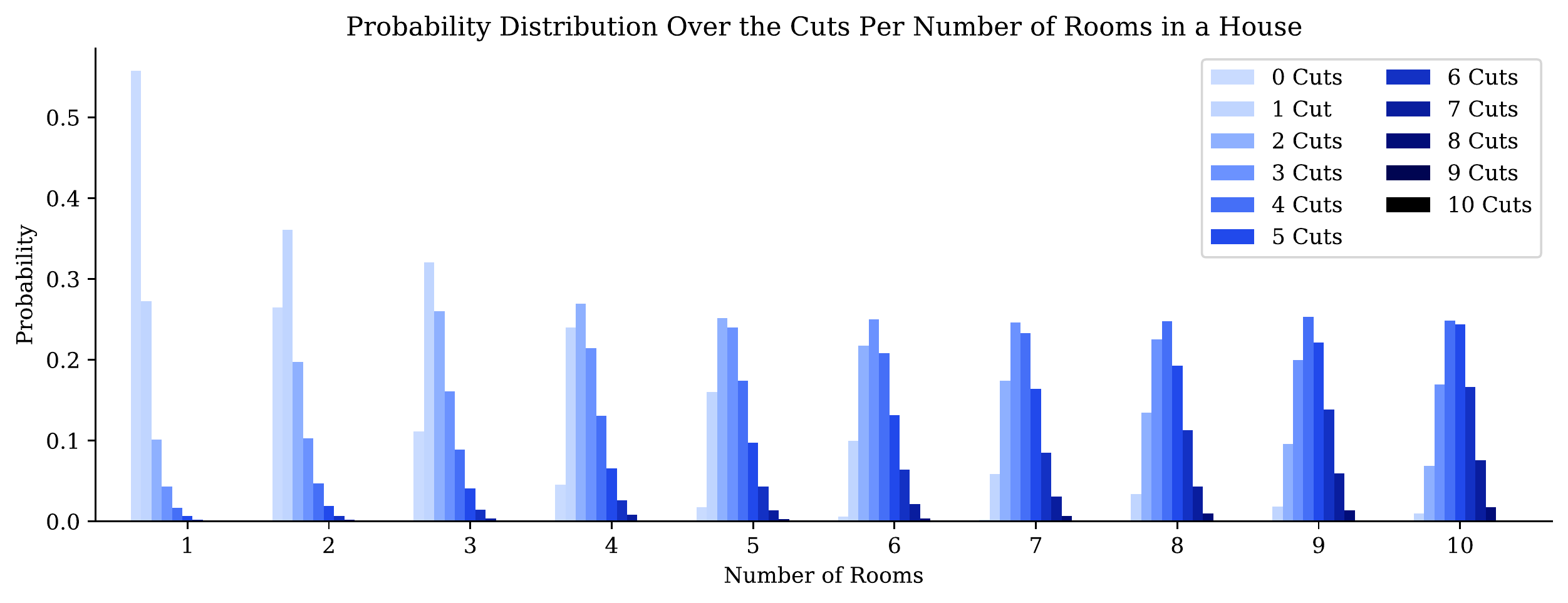}
    \caption{The probability distribution over the number of cuts, $n_c$, made to the rectangular boundary ($x_s$, $z_s$) with respect to the number of rooms in the house, $n_r$. Notice that when there are more rooms in the house, the number of cuts in the distribution increases.}
    \label{fig:ncuts}
\end{figure}

Once we have the rectangular boundary $(x_s, z_s)$, we then make several \textit{cuts} to the outside of the rooms such that the interior boundaries can take on the shape of more complex polygons. The number of cuts, $n_c$, is sampled from the distribution $n_c\sim \lfloor 10\cdot \text{Beta}(\alpha_c, \beta_c) + \nicefrac{1}{2}\rfloor$, where $\alpha_c=\nicefrac{n_r}{2}$ and $\beta_c = 6$. Figure \ref{fig:ncuts} shows the distribution that is formed with respect to the number of rooms in the house, $n_r$. When there are more rooms, the probability distribution over the number of cuts increases. Since the range of the beta distribution is $(0, 1)$, the upper bound on the number of cuts is exactly 10.

The size of each cut is a rectangle, in meters, denoted by $(c_x$, $c_z)$. Both $c_x$ and $c_z$ are sampled from integer distributions. We sample from $c_x\sim U(1, \max(2, \min(x_s - 1, \lfloor a_{\max} / 2\rfloor) - 1)$, inclusive, where $a_{\max}$ is set to $6$ representing the maximum cut area. We then sample $c_z\sim U(1, a_{\max} - c_x)$. The position of where the cut happens is anchored to one of the 4 corners of the interior boundary, where the exact corner is independently and uniformly sampled each time.

Since the size of each cut is an integer, and the rectangular boundary sizes are also integers, we can efficiently represent the interior boundary with a $(x_s, z_s)$ boolean matrix. Here, we could have $1$s representing where the inside of the interior boundary and $0$s representing the outside of the interior house boundary.

\begin{figure}[htbp]
    \centering
    \includegraphics[width=\textwidth]{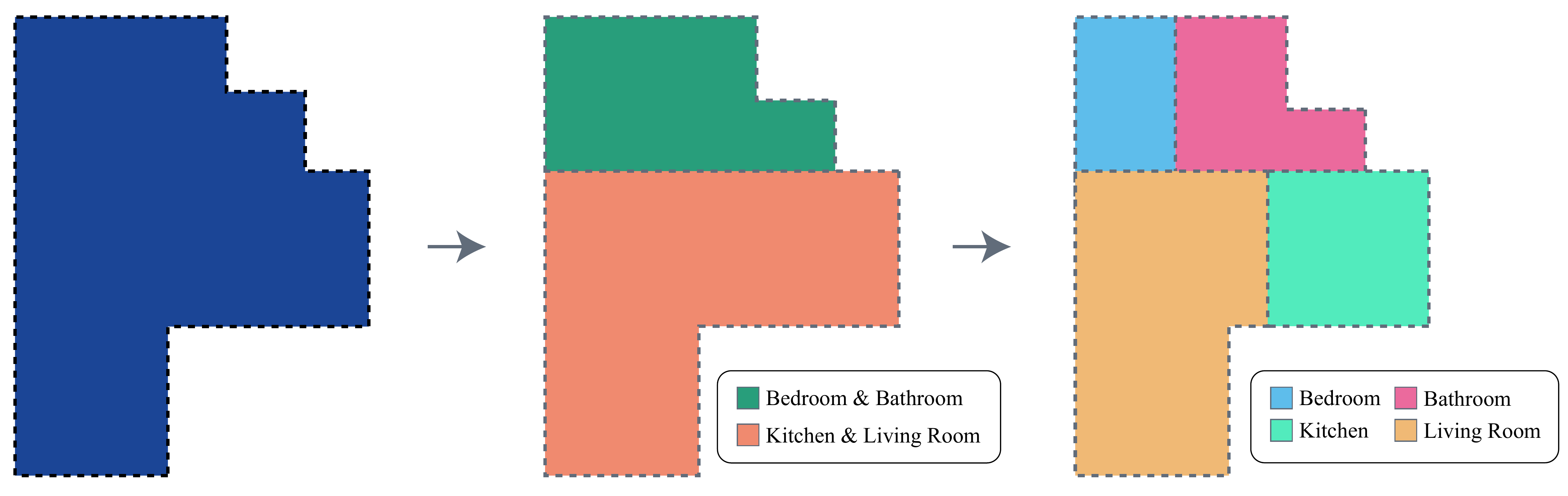}
    \caption{An example of the recursive floor plan generation algorithm, given an interior boundary and the room spec in Figure~\ref{fig:roomSpecExsA}. Here, we first divide the room into a ``bedroom \& bathroom'' and a ``kitchen \& living room'' zone. Then, within the ``bedroom \& bathroom'' zone we place both the bedroom and bathroom, and within the ``kitchen \& living Room'' zone, we place both the kitchen and living room.}
    \label{fig:recFP}
\end{figure}

Given a room spec and an interior boundary, we use the algorithm proposed in \cite{lopes2010constrained} to divide the interior boundary into rooms. The algorithm recursively subdivides the interior boundary for each subtree in the room spec. Figure \ref{fig:recFP} shows an example using Figure~\ref{fig:roomSpecExsA}'s room spec. The algorithm first divides the interior boundary into two zones, the ``bedroom \& bathroom'' zone and the ``kitchen \& living room'' zone. The ``bedroom \& bathroom'' zone then subdivides into two rooms, the bedroom and bathroom. Similarly, the ``kitchen \& living room'' zone is also subdivided into two rooms, the kitchen and living room. The growth weight is used to approximate the size of each subdivision. By recursively subdividing the zones of each subtree, we satisfy the constraint that we can traverse between child nodes of the same parent in the room spec.

Finally, we scale the entire floor plan by $s\sim U(1.6, 2.2)$. Scaling the interior boundary to be larger provides more room for the agent to be able to navigate within the houses. Using a range of values also provides more variability on the size of the houses. We set the upper bound to 2.2 based on the empirical quality of the houses, where values above that often left too much empty space.

\subsection{Connecting Rooms}

\begin{figure}[htbp]
    \centering
    \begin{subfigure}{0.32\textwidth}
        \includegraphics[width=\textwidth]{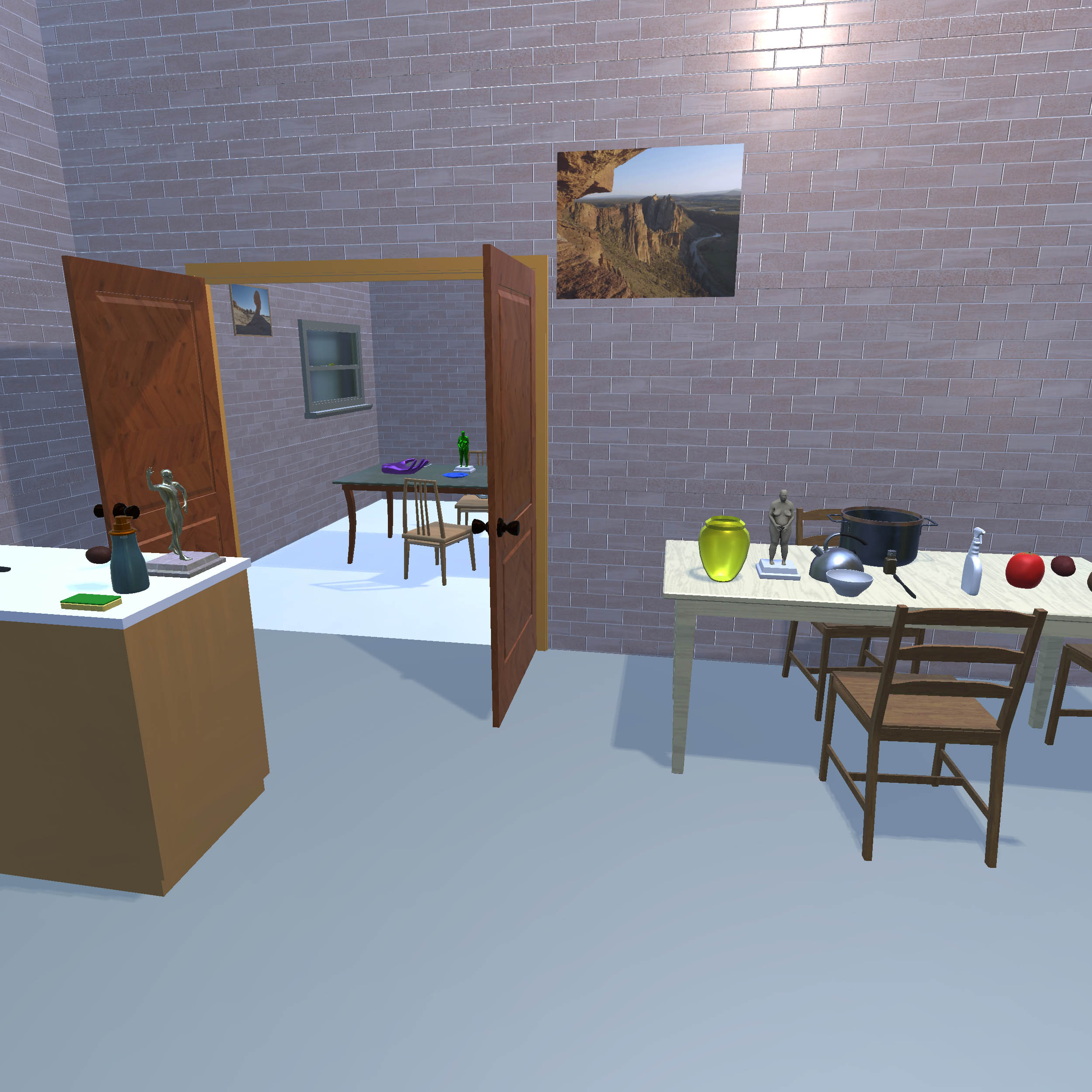}
        \caption{Doorway Connection}
    \end{subfigure}
    \begin{subfigure}{0.32\textwidth}
        \includegraphics[width=\textwidth]{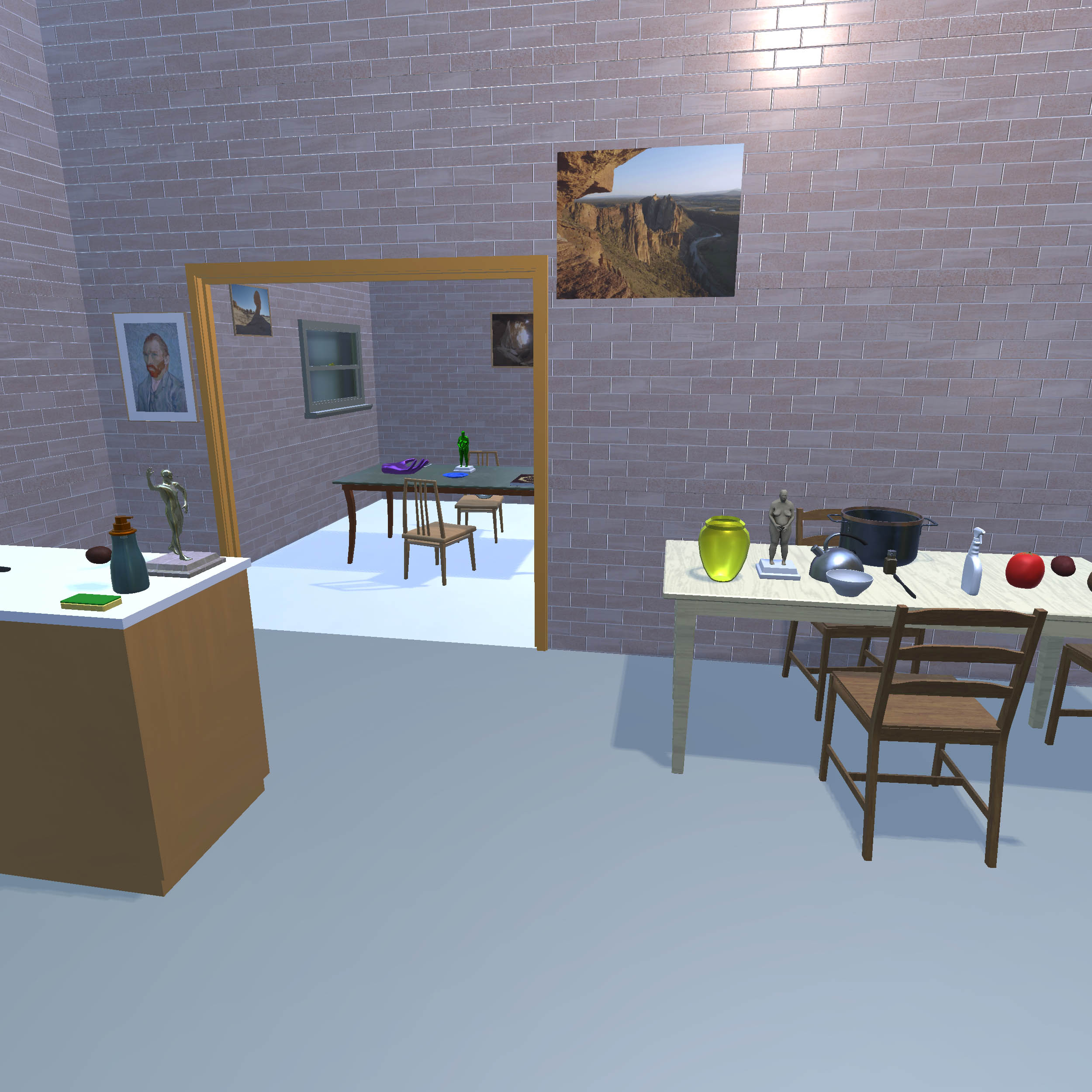}
        \caption{Door Frame Connection}
    \end{subfigure}
    \begin{subfigure}{0.32\textwidth}
        \includegraphics[width=\textwidth]{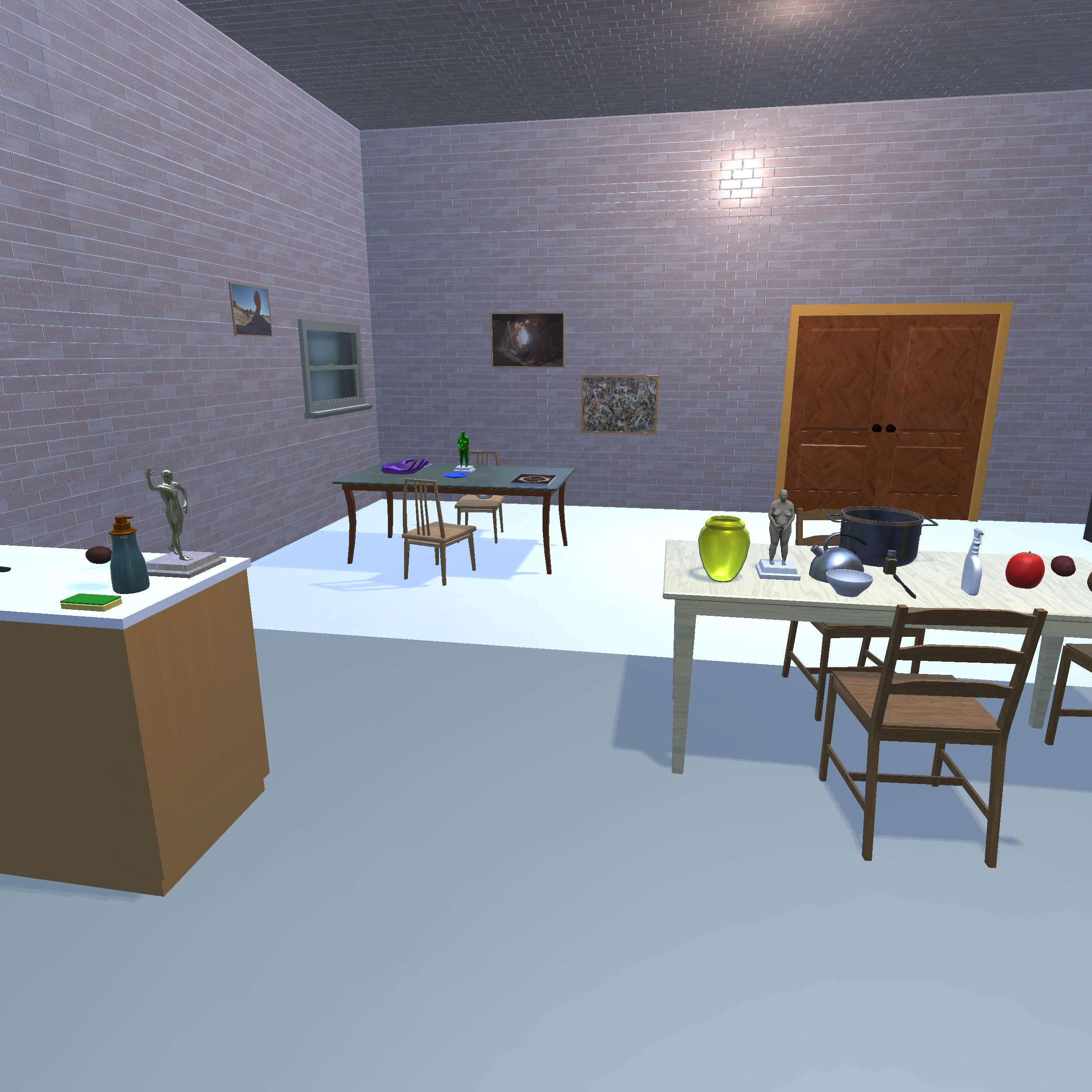}
        \caption{Open Room Connection}
    \end{subfigure}
    \caption{An example of the 3 ways to connect different rooms, using either a doorway, door frame, or open room connection.}
    \label{fig:door-connections}
\end{figure}

Figure \ref{fig:door-connections} shows the 3 types of ways adjacent rooms may be connected. Specifically, rooms may be connected using 3 different types of connections: doorways, door frames, or open room connections. We determine which rooms should have doors between them based on the constraints in the room spec. Amongst adjacent rooms that may have doors between them, subject to the constraints in the room spec, we randomly sample which rooms have doors. We also impose the constraint that neighboring rooms in the room spec may have at most 1 room connected to it.

To choose the type of connection, we consider the rooms we are connecting. Specifically, we only allow open room connections and door frame connections between kitchen and living room room types. We impose this constraint because it would be unrealistic for a room like a bathroom to be fully visible from another room. For connecting room types that do support open room connections or door frames, we annotate the probability of sampling a doorway, door frame, and open room connection. Between a kitchen and living room the probability is 0.375 for sampling both an open room connection and a door frame connection, and 0.25 for sampling a doorway connection.

If a doorway or door frame is sampled, we filter to use a valid asset that is smaller than the wall connecting the rooms. For our generation, the minimum wall size is always greater than a single door size, but occasionally the filter might remove double doors from valid doors that can be sampled as they would be too big. The placement of the door is then uniformly sampled from anywhere along the wall. For doorways, the open direction is uniformly sampled. Finally, if the open state from any 2 doorways collides, we also use rejection sampling to potentially change the open direction and modify the placement of doorways.

Each house also has a permanently closed exterior door connecting to the outside. We prioritize placing this door in kitchen and living room room types, as it is unnatural to have to go through a bathroom or bedroom to go outside. However, in the case where the room spec does not include a kitchen or living room (\textit{e.g.} if the room is a standalone bathroom), we randomly place a door to the outside in one of the remaining rooms.

\subsection{Structure Materials}

\textbf{Wall materials.} To choose the materials that make up the walls, we consider 2 families of wall materials: solid colors and texture-based materials. Our solid color materials consist of 40 unique colors of popular paint colors found in houses. We constrain ourselves to only using popular paint colors, so we do not randomize the walls to unrealistic colors such as bright green or yellow. For the texture-based materials, we annotate 122 different AI2-THOR materials to be suitable as wall materials. These include materials for brick textures, drywall textures, and tiling textures, amongst others.

Each wall in a room shares the same materials. For each room, we sample it if its materials are a solid color with $w_{\textit{solid}} \sim \text{Bernoulli}(0.5)$. It is sometimes the case in real life that all rooms in a house share the same material (\textit{e.g.} every room in an apartment is painted with white walls). We therefore also have a parameter $w_{\textit{same}} \sim \text{Bernoulli}(0.35)$ that specifies if all rooms in the house will have the same material.

\textbf{Ceiling material.} The entire ceiling of the house is always assigned to a single wall material. If $w_{\textit{same}}$, then the ceiling material is also set to the wall material. Otherwise, it is independently sampled with the same wall material sampling process.

\textbf{Floor materials.} We annotate 55 materials in AI2-THOR as floor materials. Most commonly, these materials are wood materials. For each room, we independently sample its floor material from the set of annotated floor materials. However, similar to wall materials, we independently sample $f_{\textit{same}} \sim \text{Bernoulli}(0.15)$ that specifies if all rooms in the house will have the same material.

\subsection{Ceiling Height}

\begin{figure}
    \centering
    \includegraphics[width=0.5\textwidth]{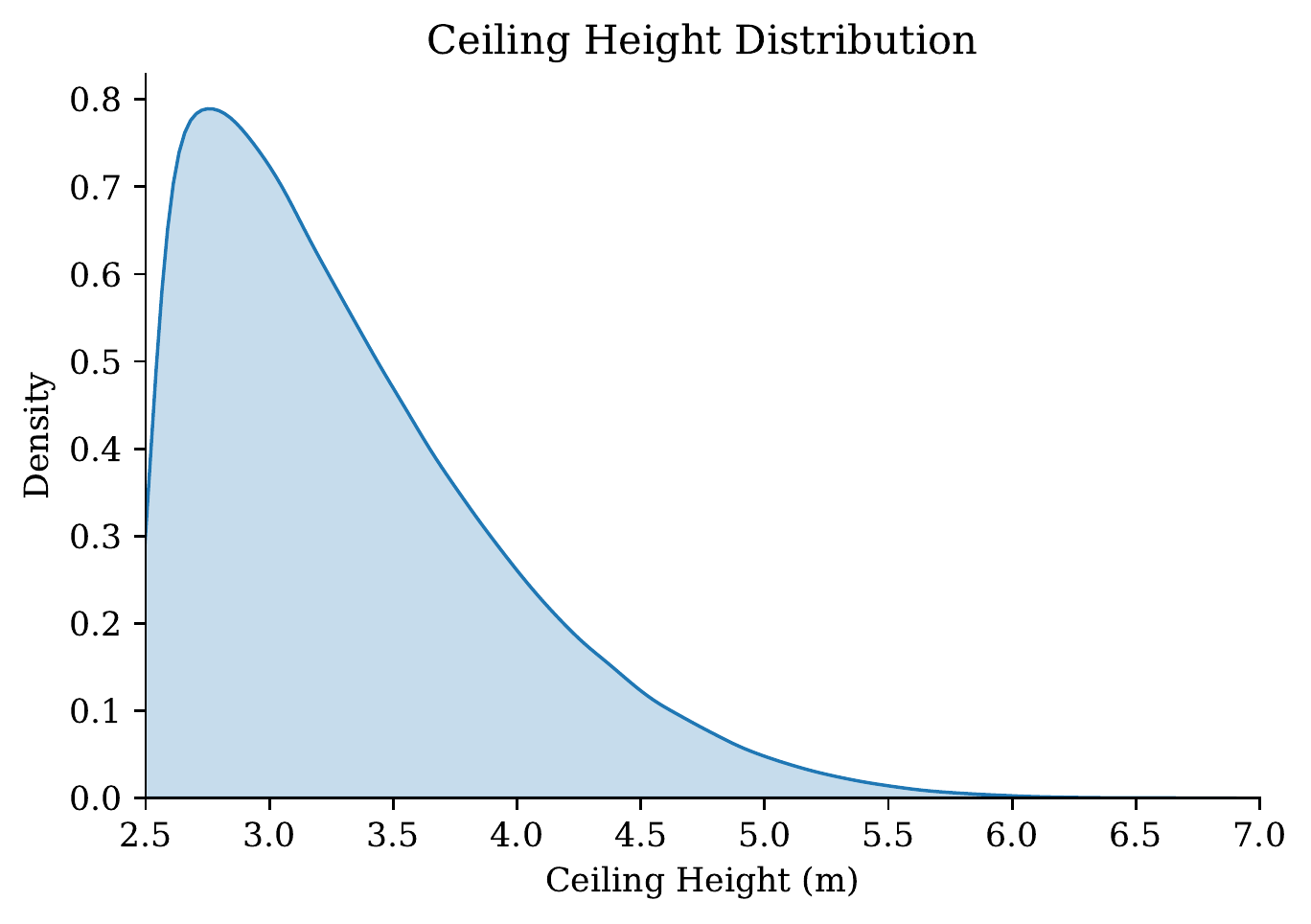}
    \caption{The distribution of the ceiling height of each house, in meters.}
    \label{fig:ceilingHeight}
\end{figure}

The ceiling height for the house, in meters, is sampled from $c_h\sim h_{\min} + (h_{\max} - h_{\min})\cdot \text{Beta}(\alpha_h, \beta_h)$, where we set $h_{\min}=2.5$, $h_{\max}=7$, $\alpha_h=1.25$, and $\beta_h=5.5$. Figure \ref{fig:ceilingHeight} shows the ceiling height distribution that is formed. All rooms in the house have the same ceiling height.

The minimum and mean values were chosen based on the typical height of an American apartment, while $\beta_h$ allows some of the train houses to have much larger ceilings.

\subsection{Lighting}

\textbf{Lighting Placement.} Each procedural house places two types of lights: a directional light and point lights. The directional light is analogous to the sun in the scene, where only 1 is placed in each scene. Light from point lights are analogous to the light emitted from lightbulbs. We place a point light in each room near the ceiling, centered at the centroid of the room's floor polygon. Using the centroid ensures that the light is always placed inside of the room, even for L-shaped rooms. Additionally, desk lamp and floor lamp objects have a point light associated with them.

\begin{figure}[htbp]
    \centering
    \begin{subfigure}{0.325\textwidth}
        \includegraphics[width=\textwidth]{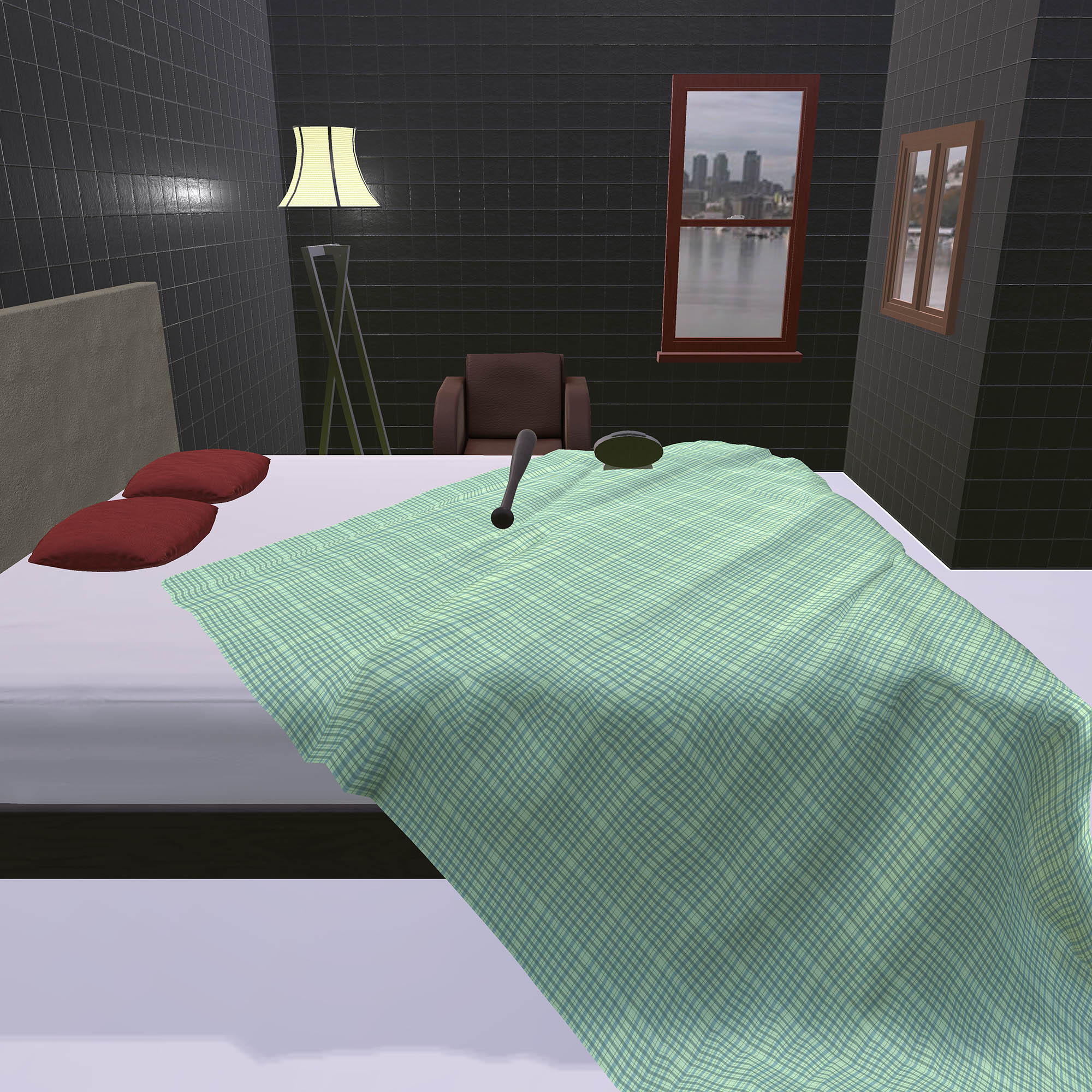}
        \caption{Midday Skybox}
    \end{subfigure}
    \begin{subfigure}{0.325\textwidth}
        \includegraphics[width=\textwidth]{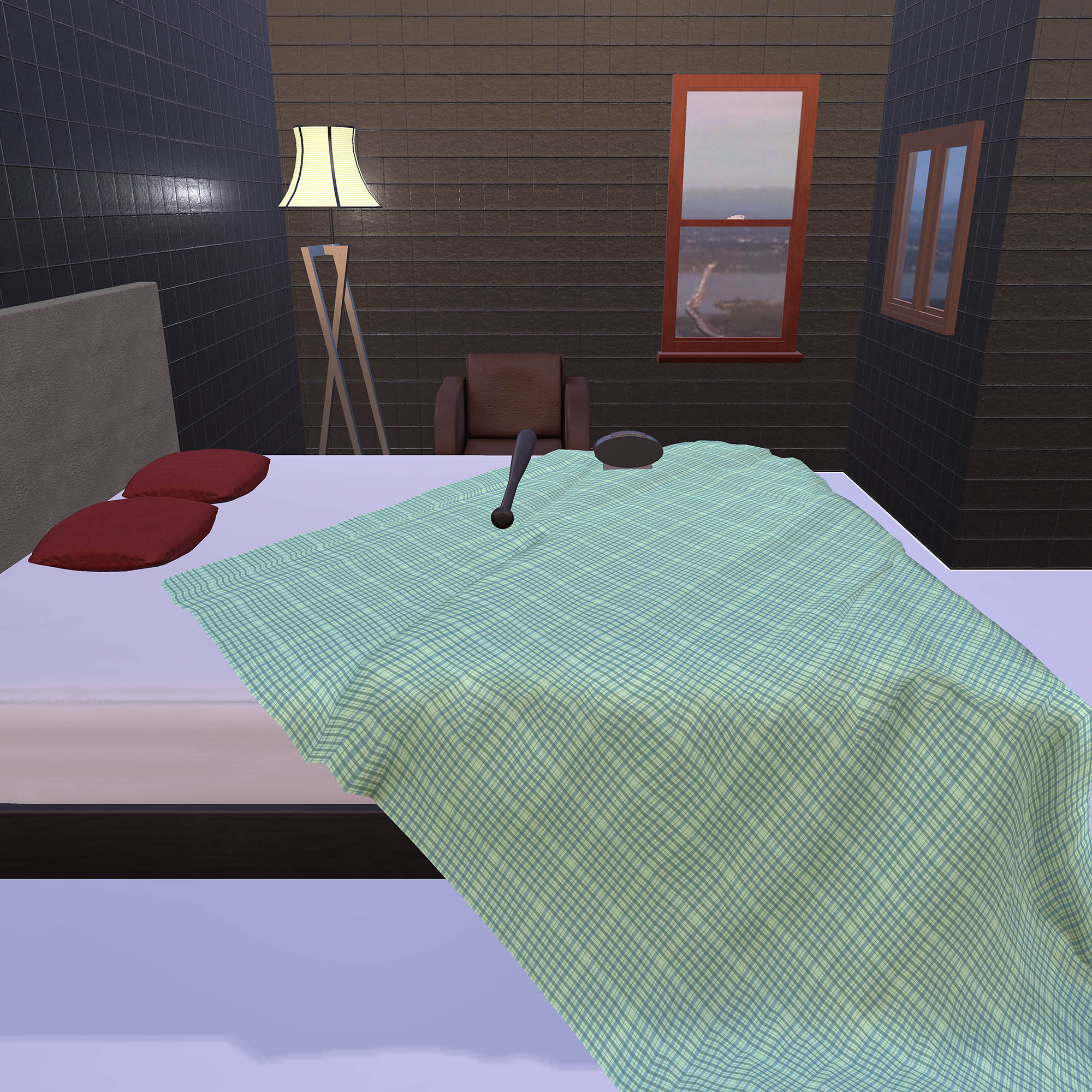}
        \caption{Golden Hour Skybox}
    \end{subfigure}
    \begin{subfigure}{0.325\textwidth}
        \includegraphics[width=\textwidth]{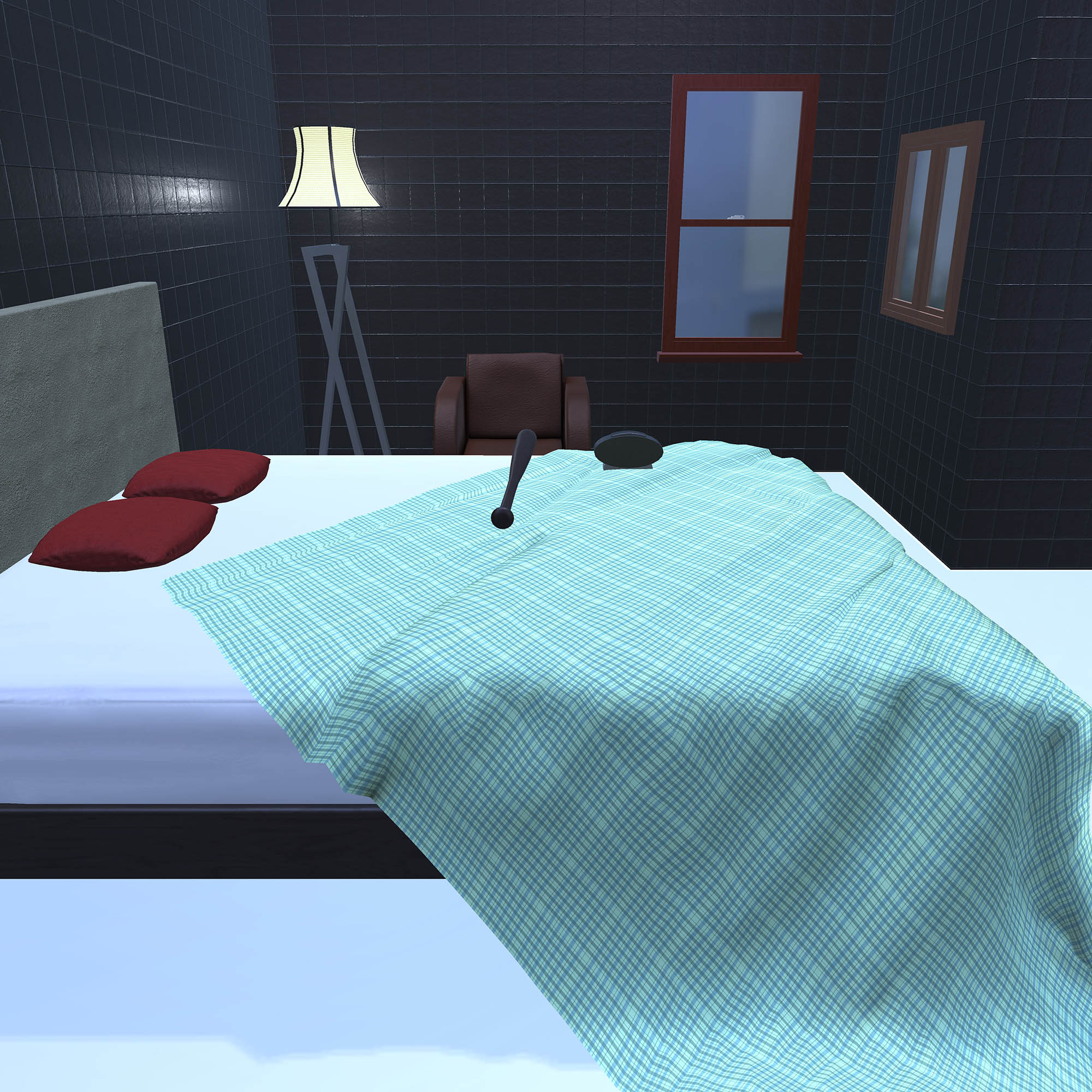}
        \caption{Blue Hour Skybox}
    \end{subfigure}
    \caption{Examples different skyboxes in a scene. Notice how the colors of the images differ and how the content outside of the window changes with the skybox.}
    \label{fig:skyboxes}
\end{figure}

\textbf{Effects by the time of day}. Skyboxes may appear at 3 different times of day: midday, golden hour, and blue hour. The time of day determines the intensity, hue, and direction of the ambient outdoor lighting. For each time of day, there exist multiple \textit{skyboxes}, which dictate the lighting of the environment. Figure \ref{fig:skyboxes} shows examples of how the time of day visually affects the scene. At this time, there are 16 midday skyboxes, 5 golden hour skyboxes, and 1 blue hour skybox, based on full 360-degree photos taken in Seattle and San Francisco.

\subsection{Object Placement}
\label{sec:op}

In this section, we discuss how objects are placed realistically in the house. We hypothesize reasonable object placement is necessary in order to train efficient agents. For instance, if a toilet could appear anywhere in the house, the agent would have a much harder search problem, leading to longer episodes, than if the toilet was always in the bathroom. Moreover, we do not want objects to appear in unnatural positions, such as a fridge facing the wall, as it would make it unnatural, and even unusable, for interaction.

Finally, we do not always want objects to spawn independently. For instance, we might want a table to be surrounded by chairs. We achieve dependant sampling by developing SAGs, which are described in the section that follows.

\subsubsection{Assets}
\label{sec:assets}

The ProcTHOR asset database consists of 1,633 interactive household assets across 108 object types (see Appendix~\ref{sec:procthorObjects} for more details). The majority of assets come from AI2-THOR. Windows, doors, and counter tops are built into the exterior of rooms in AI2-THOR, which prevents us from spawning them in as standalone assets. Thus, we have also hand-built 21 windows, 20 doors, and 33 counter tops.

\begin{figure}[htbp]
    \centering
    \includegraphics[width=1\textwidth]{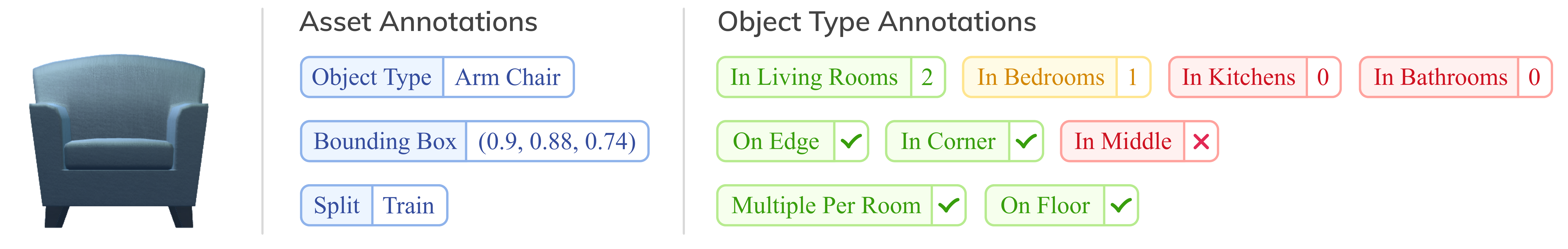}
    \caption{An example of the asset annotations used to place an arm chair asset. This particular instance is annotated with its object type, bounding box, and split. Annotations about how it is placed in the house are done at an object type level, applying to all instances of that type.}
    \label{fig:assetAnnotations}
\end{figure}

\textbf{Asset Annotations.} Our assets include several annotations that help us place them realistically in a house. Figure~\ref{fig:assetAnnotations} shows an example of the asset annotations used to place an arm chair. For an individual asset, we annotate its object type, computationally obtain its 3D bounding box, and partition assets of object types into training, validation, and testing splits. Then, we annotate how each object type might be spawned into the house. Annotating the 108 object types, as opposed to annotating the 1,633 individual assets, allows us to scale up the number of unique assets dramatically. Moreover, it does not require any new annotation to add an asset that can be grouped with an existing object type.

If instances of an object type cannot be placed independently on the floor, the rest of its annotations are not considered. For instance, we do not allow television object types to be placed alone on the floor, rather they are often placed on top of a television stand or mounted on the wall, which is discussed later in this section. Similarly, we also annotate small objects, like a fork, pen, and mug to not be placed independently on the floor. However, typical large object types, such as counter top, arm chair, or fridge object types can be placed independently on the floor.

Among the remaining object types, we annotate where and in which rooms the object type may appear. Each object type has a room weight, $r_w\in \{0, 1, 2, 3\}$, corresponding to how likely it is to appear in each room type. For each room type, a $0$ indicates the object should never appear (e.g., a fridge in a bathroom); a $1$ indicates the object may appear, but is unlikely; a $2$ indicates that the object appears quite often; and a $3$ indicates that the object nearly always appears (e.g., a bed in a bedroom). To determine where the object is placed, we annotate whether it may appear on the edge, in the corner, or in the middle of a room. For example, we annotate that a fridge can be placed on the edge or in the corner of the room, but not in the middle. We also annotate whether there can be multiple instances of an object type in a single room. Here, we annotate that multiple toilet object types cannot be in the same room, for instance.

\noindent\textbf{Asset Splits.} If an object type has over 5 unique assets, then those assets are partitioned into train, validation, and testing splits. Specifically, approximately $\nicefrac{2}{3}$ of the assets are assigned to the train split, and approximately $\nicefrac{1}{6}$ of the assets are assigned to each of the validation and testing splits. For object types that have 5 or fewer unique assets, they may appear in any split. In general, the more visual diversity an object type has, the more instances of that object type exist. For instance, there are many chair objects, but there are much fewer CD, toilet, and fork objects. Appendix~\ref{sec:procthorObjects} shows the precise count of each object type.

\subsubsection{Semantic Asset Groups (SAGs)}

\begin{figure}[htbp]
    \centering
    \begin{minipage}{0.58\linewidth}
        \includegraphics[width=1\linewidth]{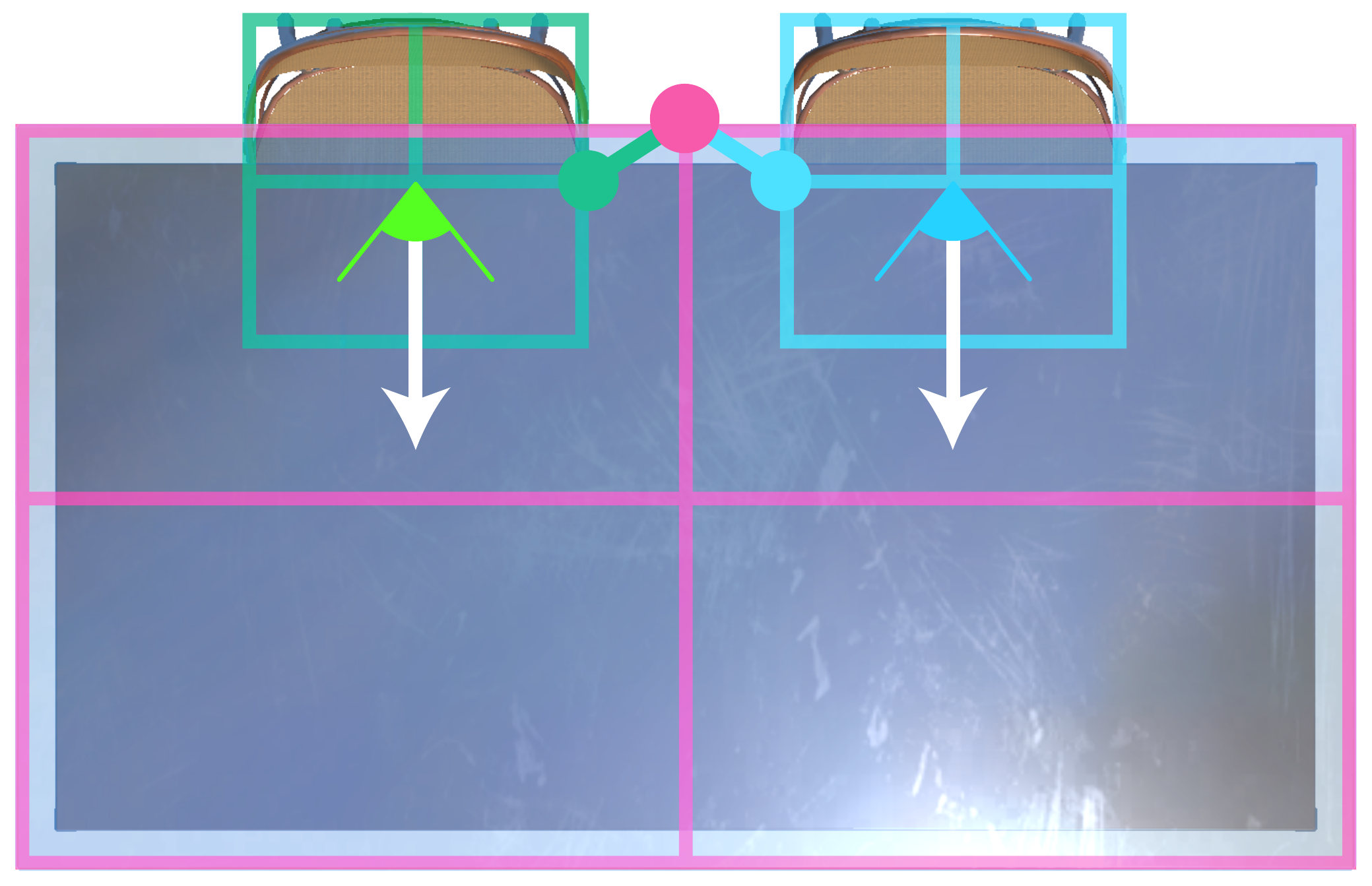}
        \subcaption{An interface for viewing SAGs showing child asset anchoring and rotational randomness.}
        \label{sub:anchor}
    \end{minipage}
    \qquad\quad
    \begin{minipage}{0.22\linewidth}
        \includegraphics[width=1\linewidth]{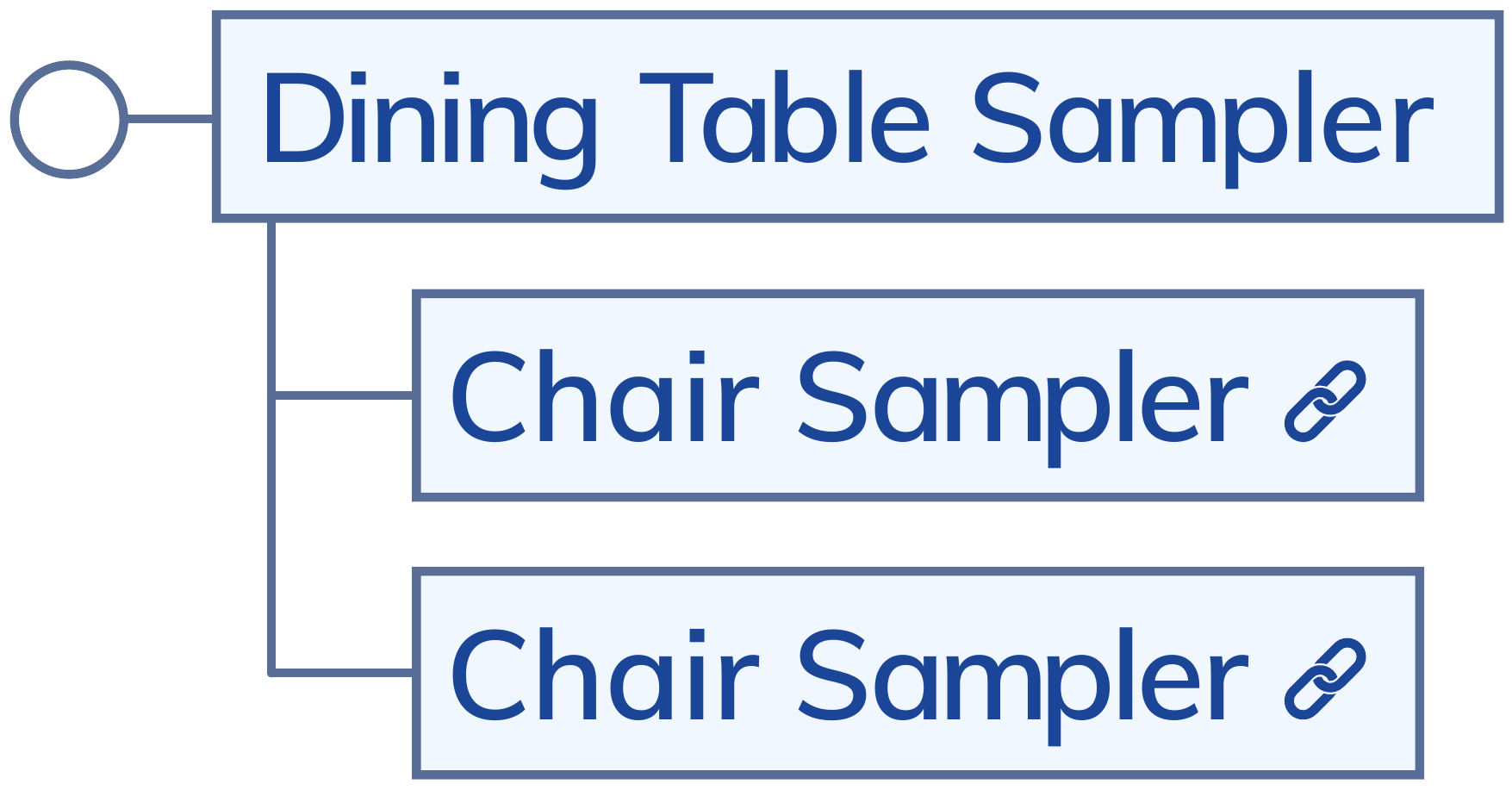}
        \subcaption{Hierarchy}
        \vspace{0.05in}

        \centering
        \includegraphics[width=0.70\linewidth]{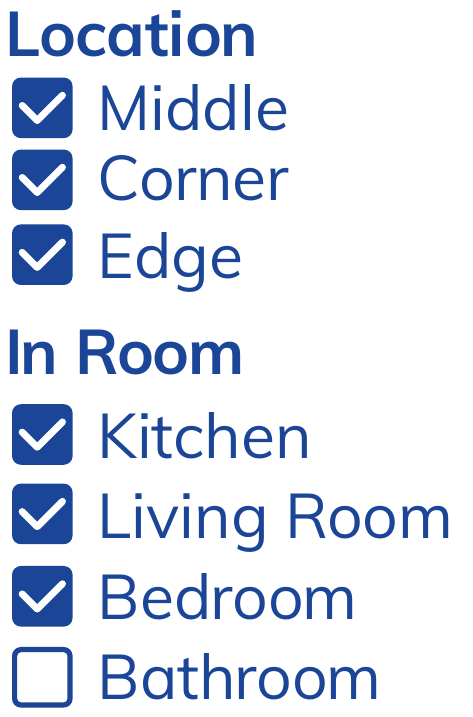}
        \vspace{-0.03in}
        \subcaption{Annotation}
    \end{minipage}
    \caption{An example of a semantic asset group (SAG), where two chair samplers are parented to a dining table sampler. Both chairs are anchored to the top middle of the table.}
    \label{fig:semGroupExamples}
\end{figure}

A \textit{Semantic Asset Group} (SAG) provides a flexible and diverse way to encode which objects may appear near each other. The power of SAGs comes in their ability to support randomized asset and rotational sampling. SAGs can be created and exported in seconds with our user-friendly drag-and-drop web interface.

Figure~\ref{fig:semGroupExamples} shows an example of how we might construct a SAG that has two chairs pushed into the side of a dining table. The SAG includes two chair samplers and a dining table sampler. Asset samplers contain a set of unique 3D modeled asset instances that may be sampled. When the SAG is instantiated, each asset sampler randomly chooses one of its instances. Asset samplers can also be linked, where multiple samplers sample the same asset instance each time. Here, linking may allow for multiple instances of the same chair to be placed at a dining table, instead of independently sampling a different chair for each sampler.

The ability to randomly sample assets to place in a SAG is incredibly expressive. For instance, consider a SAG with samplers for a TV stand, television, sofa, and arm chair. If each of these samplers can sample from just 30 different 3D modeled asset instances, then there are over $800$k unique combinations of instances that can make be sampled from that SAG. 

\begin{figure}[ht!]
    \vspace{-0.05in}
    \centering
    \begin{subfigure}[b]{0.24\textwidth}
        \captionsetup{justification=centering}
        \centering
        \includegraphics[width=0.9\textwidth]{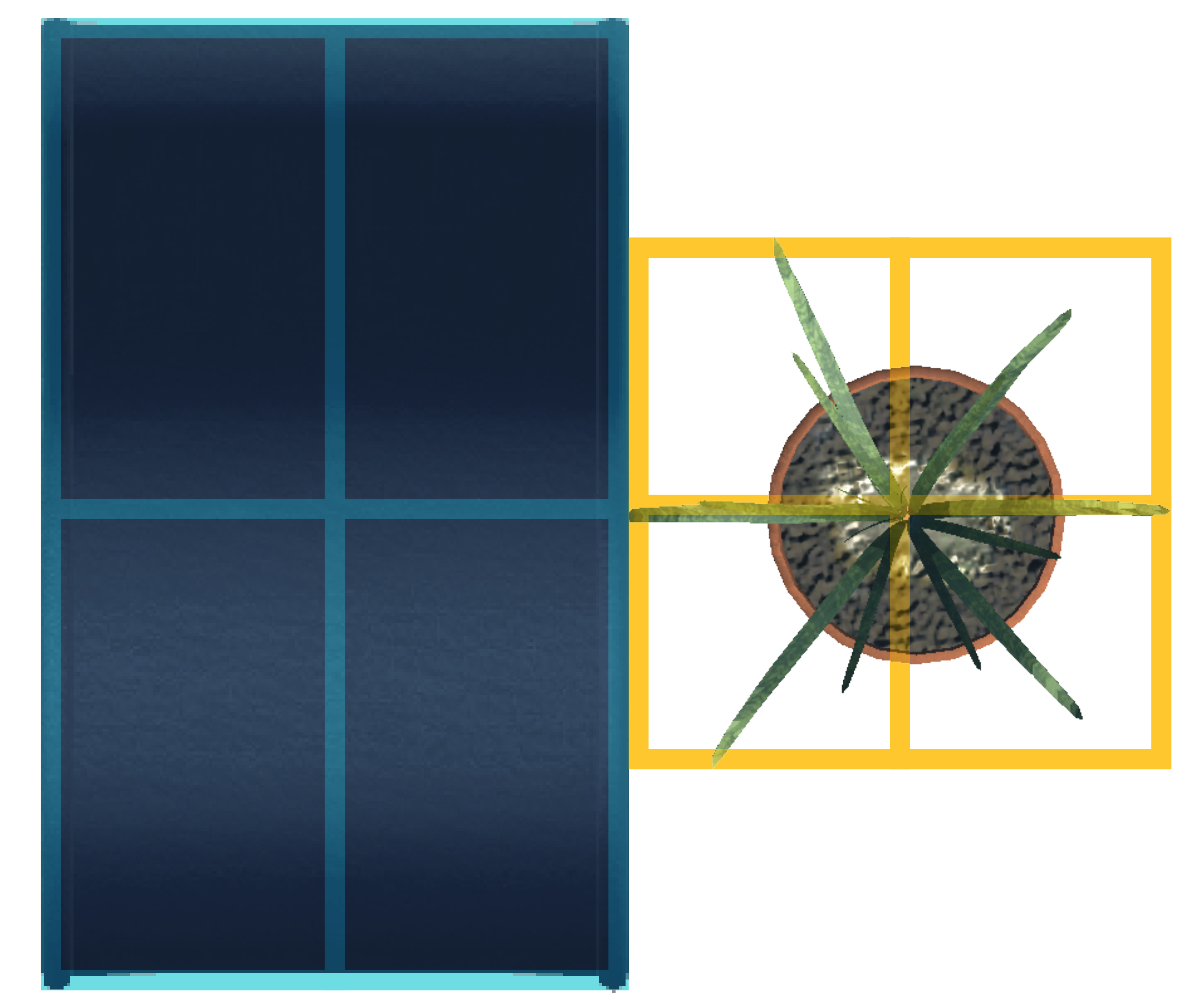}
        \caption{\\ \vspace{0.05in} \scriptsize \textsc{Center Right} Anchor \\ \textsc{Center Left} Pivot}
    \end{subfigure}
    \hfill
    \begin{subfigure}[b]{0.24\textwidth}
        \centering
        \captionsetup{justification=centering}
        \includegraphics[width=\textwidth]{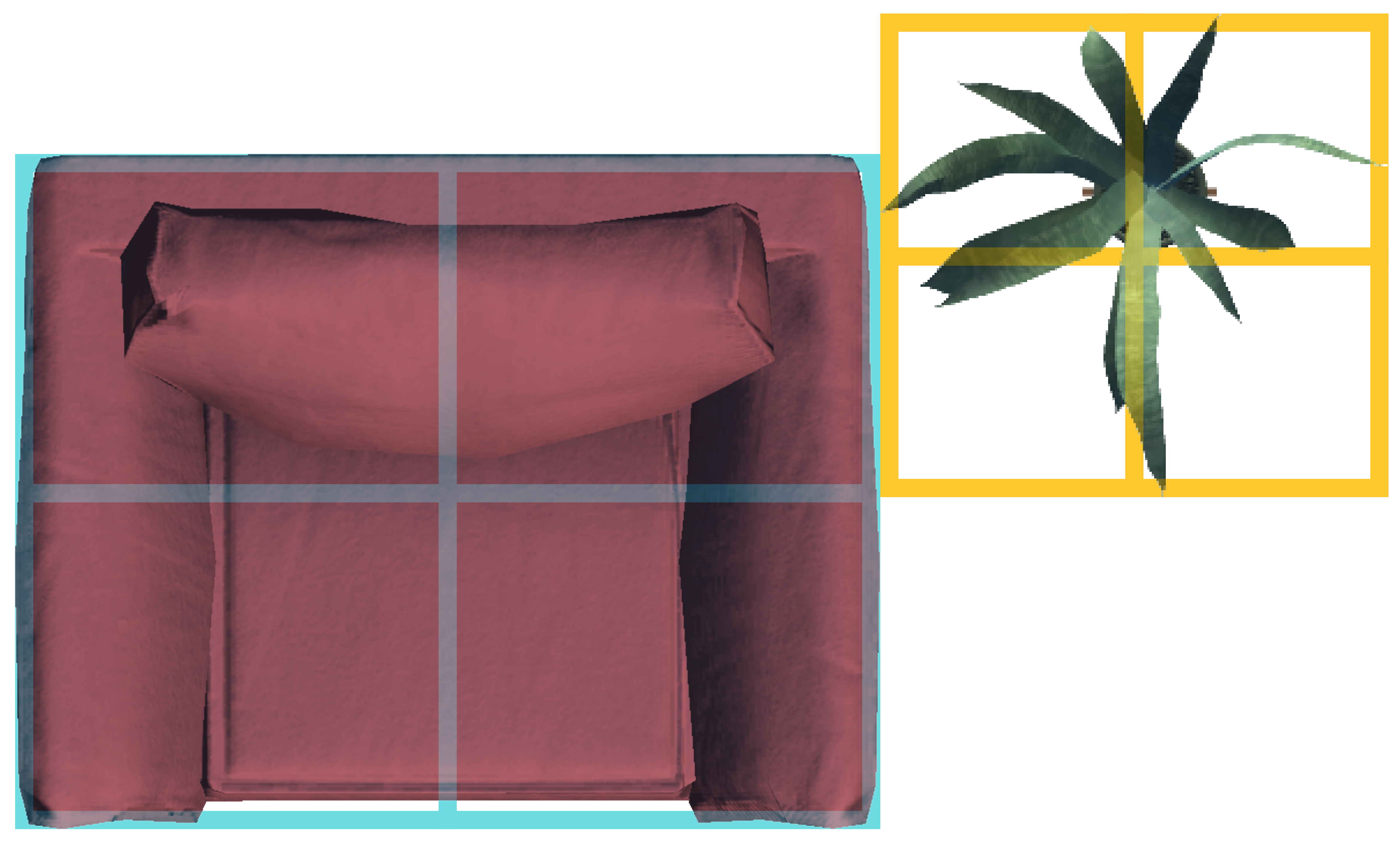}
        \vspace{0in}
        \caption{\\ \vspace{0.05in} \scriptsize \textsc{Center Right} Anchor\\\textsc{Bottom Left} Pivot}
    \end{subfigure}
    \hfill
    \begin{subfigure}[b]{0.24\textwidth}
        \centering
        \captionsetup{justification=centering}
        \includegraphics[width=0.8\textwidth]{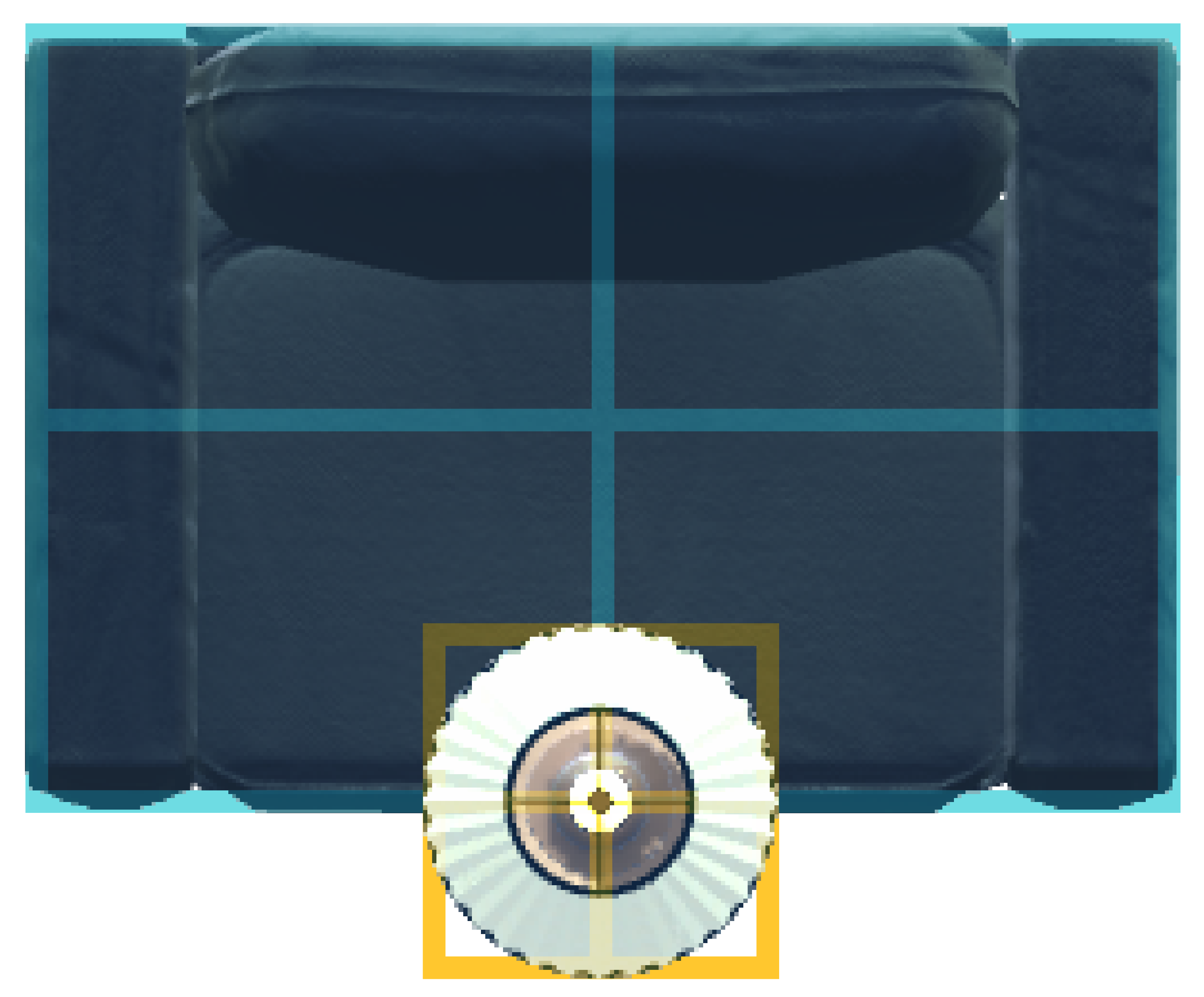}
        \caption{\\ \vspace{0.05in} \scriptsize \mbox{\textsc{Bottom Center} Anchor}\\\textsc{Center Center} Pivot}
    \end{subfigure}
    \hfill
    \begin{subfigure}[b]{0.24\textwidth}
        \centering
        \captionsetup{justification=centering}
        \includegraphics[width=0.6\textwidth]{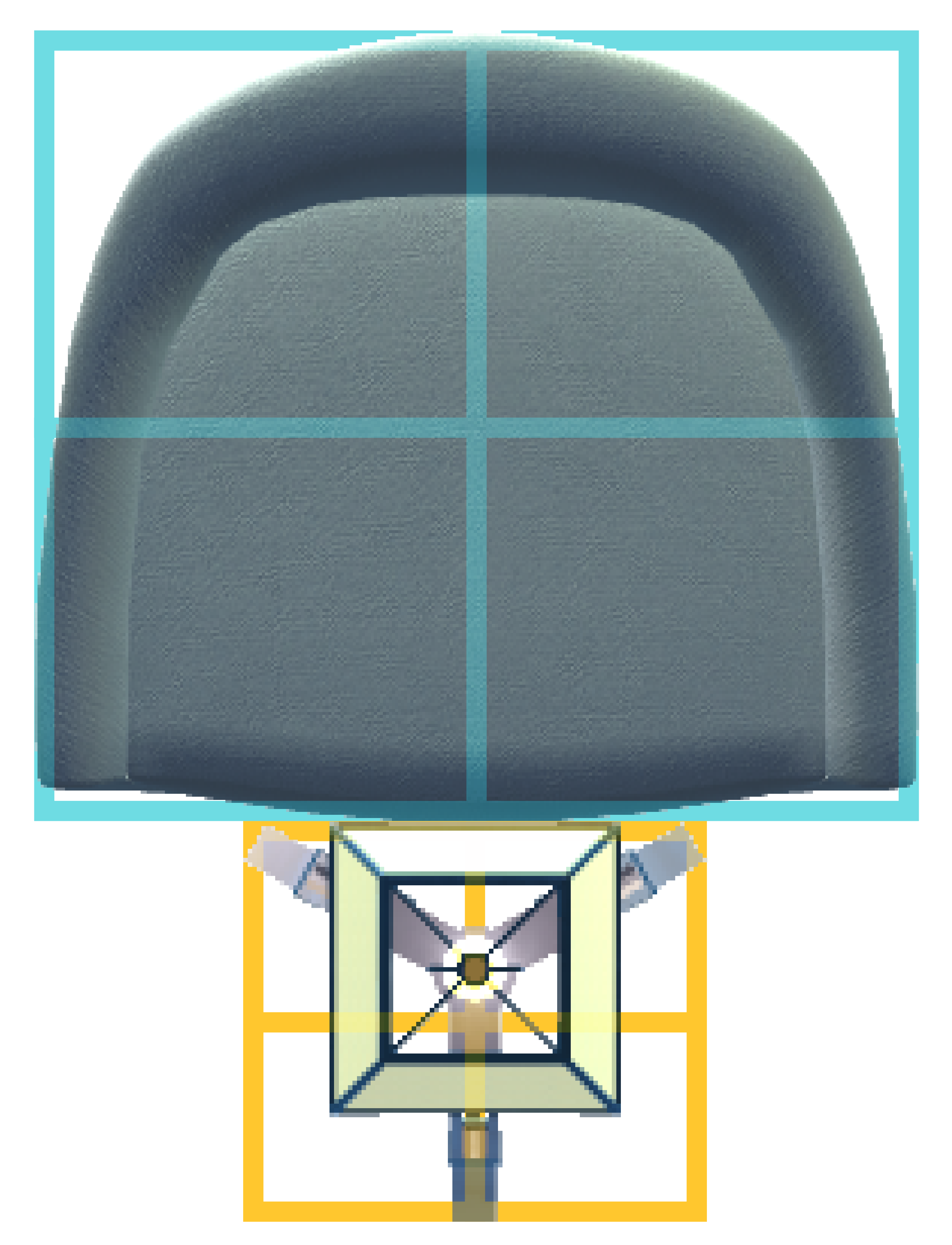}
        \caption{\\ \vspace{0.05in} \scriptsize \mbox{\textsc{Bottom Center} Anchor}\\\textsc{Top Center} Pivot}
    \end{subfigure}
    \caption{
        Instantiations of a SAG that places a plant or floor lamp sampler $\mathcal S_{c}$ around a parented arm chair sampler $\mathcal S_{p}$ with anchor and pivot position annotations. Notice that the placement from $\mathcal S_c$ reacts to the size of the asset sampled from $\mathcal S_{p}$. None of the examples have any offset.
    }
    \label{fig:plant}
    \vspace{-0.05in}
\end{figure}

Asset samplers define how assets are positioned relative to one another. SAGs are constructed by looking at instances of asset samplers from their top-down orthographic images, such as the one shown in Figure \ref{sub:anchor}. Here, both of the chair samplers are parented to the dining table sampler. Each child asset sampler is anchored to its parent asset sampler vertically in $\mathcal V = \{\textsc{Top}, \textsc{Center}, \textsc{Bottom}\}$ and horizontally in $\mathcal H = \{\textsc{Left}, \textsc{Center}, \textsc{Right}\}$. Each child asset sampler's pivot position can similarly be set vertically in $\mathcal V$ and horizontally in $\mathcal H$. For instance, in Figure \ref{sub:anchor}, both chair samplers are anchored to the parent vertically on \textsc{Top} and horizontally in the \textsc{Center}. But, the chair sampler on the left's pivot position is vertically in the \textsc{Center} and horizontally on the \textsc{Right}, whereas the chair sampler on the right's pivot position is vertically in the \textsc{Center} and horizontally on the \textsc{Left}. Figure \ref{fig:plant} shows more examples of how a plant or floor lamp sampler may be positioned around an arm chair sampler. Each child asset sampler can then have an $(x, y)$ offset, which is the distance from the parent sampler's anchor point to the child sampler's pivot position.

The motivation for the relative positioning of asset samplers is to prevent the meshes from clipping into each other. For instance, with the same SAG in Figure \ref{sub:anchor}, consider what would happen if the dining table sampler samples a table that is double the size of the current table. Instead of the chairs being stuck in a fixed global position, and effectively colliding with the new dining table, the chairs will reactively move back, and be re-positioned to remain slightly tucked under the larger table. Moreover, consider that the size of instances that are sampled from an asset sampler are often quite different. For instance, one table might be square-ish, while another is elongated. If we only used a \textsc{Center Center} pivot and an offset, one would not be able to reliably place asset samplers, containing differently sized objects, directly beside each other without it resulting in clipping.

\begin{figure}
    \centering
    \vspace{-1in}
    \begin{subfigure}[b]{0.49\textwidth}
        \centering
        \includegraphics[width=0.925\textwidth]{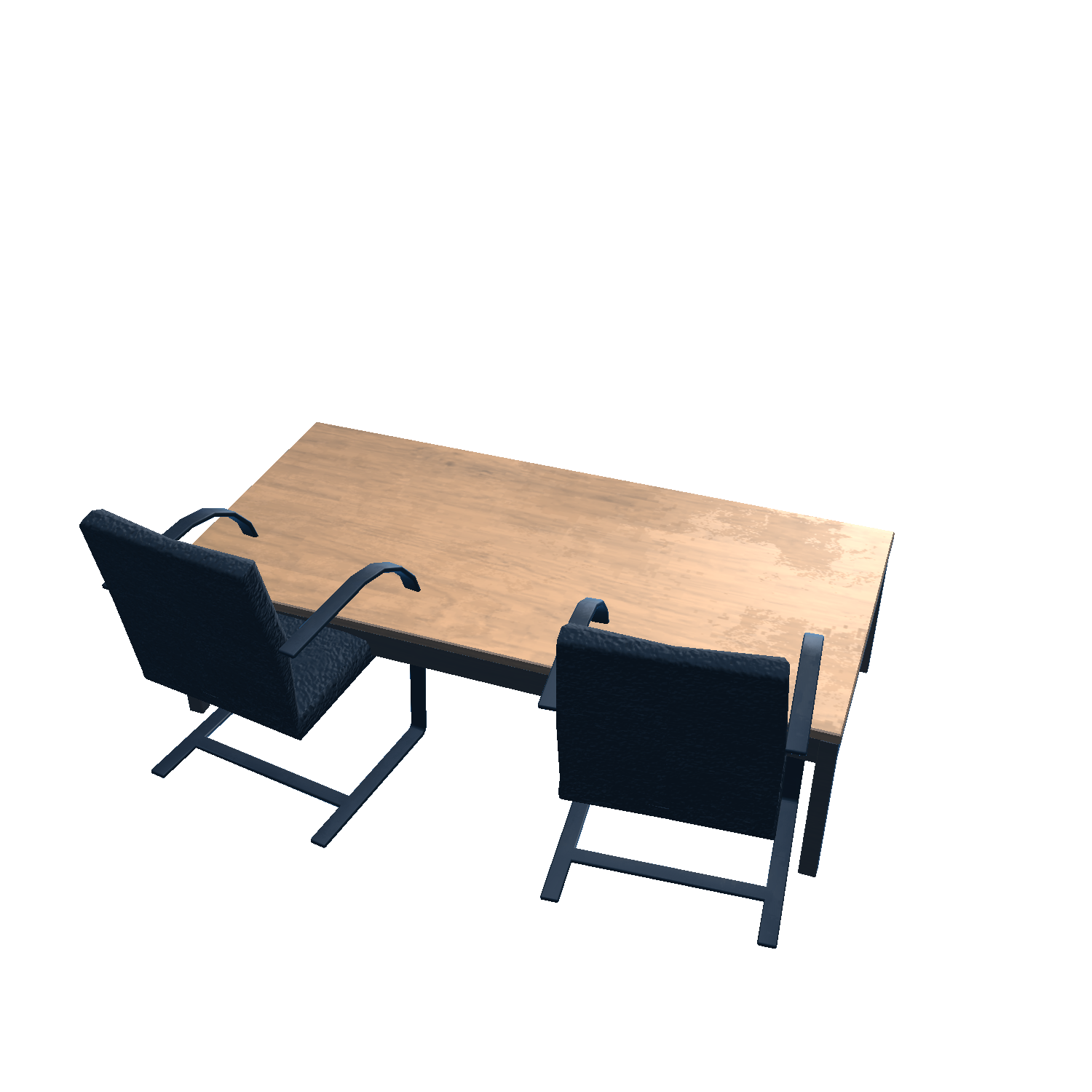}
        \vspace{-0.20in}
        \caption{\includegraphics[width=0.085in]{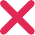} Rejected}
    \end{subfigure}
    \hfill
    \begin{subfigure}[b]{0.49\textwidth}
        \centering
        \includegraphics[width=0.75\textwidth]{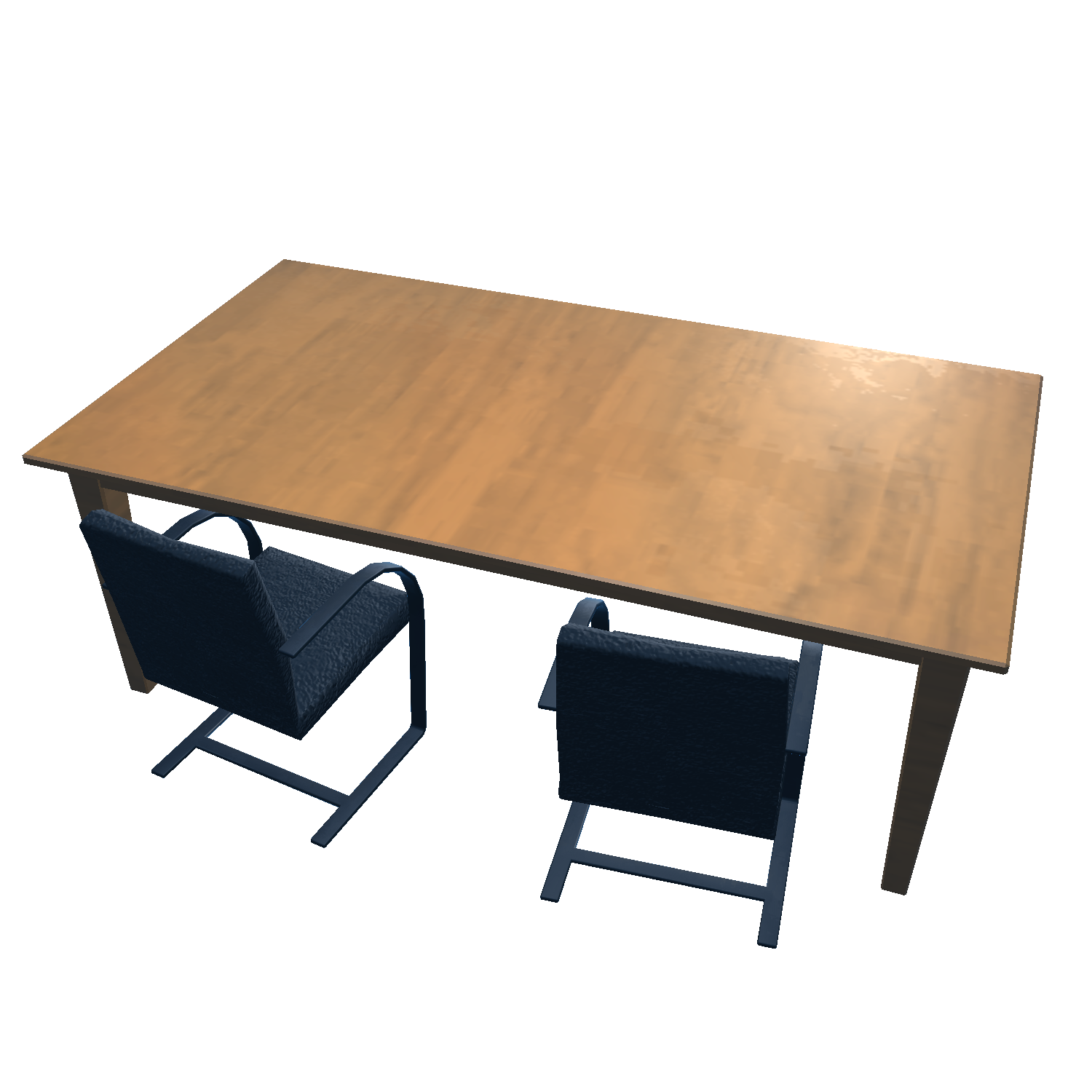}
        \vspace{-0.15in}
        \caption{\includegraphics[width=0.115in]{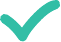} Accepted}
    \end{subfigure}
    \caption{Rejection sampling is used to make sure objects placed in SAGs do not collide. \emph{Left:} the chair collides with the dining table, and hence it is rejected; \emph{Right:} none of the objects in the instantiated SAG collide with each other, so the SAG is accepted as valid.}
    \label{fig:rej}
\end{figure}

While setting anchoring and pivot positions solves many mesh clipping issues, some cases may still arise. Figure \ref{fig:rej} shows an example, where if our dining table sampler samples a short dining table, it may clip into certain chairs. Such issues are rare in practice, but object clipping would lead to less realistic and interactive houses. To solve the clipping issue, we use rejection sampling to resample the assets of a SAG until none of the 3D meshes of the sampled assets are clipping.








In \env{}-10K, we construct 18 SAGs, which can be instantiated with over 20 million unique combinations of assets. These include semantic asset groups for chairs around tables, pillows on top of beds, sofas and arm chairs looking at a television on top of a TV stand, faucets on top of sinks, and a desk with a chair, amongst others.


%


\subsubsection{Floor Object Placement}

\begin{figure}[htbp]
    \centering
    \includegraphics[width=\textwidth]{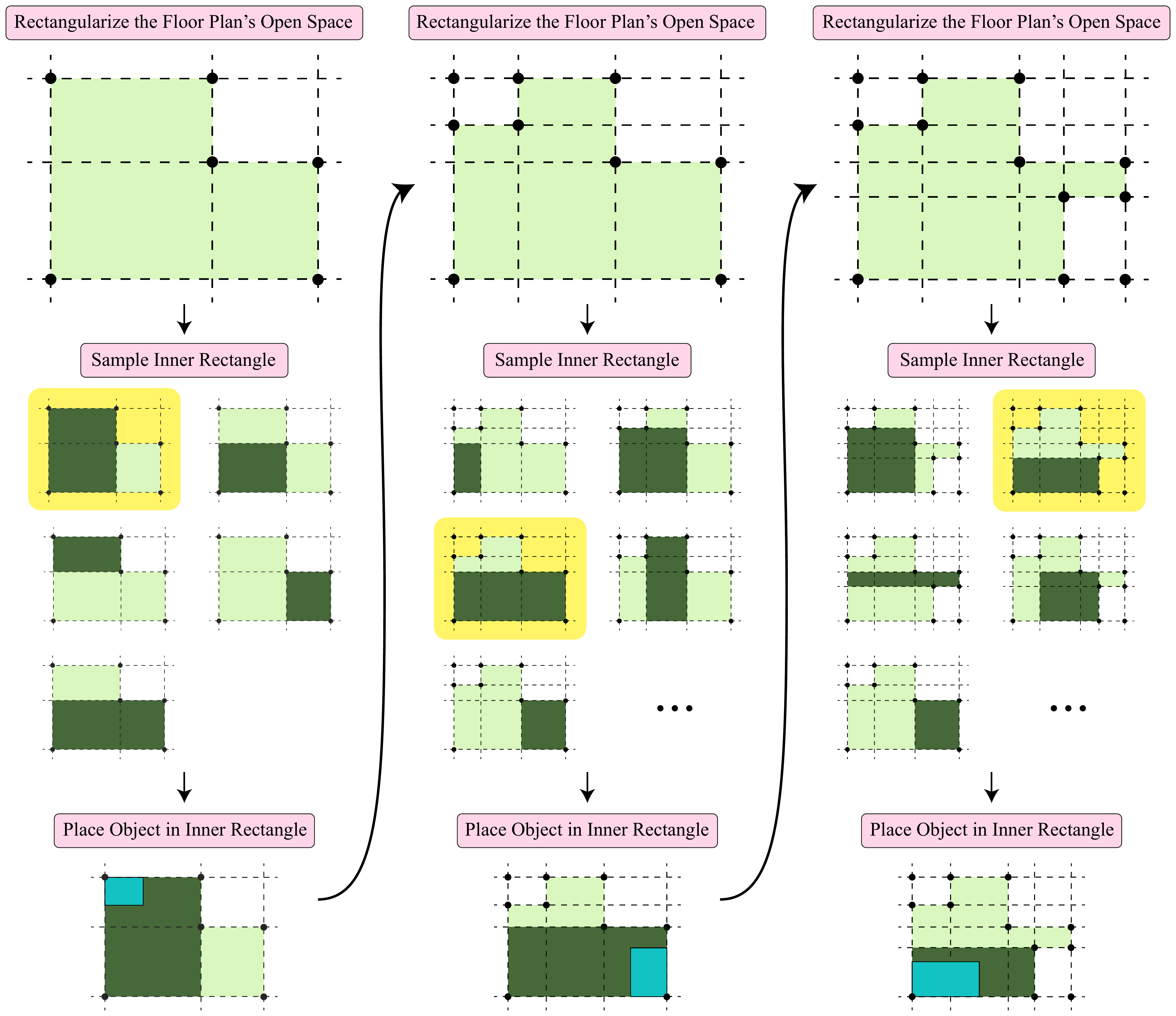}
    \caption{Diagram detailing how floor objects are placed in a room. First, we rectangularize the top-down view of the room's open floor plan by drawing horizontal and vertical dividers from each corner point. Then, we construct all possible rectangles that are formed within the dividers. We then sample one of those rectangles and place the object within that rectangle. The sampled object's top-down bounding box (with margin) is shown in blue. The bounding box is then subtracted from the open floor plan before repeating the process again.}
    \label{fig:lpo}
\end{figure}

We start object placement by first placing objects on the floor of the house. Objects are independently placed on a room-by-room basis, where we may first place objects in the bedroom and then place objects in the bathroom, without either affecting each other.

For each room, we filter the objects down into only using objects that have a room weight $r_w > 0$ in the given room type, and that have the annotation that they can be placed on the floor. Here, for instance, a chair object may have the annotation that it can be placed on the floor, but a knife object may not.


At this stage, we simplify rooms to just look at the top-down 2D bounding box that makes up the room in the floor plan. We also simplify objects to just look at its top-down 2D bounding box, of size $(o_w, o_h)$. These simplifications make it easier to determine if an object will fit in the room, specifically in a particular rectangle.

Figure \ref{fig:lpo} illustrates the iterative process of placing objects in the scene. First, the polygon forming the area left to place an object is partitioned into rectangles. The rectangles come from drawing a horizontal and vertical grid line at all corner points of the open polygon. Here, we can easily obtain the largest rectangle remaining in the open room polygon. We sample $r_\ell\sim \text{Bernoulli}(0.8)$ to determine if the next object to be placed should be placed inside of the largest rectangle. Otherwise, we randomly choose amongst all possible rectangles, weighted by the area of each rectangle.

Once we have the rectangle $(r_w, r_h)$ where the object should be placed in, we filter our objects to only those that would fit, both semantically and physically, in the rectangle. Semantically, we consider 3 scenarios: the rectangle being on the corner, edge, or middle of the room's polygon.

If any of the rectangle's corners is in a corner of the room, then we will place an object in that corner of the room. If multiple of the rectangle's corners are in a corner of the room, then we uniformly sample a corner amongst one of those corners.

\begin{figure}[htbp]
    \centering
    \begin{subfigure}{0.19\textwidth}
        \centering
        \vspace{1.055in}
        \includegraphics[width=\linewidth]{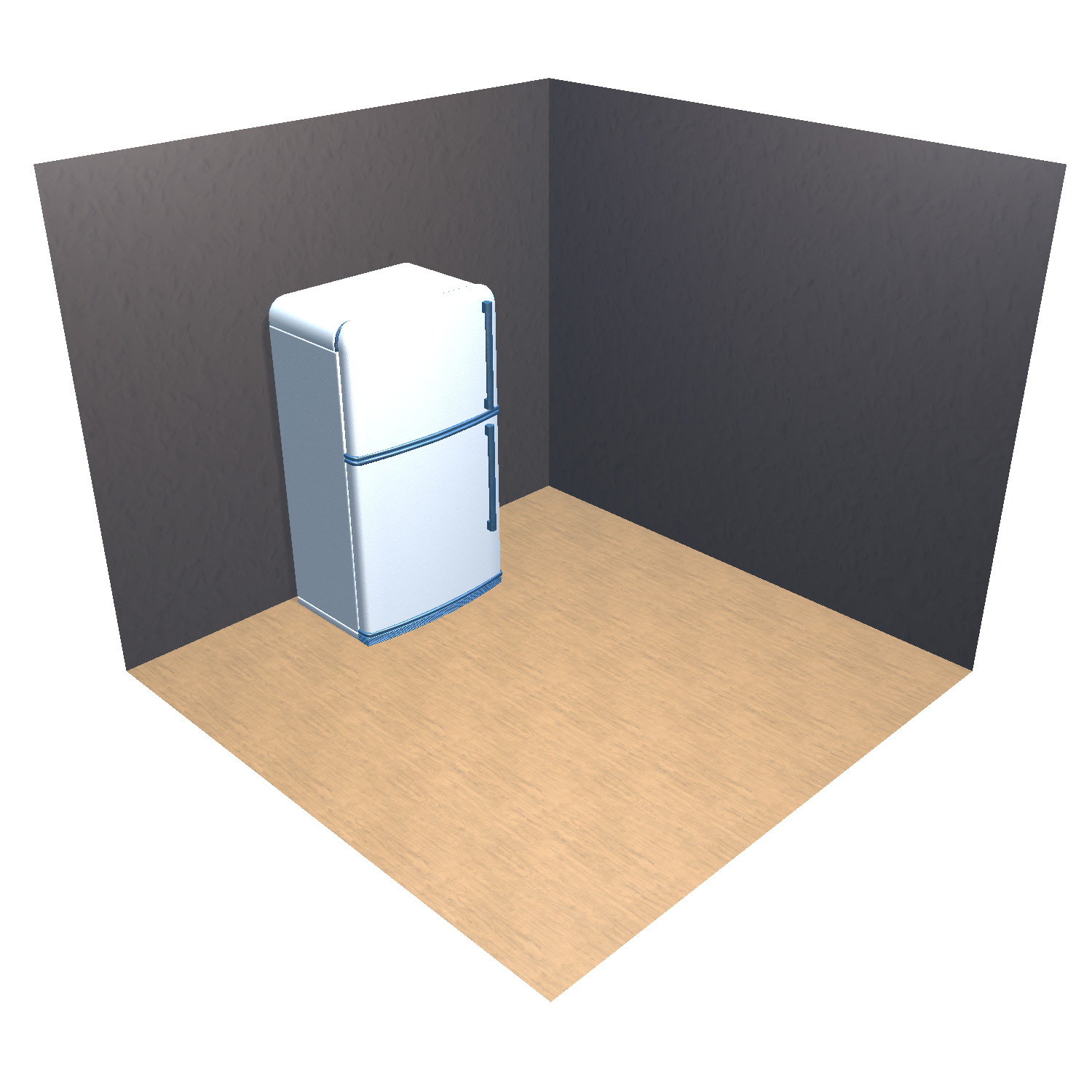}
        \caption{Edge Rotations}
        \label{fig:edgeR}
    \end{subfigure}
    \qquad\qquad
    \begin{subfigure}{0.19\textwidth}
        \centering
        \includegraphics[width=\linewidth]{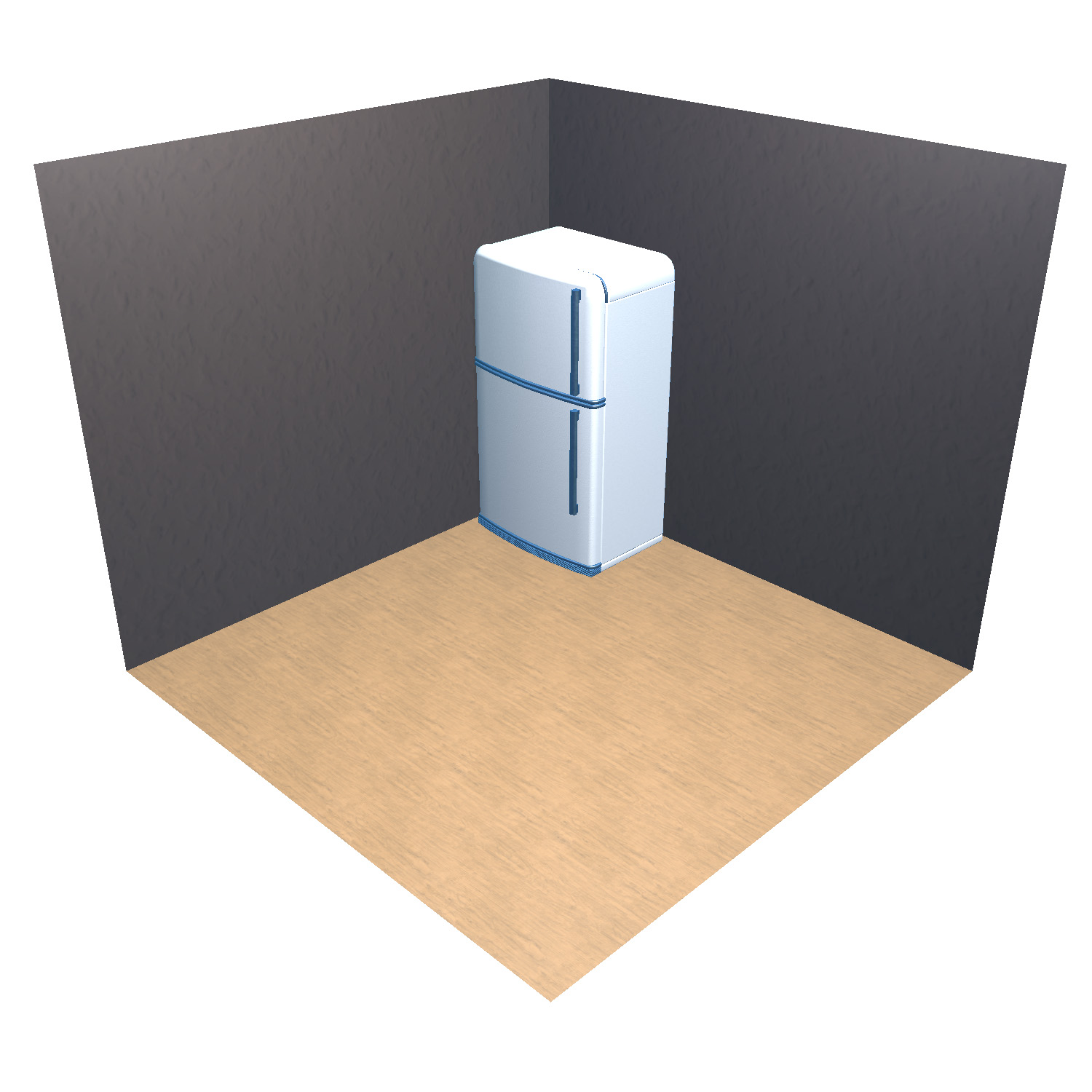}\\
        \includegraphics[width=\linewidth]{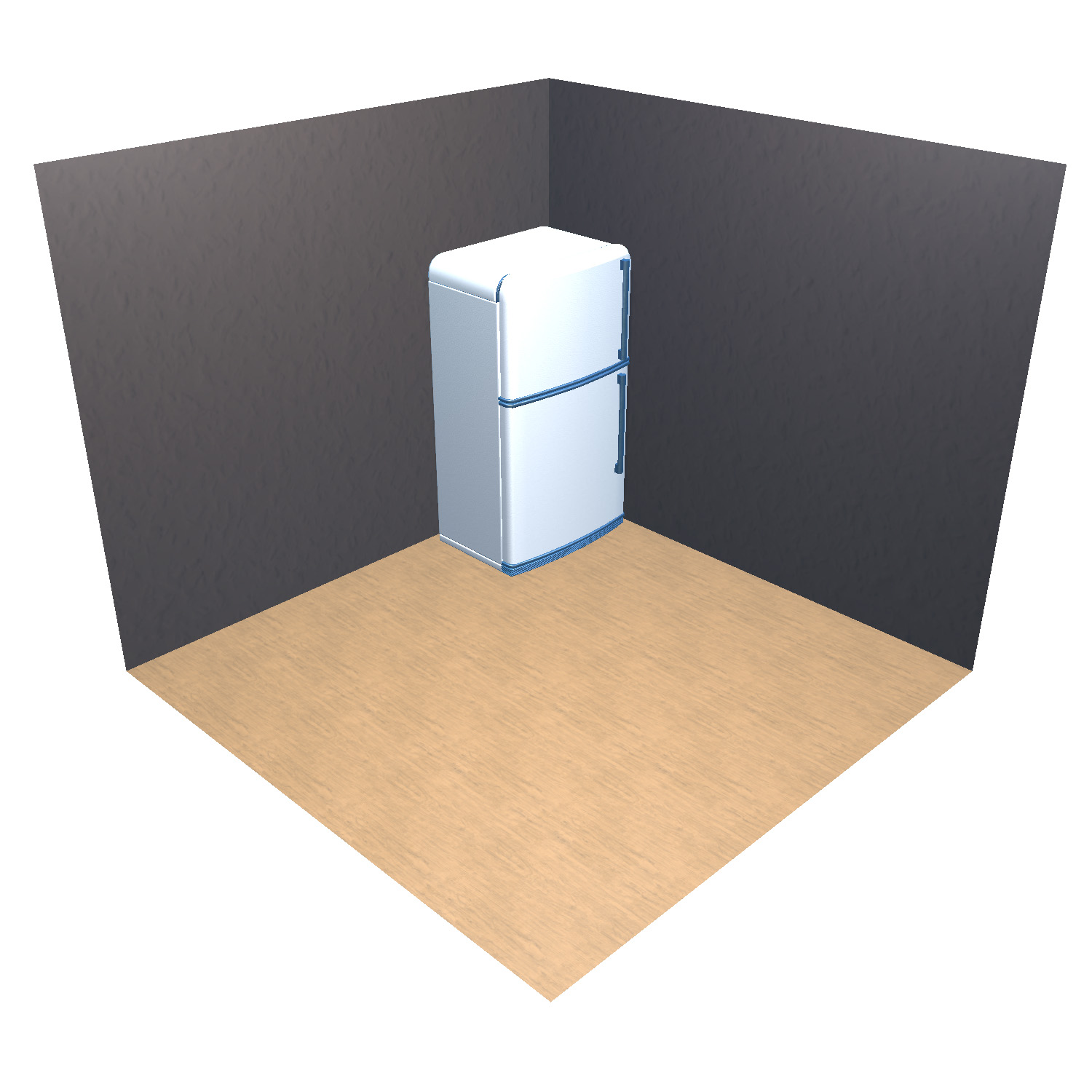}
        \caption{Corner Rotations}
        \label{fig:cornerRots}
    \end{subfigure}
    \qquad\qquad
    \begin{subfigure}{0.39\textwidth}
        \centering
        \includegraphics[width=0.48\linewidth]{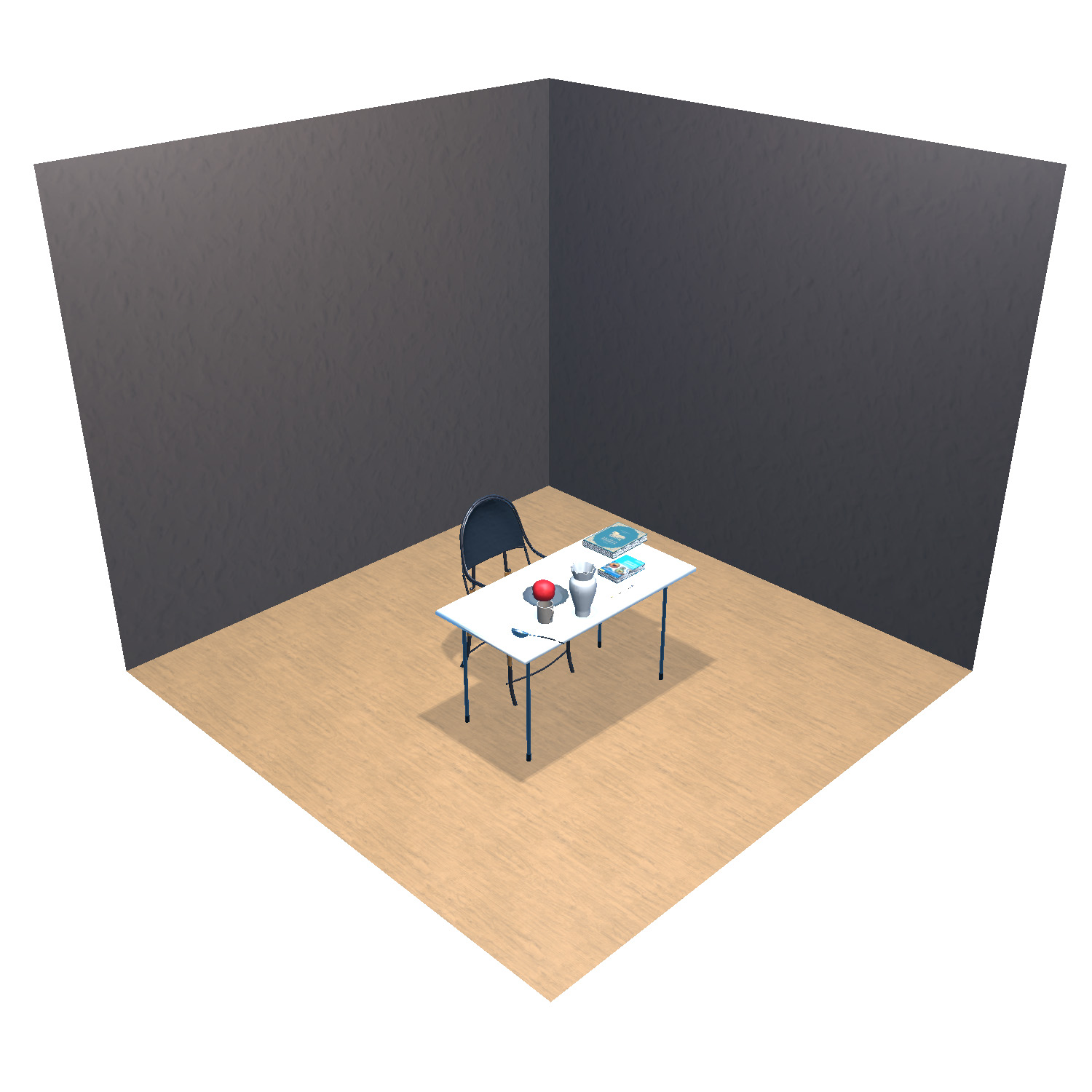}
        \includegraphics[width=0.48\linewidth]{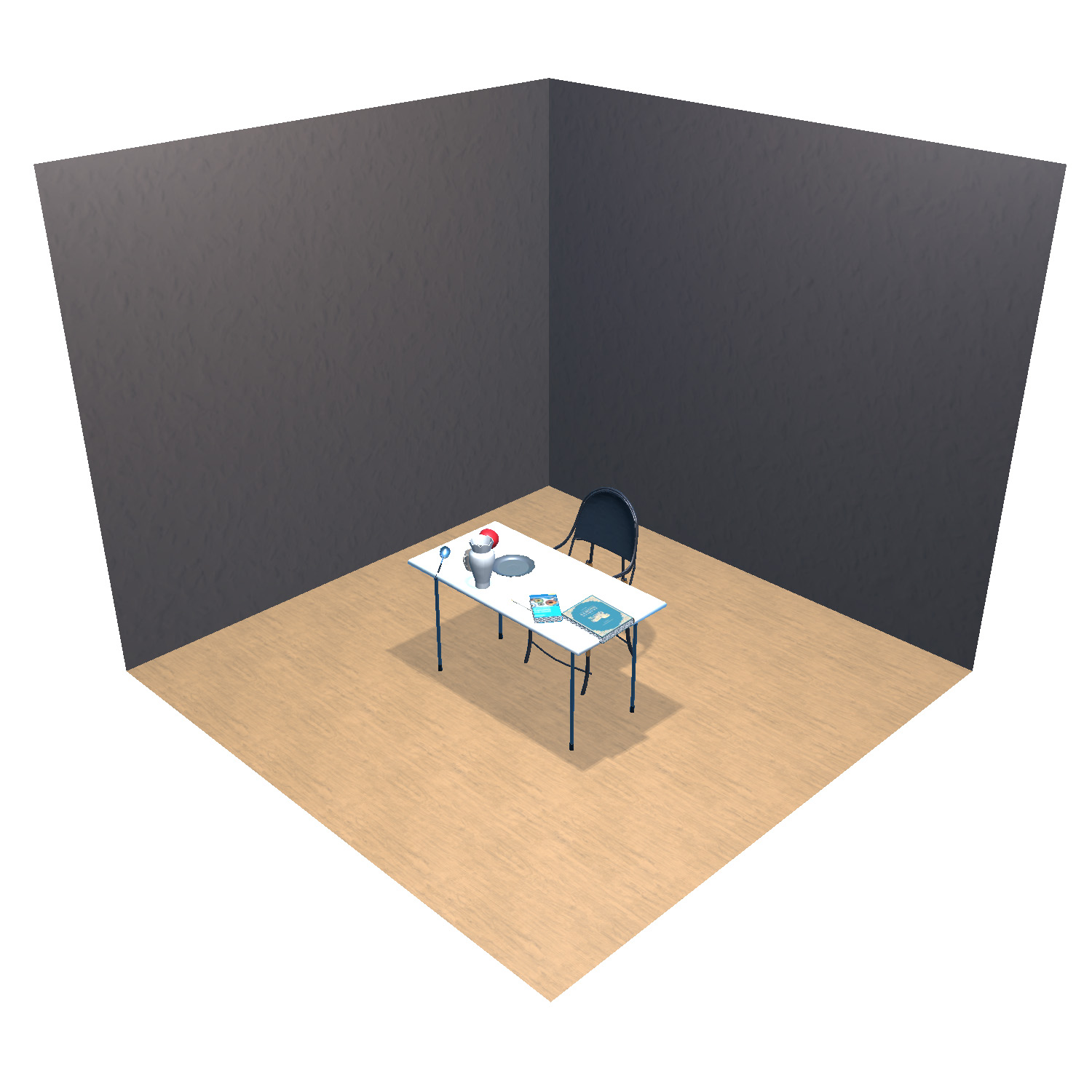}
        \includegraphics[width=0.48\linewidth]{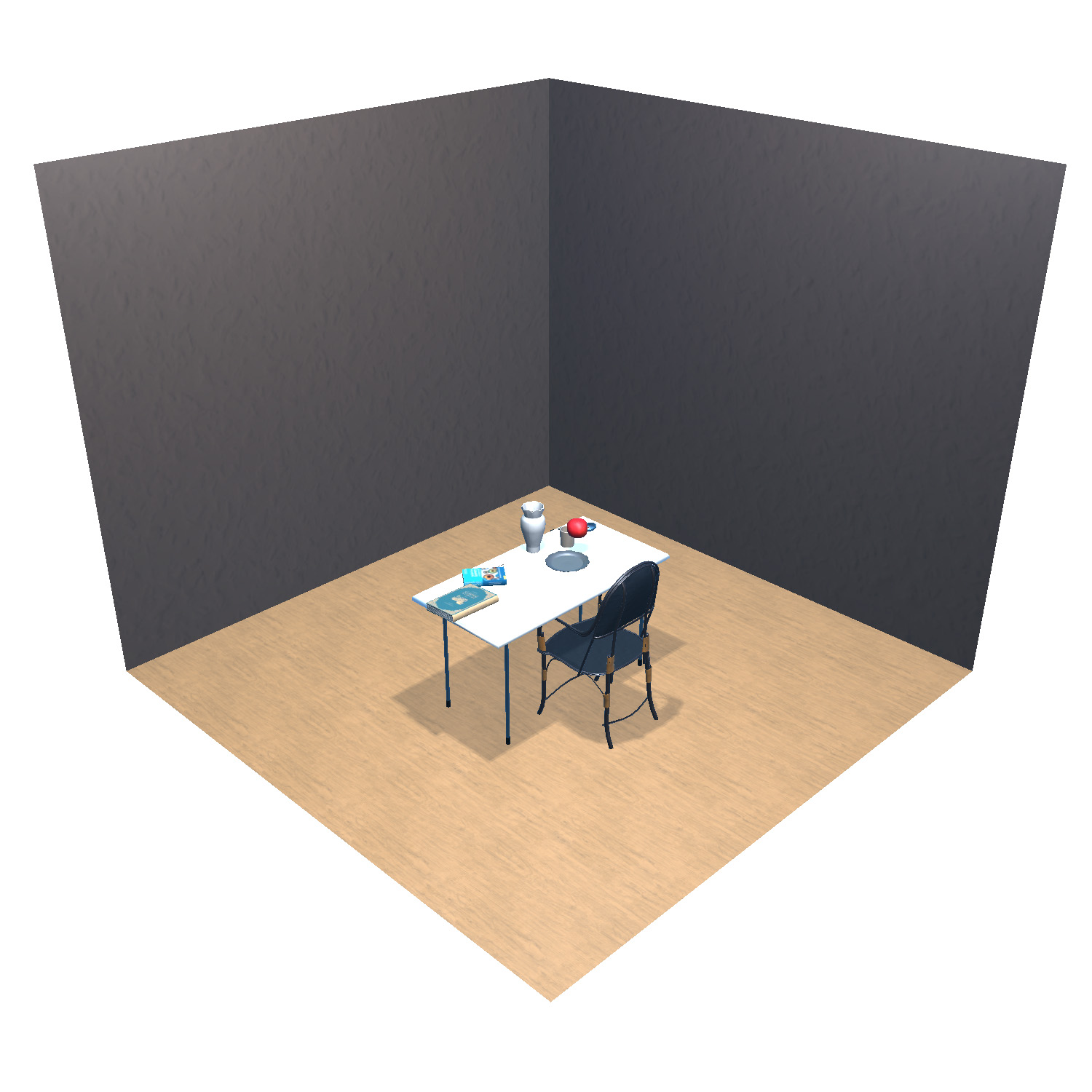}
        \includegraphics[width=0.48\linewidth]{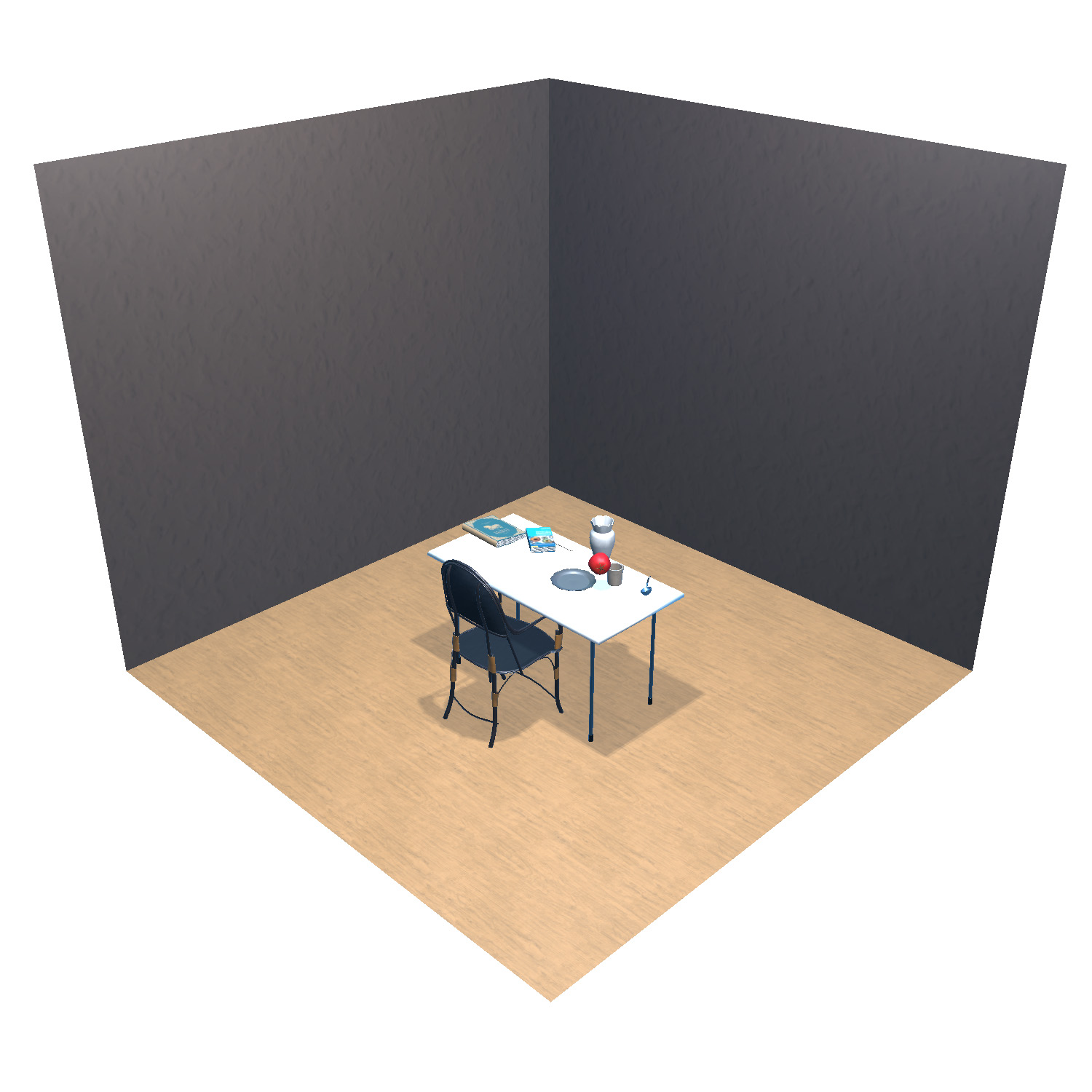}
        \caption{Middle Rotations}
        \label{fig:midR}
    \end{subfigure}
    \caption{Valid rotations of objects when placed on the edge, corner, and middle of the room. Objects placed on the edge or corner of the room always have their backs to the wall. Objects in the middle of the scene can be rotated in any direction. By constraining rotations of objects, we ensure an object on the edge of the room, such as a fridge or drawer, can still be opened.}
    \label{fig:objRotations}
\end{figure}

Now, we will filter down objects and asset groups to only consider:
\begin{enumerate}[leftmargin=0.25in]
    \item Those that are annotated specifying that they can be placed in the corner of the room. For example, we might annotate a fridge to be placed in the corner of the room, but we might not annotate a SAG consisting of a dining table to be placed in the corner of the room.
    \item The annotated split of the asset instance matches the current split of the generated house. See Appendix~\ref{sec:assets} which talks about asset splits to create train/val/test homes.
    \item The top-down bounding box of the object (with margin) must fit within the chosen rectangle. For a corner object, Figure \ref{fig:cornerRots} shows the 2 valid rotations that this object may take on. Specifically, the back of the object may be against either wall. Then, we filter down remaining objects to only use those where the object's bounding box fits within the rectangle's bounding box; that is, $(o_h + w_{\textit{pad}}\leq r_w \text{ and } o_w + w_{\textit{pad}} \leq r_h) \text{ or } (o_h + w_{\textit{pad}} \leq r_h \text{ and } o_w + w_{\textit{pad}} \leq r_w)$. If both conditions are valid, we uniformly choose one of the rotations of the object's bounding box.

    We add margin around objects to make sure it is always possible to navigate around them. Objects to be placed in the middle of the room have $m_{\textit{pad}}=0.35$ meters of margin on each side. Objects on the edge or corner of the room have $w_{\textit{pad}} = 0.5$ meters of margin only in front of the object, which enables objects to be placed directly beside it. 
\end{enumerate}

We sample an object or asset group that satisfies all of the previous conditions. If there are no objects or asset groups that satisfy all conditions, we continue to the next iteration and remove the selected rectangle from consideration. We slightly prioritize placing asset groups over standalone assets when possible. Once we have chosen an object or asset group, the bounding box with margin is then anchored to the corner of the rectangle, and hence to the corner of the room. We then subtract the object's bounding box, with margin, from the open polygon representing the space remaining in the room before doing the same process again.

If the rectangle is along the edge, we sample $r_{\textit{edge}} \sim \text{Bernoulli}(0.7)$ to determine if we should try to place an object on the edge of the rectangle, or if we should try and place it in the middle. If the rectangle is not along the edge or on the corner of the room, then we will always try to place an object in the middle of it. We use a similar filtering process, as the one described with edge rectangles, to filter down objects to those that only fit within the bounds of the rectangle. However, as depicted in Figure~\ref{fig:edgeR} and Figure~\ref{fig:midR}, edge objects can only have their backs to the wall, and middle objects can be rotated in any 90-degree rotation.

The iterative process of sampling a rectangle from the open polygon of the room, placing an object in that rectangle, and subtracting the bounding box formed by the object in the rectangle, continues on for $r_{i}$, where $r_i$ is sampled from
\begin{equation}
    r_i\sim \begin{cases}
        1 & p=\nicefrac{1}{200}\\
        4 & p=\nicefrac{2}{200}\\
        5 & p=\nicefrac{4}{200}\\
        6 & p=\nicefrac{20}{200}\\
        7 & p=\nicefrac{173}{200}
    \end{cases}.
\end{equation}
Sampling $r_i$ allows us to infrequently have rooms in the house where there are very few objects, which is sometimes the case in real-world homes. It should also be noted that there can be more than $r_i$ objects on the floor of the scene if some objects in the scene are in SAGs.

By iteratively choosing the largest, or near largest, rectangle in the room's open polygon, placing an object in it, and subtracting the object's bounding box with margin from the open room polygon, we enable great coverage across the entirety of the room, and hence the entirety of the house.

\subsubsection{Wall Object Placement}

\begin{figure}
    \centering
    \begin{subfigure}{0.325\textwidth}
        \includegraphics[width=\textwidth]{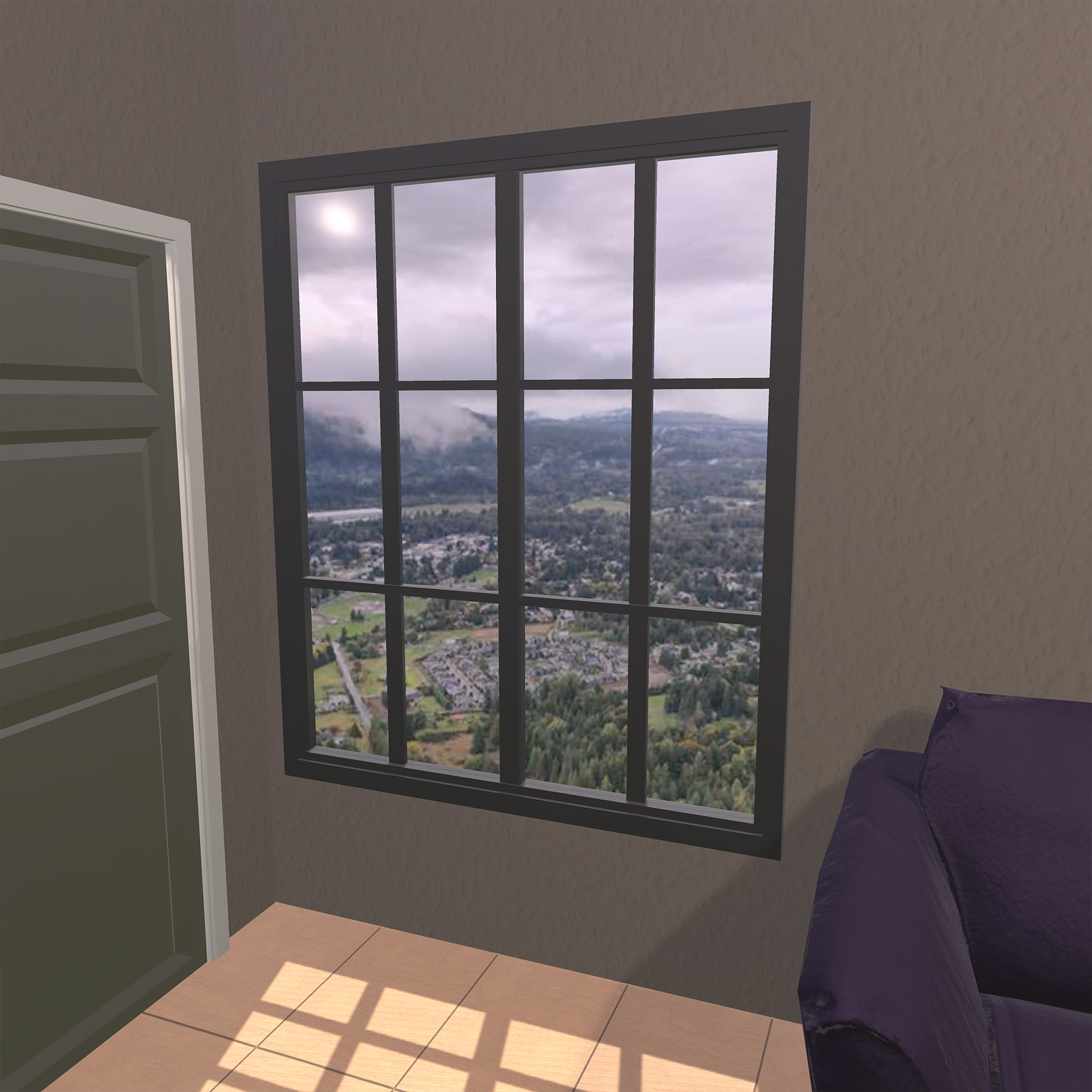}
        \caption{Window}
    \end{subfigure}
    \begin{subfigure}{0.325\textwidth}
        \includegraphics[width=\textwidth]{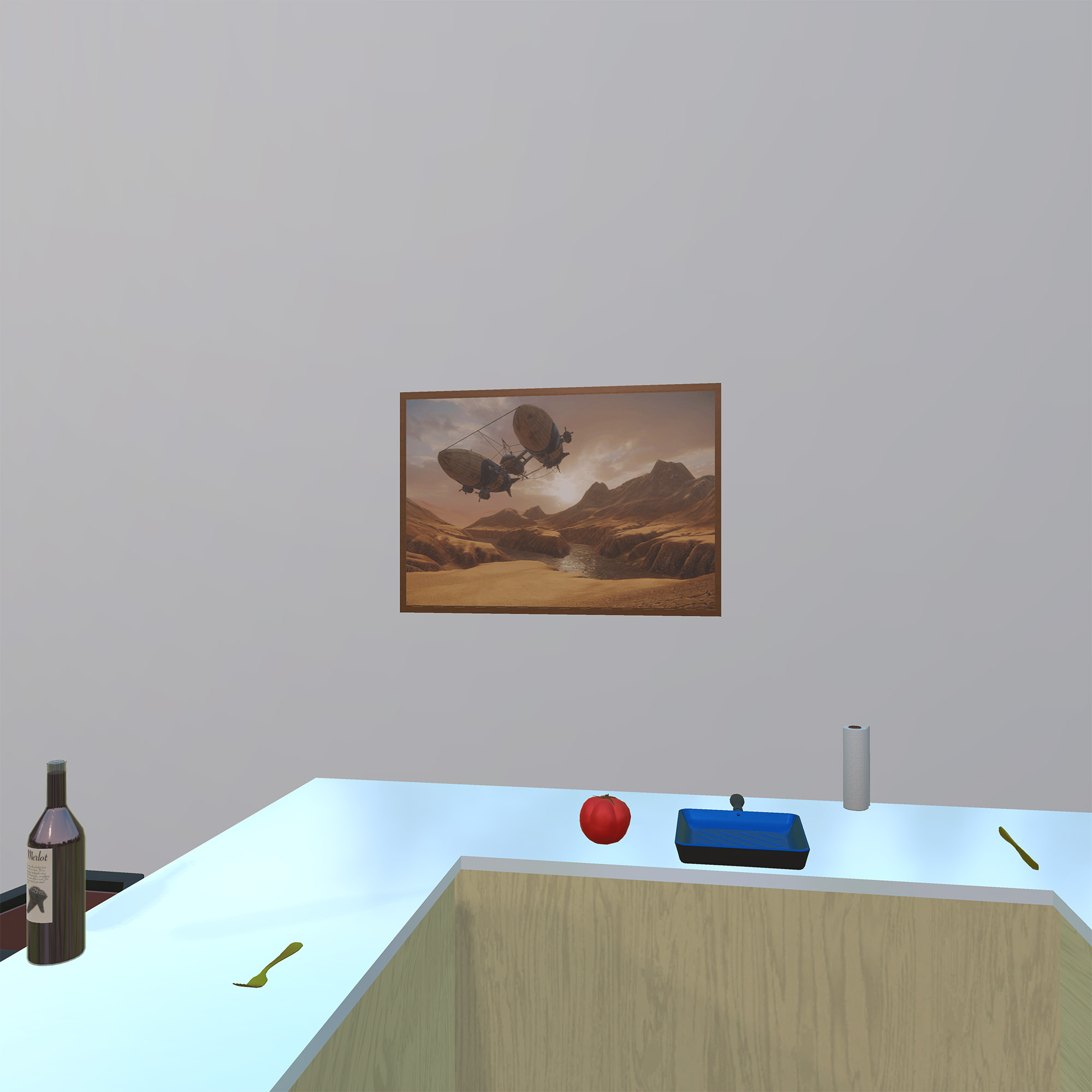}
        \caption{Painting}
    \end{subfigure}
    \begin{subfigure}{0.325\textwidth}
        \includegraphics[width=\textwidth]{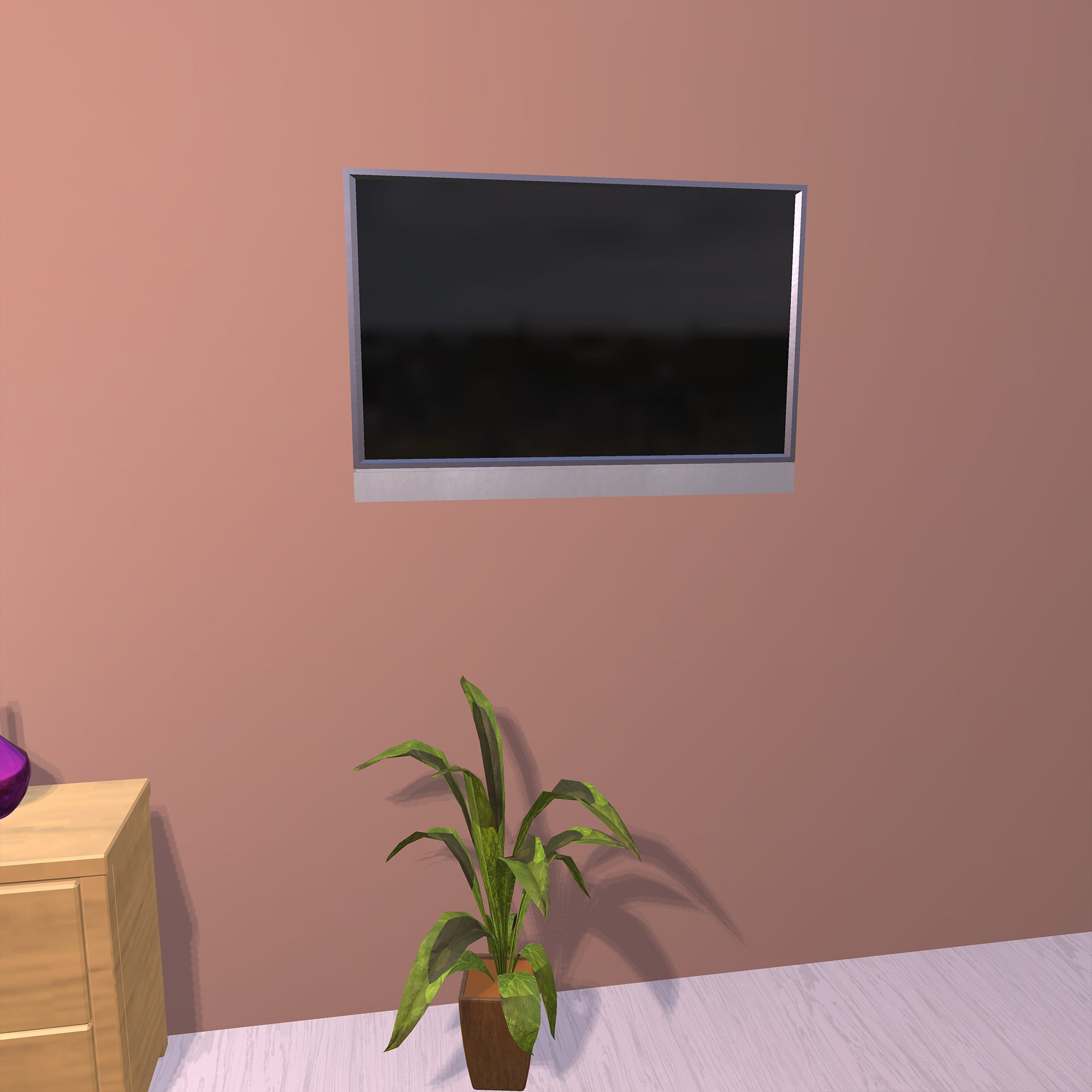}
        \caption{Television}
    \end{subfigure}
    \caption{Examples of objects placed on the wall of a house.}
    \label{fig:wall-objects}
\end{figure}

After placing objects on floors, we then place objects on walls. We currently place window, painting, and television objects on the walls. Figure \ref{fig:wall-objects} shows some examples. Window and television objects may appear in kitchen, living room, and bedroom room types. Paintings may appear in any room type.

\textbf{Windows.} Window objects are the first objects we place on the walls of the house. We only consider placing a window on walls that are connected to the outside of the house, such that we do not place a window between two indoor rooms. For each kitchen, living room, and bedroom in the house, we sample
\begin{equation}
    n_w\sim \begin{cases}
        0 & p=0.125\\
        1 & p=0.375\\
        2 & p=0.5
    \end{cases}
\end{equation}
maximum window objects to be placed.

For each wall in a given room, we look at the segment formed by each edge connecting 2 adjacent corners. If there is a floor object placed along that edge (or corner) of the wall, we subtract it from the segment. Here, the segment may break into different segments, where each segment is treated just like the original one. If the length of any segment is smaller than the minimum window size in the split, we remove the segment. We then use a uniform sample over the remaining segments, weighted by their lengths, to determine where to place the window. If no segments are longer than the smallest window, we move on to the next room in the house. A window smaller than the length of the segment is then uniformly placed somewhere along the sampled segment. The window is vertically centered along the wall between the floor and $w_{\max}=\min(3, c_h)$. All segments along the wall where the window was placed are removed from future sampling calls, and we continue this process $n_w$ times.

\textbf{Paintings.} Painting objects are placed on the walls after window objects. They may be placed in any room. The maximum number of painting objects that are attempted to be placed in each room is sampled from
\begin{equation}
    n_p\sim\begin{cases}
        0 & p=0.05\\
        1 & p=0.1\\
        2 & p=0.5\\
        3 & p=0.25\\
        4 & p=0.1
    \end{cases}.
\end{equation}

The placement of painting objects is similar to the placement of window objects. However, multiple painting objects may be placed along the same wall, so instead of removing the entire wall segment after an object is placed on it, we subtract the width of the painting from the segment. Moreover, we also allow painting objects to be placed above edge floor objects if the height of the edge object is less than 1.15 meters. Here, this allows for a painting to be above an object like a counter top, but not behind a taller object like a fridge.

The vertical position of each painting is sampled at $o_y\sim w_{\min} + (w_{\max} - w_{\min})\cdot \text{Beta}(12, 12)$, where $w_{\min}$ is the maximum height of a floor object along the wall line. Here, we allow a painting to be placed above an object along the wall of the room, such as placing it above a counter top. Sampling from $\text{Beta}(12, 12)$ allows for some randomness in the sampling process while still having a large density near the center.

\textbf{Televisions.} Television wall objects may only be placed in living room, kitchen, and bedroom room types. Only 1 wall television may be placed in each room. From our annotations, television objects cannot be placed standalone on the floor. However, a television is often placed in a SAG, on top of an object like a TV stand. So as to not place too many television objects in the same room, we only filter by rooms that do not have a television object already in them. Amongst the remaining rooms, if the room type is a living room, we sample $\text{Bernoulli}(0.8)$ if we should try placing a wall television in the room. For kitchen and bedroom room types, we sample from $\text{Bernoulli}(0.25)$ and $\text{Bernoulli}(0.4)$, respectively. We only consider television objects that could be mounted to a wall (\textit{i.e.} they do not have a base that is sticking out of the object). Television wall objects sample from the same vertical position distribution as painting objects, and follow the same placement on the walls as painting objects.

\subsubsection{Surface Object Placement}
\label{sec:SurfOP}


\begin{figure}
    \centering
    \includegraphics[width=0.5\textwidth]{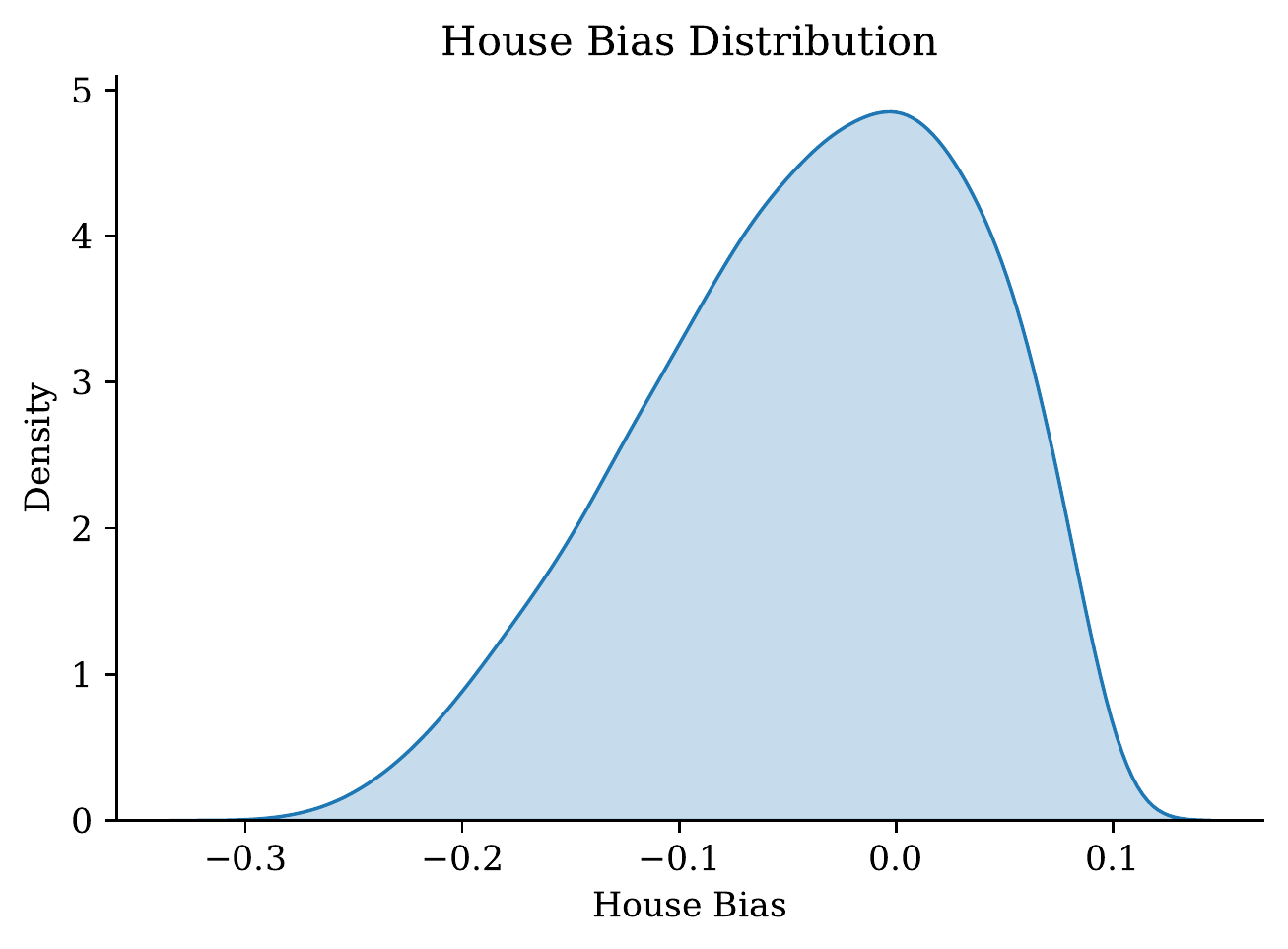}
    \caption{The house bias distribution $b_{\textit{house}}$ that offsets the probability of attempting to spawn an object in a receptacle.}
    \label{fig:house-bias}
\end{figure}

After placing objects on the floor and wall of the house, we focus on placing objects on the surface of the floor objects just placed. For example, we may place objects like a coffee machine, plate, or knife on of a receptacle like a counter top.

For each receptacle object, we approximate the probability that each object type appears on its surface. We use the hand-modeled AI2-iTHOR or RoboTHOR rooms to obtain these approximations. Here, we compute  the total number of times each object type is on the receptacle type and divide it by the total number of times the receptacle type appears across the scenes.

For each receptacle placed on the floor, we look at the probability of each object type $p_{\textit{spawn}}$ that it has been placed on that receptacle. We then iterate over the object types that may be on the receptacle. For each object type, we try spawning it on the receptacle if $\text{Bernoulli}(p_{\textit{spawn}} + b_{\textit{house}} + b_{\textit{receptacle}} + b_{\textit{object}})$, where

\begin{itemize}[leftmargin=0.25in]
    \item $b_{\textit{house}}$ denotes the additional bias of how likely objects are to be spawned on receptacles in this particular house. Each house samples
    \begin{equation}
        b_{\textit{house}}\sim (b_{\textit{house-max}} - b_{\textit{house-min}}) \cdot \text{Beta}(3.5, 1.9) + b_{\textit{house-min}},
    \end{equation}
    where $b_{\textit{house-min}} = - 0.3$ and $b_{\textit{house-max}}=0.1$. Figure \ref{fig:house-bias} shows the distribution that $b_{\textit{house}}$ forms. Using a house bias allows for some houses to be much cleaner or dirtier than others, whereas cleaner houses would have more objects put away that are not on receptacles. 
    \item $b_{\textit{receptacle}}$ denotes the additional bias of how likely an object is to be spawned on a receptacle. The default receptacle bias is $0.2$, which is only overwritten by shelving unit ($0.4$ bias), counter top ($0.2$ bias), arm chair ($0$ bias), and chair ($0$ bias). Receptacle biases were manually set based on the empirical quality of the houses.
    \item $b_{\textit{object}}$ denotes the additional bias of how likely a particular object is to spawn in the scene. By default, $b_{\textit{object}}$ is set to $0$, and overwritten by house plant ($0.25$ bias), basketball ($0.2$ bias), spray bottle ($0.2$ bias), pot ($0.1$ bias), pan ($0.1$ bias), bowl ($0.05$ bias), and baseball bat ($0.1$ bias). Object biases were also manually set based on the empirical quality of the houses to ensure more target objects appear in each of the procedurally generated houses.
\end{itemize}

Note that $p_{\textit{spawn}} + b_{\textit{house}} + b_{\textit{receptacle}} + b_{\textit{object}}$ may be greater than 1, in which case we will always try to spawn the object on the receptacle, or less than 0, where we will never try to spawn the object on the receptacle.

To attempt to spawn an object of a given type on a receptacle, we will sample an instance of that object type and randomly try $n_{\textit{pa}}=5$ poses of the object to try and fit the object instance on the receptacle. If the object instance fits and does not collide with another object, we keep it there. Otherwise, we try another pose of the object on the receptacle until we reach $n_{\textit{pa}}$ attempted poses. If none of the attempted poses work, we continue on to the next object type that may be on the receptacle.

If the first object of a given type is placed successfully on a receptacle, we attempt to place $n_{\textit{or}}\sim\min(s_{\max}, \text{Geom}(p_{\textit{spawn}}) - 1) - 1$ more objects of that type given type on the receptacle. Here, $s_{\max}$ is set to $3$, representing the maximum number of objects of a type that may be on a receptacle. We ignore the biases to not have too many objects of a given type on the same receptacle.

\subsection{Material and Color Randomization}

\begin{figure}
    \centering
    \begin{subfigure}{\textwidth}
        \centering
        \includegraphics[width=0.24\textwidth]{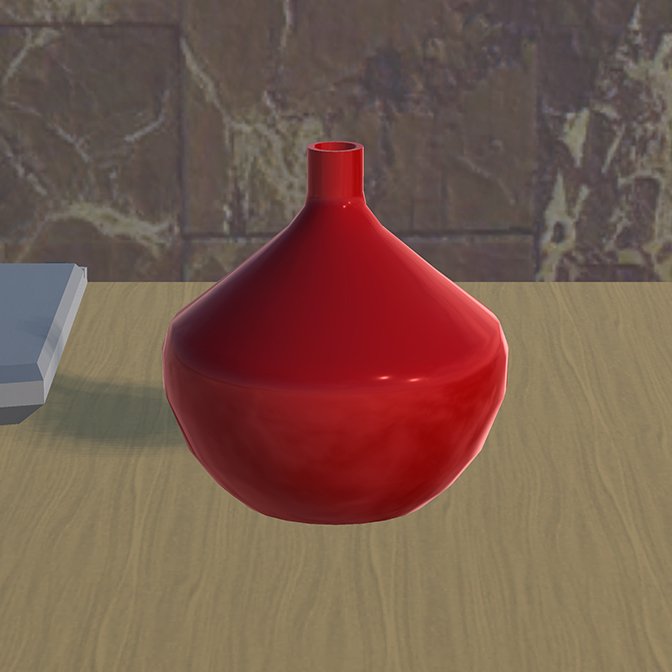}
        \includegraphics[width=0.24\textwidth]{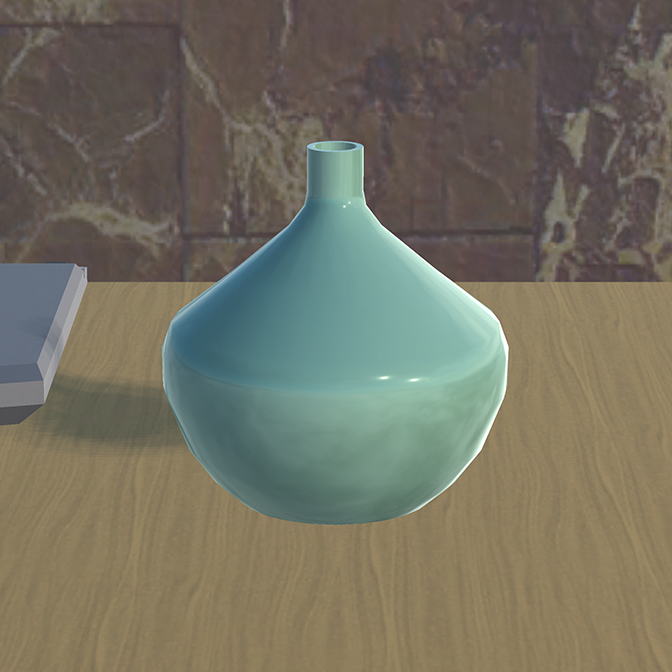}
        \includegraphics[width=0.24\textwidth]{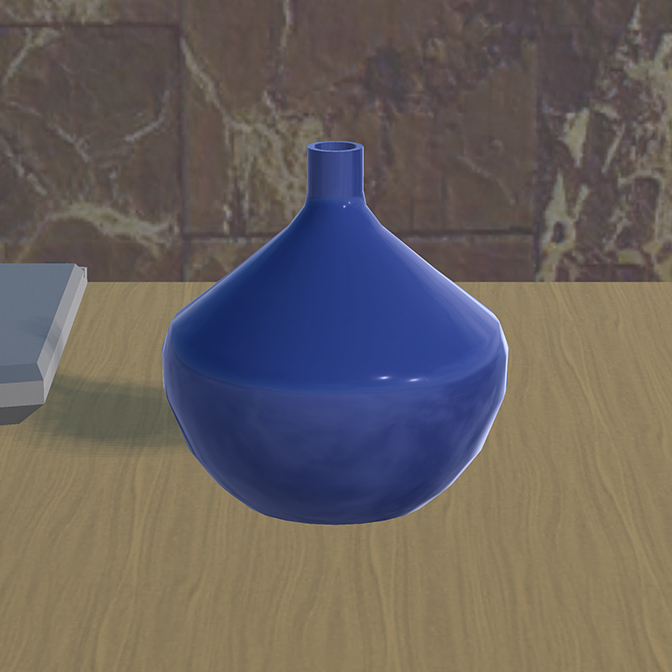}
        \includegraphics[width=0.24\textwidth]{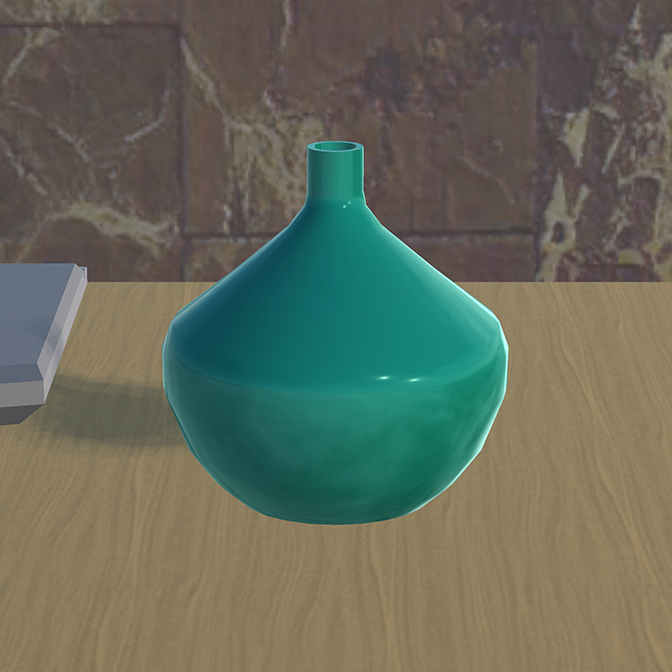}
        \caption{Examples of color randomization for a vase object. The original color is shown on the left. Notice that the vase still looks realistic with many possible colors.\\[-0.05in]}
        \label{fig:colorRand}
    \end{subfigure}
    \begin{subfigure}{\textwidth}
        \centering
        \includegraphics[width=0.24\textwidth]{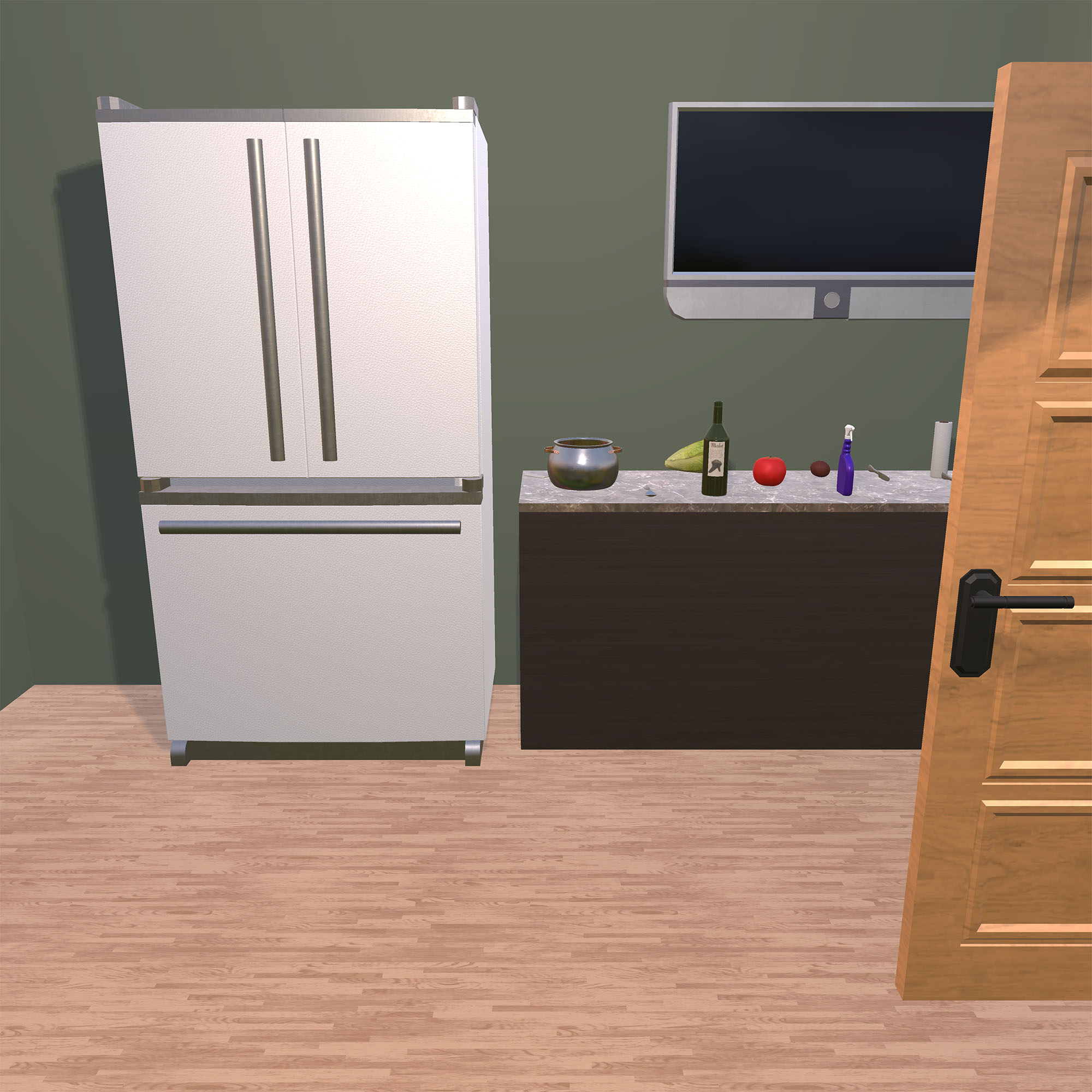}
        \includegraphics[width=0.24\textwidth]{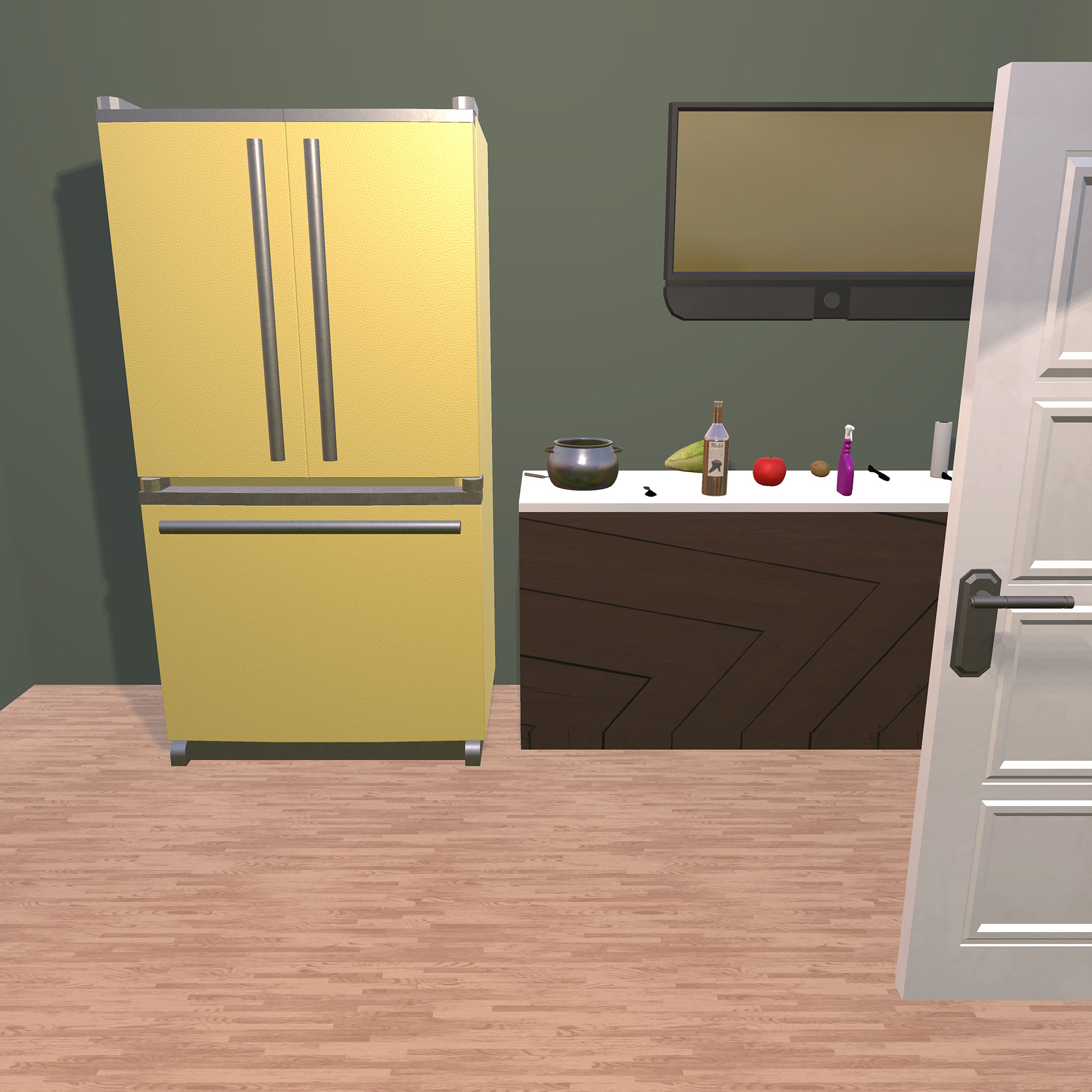}
        \includegraphics[width=0.24\textwidth]{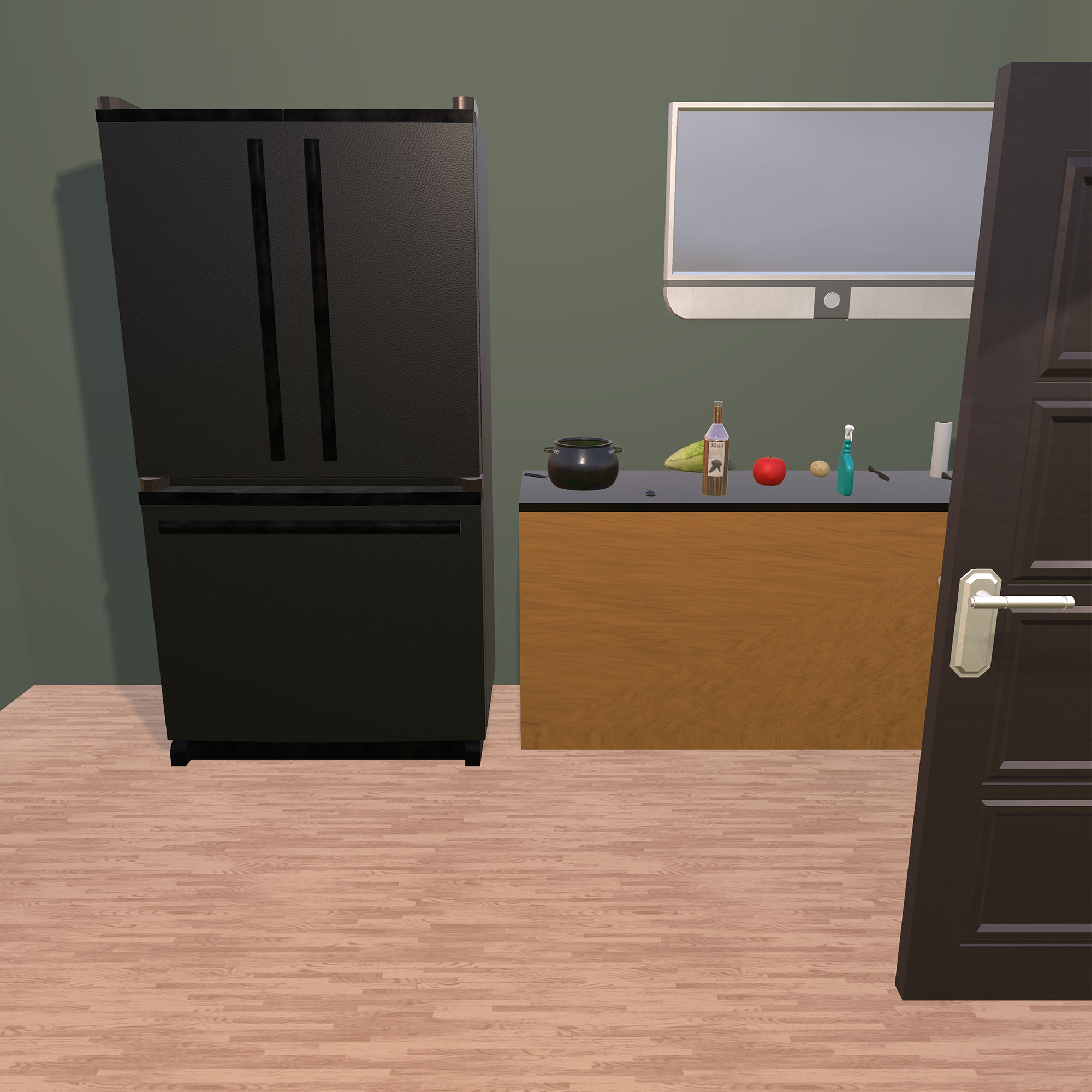}
        \includegraphics[width=0.24\textwidth]{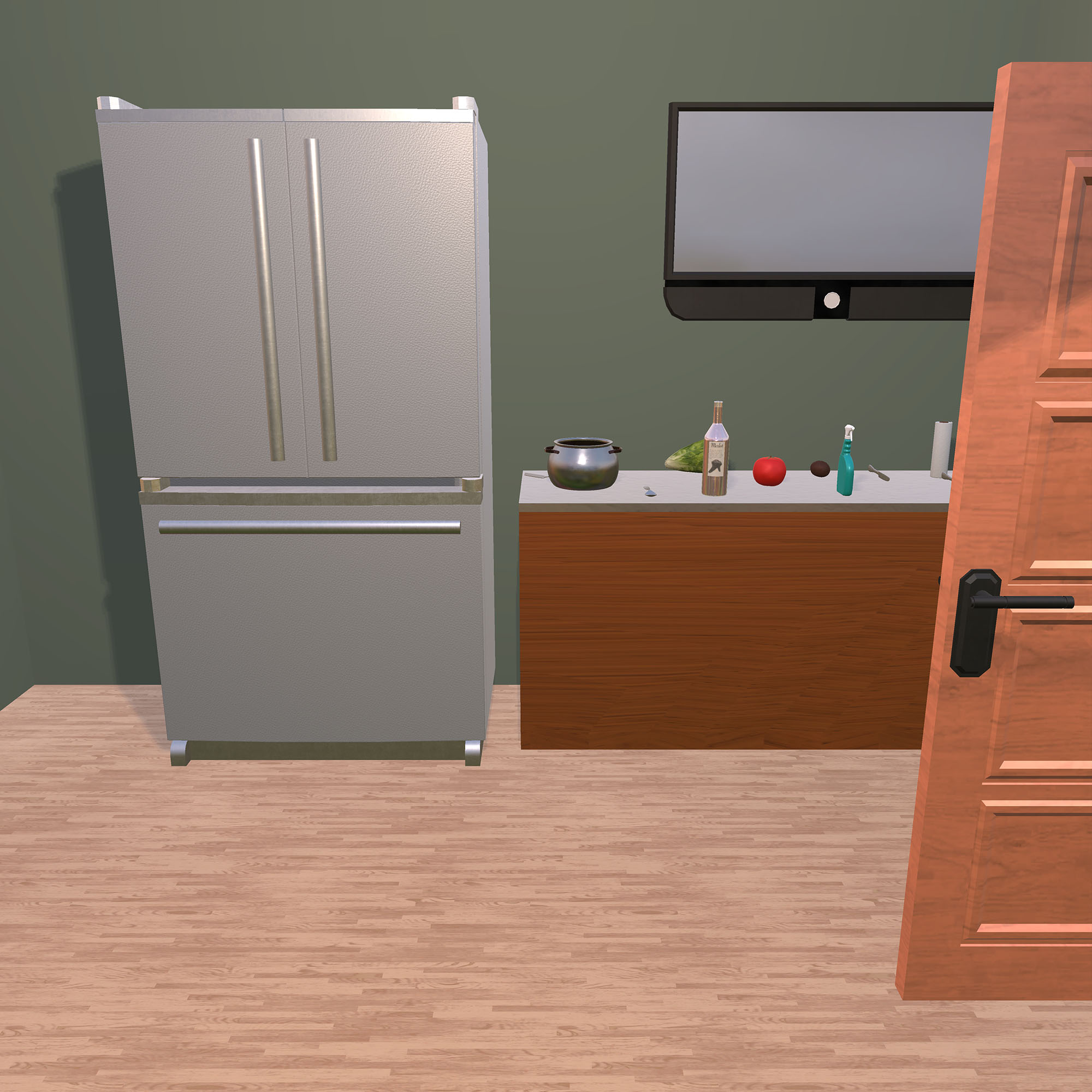}
        \caption{Examples of material randomization in ProcTHOR. Notice that only the objects randomize in materials, where the walls, floor, and ceiling remain the same.}
        \label{fig:materialRand}
    \end{subfigure}
    \caption{Examples of color randomization and material randomization in ProcTHOR.}
\end{figure}

Several object types may have their color randomized to a randomly sampled RGB value. Specifically, for each vase, statue, or bottle in the scene, we independently sample from $r_c\sim \text{Bernoulli}(0.8)$ to determine if we should randomize the object's color. These objects were chosen because they all still looked natural as any solid color. Figure \ref{fig:colorRand} shows some examples of randomizing the color of a vase.

For each training episode, we sample from $r_m\sim \text{Bernoulli}(0.8)$ to determine if we should randomize the default object materials in the scene. Wall, ceiling, and floor materials are left untouched to preserve $w_{\textit{solid}}$ and $w_{\textit{same}}$ sampling parameters. Materials are only randomized within semantically similar classes, which ensures objects still look and behave like the class they represent. For instance, an apple will not swap materials with an orange. Figure \ref{fig:materialRand} shows some examples of randomizing the materials in the scene.


\subsection{Object States}

\begin{figure}[htbp]
    \centering
    \begin{subfigure}{\textwidth}
        \centering
        \includegraphics[width=0.32\textwidth]{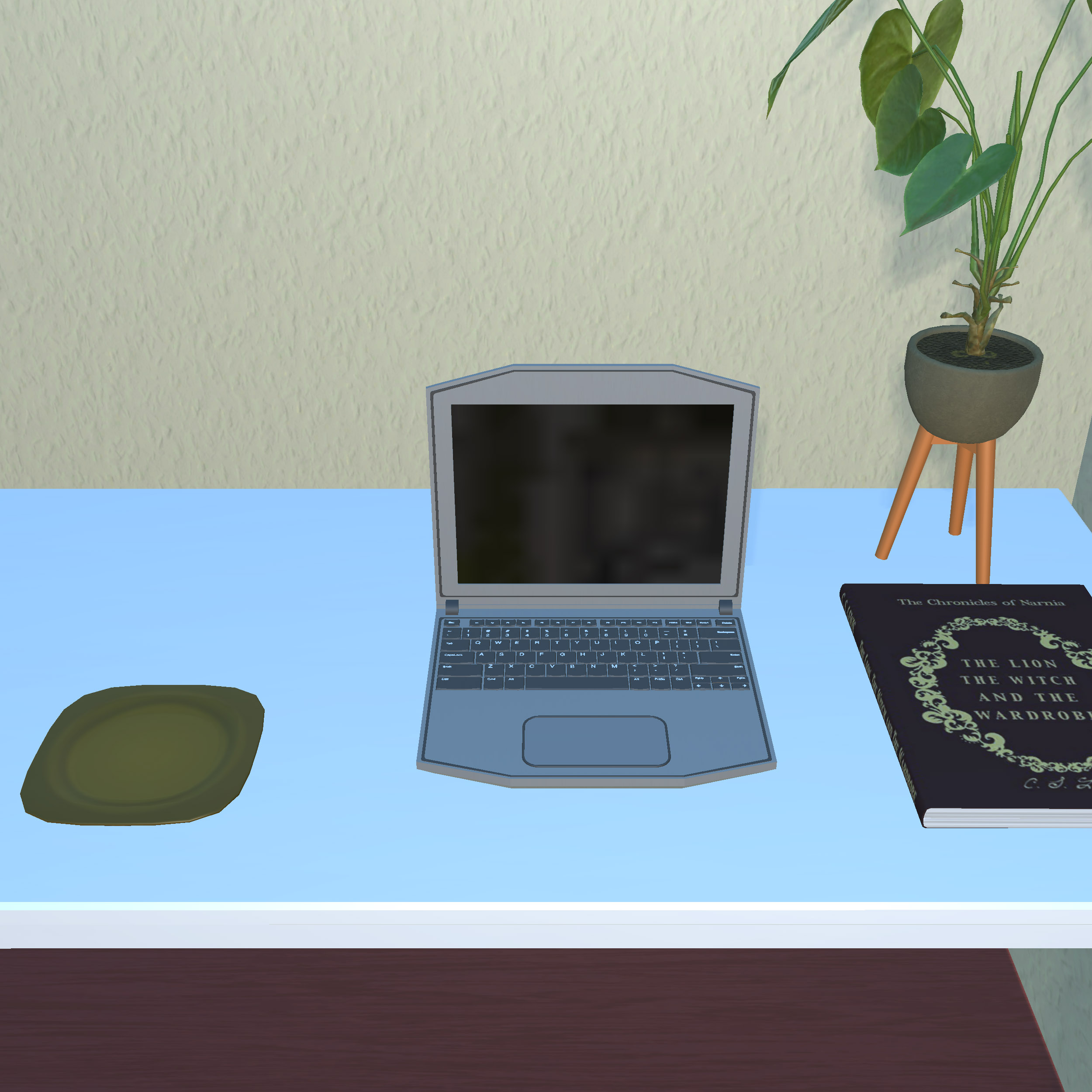}
        \includegraphics[width=0.32\textwidth]{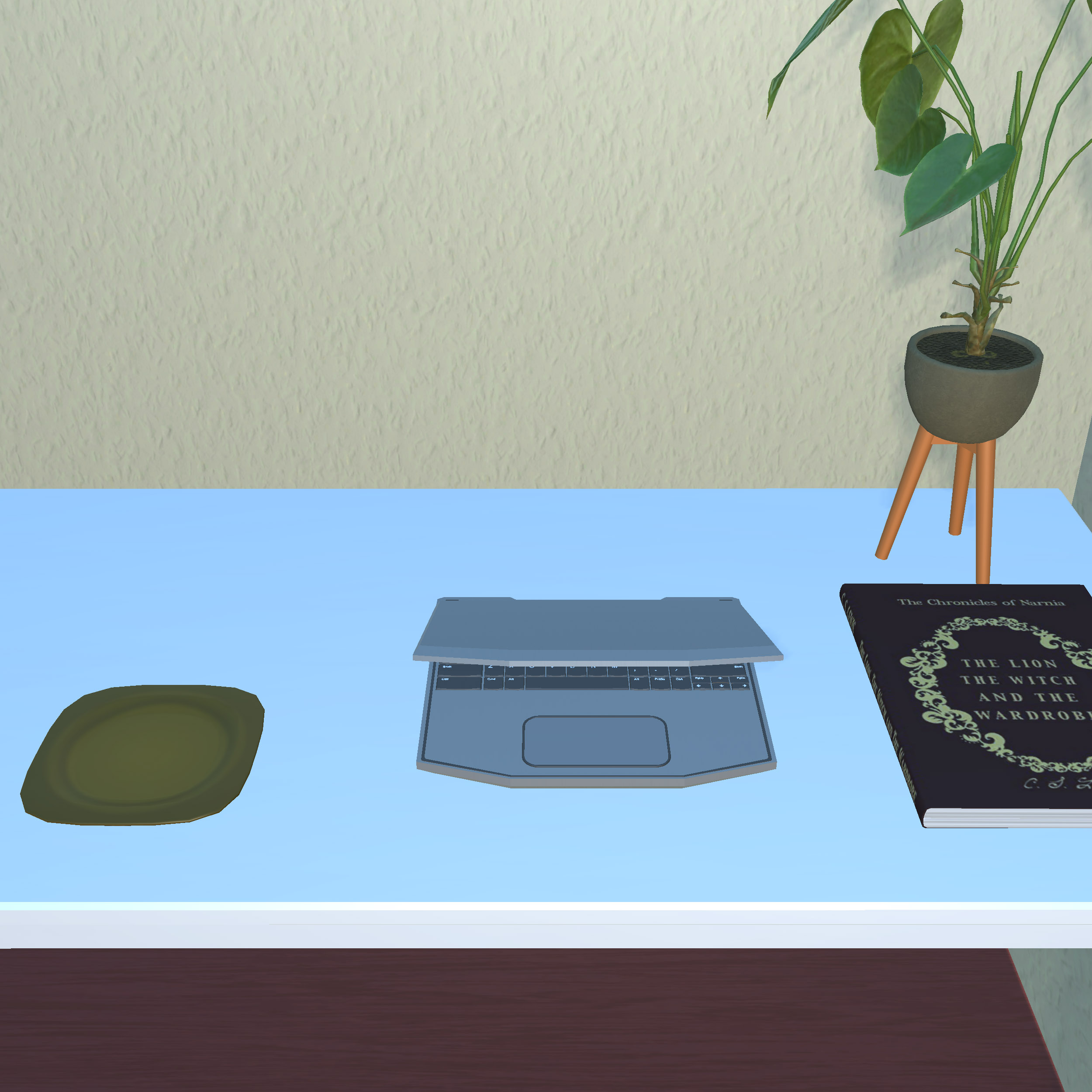}
        \includegraphics[width=0.32\textwidth]{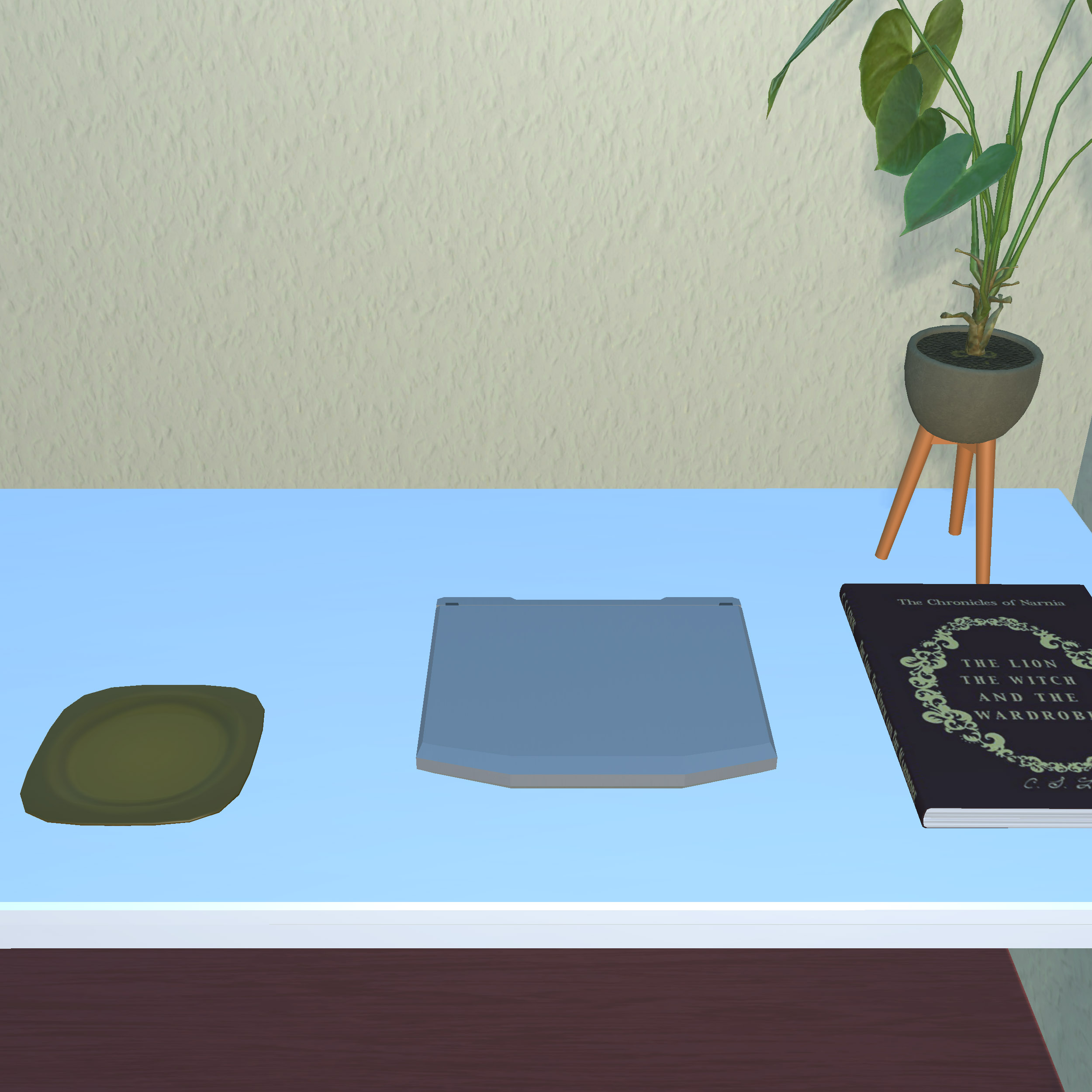}
        \caption{Openness state randomness example with a laptop.}
        \label{fig:openness}
    \end{subfigure}
    \\[0.05in]
    \begin{subfigure}{\textwidth}
        \centering
        \includegraphics[width=0.32\textwidth]{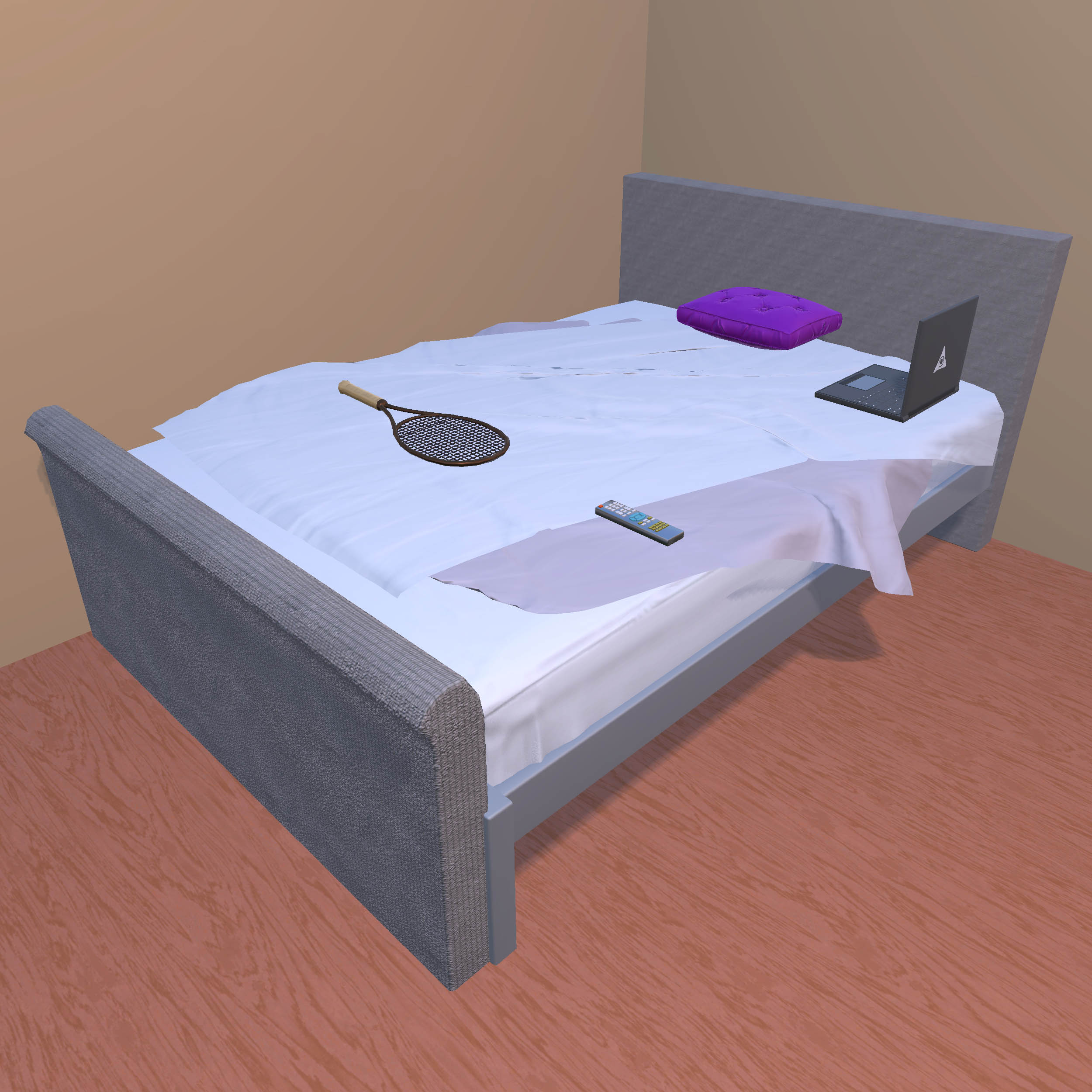}
        \includegraphics[width=0.32\textwidth]{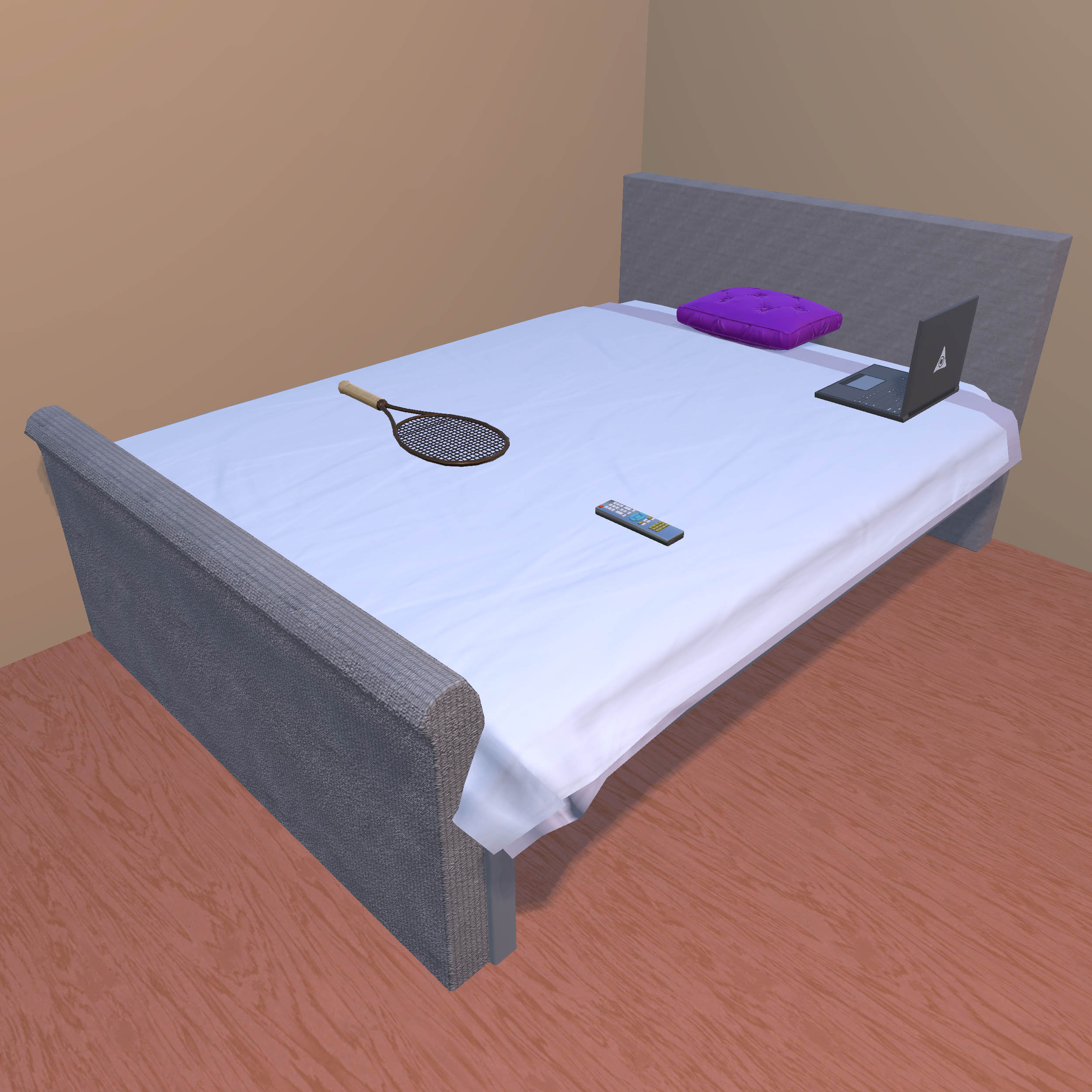}
        \caption{Clean state randomness example with a bed.}
        \label{fig:clean}
    \end{subfigure}
    \\[0.05in]
    \begin{subfigure}{\textwidth}
        \centering
        \includegraphics[width=0.32\textwidth]{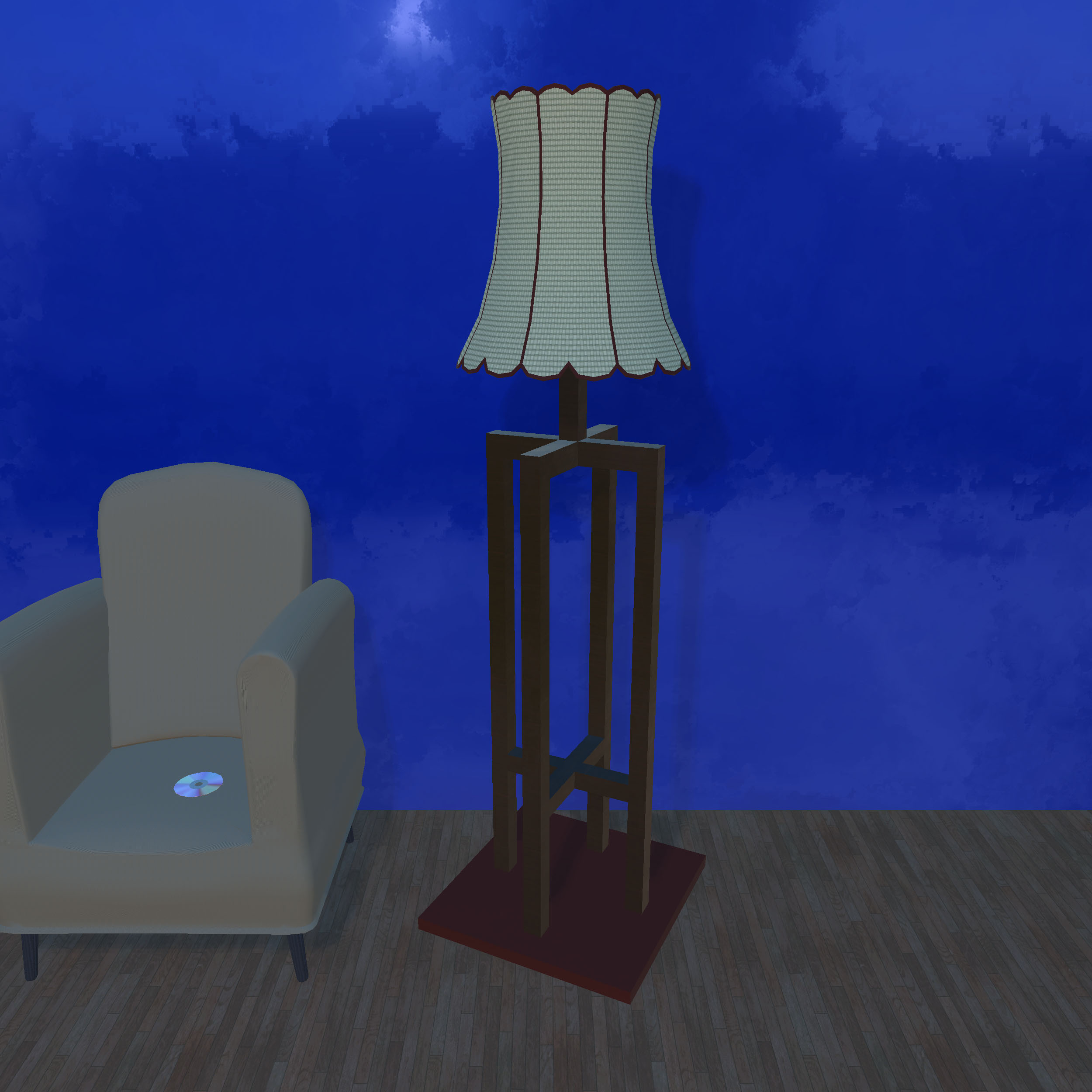}
        \includegraphics[width=0.32\textwidth]{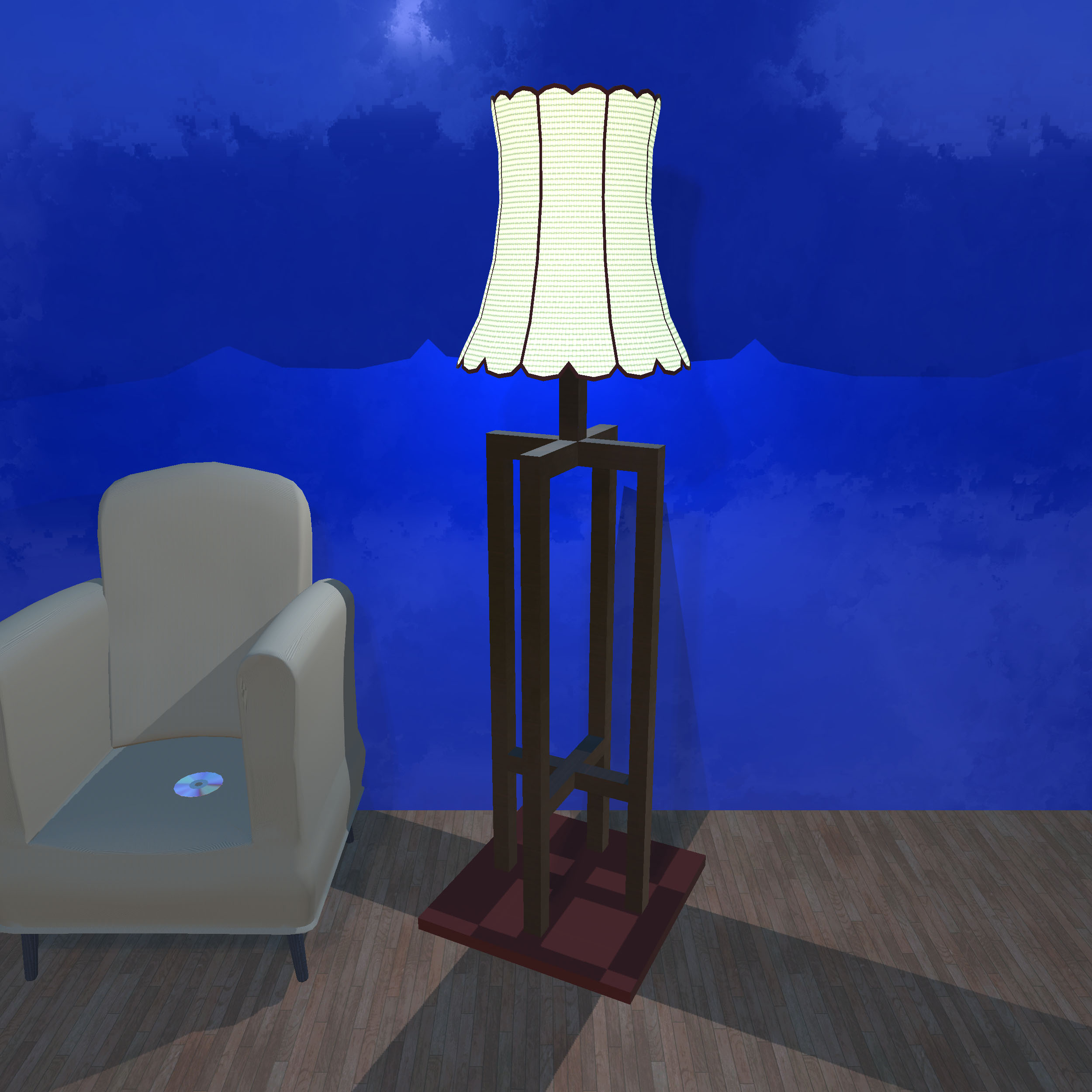}
        \caption{On or off state randomness with a floor lamp.}
        \label{fig:toggle}
    \end{subfigure}
    \\[0.05in]
    \caption{Examples of object state randomness.}
    \label{fig:stateRand}
\end{figure}

We randomize object states to expose the agent to more diverse objects during training. For instance, instead of always having an open laptop or a clean bed, we randomize the openness of each laptop and if each bed is clean or dirty. Figure \ref{fig:stateRand} shows some examples. Our current set of state randomizations include:

\begin{itemize}[leftmargin=0.25in]
    \item \textbf{Toggling objects.} Floor lamp and desk lamp object types have their state toggled on or off.
    \item \textbf{Cleaning or dirtying objects.} Bed object types may appear as either clean or dirty.
    \item \textbf{Opening or closing objects.} Box and laptop object types may 
\end{itemize}

toggling objects on or off (for floor lamp and desk lamp object types), setting objects to clean or dirty (for bed object types), and openness randomizations (for box and laptop object types).


\subsection{Validator}

Once a house is generated, we use a validator to make sure that the agent can successfully navigate to each room in the house, without modifying the scene through interaction (\eg moving an object out of the way). Specifically, we first make sure the agent can teleport to a location inside the house. Then, from that position, we perform a BFS over neighboring positions on a $0.25\times0.25$ meter grid to obtain all reachable positions from the agent's current position. The validator checks to make sure that every room in the house has at least 5 reachable positions on the grid. If the validator fails, we resample a new house using the same room spec, so as to not change the distribution of room specs that we sample from.

\subsection{Limitations and Future Work}

ProcTHOR-10K only uses 1-floor houses. We plan to support multi-floor houses in ProcTHOR-v2.0. This will allow us to capture a wider range of houses and provide better fine-tuning results. Additionally, we plan to scale up our asset databases by leveraging many open-source 3D asset databases, such as ABO~\cite{collins2021abo}, PartNet~\cite{Mo_2019_CVPR}, ShapeNet~\cite{chang2015shapenet}, Google Scanned Objects~\cite{downs2022google}, and CO3D~\cite{reizenstein2021common}, among others.

\section{\env{} Datasheet}
\label{sec:datasheet}

\begin{longtable}{p{0.35\linewidth} |p{0.6\linewidth} }
    \bottomrule
        \multicolumn{2}{c}{\rule{0in}{0.2in}\textbf{Motivation}\vspace{0.10in}}\\
    \toprule
     For what purpose was the dataset created? & The dataset was created to enable the training of simulated embodied agents in substantially more diverse environments.\\[0.15in]
     \midrule
     Who created and funded the dataset? &
      This work was created and funded by the PRIOR team at Allen Institute for AI. See the contributions section for specific details.\\
     
    \bottomrule
        \multicolumn{2}{c}{\rule{0in}{0.2in}\textbf{Composition}\vspace{0.10in}}\\
    \toprule
    What do the instances that comprise the dataset represent? & Each house is specified as a JSON file, which specifies how to populate a 3D Unity scene in AI2-THOR. 


    \\[0.15in]
    \midrule
    How many instances are there in total (of each type, if appropriate)? & There are 10K houses released in the dataset, along with the code to sample substantially more. Section~\ref{sec:analysis} shows the distribution of houses in \env{}-10K. \\[0.15in]
    \midrule
    Does the dataset contain all possible instances or is it a sample (not necessarily random) of instances from a larger set? & We make 10K houses available, but more houses can easily be sampled with the procedural generation scripts.\\[0.15in]
    \midrule
    What data does each instance consist of? & Each house is specified as a JSON file, which precisely describes how our AI2-THOR build should create the house. The procedurally generated JSON files are typically several thousand lines long.\\[0.15in]
    \midrule
    Is there a label or target associated with each instance? & No.\\[0.15in]
    \midrule
    Is any information missing from individual instances? & No.\\[0.15in]
    \midrule
    Are relationships between individual instances made explicit (e.g., users' movie ratings, social network links)? & Each house is generated independently, meaning there are no relationships between the houses.\\[0.15in]
    \midrule
    Are there recommended data splits? & Yes. See Appendix~\ref{sec:assets}.\\[0.15in]
    \midrule
    Are there any errors, sources of noise, or redundancies in the dataset? & No.\\[0.15in]
    \midrule
    Is the dataset self-contained, or does it link to or otherwise rely on external resources (e.g., websites, tweets, other datasets)? & The dataset is self-contained.\\[0.15in]
    \midrule
    Does the dataset contain data that might be considered confidential? & No.\\[0.15in]
    \midrule
    Does the dataset contain data that, if viewed directly, might be offensive, insulting, threatening, or might otherwise cause anxiety? & No.\\
    \bottomrule
        \multicolumn{2}{c}{\rule{0in}{0.2in}\textbf{Collection Process}\vspace{0.10in}}\\
    \toprule
    How was the data associated with each instance acquired? & Each house was procedurally generated. See Appendix~\ref{sec:procthorObjects}.\\[0.15in]
    \midrule
    If the dataset is a sample from a larger set, what was the sampling strategy? & The dataset consists of 1 million houses sampled from the procedural generation scripts.\\[0.15in]
    \midrule
    Who was involved in the data collection process? & The authors were the only people involved in constructing the dataset.\\[0.15in]
    \midrule
    Over what timeframe was the data collected? & Data was collected between the end of 2021 and the beginning of 2022.\\[0.15in]
    \midrule
    Were any ethical review processes conducted? & No.\\
    \bottomrule
        \multicolumn{2}{c}{\rule{0in}{0.2in}\textbf{Preprocessing/Cleaning/Labeling}\vspace{0.10in}}\\
    \toprule
    Was any preprocessing/cleaning/labeling of the data done? & Section~\ref{sec:op} describes the labeling that was done to make the assets spawn in realistic places.\newline
    
    We have also gone through every asset in the asset database to make sure the pivots for each asset are facing a consistent direction.\\[0.15in]
    \midrule
    Was the ``raw'' data saved in addition to the preprocessed/cleaned/labeled data? & There is no raw data associated with the house JSON files.\\[0.15in]
    \midrule
    Is the software that was used to preprocess/clean/label the data available? & The code to generate the houses is made available.\\
    \bottomrule
        \multicolumn{2}{c}{\rule{0in}{0.2in}\textbf{Uses}\vspace{0.10in}}\\
    \toprule
    Has the dataset been used for any tasks already? & Yes. See Section~\ref{sec:experiments} of the paper.\\[0.15in]
    \midrule
    What (other) tasks could the dataset be used for? & The houses can be used in a wide variety of interactive tasks in embodied AI and computer vision.\newline

    Any task that can be performed in AI2-THOR can be performed in ProcTHOR. For instance, in embodied AI, the houses may be used for navigation \cite{khandelwal2021simple, perez2021robot, wortsman2019learning, zhu2017target, wijmans2019dd, yitzhak2022clip, luo2022stubborn, Zheng2022TowardsOp}, multi-agent interaction \cite{jain2020cordial, jain2019two, team2021open}, rearrangement and interaction \cite{weihs2021visual, gadre2022continuous, gan2021threedworld, Chitnis2021LearningNe, Srivastava2021BEHAVIORBF}, manipulation \cite{manipulathor, ni2021towards, ehsani2022object, Xia2021ReLMoGenIM}, Sim2Real transfer \cite{Deitke2020RoboTHORAO, kadian2020sim2real, kumar2021rma}, embodied vision-and-language \cite{shridhar2020alfred, padmakumar2021teach, huang2022language, krantz2020beyond, gordon2018iqa, karamcheti2020learning}, audio-visual navigation \cite{chen2020soundspaces, gan2020look, chen2021savi}, and virtual reality interaction \cite{wu2021communicative, murnane2021simulator, higgins2022head}, among others.\newline

    In the broader field of computer vision, the dataset may be used to study object detection \cite{kotar2022interactron}; NeRFs \cite{mildenhall2020nerf, tancik2022block, greff2022kubric, li20223d}; segmentation, depth, and optimal flow estimation \cite{Feng2021DeepMO, greff2022kubric}; generative modeling \cite{kim2020learning, koh2021pathdreamer, koh2022simple}; occlusion reasoning \cite{ehsani2018segan}; and pose estimation \cite{Charco2021CameraPE}, among others.\newline

    Our framework for loading in procedurally generated houses from a JSON spec also enables the study of scene clutter generation, building more realistic procedurally generated homes, and the development of synthetically generated spaces to train embodied agents in factories \cite{narang2022factory}, offices, grocery stores \cite{mata2022standardsim}, and full procedurally generated cities.\\[0.15in]
    \midrule
    Is there anything about the composition of the dataset or the way it was collected and preprocessed/cleaned/labeled that might impact future uses? & No.\\[0.15in]
    \midrule
    Are there tasks for which the dataset should not be used? & Our dataset may be used for both commercial and non-commercial purposes.\\
    \bottomrule
        \multicolumn{2}{c}{\rule{0in}{0.2in}\textbf{Distribution}\vspace{0.10in}}\\
    \toprule
    Will the dataset be distributed to third parties outside of the entity on behalf of which the dataset was created? & Yes. We plan to make the entirety of the work open-source, including the code used to generate and load houses, the initial static dataset of 1 million procedurally generated house JSON files, and the asset and material databases.\\[0.15in]
    \midrule
    How will the dataset be distributed? & The static house JSON files will be distributed with a custom Python package.\newline
    
    The code, asset, and material databases will be distributed on GitHub.\\[0.15in]
    \midrule
    Will the dataset be distributed under a copyright or other intellectual property (IP) license, and/or under applicable terms of use (ToU)? & The house dataset, 3D asset database, and generation code will be released under the Apache 2.0 license. \\[0.15in]
    \midrule
    Have any third parties imposed IP-based or other restrictions on the data associated with the instances? & No.\\[0.15in]
    \midrule
    Do any export controls or other regulatory restrictions apply to the dataset or to individual instances? & No.\\
    \bottomrule
        \multicolumn{2}{c}{\rule{0in}{0.2in}\textbf{Maintenance}\vspace{0.10in}}\\
    \toprule
    Who will be supporting/hosting/maintaining the dataset? & The authors will be providing support, hosting, and maintaining the dataset.\\[0.15in]
    \midrule
    How can the owner/curator/manager of the dataset be contacted? & For inquiries, email <mattd@allenai.org>.\\[0.15in]
    \midrule
    Is there an erratum? & We will use GitHub issues to track issues with the dataset.\\[0.15in]
    \midrule
    Will the dataset be updated? & We expect to continue adding support for new features to continue to make procedurally generated houses even more diverse and realistic. We also intend to support new tasks in the future.\\[0.15in]
    \midrule
    If the dataset relates to people, are there applicable limits on the retention of the data associated with the instances (e.g., were the individuals in question told that their data would be retained for a fixed period of time and then deleted)? & The dataset does not relate to people. \\[0.15in]
    \midrule
    Will older versions of the dataset continue to be supported/hosted/maintained? & Yes. Revision history will be available for older versions of the dataset.\\[0.15in]
    \midrule
    If others want to extend/augment/build on/contribute to the dataset, is there a mechanism for them to do so? & Yes. The work will be open-sourced and we intend to provide support to help others use and build upon the dataset.\\[0.15in]
\bottomrule
    \caption{A datasheet \cite{Gebru2021DatasheetsFD} for \env{} and \env{}-10K.}
\end{longtable}

\newpage

\section{\elienv}

\begin{figure}[htbp]
    \vspace{-0.1in}
    \centering
    \includegraphics[width=\textwidth]{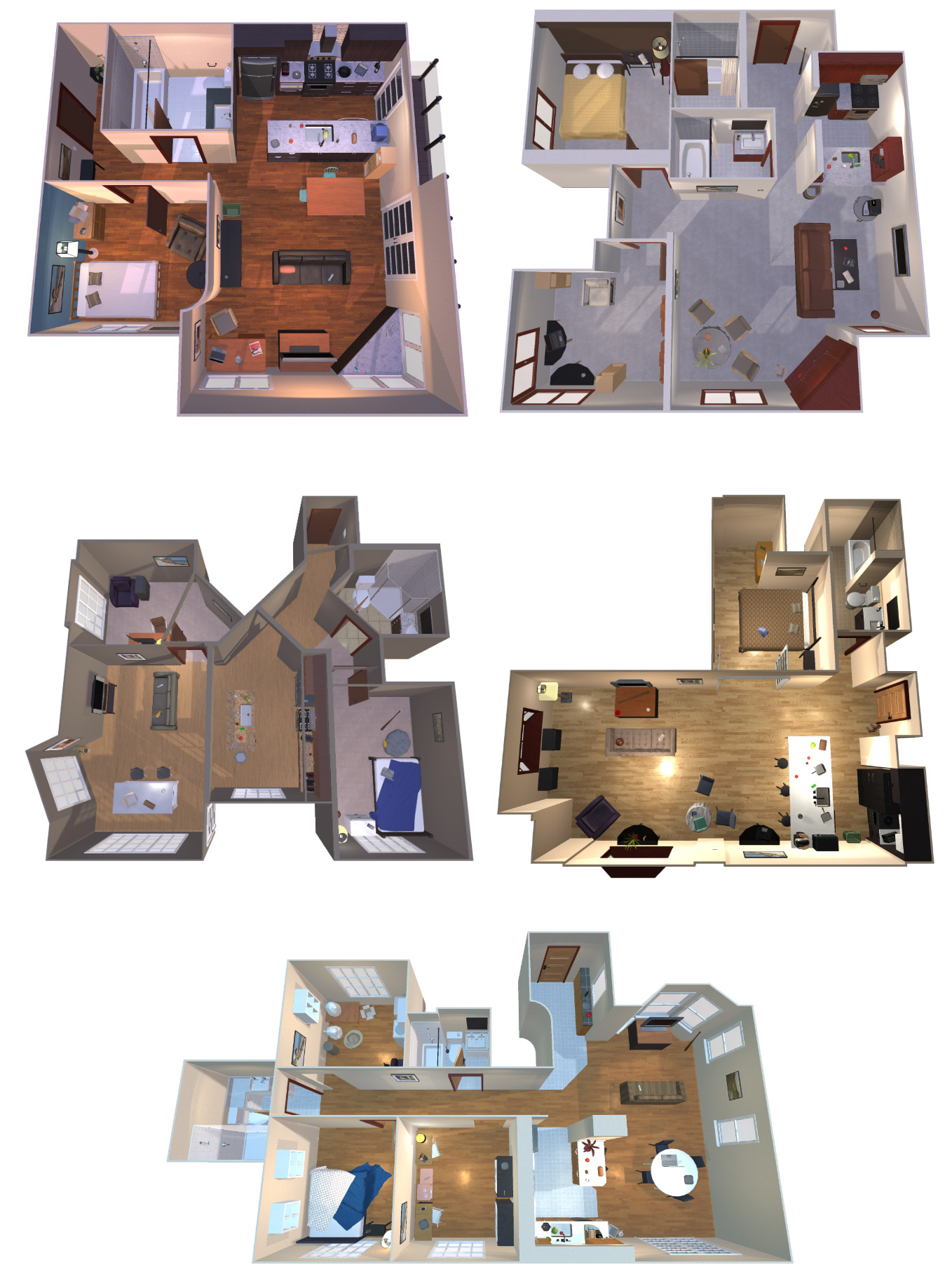}
    \caption{Top-down images of the 5 custom-built interactive validation houses in \elienv{}. The goal of these houses is to evaluate interactive agents in more realistic and larger home environments.}
    \label{fig:elienv}
\end{figure}

\pagebreak

\subsection{Datasheet}
\label{sec:eli-datasheet}

\begin{longtable}{p{0.35\linewidth} |p{0.6\linewidth} }
    \bottomrule
        \multicolumn{2}{c}{\rule{0in}{0.2in}\textbf{Motivation}\vspace{0.10in}}\\
    \toprule
     For what purpose was the dataset created? & \elienv{} was created to enable the evaluation of embodied agents in large, realistic, and interactive household environments.\\[0.15in]
     \midrule
     Who created and funded the dataset? &
     This work was created and funded by the PRIOR team at Allen Institute for AI. See the contributions section for specific details.\\
     
    \bottomrule
        \multicolumn{2}{c}{\rule{0in}{0.2in}\textbf{Composition}\vspace{0.10in}}\\
    \toprule
    What do the instances that comprise the dataset represent? & Instances of the dataset comprise interactive 3D houses that were built in Unity and can be used with our custom build of the AI2-THOR API.
    \\[0.15in]
    \midrule
    How many instances are there in total (of each type, if appropriate)? & There are 10 total houses, comprising 5 validation houses and 5 testing houses.\\[0.15in]
    \midrule
    Does the dataset contain all possible instances or is it a sample (not necessarily random) of instances from a larger set? & The dataset is self-contained.\\[0.15in]
    \midrule
    What data does each instance consist of? & Each instance of a house is a Unity scene, which includes data such as the placement of objects, lighting, and texturing.\\[0.15in]
    \midrule
    Is there a label or target associated with each instance? & No.\\[0.15in]
    \midrule
    Is any information missing from individual instances? & No.\\[0.15in]
    \midrule
    Are relationships between individual instances made explicit (e.g., users' movie ratings, social network links)? & Each house was independently created.\\[0.15in]
    \midrule
    Are there recommended data splits? & Yes. The houses themselves are partitioned as 5 validation houses and 5 testing houses. The assets placed in the house follow the same train/val/test splits used in \env{}-10K.\\[0.15in]
    \midrule
    Are there any errors, sources of noise, or redundancies in the dataset? & No.\\[0.15in]
    \midrule
    Is the dataset self-contained, or does it link to or otherwise rely on external resources (e.g., websites, tweets, other datasets)? & The dataset is self-contained.\\[0.15in]
    \midrule
    Does the dataset contain data that might be considered confidential? & No.\\[0.15in]
    \midrule
    Does the dataset contain data that, if viewed directly, might be offensive, insulting, threatening, or might otherwise cause anxiety? & No.\\
    \bottomrule
        \multicolumn{2}{c}{\rule{0in}{0.2in}\textbf{Collection Process}\vspace{0.10in}}\\
    \toprule
    How was the data associated with each instance acquired? & Each house was professionally hand-modeled by 3D artists. Most objects placed in the hosues come from the \env{} asset database. However, countertops, showers, and many cabinets were custom built.\\[0.15in]
    \midrule
    If the dataset is a sample from a larger set, what was the sampling strategy? & The dataset consists of 1 million houses sampled from the procedural generation scripts.\\[0.15in]
    \midrule
    Over what timeframe was the data collected? & The houses were built towards the beginning of 2022.\\[0.15in]
    \midrule
    Were any ethical review processes conducted? & No.\\
    \bottomrule
        \multicolumn{2}{c}{\rule{0in}{0.2in}\textbf{Preprocessing/Cleaning/Labeling}\vspace{0.10in}}\\
    \toprule
    Was any preprocessing/cleaning/labeling of the data done? & No.\\[0.15in]
    \midrule
    Was the ``raw'' data saved in addition to the preprocessed/cleaned/labeled data? & There is no raw data associated with the \elienv{} houses.\\[0.15in]
    \midrule
    Is the software that was used to preprocess/clean/label the data available? & Yes. We will open-source the \elienv{} houses and they can be opened and viewed in Unity.\\
    \bottomrule
        \multicolumn{2}{c}{\rule{0in}{0.2in}\textbf{Uses}\vspace{0.10in}}\\
    \toprule
    Has the dataset been used for any tasks already? & Yes. Please see Section~\ref{sec:experiments} of the paper.\\[0.15in]
    \midrule
    What (other) tasks could the dataset be used for? & The tasks can be used for any type of navigation and interaction tasks in embodied AI. The houses are built into our build of AI2-THOR, meaning \elienv{} can work with any task that can be performed in AI2-THOR.\newline
    
    We especially think \elienv{} will be useful as an evaluation suite for evaluating different sets of \env{} tasks and evaluating agents trained on different sets of procedurally generated houses.\\[0.15in]
    \midrule
    Is there anything about the composition of the dataset or the way it was collected and preprocessed/cleaned/labeled that might impact future uses? & No.\\[0.15in]
    \midrule
    Are there tasks for which the dataset should not be used? & Our dataset may be used for both commercial and non-commercial purposes.\\
    \bottomrule
        \multicolumn{2}{c}{\rule{0in}{0.2in}\textbf{Distribution}\vspace{0.10in}}\\
    \toprule
    Will the dataset be distributed to third parties outside of the entity on behalf of which the dataset was created? & Yes. All houses in \elienv{} will be released to the open-source community and available through our build of the AI2-THOR Python API.\\[0.15in]
    \midrule
    How will the dataset be distributed? & The houses will be distributed on GitHub and available to open as Unity scenes.\\[0.15in]
    \midrule
    Will the dataset be distributed under a copyright or other intellectual property (IP) license, and/or under applicable terms of use (ToU)? & \elienv{} will be released under the Apache 2.0 license.\\[0.15in]
    \midrule
    Have any third parties imposed IP-based or other restrictions on the data associated with the instances? & No.\\[0.15in]
    \midrule
    Do any export controls or other regulatory restrictions apply to the dataset or to individual instances? & No.\\
    \bottomrule
        \multicolumn{2}{c}{\rule{0in}{0.2in}\textbf{Maintenance}\vspace{0.10in}}\\
    \toprule
    Who will be supporting/hosting/maintaining the dataset? & The authors will be providing support, hosting, and maintaining the dataset.\\[0.15in]
    \midrule
    How can the owner/curator/manager of the dataset be contacted? & \textit{Omitted for anonymous review.}\\[0.15in]
    \midrule
    Is there an erratum? & We will use GitHub issues to track issues with the dataset once it is published.\\[0.15in]
    \midrule
    Will the dataset be updated? & \elienv{} is currently in maintenance mode and we do not expect it to update much from its current state. However, we plan to actively support future AI2-THOR functionalities in \elienv, such as support for new robots, more advanced interaction capabilities, and bug fixes.\\[0.15in]
    \midrule
    If the dataset relates to people, are there applicable limits on the retention of the data associated with the instances (e.g., were the individuals in question told that their data would be retained for a fixed period of time and then deleted)? & The dataset does not relate to people. \\[0.15in]
    \midrule
    Will older versions of the dataset continue to be supported/hosted/maintained? & Yes. Revision history will be available in the GitHub repository.\\[0.15in]
    \midrule
    If others want to extend/augment/build on/contribute to the dataset, is there a mechanism for them to do so? & Yes. The work will be open-sourced and we intend to provide support to help others use and build upon the dataset.\\[0.15in]
\bottomrule
    \caption{A datasheet \cite{Gebru2021DatasheetsFD} for the artist-designed \elienv{} houses.}
\end{longtable}

\subsection{Analysis}

\begin{figure}[htbp]
    \centering
    \includegraphics[width=\textwidth]{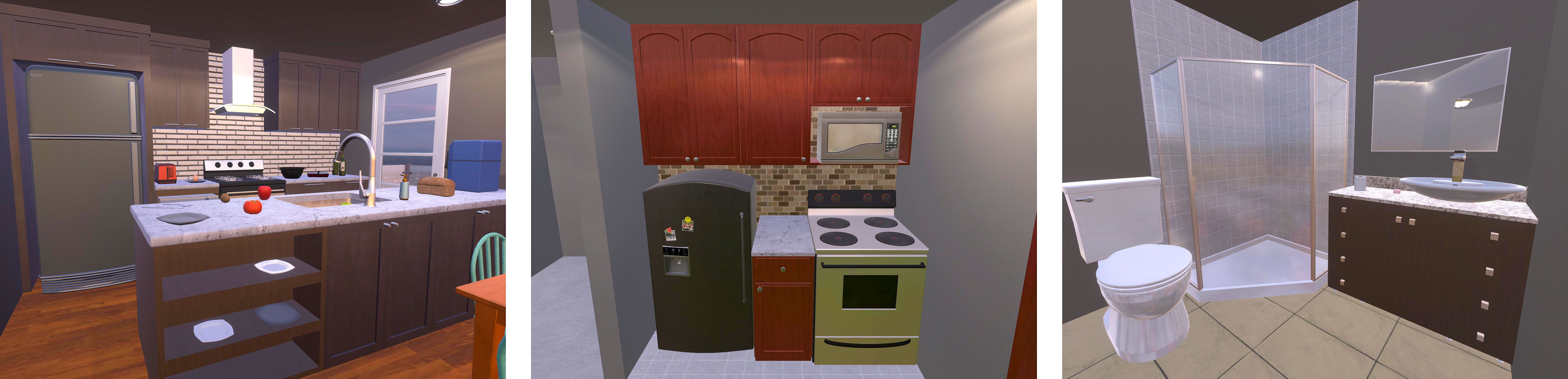}
    \caption{Examples of images inside of 2 hand-modeled kitchens and 1 hand-modeled bathroom from \elienv{} validation.}
    \label{fig:eliExamples}
\end{figure}

\elienv{} consists of 10 remarkably high-quality large interactive 3D houses. Figure \ref{fig:elienv} shows top-down images of the 5 validation houses. Figure \ref{fig:eliExamples} shows some examples of images taken inside of 2 kitchens and a bedroom from \elienv{} validation.

\begin{figure}[htbp]
    \centering
    \includegraphics[width=\textwidth]{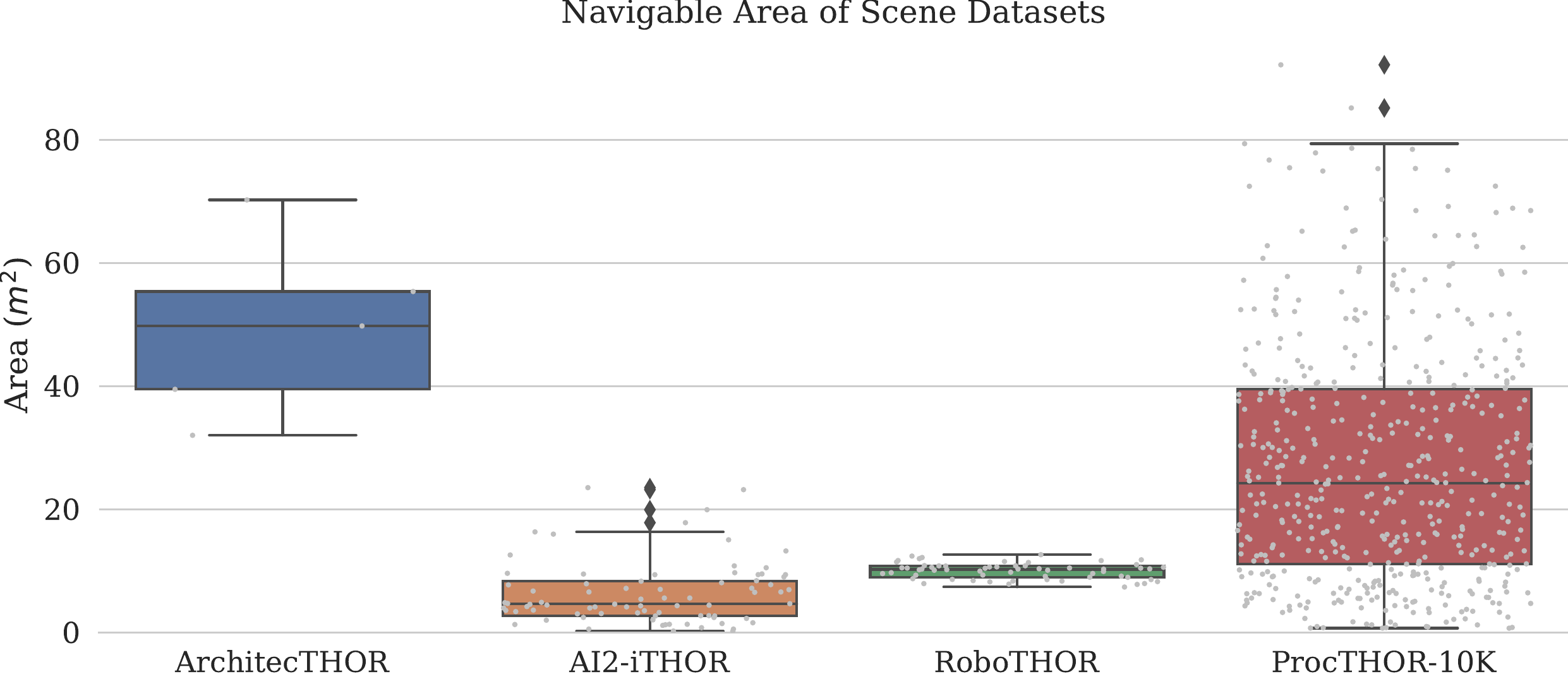}
    \caption{Box plots of the navigable areas for \elienv{} compared to AI2-iTHOR, RoboTHOR, and \env{}-10K. Validation scenes were used to calculate the data for \elienv{}, and training scenes were used to calculate the data for AI2-iTHOR, RoboTHOR, and \env{}-10K.}
    \label{fig:archNav}
\end{figure}

\elienv{} was built to be much larger than AI2-iTHOR and RoboTHOR. Figure \ref{fig:archNav} shows the size comparisons between comparable hand-built scene datasets in AI2-iTHOR and RoboTHOR, measured in navigable area. Notice that the navigable area in \elienv{} is substantially larger than in those. The figure also shows the navigable areas in \env{}-10K span the spectrum of navigable areas between AI2-iTHOR, RoboTHOR, and \elienv{}.

\begin{figure}[htbp]
    \centering
    \includegraphics[width=\textwidth]{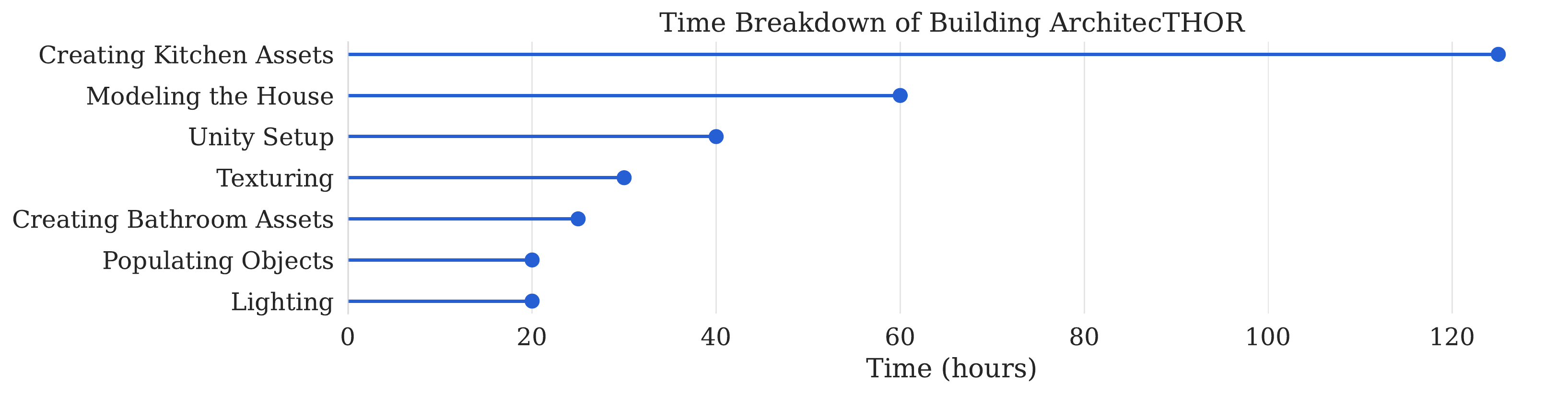}
    \caption{Cumulative time breakdown of the development of \elienv{} across 3D artists.}
    \label{fig:eliTime}
\end{figure}

In total, the creation of the 10 houses in \elienv{} took approximately 320 hours of cumulative work by professional 3D artists. Figure \ref{fig:eliTime} shows the time breakdown of which parts of the process took the longest. In particular, the creation of custom assets for the kitchen, such as modeling each of the countertops and cabinets, took the longest amount of time, followed by modeling the 3D structure of house.

\pagebreak

\newpage

\section{Input Modalities}

\begin{figure}[hb]
    \vspace{-0.1in}
    \centering
    \begin{subfigure}{0.405\textwidth}
        \centering
        \includegraphics[width=\textwidth]{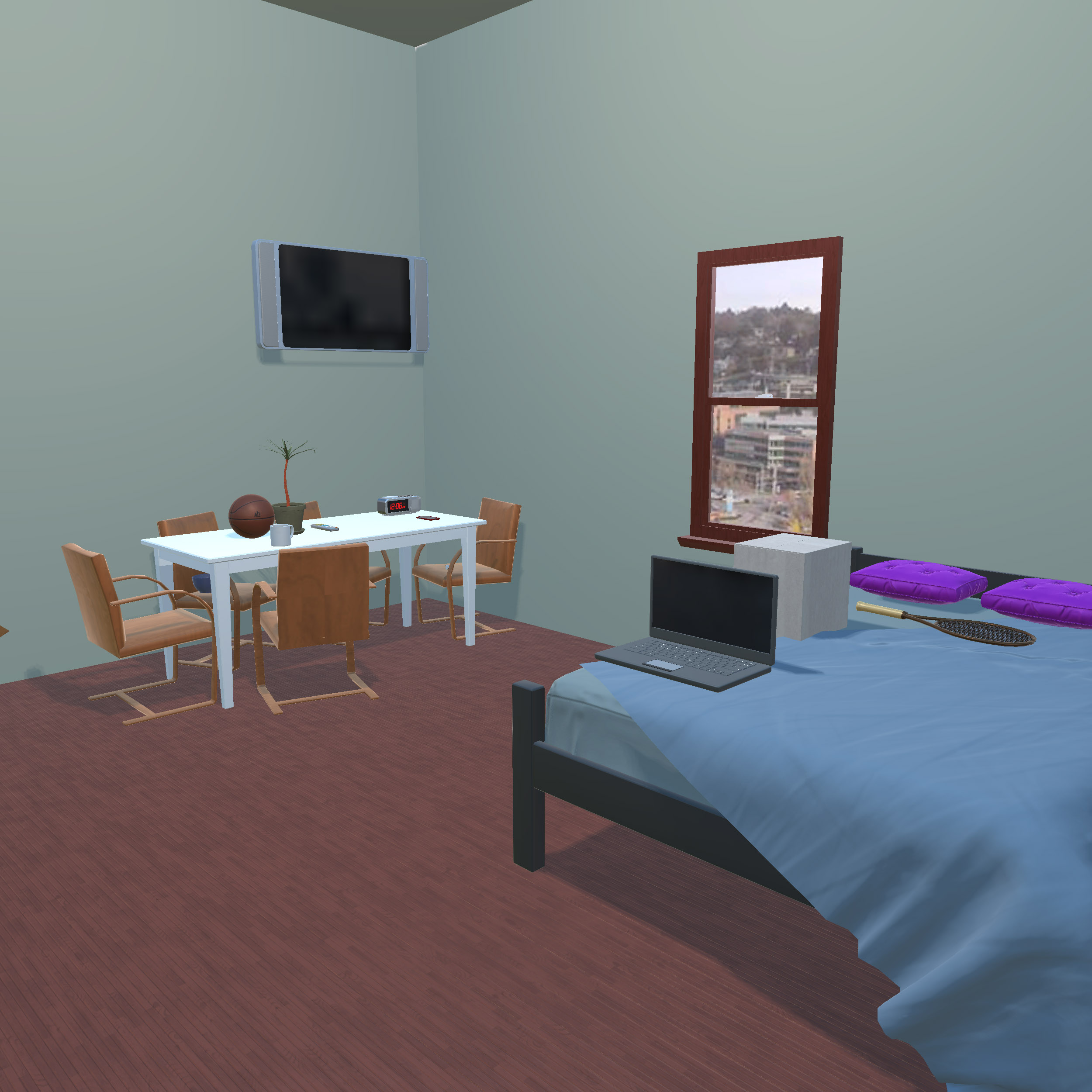}
        \caption{RGB}
    \end{subfigure}
    \begin{subfigure}{0.405\textwidth}
        \centering
        \includegraphics[width=\textwidth]{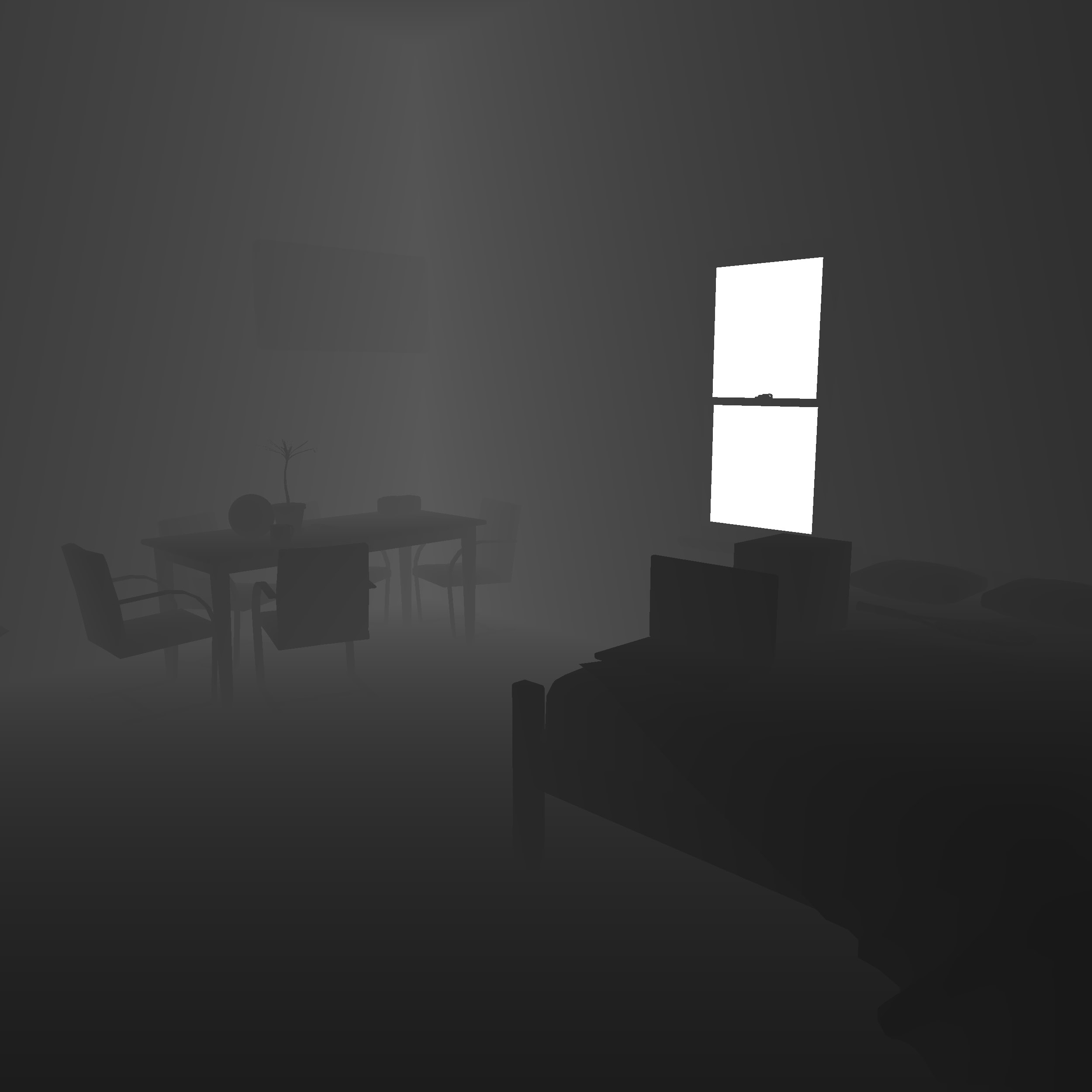}
        \caption{Depth}
    \end{subfigure}\\[0.05in]
    \begin{subfigure}{0.405\textwidth}
        \centering
        \includegraphics[width=\textwidth]{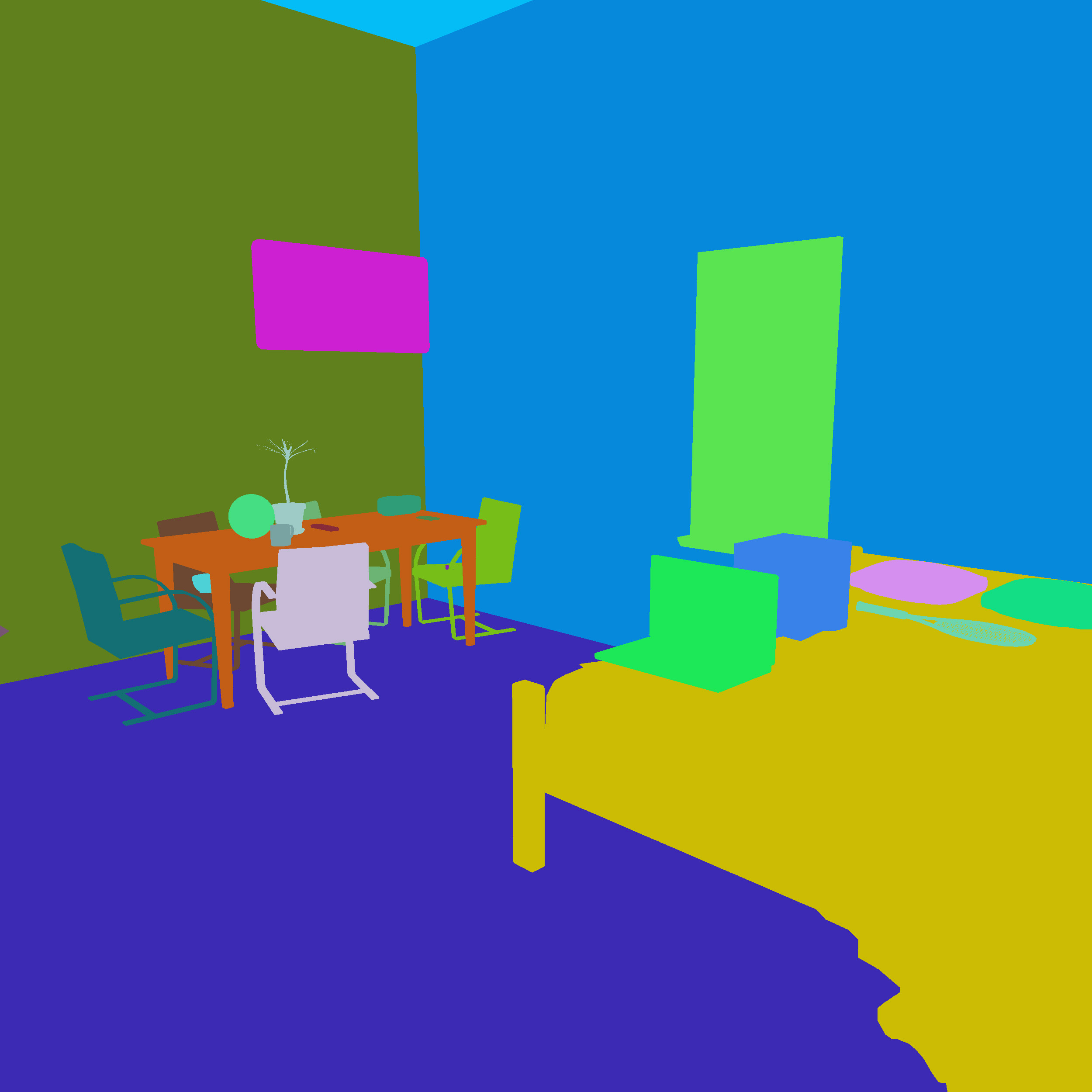}
        \caption{Instance Segmentation}
    \end{subfigure}
    \begin{subfigure}{0.405\textwidth}
        \centering
        \includegraphics[width=\textwidth]{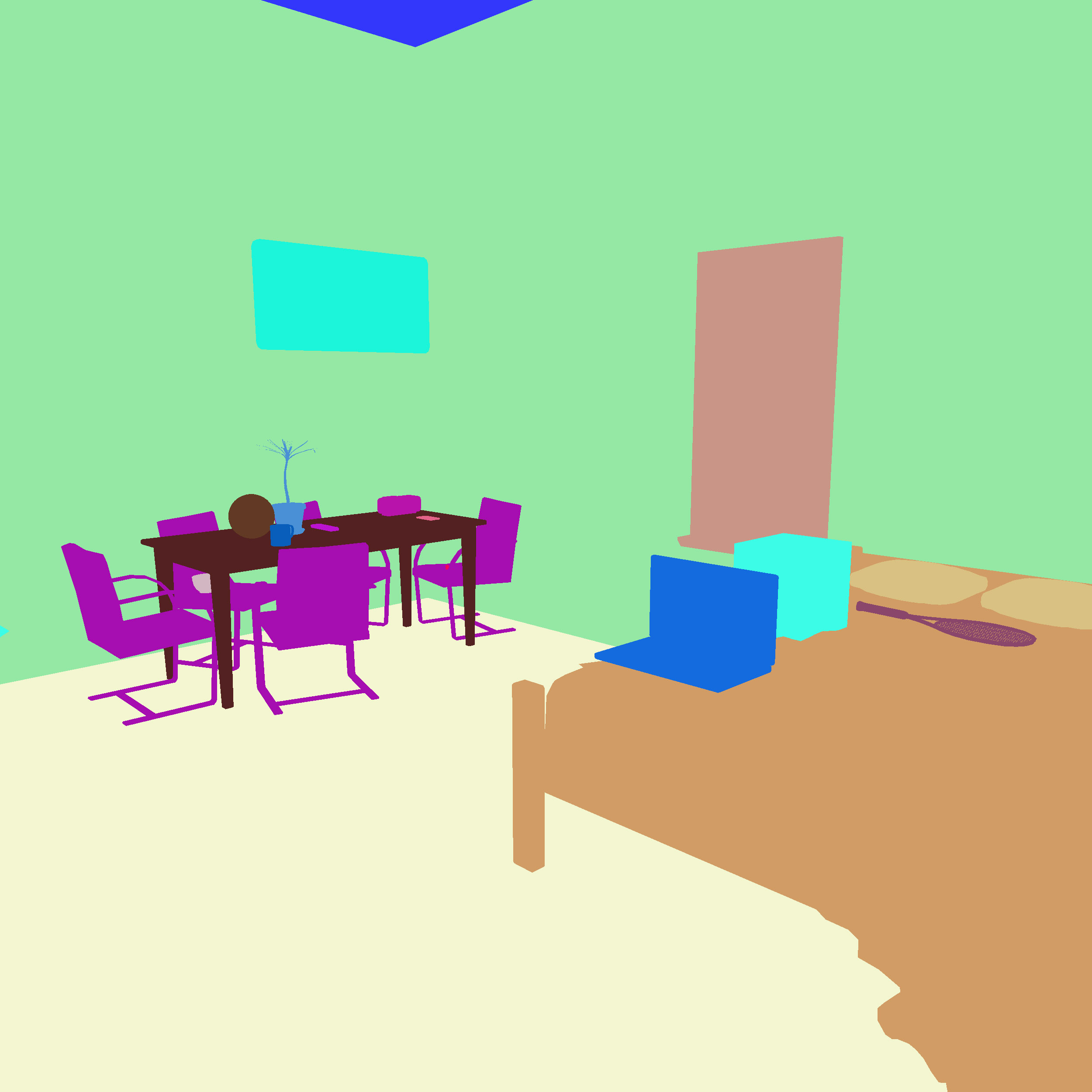}
        \caption{Semantic Segmentation}
    \end{subfigure}\\[0.05in]
    \begin{subfigure}{0.405\textwidth}
        \centering
        \includegraphics[width=\textwidth]{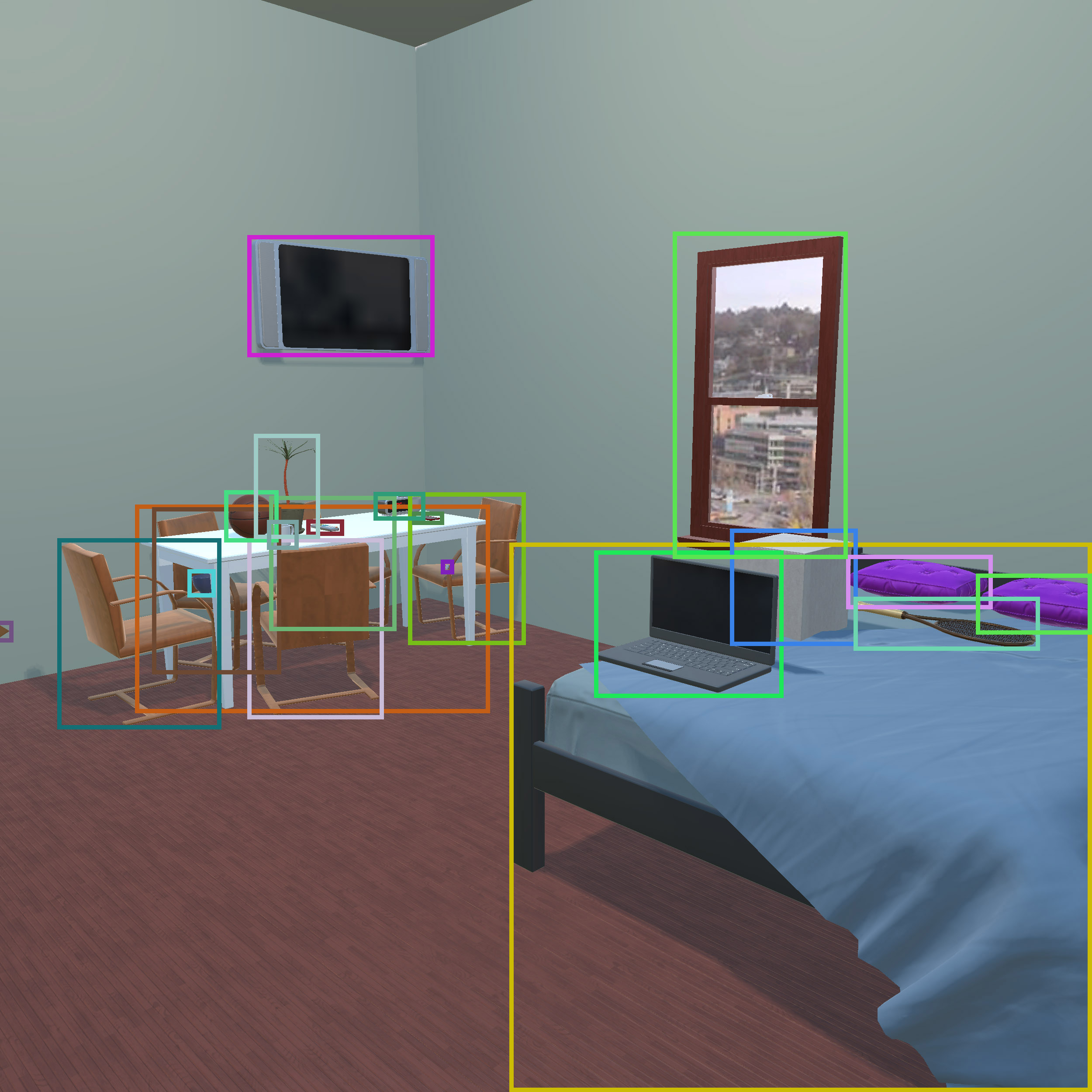}
        \caption{Bounding Box Annotations}
    \end{subfigure}
    \begin{subfigure}{0.405\textwidth}
        \centering
        \includegraphics[width=\textwidth]{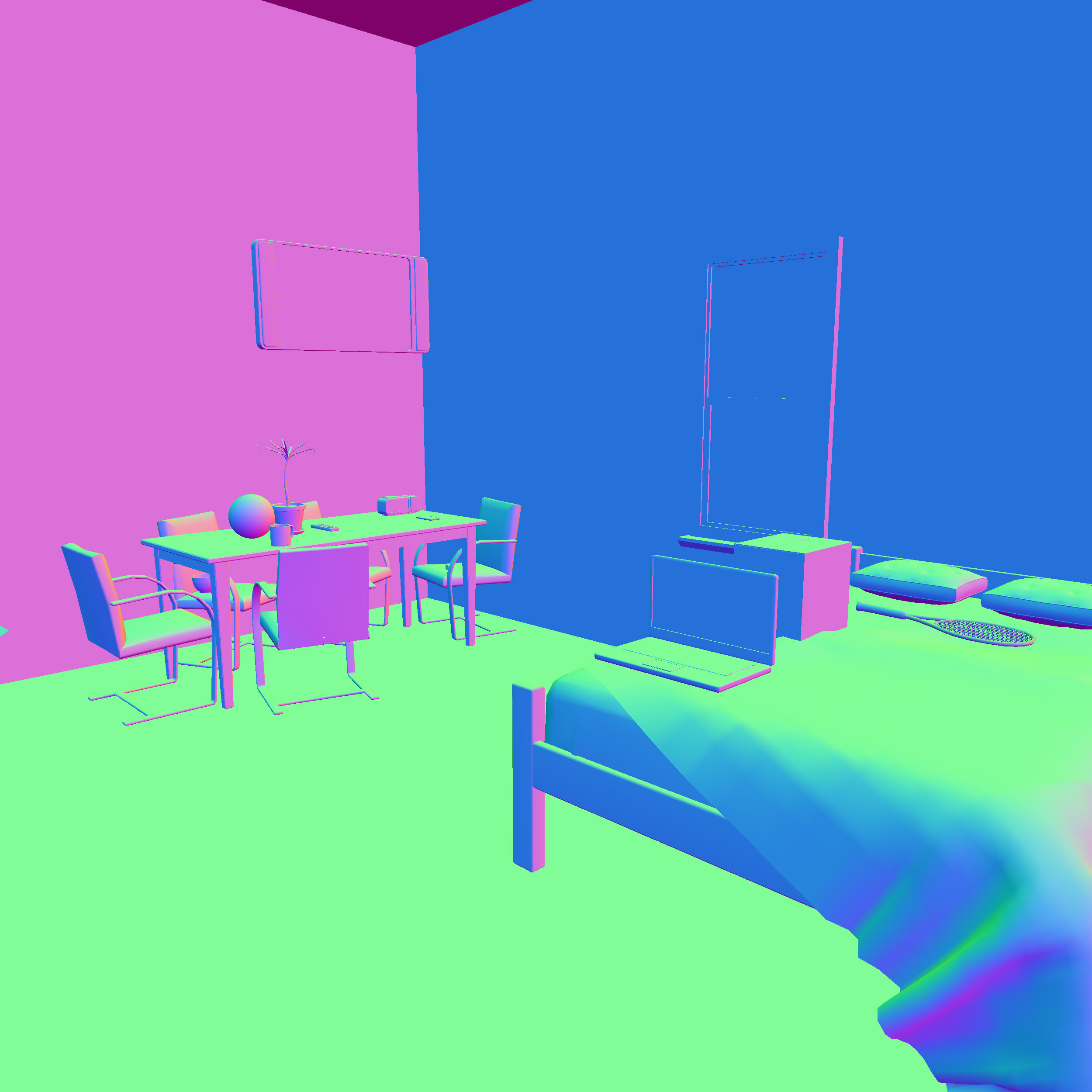}
        \caption{Surface Normals}
    \end{subfigure}\\[0.00in]
    \caption{Examples of image-based modalities available in ProcTHOR include RGB, depth, instance segmentation, semantic segmentation, bounding box annotations, and surface normals. More image modalities can be added by modifying the Unity backend.}
    \label{fig:modalities}
\end{figure}

\newpage





\section{Experiment details}

This section discusses the training details used for our experiments. We discuss baselines, \env{} pre-training, and environment-specific fine-tuning details for the tasks of ObjectNav, ArmPointNav, and rearrangement.


\subsection{ObjectNav experiments}

For ObjectNav experiments, agents are given a target object type (\eg a bed) and are tasked with finding a path in the environment that navigates to that target object type. The task setup matches what is commonly used in embodied AI \cite{Deitke2020RoboTHORAO, batra2020objectnav, khandelwal2021simple, Ramrakhya2022HabitatWebLE}, although we only utilize forward-facing egocentric RGB images at each time step. All ObjectNav experiments are trained with a simulated LoCoBot (Low Cost Robot) agent \cite{locobot}. The task and training details are described below.

\paragraph{Evaluation.}
Following \cite{Anderson2018OnEO}, an ObjectNav task is considered successful if all of the following conditions are met:
\begin{enumerate}[leftmargin=0.25in]
    \item The agent terminates the episode by issuing the \textsc{Done} action.
    \item The target object type is within a distance of 1 meter from the agent's camera.
    \item The object is visible in the final frame from the agent's camera. For instance, if (1) and (2) are satisfied, and the agent is looking in the direction of the object, but the target object is occluded behind a wall, then the task is unsuccessful. Similarly, if the target object type is located in the opposite direction of where the agent is looking, then the task will be unsuccessful.
\end{enumerate}

We also use SPL to evaluate the efficiency of the agent's trajectory to the target object. SPL is defined and discussed in \cite{Anderson2018OnEO, batra2020objectnav}. A house may have multiple instances of objects for a given type that the agent can successfully reach. For instance, a house may have multiple bedrooms, where each bedroom includes a bed. Here, if the agent navigates to any of the beds, the episode is successful. To calculate SPL in these scenarios, the shortest path length for the task is the minimum shortest path length from the starting position of the agent to any of the reachable target objects of the given type, regardless of which instance the agent navigates towards.

\paragraph{Actions.}
For each of the trained models, we use a discrete action space consisting of 6 actions, which is shown in Table \ref{tab:ONActions}. Following common practice \cite{Deitke2020RoboTHORAO, kadian2020sim2real}, we use stochastic actuation to better simulate noise in the real world.

\begin{table}[htbp]
    \centering
    \begin{tabular}{ p{0.2\textwidth} p{0.7\textwidth} }
        \toprule
        \textbf{Action}\qquad\qquad\qquad & \textbf{Description}\\
        \midrule
        \textsc{MoveAhead} & Attempts to move the agent forward by $\delta_m\sim\mathcal N(\mu=0.25, \sigma=0.01)$ meters from its current facing direction. If moving the agent forward by $\delta_m$ meters results in a collision in the scene (\eg there is a wall directly in-front of the agent within $\delta_m$ meters), the action fails and the agent's position remains unchanged. \\~\\
        \textsc{RotateRight}\newline\textsc{RotateLeft} & Rotates the agent rightwards or leftwards from its current forward facing direction by $\delta_r\sim\mathcal N(\mu=30, \sigma=0.5)$ degrees.\\~\\
        \textsc{LookUp}\newline\textsc{LookDown} & Tilts the agent's camera up or down by 30 degrees.\\~\\
        \textsc{Done} & A signal from the agent to terminate the episode and evaluate the trajectory from its current state. Discussed in \cite{Anderson2018OnEO}.\\
        \toprule\\[-0.05in]
    \end{tabular}
    \caption{The action space for ObjectNav experiments.}
    \label{tab:ONActions}
\end{table}

\paragraph{Model.} We use the relatively simple EmbCLIP~\cite{khandelwal2021simple} training setup for training all ObjectNav experiments. Table~\ref{tab:onhps} shows the hyperparameters used during training, which are adapted from~\cite{khandelwal2021simple}. Except for the ``ProcTHOR+Large'' model trained for HM3D (described below), we otherwise use the same model architecture across ObjectNav experiments. Namely, at each time step, the agent receives a $3\times 224\times224$ egocentric RGB image from its camera. The image is processed with a frozen RN50 CLIP-ResNet visual encoder \cite{Radford2021LearningTV} to produce a $2048\times 7\times7$ visual embedding, $\mathbf V_t$. The embedding is compressed through a 2-layer CNN (going from $2048$ to $128$ to $32$ channels) with $1\times1$ convolutions \cite{szegedy2015going} to obtain a $32\times 7\times 7$ tensor, $\mathbf V^\prime_t$.

The target object type is represented as an integer in $\{0, 1, \ldots, T\}$, where $T$ is the number of target object types used during training. We use an embedding of $t$ to obtain a 32-dimensional vector. The vector is resized to be a $32\times1\times1$ tensor. The tensor is then expanded to be of size $32\times7\times7$, to form our goal target object type embedding $\mathbf G_t$, where the $32\times1\times1$ tensor is copied $7\times7$ times.

We concatenate $\mathbf V^\prime_t$ and $\mathbf G_t$ to form a $64\times7\times7$ tensor, which is compressed with a 2-layer CNN to form a $32\times7\times7$ tensor, $\mathbf Z_t$. The tensor $\mathbf Z$ is flattened to form a 1568 dimensional vector, $\mathbf z_t$. Following \cite{ni2021towards}, we use an embedding of the previous action, represented as an integer in $\{0, 1, \ldots, 5\}$, to obtain a 6 dimensional vector $\mathbf a_{t-1}$. We concatenate $\mathbf z_t$ and $\mathbf a_{t-1}$ to form a 1574 dimensional vector $\mathbf x_t$. The vector $\mathbf x_t$ is passed through a 1-layer GRU \cite{cho2014learning, chung2014empirical} with a hidden belief state $\mathbf b_{t-1}$, of size 512, to obtain $\mathbf b_t$.

Using an actor-critic formulation, the 512-dimensional belief state $\mathbf b_t$ is passed through a 1-linear layer, representing the \textit{actor}, to get a 6-dimensional vector, where each entry represents an action. The 6-dimensional vector is passed through a softmax function to obtain the agent's policy $\pi$ (\ie the probability distribution over the action space). We sample from $\pi$ to choose the next action. We also pass the belief state $\mathbf b_t$ through a separate 1-linear layer, representing the \textit{critic} to obtain the scalar $v$, estimating the value of the current state.

The ``ProcTHOR+Large'' is similar to the above except we: (1) use the larger RN50x16 CLIP-ResNet model, (2) use a 1024-dimensional hidden belief state in our GRU, and (3) input images to the model at a $512{\times}384$ resolution.

\begin{table}[htbp]
    \centering
    \begin{tabular}{ l l }
        \toprule
        \textbf{Hyperparameter}\qquad\qquad\qquad & \textbf{Value}\\
        \midrule
        Discount factor ($\gamma$) & 0.99 \\
        GAE parameter ($\lambda$) & 0.95 \\
        Value loss coefficient & 0.5 \\
        Entropy loss coefficient & 0.01 \\
        Clip parameter ($\epsilon$) & 0.1 \\
        Rollout timesteps & 20\\
        Rollouts per minibatch & 1\\
        Learning rate & 3e-4\\
        Optimizer & Adam \cite{kingma2014adam} \\
        Gradient clip norm & 0.5 \\
        \toprule\\[-0.05in]
    \end{tabular}
    \caption{Training hyperparameters for ObjectNav experiments.}
    \label{tab:onhps}
\end{table}

\paragraph{Training.} Each agent is trained using DD-PPO~\cite{schulman2017proximal, wijmans2019dd}, using a clip parameter $\epsilon=0.1$, an entropy loss coefficient of 0.01, and a value loss coefficient of 0.5. Agents are trained to maximize the cumulative discounted rewards $\sum_{t=0}^H\gamma^t\cdot r_t$, where we set the discount factor $\gamma$ to $0.99$ and the episode's horizon $H$ to 500 steps. We also employ GAE~\cite{schulman2015high} parameterized by $\lambda=0.95$. 

\paragraph{Reward.} The reward function follows that of \cite{khandelwal2021simple}. Specifically, at each time step, it is calculated as $r_t = \max(0, \min\Delta_{0:t-1} - \Delta_{t}) + s_t - \rho$, where:

\begin{itemize}[leftmargin=0.25in]
    \item $\min\Delta_{0:t-1}$ is the minimum L2 distance from the agent to any of the reachable instances of the target object type that the agent has observed over steps $\{0,1,\ldots,t-1\}$.
    \item $\Delta_t$ is the current L2 distance from the agent to the nearest reachable instance of the target object type.
    \item $s_t$ is the reward for successfully completing the episode. If the agent takes the \textsc{Done} action and the episode is deemed successful, then $s_t$ is $10$. Otherwise, it is $0$.
    \item $\rho$ is the step penalty that encourages the agent to finish the episode quickly. It is set to $0.01$.
\end{itemize}


\paragraph{ProcTHOR pre-training.} We pre-train our ObjectNav agents on the full set of 10k training houses in \env{}-10K.\footnote{When training the ``ProcTHOR+Large'' model used in the HM3D challenge, we use a modified set of 10K houses, see below for details.} We pre-train with all $T=16$ target object types, which are shown in Table~\ref{tab:targetObjectTypes}. The agent is trained for 423 million steps, although by 200 million steps, the agent has reached 90\% of its peak performance. We used multi-node training to train on 3 AWS g4dn.12xlarge machines, which takes approximately 5 days to complete.

\newcommand\y{\includegraphics[width=0.135in]{figures/success.pdf}}
\newcommand\n{\includegraphics[width=0.097in]{figures/failure.pdf}}

\begin{table}[htbp]
    \centering
    \begin{tabular}{lcccc}
        \toprule
        Object Type & RoboTHOR & HM3D-Semantics & AI2-iTHOR & \elienv{}\\
        \midrule
        Alarm Clock & \y & \n &\y &\y \\[0.035in]
        Apple & \y & \n & \y & \y\\[0.035in]
        Baseball Bat & \y & \n & \y& \y \\[0.035in]
        Basketball & \y & \n & \y& \y \\[0.035in]
        Bed & \n & \y & \y& \y \\[0.035in]
        Bowl & \y & \n & \y& \y \\[0.035in]
        Chair & \n & \y & \y& \y \\[0.035in]
        Garbage Can & \y & \n & \y& \y \\[0.035in]
        House Plant & \y & \y & \y& \y \\[0.035in]
        Laptop & \y & \n & \y& \y \\[0.035in]
        Mug & \y & \n & \y& \y \\[0.035in]
        Sofa & \n & \y & \y& \y \\[0.035in]
        Spray Bottle & \y & \n & \y& \y \\[0.035in]
        Television & \y & \y & \y& \y \\[0.035in]
        Toilet & \n & \y & \y& \y \\[0.035in]
        Vase & \y& \n & \y& \y \\
     \toprule\\[-0.05in]
    \end{tabular}
    \caption{The target objects that are used for each ObjectNav task.}
    \label{tab:targetObjectTypes}
\end{table}

\emph{Sampling target object types.} To sample the target object type for a given episode, we restrict ourselves to only sampling target object types that have a possibility of leading to a successful episode. For instance, even if there is an object like an apple in the scene, it might be located in the fridge, and so if it was used as a target object, the agent would never succeed because the object would never appear visible in the frame (without any manipulation actions). Therefore, we impose a constraint that the target object must be visible without any form of manipulation.

For each house, we use an approximation to determine the set of target object instances that the agent can successfully reach, without any manipulation. Specifically, we start by teleporting the agent into the house, and then perform a BFS over a $0.25\times 0.25$ meter grid to obtain the reachable positions in the scene. A position is considered reachable if teleporting to it would not cause any collisions with any other objects, and the agent is successfully placed on the floor. Then, for each candidate instance of every target object type, we look at the nearest 6 reachable agent positions $\langle x^{(a)}, z^{(a)}\rangle$ to the candidate object instance's center position. For each reachable agent position, we perform a raycast from the agent's camera height $y^{(a)}$ to up to 6 random \textit{visibility points} on the object $\langle x^{(o)}, y^{(o)}, z^{(o)}\rangle$. Each object is annotated with visibility points, which are used as a fast approximation to determine if an object is visible with just using a few raycasts, instead of using full segmentation masks. If any of the raycasts from the agent's reachable position to the object's visibility point do not have any collisions with other objects (\eg the raycast does not collide with the outside of the fridge), and the L2 distance between $\langle x^{(o)}, y^{(o)}, z^{(o)}\rangle$ and $\langle x^{(a)}, y^{(a)}, z^{(a)}\rangle$ is less than 1 meter, then the object instance is considered successfully reachable by the agent.

To choose a target object type, we use an $\epsilon$-greedy sampling method. Specifically, with a probability of $\epsilon=0.2$, we randomly sample a target object type that has at least 1 reachable object instance in a given house. With a probability of $1-\epsilon$, the target object type is the target object type that has been most infrequently sampled in the training process. Since some objects appear much more frequently than others (\eg beds appear in many more houses than baseball bats), sampling based on the least commonly sampled target object types allows us to maintain a more uniform distribution of sampled target object types.

\paragraph{RoboTHOR.} RoboTHOR is evaluated in both a 0-shot and fine-tuned setting. For 0-shot, we take the pre-trained model on \env{}-10K and run it on the RoboTHOR evaluation tasks. For fine-tuning, we reduce $T$ to the 12 RoboTHOR target object types, shown in Table~\ref{tab:targetObjectTypes} and train on the 60 provided training scenes. We fine-tune for 29 million steps, before validation performance starts to go down, on a machine with 8 NVIDIA Quadro RTX 8000 GPUs. Fine-tuning took about 7 hours to complete.

\paragraph{HM3D-Semantics.} We evaluate on HM3D-Semantics in both a 0-shot and fine-tuned setting using the ``ProcTHOR'' and ``ProcTHOR+Large'' architectures described above, these two architectures have slightly different pretraining strategies. 

\noindent \emph{``ProcTHOR'' model.} For 0-shot, we take the pre-trained model on \env{}-10K, and run it on the HM3D-Semantics evaluation tasks. For fine-tuning, we reduce $T$ to the 6 target object types used in HM3D-Semantics (see Table~\ref{tab:targetObjectTypes}) and train on the 80 provided training houses. We use an early checkpoint from \env{} pre-training, specifically from after 220 million steps. We performed fine-tuning on a machine with 8 NVIDIA RTX A6000 GPUs for approximately 220M steps, which took about 43 hours to complete.

\noindent \emph{``ProcTHOR+Large'' model.} We pre-train this model using \env{}\textsc{Large}-10K a variant of \env{}-10K with houses sampled to better align to the distribution of houses in HM3D. In particular, \env{}\textsc{Large}-10K contains 10K procedurally generated houses each of which contains between 4 and 10 rooms (houses in \env{}\textsc{Large}-10K thus tend to be much larger than houses in \env{}-10K). Moreover, during pretraining we only train our agent to navigate to the 6 object categories used in HM3D-Semantics. Fine-tuning is done identically as above. We use an early checkpoint from \env{} pre-training, specifically from after 125 million steps. We performed fine-tuning on a machine with 8 NVIDIA RTX A6000 GPUs for approximately 185M steps taking 85 hours to complete.

\paragraph{AI2-iTHOR.} Similar to RoboTHOR and HM3D-Semantics, we use AI2-iTHOR for both 0-shot and fine-tuning. For 0-shot, we take the pre-trained model on \env{}-10K, and run it on the AI2-iTHOR evaluation tasks. Since the AI2-iTHOR evaluation tasks use the full set of target objects used during \env{} pre-training, we do not need to update $T$. For fine-tuning, we use a machine with 8 TITAN V GPUs. We fine-tune for approximately 2 million steps before validation performance starts to go down, which takes about 1.5 hours to complete.

\paragraph{ArchitecTHOR.} Since \elienv{} does not include any training scenes, we only use it for evaluation of the \env{} pre-trained model. As shown in Table~\ref{tab:targetObjectTypes}, \elienv{} evaluation uses the full-set of target object types that are used during \env{} pre-training.

\subsection{ArmPointNav experiments}

In ArmPointNav, we followed the same architecture as \cite{manipulathor}. The task is to move a target object from a starting location to a goal location using the relative location of the target in the agent's coordinate frame. The visual input is encoded using 3 convolutional layers followed by a linear layer to obtain a $512$ feature vector. The 3D relative coordinates, specifying the targets, are embedded using three linear layers to a $512$ embedding which combined with the visual encoding is input to the GRU. The agent is allowed to take up to 200 steps or the episode will automatically fail. 

\begin{table}[htbp]
    \centering
    \begin{tabular}{ l l }
        \toprule
        \textbf{Hyperparameter}\qquad\qquad\qquad & \textbf{Value}\\
        \midrule
        Learning rate & 3e-4\\
        Gradient steps & 128\\
        Discount factor ($\gamma$) & 0.99 \\
        GAE parameter ($\lambda$) & 0.95 \\
        Gradient clip norm & 0.5 \\
        Rotation Degrees & 45\\
        Step penalty & -0.01\\
        Number of RNN Layers & 1\\
        Rollouts per minibatch & 1\\
        Optimizer & Adam \cite{kingma2014adam} \\
        \toprule\\[-0.05in]
    \end{tabular}
    \caption{Training hyperparameters for ArmPointNav experiments.}
    \label{tab:apnHp}
\end{table}

\paragraph{ProcTHOR pre-training.} We pre-train our model on a subset of 7000 houses, on 58 object categories. For each episode, we move the agent to a random location, randomly choose an object in the room that is pickupable, and randomly select a target location. We train our model for 100M frames, running on 4 AWS g4dn.12xlarge machines. Running on a total of 16 GPUs and 192 CPU cores took 3 days of training. Table \ref{tab:apnHp} shows the hyperparameters used for pre-training.

\paragraph{AI2-iTHOR evaluation.} We evaluate our model on 20 test rooms of AI2-THOR (5 kitchens, 5 living rooms, 5 bedrooms, 5 bathrooms), on a subset of 28 object categories for a total of 528 tasks. We attempted to perform fine-tuning on AI2-iTHOR, but none of the fine-tuning models performed better than the zero-shot model trained with \env{} pre-training.

\subsection{Rearrangement experiments}

Following \cite{weihs2021visual,khandelwal2021simple}, we use imitation learning (IL) to train all models for the 1-phase modality of the task. We divide the full training of the final model into two stages: pre-training in \env{} and fine-tuning in AI2-iTHOR.


\begin{table}[htbp]
    \centering
    \begin{tabular}{ l l }
        \toprule
        \textbf{Hyperparameter}\qquad\qquad\qquad & \textbf{Value}\\
        \midrule
        Rollout timesteps & 64\\
        Batch size & 7,680 \\
        Learning rate & $7.4\cdot10^{-4}$\\
        Optimizer & Adam \cite{kingma2014adam} \\
        Gradient clip norm & 0.5 \\
        BC$^{\text{tf}=1}$ steps & 200,000 \\
        DAgger steps & 2,000,000 \\
        \toprule\\[-0.05in]
    \end{tabular}
    \caption{ProcTHOR pre-training hyperparameters for Rearrange experiments.}
    \label{tab:rearHp}
\end{table}

\paragraph{ProcTHOR pre-training.} We pre-train our model on a subset of 2,500 one and two-room \env{}-10K houses where a number of 1 to 5 objects are shuffled from their target poses in each episode, including two shuffle modalities: different openness degree (at most one object in an episode) and a different location (up to five objects in an episode). For each house, 20 episodes are sampled such that all shuffled objects are in the same room where the agent is initially spawned. We train with $2\cdot10^{5}$ steps of teacher forcing and 2 million steps of dataset aggregation \cite{Ross2011ARO}, followed by about 180 million steps of behavior cloning. We use a small set of 200 episodes sampled from 20 validation houses unseen during training to select a checkpoint to evaluate every 5 million steps.

Running on 6 AWS g4dn.12xlarge (totaling 24 GPUs and 288 virtual CPU cores), pre-training with 240 parallel simulations took 4 days. Table~\ref{tab:rearHp} shows the hyperparameters used during pre-training.

\paragraph{AI2-iTHOR fine-tuning.} We use the training dataset provided by \cite{roomr-challenge} (4,000 episodes over 80 single-room scenes), and a small subset of 200 episodes from the also provided full validation set to perform model selection. We fine-tune for 3 million steps with 64-step long rollouts, 6 additional million steps with 96-step long rollouts, and another 6 million steps with 128-step long rollouts.

Running on 8 Titan X GPUs and 56 virtual CPU cores, fine-tuning with 40 parallel simulations took 16 hours.

\section{Performance Benchmark}

To calculate the FPS performance benchmark shown in the Analysis section, we partitioned houses into small houses (1-3 room houses) and large houses (7-10 room houses). For the navigation benchmark, we perform move and rotate actions. For the interaction benchmark, we performing a pushing object action. For querying the environment for data, we obtain a piece of metadata from the environment that is not commonly provided at each time step (\eg checking the dimensions of the agent). At each time step, we render a single $3\times224\times224$ RGB image from the agent's egocentric perspective. Experiments were conducted on a server with 8 NVIDIA Quadro RTX 8000 GPUs. We employ 15 processes for the single GPU tests and 120 processes for the 8 GPU tests, evenly divided across the GPUs. Table~\ref{tab:baseFps} shows the comparisons to AI2-iTHOR and RoboTHOR.

\begin{table}
    \centering
    \small
    \resizebox{\textwidth}{!}{%
    \begin{tabular}{l cc c cc c cc}
        \toprule
        & \multicolumn{2}{c}{Navigation FPS} && \multicolumn{2}{c}{Isolated Interaction FPS} && \multicolumn{2}{c}{Environment Query FPS} \\
        \cmidrule{2-3}\cmidrule{5-6}\cmidrule{8-9}
        Compute & AI2-iTHOR & RoboTHOR && AI2-iTHOR & RoboTHOR && AI2-iTHOR & RoboTHOR \\
        \midrule
        8 GPUs & 5,779{\scriptsize$\pm$189} & 9,195{\scriptsize$\pm$294} && 5,411{\scriptsize$\pm$190} & 6,331{\scriptsize$\pm$137} && 463,446{\scriptsize$\pm$18,577} & 412,550{\scriptsize$\pm$21,806} \\
        1 GPU & 1,316{\scriptsize$\pm$19} & 1,648{\scriptsize$\pm$11} && 1,451{\scriptsize$\pm$72} & 1,539{\scriptsize$\pm$5} && 169,092{\scriptsize$\pm$4,232} & 163,660{\scriptsize$\pm$3,336}\\
        1 Process & 180{\scriptsize$\pm$9} & 340{\scriptsize$\pm$26} && 141{\scriptsize$\pm$2} & 217{\scriptsize$\pm$1} &&
        15,584{\scriptsize$\pm$156} & 15,578{\scriptsize$\pm$164}\\
        
        \cmidrule{2-3}\cmidrule{5-6}\cmidrule{8-9}
         & \env{}-S & \env{}-L && \env{}-S & \env{}-L && \env{}-S & \env{}-L \\
        \cmidrule{2-3}\cmidrule{5-6}\cmidrule{8-9}

        8 GPUs & 8,599{\scriptsize$\pm$359} & 3,208{\scriptsize$\pm$127} && 6,488{\scriptsize$\pm$250} & 2,861{\scriptsize$\pm$107} && 480,205{\scriptsize$\pm$19,684} & 433,587{\scriptsize$\pm$18,729} \\
        1 GPU & 1,427{\scriptsize$\pm$74} & 6,280{\scriptsize$\pm$40} && 1,265{\scriptsize$\pm$71}  & 597{\scriptsize$\pm$37} && 160,622{\scriptsize$\pm$2,846} & 157,567{\scriptsize$\pm$2,689}\\
        1 Process & 240{\scriptsize$\pm$69} & 115{\scriptsize$\pm$19} && 180{\scriptsize$\pm$42} & 93{\scriptsize$\pm$15} &&
        14,825{\scriptsize$\pm$199} & 14,916{\scriptsize$\pm$186}\\
        \bottomrule
    \end{tabular}%
    }
    \vspace{0.05in}
    \caption{Comparing performance benchmarks in \env{} to baselines in AI2-iTHOR and RoboTHOR. FPS for navigation, interaction, and querying the environment for data. \env{}-S and \env{}-L denotes small and large \env{} houses, respectively.}
    \label{tab:baseFps}
    \vspace{-0.2in}
\end{table}

\section{Broader Impact}

This work focuses on increasing the generalization abilities of robotic agents on various tasks. We specifically focus on robots that operate in household environments. More capable robotic agents can help improve the lives of many by assisting with cooking, cleaning, and providing social interaction. Furthermore, robots can provide a wide range of health benefits. For example, they could give domestic assistance to individuals with physical and mental disabilities and the elderly. They could provide social and emotional support to children, adolescents, and adults, such as delivering personalized educational content, reducing loneliness, and counseling in times of crisis. We can also use home-assisted robots to monitor and provide feedback on people's physical activity, sleep, and diet.

However, the adoption of home-assisted robots could have several undesirable social consequences. One is that home-assisted social robots may lead individuals to become more dependant on robots for companionship and care, leading to increased social isolation and loneliness. Another concern is that they may exacerbate existing inequities, as those who can afford to buy and maintain robots will have access to care and assistance that those who cannot will not. Furthermore, because robots would have access to sensitive information about people's daily lives, they could threaten privacy and security. Finally, robots have the potential to be exploited for malicious intent, such as for mass surveillance or being used for autonomous warfare. As a community, we need to work to reduce the risks of social robots while maximizing the benefits for the common good.

\end{document}